\documentclass[runningheads]{llncs}
\usepackage{graphicx}
\usepackage{amsmath,amssymb} 
\usepackage{color}
\usepackage{subfigure}
\usepackage[width=122mm,left=12mm,paperwidth=146mm,height=193mm,top=12mm,paperheight=217mm]{geometry}
\usepackage{url}
\urldef{\mailsa}\path{lizhen36@hku.hk, yizhouy@acm.org}
\urldef{\mailsb}\path{ganyk@mail2.sysu.edu.cn, xdliang328@gmail.com}
\urldef{\mailsc}\path{chengh9@mail.sysu.edu.cn, linliang@ieee.org}

\begin{document}
\pagestyle{headings}
\mainmatter

\title{LSTM-CF: Unifying Context Modeling and Fusion with LSTMs for RGB-D Scene Labeling} 

\titlerunning{LSTM-CF: Unifying Context Modeling and Fusion with LSTMs}

\authorrunning{Z. Li, Y. Gan, X. Liang, Y. Yu, H. Cheng, and L. Lin}

\author{Zhen Li$^1$\and Yukang Gan$^2$\and Xiaodan Liang$^2$\and Yizhou Yu$^1$\and \\Hui Cheng$^2$\and Liang Lin$^2$\thanks{The corresponding author is Liang Lin. This work was support by Projects on Faculty/ Student Exchange and Collaboration Scheme between the Higher Education in Hong Kong and the Mainland, Guangzhou Science and Technology Program under grant 1563000439, and Fundamental Research Funds for the Central Universities.}}


\institute{$^1$ Department of Computer Science, The University of Hong Kong, Hong Kong, \mailsa\\
           $^2$ School of Data and Computer Science, Sun Yat-sen University, Guangzhou, \mailsb\\ \mailsc}

\maketitle

\begin{abstract}
Semantic labeling of RGB-D scenes is crucial to many intelligent applications including perceptual robotics. It generates pixelwise and fine-grained label maps from simultaneously sensed photometric (RGB) and depth channels. This paper addresses this problem by i) developing a novel Long Short-Term Memorized Context Fusion (LSTM-CF) Model that captures and fuses contextual information from multiple channels of photometric and depth data, and ii) incorporating this model into deep convolutional neural networks (CNNs) for end-to-end training. Specifically, contexts in photometric and depth channels are, respectively, captured by stacking several convolutional layers and a long short-term memory layer; the memory layer encodes both short-range and long-range spatial dependencies in an image along the vertical direction. Another long short-term memorized fusion layer is set up to integrate the contexts along the vertical direction from different channels, and perform bi-directional propagation of the fused vertical contexts along the horizontal direction to obtain true 2D global contexts. At last, the fused contextual representation is concatenated with the convolutional features extracted from the photometric channels in order to improve the accuracy of fine-scale semantic labeling. Our proposed model has set a new state of the art, i.e., ${\bf{48.1}}\%$ and ${\bf{49.4}}\%$ average class accuracy over $37$ categories (${\bf{2.2}}\%$ and ${\bf{5.4}}\%$ improvement) on the large-scale SUNRGBD dataset and the NYUDv$2$ dataset, respectively.

\end{abstract}

\section{Introduction}

\begin{figure}[t]
\centering
\includegraphics[height=4cm]{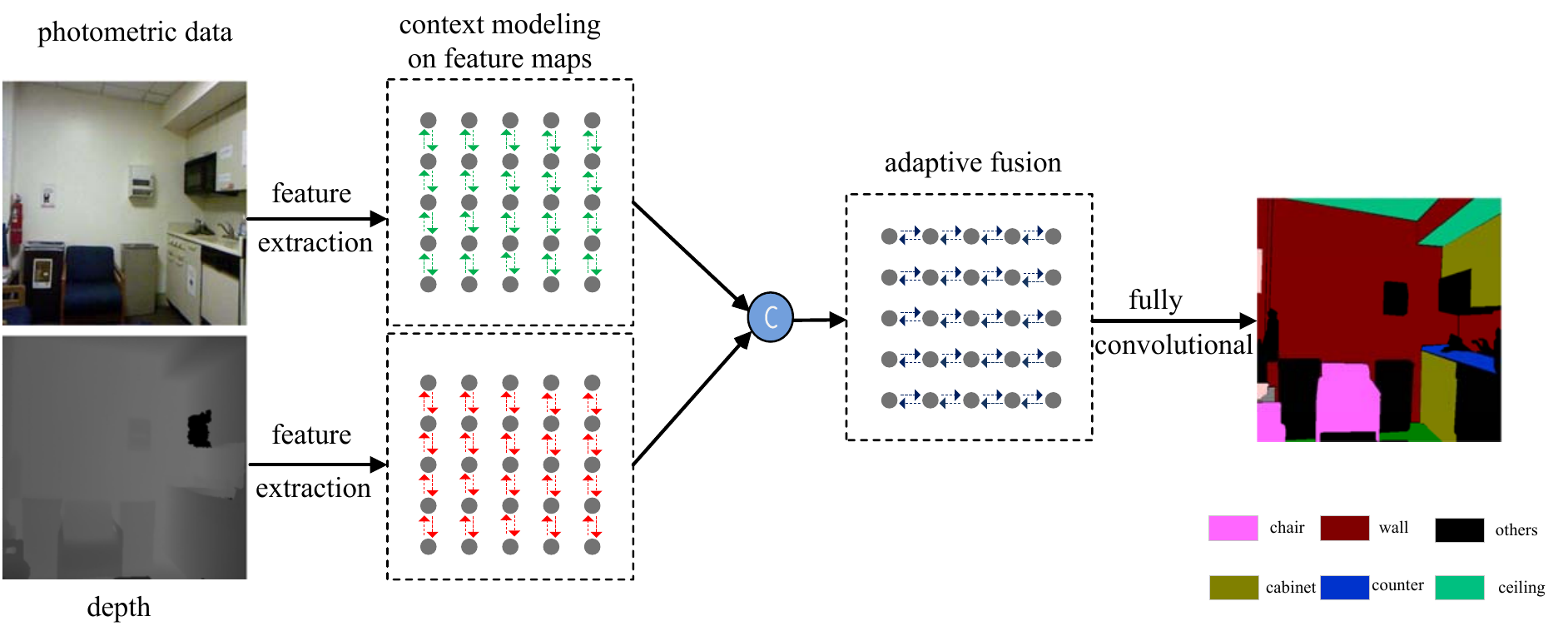}
\caption{An illustration of global context modeling and fusion for RGB-D images. Our LSTM-CF model first captures vertical contexts through a memory network layer encoding short- and long-range spatial dependencies along the vertical direction. After a concatenation operation (denoted by ``C'') over photometric and depth channels, our model utilizes another memory network layer to fuse vertical contexts from all channels in a data-driven way and performs bi-directional propagation along the horizontal direction to obtain true 2D global contexts. Best viewed in color.}
\label{fig:motivation}
\end{figure}

Scene labeling, also known as semantic scene segmentation, is one of the most fundamental problems in computer vision. It refers to associating every pixel in an image with a semantic label, such as table, road and wall, as illustrated in Fig. \ref{fig:motivation}. High-quality scene labeling can be beneficial to many intelligent tasks, including robot task planning~\cite{wu2014hierarchical}, pose estimation~\cite{hinterstoisser2012model}, plane segmentation~\cite{holz2011real}, context-based image retrieval~\cite{schuster2015generating}, and automatic photo adjustment~\cite{YanZWPY16}.

Previous work on scene labeling can be divided into two categories according to their target scenes: indoor and outdoor scenes. Compared with outdoor scene labeling \cite{farabet2013learning,gould2009decomposing,tighe2010superparsing}, indoor scene labeling is more challenging due to a larger set of semantic labels, more severe object occlusions, and more diverse object appearances \cite{khan2015integrating}. For example, indoor object classes, such as beds covered with different sheets and various appearances of curtains, are much harder to characterize than outdoor classes, e.g., roads, buildings, and sky, through photometric channels only. Recently, utilizing depth sensors to augment RGB data have effectively improved the performance of indoor scene labeling because the depth channel complements photometric channels with structural information. Nonetheless, two key issues remain open in the literature of RGB-D scene labeling.

(I) {\bf How to effectively represent and fuse the coexisting depth and photometric (RGB) data} For data representation, a batch of sophisticated hand-crafted features have been developed in previous methods. Such hand-crafted features are somewhat ad hoc and less discriminative than those RGB-D representations learned using convolutional neural networks (CNNs)~\cite{long2015fully,zheng2015conditional,chen2014semantic,gupta2014learning,song2015deep}. However, in these CNN-related works, the fusion of depth and photometric data has often been oversimplified. For instance, in \cite{gupta2014learning,song2015deep}, two independent CNNs are leveraged to extract features from depth and photometric data separately, and such features are simply concatenated before used for final classification. Overlooking the strong correlation between depth and photometric channels could inevitably harm semantic labeling.

(II) {\bf How to capture global scene contexts during feature learning} Current CNN-based scene labeling approaches can only capture local contextual information for every pixel due to their restricted receptive fields, resulting in suboptimal labeling results. In particular, long-range dependencies sometimes play a key role in distinguishing among different objects having similar appearances, e.g., labeling ``ceiling'' and ``floor'' in Fig. \ref{fig:motivation}, according to the global scene layout. To overcome this issue, graphical models, such as a conditional random field \cite{khan2015integrating,zheng2015conditional} or a mean-field approximation~\cite{liu2015semantic}, have been applied to improve prediction results in a post-processing step. These methods, however, separate context modeling from convolutional feature learning, which may give rise to suboptimal results on complex scenes due to less discriminative feature representation \cite{liang2015semantic}. An alternative class of methods adopts cascaded recurrent neural networks (RNNs) with gate structures, e.g., long short-term memory (LSTM) networks, to explicitly strengthen context modeling~\cite{liang2015semantic,byeon2015scene,pinheiro2014recurrent}. In these methods, the long- and short-range dependencies can be well memorized by sequentially running the network over individual pixels.

To address the aforementioned challenges, this paper proposes a novel Long Short-Term Memorized Context Fusion (LSTM-CF) model and demonstrates its superiority in RGB-D scene labeling. Fig. \ref{fig:motivation} illustrates the brief idea of using memory networks for context modeling and fusion of different channels. Our LSTM-CF model captures $2$D dependencies within an image by exploiting the cascaded bi-directional vertical and horizontal RNN models as introduced in \cite{visin2015renet}.

Our method constructs HHA images~\cite{gupta2014learning} for the depth channel through geometric encoding, and uses several convolutional layers for extracting features. Inspired by~\cite{visin2015renet}, these convolutional layers are followed by a memorized context layer to model both short-range and long-range spatial dependencies along the vertical direction. For photometric channels, we generate convolutional features using the Deeplab network~\cite{chen2014semantic}, which is also followed by a memorized context layer for context modeling along the vertical direction. Afterwards, a memorized fusion layer is set up to integrate the contexts along the vertical direction from both photometric and depth channels, and perform bi-directional propagation of the fused vertical contexts along the horizontal direction to obtain true 2D global contexts. Considering the features differences, e.g., signal frequency and other characteristics (color/geometry) \cite{song2015sun}, our fusion layer facilitates deep integration of contextual information from multiple channels in a data-driven manner rather than simply concatenating different feature vectors. Since photometric channels usually contain finer details in comparison to the depth channel \cite{song2015sun}, we further enhance the network with cross-layer connections that append convolutional features of the photometric channels to the fused global contexts before the final fully convolutional layer, which predicts pixel-wise semantic labels. Various layers in our LSTM-CF model are tightly integrated, and the entire network is amenable to end-to-end training and testing.

In summary, this paper has the following contributions to the literature of RGB-D scene labeling.
\begin{itemize}
\item It proposes a novel Long Short-Term Memorized Context Fusion (LSTM-CF) Model, which is capable of capturing image contexts from a global perspective and deeply fusing contextual information from multiple sources (i.e., depth and photometric channels).

\item It proposes to jointly optimize LSTM layers and convolutional layers for achieving better performance in semantic scene labeling. Context modeling and fusion are incorporated into the deep network architecture to enhance the discriminative power of feature representation. This architecture can also be extended to other similar tasks such as object/part parsing.

\item It is demonstrated on the large-scale SUNRGBD benchmark (including $10355$ images) and canonical NYUDv2 benchmark that our method outperforms existing state-of-the-art methods. In addition, it is found that our scene labeling results can be leveraged to improve the groundtruth annotations of newly captured $3943$ RGB-D images in SUNRGBD dataset.
\end{itemize}

\section{Related work}
{\bf{Scene Labeling:}} Scene labeling has caught researchers' attention frequently \cite{farabet2013learning,zheng2015conditional,chen2014semantic,liang2015semantic,byeon2015scene,pinheiro2014recurrent,kumar2010efficiently} in recent years. Instead of extracting features from over-segmented images, recent methods usually utilize powerful CNN layers as the feature extractor, taking advantage of fully convolutional networks (FCNs) \cite{long2015fully} and its variants \cite{kendall2015bayesian} to obtain pixel-wise dense features. Another main challenge for scene labeling is the fusion of local and global contexts, i.e., taking advantage of global contexts to refine local decisions. For instance, \cite{farabet2013learning} exploits families of segmentations or trees to generate segment candidates. \cite{lempitsky2011pylon} utilizes an inference method based on graph cut to achieve image labeling. A pixel-wise conditional random forest is used in \cite{zheng2015conditional,chen2014semantic} to directly optimize a deep CNN-driven cost function. Most of the above models improve accuracy through carefully designed processing on the predicted confidence map instead of proposing more powerful discriminative features, which usually results in suboptimal prediction results \cite{liang2015semantic}. The topological structure of recurrent neural networks (RNNs) is used to model short- and long-range dependencies in \cite{liang2015semantic,pinheiro2014recurrent}. In \cite{byeon2015scene}, a multi-directional RNN is leveraged to extract local and global contexts without using a CNN, which is well suited for low-resolution and relatively simple scene labeling problems. In contrast, our model can jointly optimize LSTM layers and convolutional layers to explicitly improve discriminative feature learning for local and global context modeling and fusion.

{\bf{Scene Labeling in RGB-D images:}} With more and more convenient access to affordable depth sensors, scene labeling in RGB-D images \cite{khan2015integrating,gupta2014learning,song2015deep,ren2012rgb,gupta2015indoor,wang2015unsupervised} enables a rapid progress of scene understanding. Various sophisticated hand-crafted features are utilized in previous state-of-the-art methods. Specifically, kernel descriptions based on traditional multi-channel features, such as color, depth gradient, and surface normal, are used as photometric and depth features \cite{ren2012rgb}. A rich feature set containing various traditional features, e.g., SIFT, HOG, LBP and plane orientation, are used as local appearance features and plane appearance features in \cite{khan2015integrating}. HOG features of RGB images and HOG+HH (histogram of height) features of depth images are extracted as representations in \cite{gupta2015indoor} for training successive classifiers. In \cite{husain2017combining}, proposed distance-from-wall features are exploited to improve scene labeling performance.
In addition, an unsupervised joint feature learning and encoding model is proposed for scene labeling in \cite{wang2015unsupervised}. However, due to the limited number of RGB-D images, deep learning for scene labeling in RGB-D images was not as appealing as that for RGB images. The release of the SUNRGBD dataset, which includes most of the previously popular datasets,
may have changed this situation~\cite{gupta2014learning,song2015deep}.

Another main challenge imposed by scene labeling in RGB-D images is the fusion of contextual representations of different sources (i.e., depth and photometric data). For instance, in \cite{gupta2014learning,song2015deep}, two independent CNNs are leveraged to extract features from the depth and photometric data separately, which are then simply concatenated for class prediction. Ignoring the strong correlation between depth and photometric channels usually negatively affects semantic labeling. In contrast, instead of simply concatenating features from multiple sources, the memorized fusion layer in our model facilitates the integration of contextual information from different sources in a data-driven manner,

{\bf{RNN for Image Processing:}} Recurrent neural networks (RNNs) represent a type of neural networks with loop connections \cite{schmidhuber1989local}. They are designed to capture dependencies across a distance larger than the extent of local neighborhoods. In previous work, RNN models have not been widely used partially due to the difficulty to train such models, especially for sequential data with long-range dependencies~\cite{Bengio1994}. Fortunately, RNNs with gate and memory structures, e.g., long short-term memory (LSTM)~\cite{Schmidhuber1997}, can artificially learn to remember and forget information by using specific gates to control the information flow. Although RNNs have an outstanding capability to capture short-range and long-range dependencies, there exist problems for applying RNNs to image processing due to the fact that, unlike data in natural language processing (NLP) tasks, images do not have a natural sequential structure. Thus, different strategies have been proposed to overcome this problem. Specifically, in \cite{visin2015renet}, cascaded bi-directional vertical and horizonal RNN layers are designed for modeling 2D dependencies in images. A multi-dimensional RNN with LSTM unit has been applied to handwriting~\cite{graves2009offline}. A parallel multi-dimensional LSTM for image segmentation has been proposed in \cite{stollenga2015parallel}. In this paper, we propose an LSTM-CF model consisting of memorized context layers and a memorized fusion layer to capture image contexts from a global perspective and fuse contextual representations from different sources.

\section{LSTM-CF Model}
\begin{figure}[t]
\centering
\includegraphics[height=5.0cm]{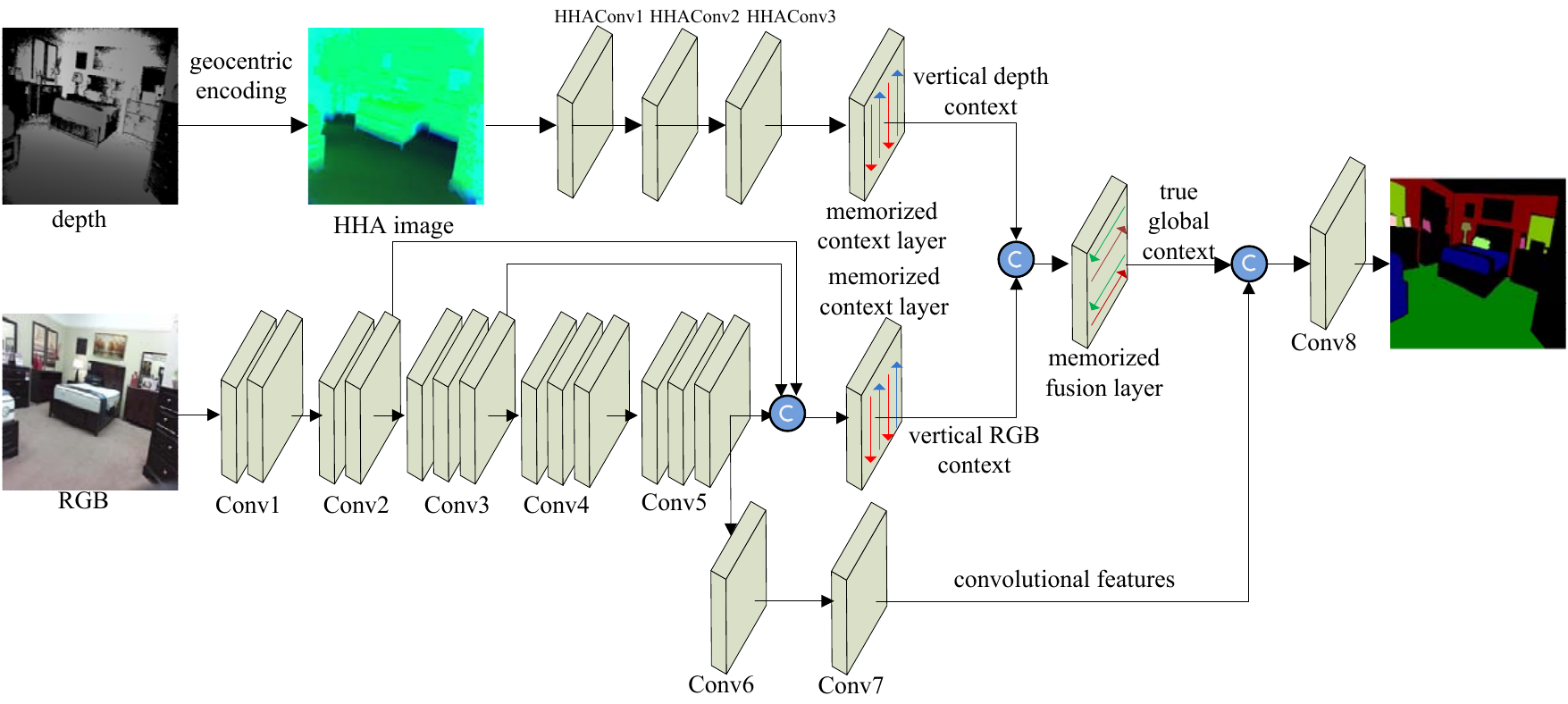}
\caption{Our LSTM-CF model for RGB-D scene labeling. The input consists of both photometric and depth channels. Vertical contexts in photometric and depth channels are computed in parallel using cascaded convolutional layers and a memorized context layer. Vertical photometric (color) and depth contexts are fused and bi-directionally propagated along the horizontal direction via another memorized fusion layer to obtain true 2D global contexts. The fused global contexts and the final convolutional features of photometric channels are then concatenated together and fed into the final convolutional layer for pixel-wise scene labeling. ``C'' stands for the concatenation operation.}
\label{fig:pipeline}
\end{figure}

As illustrated in Fig. \ref{fig:pipeline}, our end-to-end LSTM-CF model for RGB-D scene labeling consists of four components, layers for vertical depth context extraction, layers for vertical photometric context extraction, a memorized fusion layer for incorporating vertical photometric and depth contexts as true 2D global contexts, and a final layer for pixel-wise scene labeling given concatenated convolutional features and global contexts. The inputs to our model include both photometric and depth images. The path for extracting global contexts from the photometric image consists of multiple convolutional layers and an extra memorized context layer. On the other hand, the depth image is first encoded as an HHA image, which is fed into three convolutional layers \cite{song2015deep} and an extra memorized context layer for global depth context extraction. The other component, a memorized fusion layer, is responsible for fusing previously extracted global RGB and depth contexts in a data-driven manner. On top of the memorized fusion layer, the final convolutional feature of photometric channels and the fused global context are concatenated together and fed into the final fully convolutional layer, which performs pixel-wise scene labeling with the softmax activation function.

\subsection{Memorized Vertical Depth Context}
Given a depth image, we use the HHA representation proposed in \cite{gupta2014learning} to encode geometric properties of the depth image in three channels, i.e., disparity, surface normal and height. Different from \cite{gupta2014learning}, the encoded HHA image in our pipeline is fed into three randomly initialized convolutional layers (to obtain a feature map with the same resolution as that in the RGB path) instead of layers taken from the model pre-trained on the ILSVRC$2012$ dataset. This is because the color distribution of HHA images is different from that of natural images (see Fig. \ref{fig:pipeline}) according to \cite{song2015sun}. One top of the third convolutional layer (i.e., HHAConv$3$), there is an extra memorized context layer from Renet~\cite{visin2015renet}, which performs bi-directional propagation of local contextual features from the convolutional layers along the vertical direction. For better understanding, we denote the feature map HHAConv$3$ as $F=\{f_{i,j}\}$, where $F \in \mathbb{R}^{w \times h \times c}$ with $w, h$ and $c$ representing the width, height and the number of channels. Since we perform pixel-wise scene labeling, every patch in this Renet layer only contains a single pixel. Thus,  vertical memorized context layer (here we choose LSTM as recurrent unit) can be formulated as
\begin{align}
&h^f_{i,j} = \text{LSTM}(h^f_{i,j-1}, f_{i,j}),~~~~ \text{for}~~ j= 1, \ldots, h \\
&h^b_{i,j} = \text{LSTM}(h^b_{i,j+1}, f_{i,j}),~~~~ \text{for}~~ j= h, \ldots, 1,
\end{align}
where $h^f$ and $h^b$ stand for the hidden states of the forward and backward LSTM. In the forward LSTM, the unit at pixel $(i,j)$ takes $h^f_{i,j-1} \in \mathbb{R}^{d}$ and $f_{i,j}\in \mathbb{R}^{c}$ as input, and its output is calculated as follows according to \cite{Schmidhuber1997}. The operations in the backward LSTM can be defined similarly.
\begin{align}
&\text{gate}_i = \delta(W_{if}f_{i,j}+W_{ih}h^f_{i,j-1}+b_i) \nonumber\\
&\text{gate}_f = \delta(W_{ff}f_{i,j}+W_{fh}h^f_{i,j-1}+b_f) \nonumber\\
&\text{gate}_o = \delta(W_{of}f_{i,j}+W_{oh}h^f_{i,j-1}+b_o) \nonumber\\
&\text{gate}_c = \tanh(W_{cf}f_{i,j}+W_{ch}h^f_{i,j-1}+b_c) \nonumber\\
&c_{i,j} = \text{gate}_f \odot c_{i,j-1} + \text{gate}_i \odot \text{gate}_c \nonumber\\
&h^f_{i,j} = \tanh(\text{gate}_o \odot c_{i,j})
\end{align}
Finally, pixel-wise vertical depth contexts are collectively represented as a map, $C_{\text{depth}} \in \mathbb{R}^{w \times h \times {2d}}$, where $2d$ is the total number of output channels from the vertical memorized context layer.

\subsection{Memorized Vertical Photometric Context}
In the component for extracting global RGB contexts, we adapt the Deeplab architecture proposed in \cite{chen2014semantic}. Different from existing Deeplab variants, we concatenate features at three different scales to enrich the feature representation. This is inspired by the network architecture in \cite{li2016deep}. Specifically, since there exists hole operations in Deeplab convolutional layers, feature maps from Conv$2\_2$, Conv$3\_3$ and Conv$5\_3$ have sufficient initial resolutions. They can be further elevated to the same resolution using interpolation. Corresponding pixel-wise features from these three elevated feature maps are then concatenated together before being fed into the subsequent memorized fusion layer, which again performs bi-directional propagation to produce vertical photometric contexts. Here pixel-wise vertical photometric contexts can also be represented as a map, $C_{\text{RGB}} \in \mathbb{R}^{w \times h \times {2d}}$, which has the same dimensionalities as the map for vertical depth contexts.

\subsection{Memorized Context Fusion}
So far vertical depth and photometric contexts are computed independently in parallel. Instead of simply concatenating these two types of contexts, the memorized fusion layer, which performs horizontal bi-directional propagation from Renet, is exploited for adaptively fusing vertical depth and RGB contexts in a data-driven manner, and the output from this layer can be regarded as the fused representation of both types of contexts. Such fusion can generate more discriminative features through end-to-end training. The input and output dimensions of the fusion layer are set to $\mathbb{R}^{w \times h \times {4d}}$ and $\mathbb{R}^{w \times h \times {2d}}$, respectively.

Note that there are two separate memorized context layers in the photometric and depth paths of our architecture. Since the memorized context layer and the memorized fusion layer are two symmetric components of the original Renet~\cite{visin2015renet}, a more natural and symmetric alternative would have a single memorized context layer preceding the memorized fusion layer in our model (i.e., whole structure of Renet including cascaded bi-directional vertical and horizonal memorized layer) and let the memorized fusion layer incorporate the features from the RGB and depth paths. Nonetheless, in our experiments, this alternative network architecture gave rise to slightly worse performance.

\subsection{Scene Labeling}
Between photometric and depth images, photometric images contain more details and semantic information that can help scene labeling in comparison with sparse and discontinuous depth images \cite{song2015deep}. Nonetheless, depth images can provide auxiliary geometric information for improving scene labeling performance. Thus, we design a cross-layer combination that integrates pixel-wise convolutional features (i.e., Conv$7$ in Fig. \ref{fig:pipeline}) from the photometric image with fused global contexts from the memorized fusion layer as the final pixel-wise features, which are fed into the last fully convolutional layer with softmax activation to perform scene labeling at every pixel location.

\section{Experimental Results}
\subsection{Experimental Setting}
{\bf{Datasets:}} We evaluate our proposed model for RGB-D scene labeling on three public benchmarks, SUNRGBD, NYUDv$2$ and SUN$3$D. SUNRGBD~\cite{song2015sun} is the largest dataset currently available, consisting of $10355$ RGB-D images captured from four different depth sensors. It includes most previous datasets, such as NYUDv$2$ depth \cite{silberman2012indoor}, Berkeley B$3$DO \cite{janoch2013category}, and SUN3D \cite{xiao2013sun3d}, as well as $3943$ newly captured RGB-D images \cite{song2015sun}. $5285$ of these images are predefined for training and the remaining $5050$ images constitute the testing set~\cite{song2015deep}.\\
{\bf{Implementation Details:}} In our experiments, a slightly modified Deeplab pipeline~\cite{chen2014semantic} is adopted as the basic network in our RGB path for extracting convolutional feature maps because of its high performance. It is initialized with the publicly available VGG-$16$ model pre-trained on ImageNet. For the purpose of pixel-wise scene labeling, this architecture transforms the last two fully connected layers in the standard VGG-$16$ to convolutional layers with $1\times 1$ kernels. For the parallel depth path, three randomly initialized CNN layers with max pooling are leveraged for depth feature extraction. In each path, on top of the aforementioned convolutional network, a vertically bi-directional LSTM layer implements the memorized context layer, and models both short-range and long-range spatial dependencies. Then, another horizontally bi-directional LSTM layer implements the memorized fusion layer, and is used to adaptively integrate the global contexts from the two paths. In addition, there is a cross-layer combination of final convolutional features (i.e., Conv$7$) and the integrated global representation from the horizontal LSTM layer.

Since the SUNRGBD dataset was collected by four different depth sensors, each input image is cropped to $426\times 426$ (the smallest resolution of these four sensors)~\cite{song2015deep}. During fine-tuning, the learning rate for newly added layers, including HHAConv$1$, HHAConv$2$, HHAConv$3$, the memorized context layers, the memorized fusion layer and Conv$8$, is initialized to $10^{-2}$, and the learning rate for those pre-trained layers of VGG-$16$ is initialized to $10^{-4}$. All weights in the newly added convolutional layers are initialized using a Gaussian distribution with a standard deviation equal to $0.01$, and the weights in the LSTM layers are randomly initialized with a uniform distribution over $[-0.01, 0.01]$. The number of hidden memory cells in a memorized context layer or a memorized fusion layer is set to $100$, and the size of feature maps is $54\times54$. We train all the layers in our deep network simultaneously using SGD with a momentum $0.9$, the batch size is set to one (due to limited GPU memory) and the weight decay is $0.0005$. The entire deep network is implemented on the publicly available platform Caffe \cite{jia2014caffe} and is trained on a single NVIDIA GeForce GTX TITAN X GPU with $12$GB memory \footnote{LSTM-CF model is publicly available at: https://github.com/icemansina/LSTM-CF}. It takes about 1 day to train our deep network. In the testing stage, an RGB-D image takes $0.15$s on average, which is significantly faster than pervious methods, i.e., the testing time in \cite{khan2015integrating,ren2012rgb} is around $1.5$s.

\subsection{Results and Comparisons}
According to \cite{song2015deep,kendall2015bayesian}, performance is evaluated by comparing class-wise Jaccard Index, i.e., $n_{ii}/t_i$, and average Jaccard Index, i.e., $(1/n_{cl})\sum_i n_{ii}/t_i$, where $n_{ij}$ is the number of pixels annotated as class $i$ and predicted to be class $j$, $n_{cl}$ is the number of different classes, and $t_i=\sum_j n_{ij}$ is the total number of pixels annotated as class $i$ \cite{long2015fully}.\\
\begin{table}[t]
\centering
\caption{Comparison of scene labeling results on SUNRGBD using class-wise and average Jaccard Index. We compare our model with results reported in \protect\cite{song2015sun}, \protect\cite{liu2011sift}, \protect\cite{ren2012rgb} and previous state-of-the-art result in \protect\cite{kendall2015bayesian}. Boldface numbers mean best performance.}
\label{table:sunrgbd}
\resizebox{1.0\textwidth}{!}
{
\begin{tabular}{|l|l|l|l|l|l|l|l|l|l|l|l|l|l|l|l|l|l|l|l|}
\hline
~          & Wall          & floor         & cabinet       & bed           & chair         & sofa          & table         & door          & window        & bookshelf     & picture       & counter       & blinds        & desk          & shelves       & curtain       & dresser       & pillow        & mirror        \\ \hline
\cite{song2015sun}& 37.8          & 45.0          & 17.4          & 21.8          & 16.9          & 12.8          & 18.5          & 6.1           & 9.6           & 9.4           & 4.6           & 2.2           & 2.4           & 7.3           & 1.0           & 4.3           & 2.2           & 2.3           & 6.9           \\ \hline
\cite{song2015sun}   & 32.1          & 42.6          & 2.9           & 6.4           & 21.5          & 4.1           & 12.5          & 3.4           & 5.0           & 0.8           & 3.3           & 1.7           & 14.8          & 2.0           & 15.3          & 2.0           & 1.4           & 1.2           & 0.9           \\ \hline
\cite{song2015sun}   & 36.4          & 45.8          & 15.4          & 23.3          & 19.9          & 11.6          & 19.3          & 6.0           & 7.9           & 12.8          & 3.6           & 5.2           & 2.2           & 7.0           & 1.7           & 4.4           & 5.4           & 3.1           & 5.6           \\ \hline
\cite{liu2011sift}   & 38.9          & 47.2          & 18.8          & 21.5          & 17.2          & 13.4          & 20.4          & 6.8           & 11.0          & 9.6           & 6.1           & 2.6           & 3.6           & 7.3           & 1.2           & 6.9           & 2.4           & 2.6           & 6.2           \\ \hline
\cite{liu2011sift} & 33.3          & 43.8          & 3.0           & 6.3           & 22.3          & 3.9           & 12.9          & 3.8           & 5.6           & 0.9           & 3.8           & 2.2           & 32.6          & 2.0           & 10.1          & 3.6           & 1.8           & 1.1           & 1.0           \\ \hline
\cite{liu2011sift} & 37.8          & 48.3          & 17.2          & 23.6          & 20.8          & 12.1          & 20.9          & 6.8           & 9.0           & 13.1          & 4.4           & 6.2           & 2.4           & 6.8           & 1.0           & 7.8           & 4.8           & 3.2           & 6.4           \\ \hline
\cite{ren2012rgb}       & 43.2          & 78.6          & 26.2          & 42.5          & 33.2          & 40.6          & 34.3          & 33.2          & 43.6          & 23.1          & 57.2 & 31.8          & 42.3          & 12.1          & \textbf{18.4} & 59.1          & 31.4          & 49.5 & 24.8          \\ \hline
\cite{kendall2015bayesian}      & \textbf{80.2} & \textbf{90.9} & \textbf{58.9} & \textbf{64.8} & \textbf{76.0} & \textbf{58.6} & \textbf{62.6} & \textbf{47.7} & \textbf{66.4} & {31.2} & \textbf{63.6}          & 33.8 & 46.7 & 19.7 & 16.2           & \textbf{67.0} & \textbf{42.3} & \textbf{57.1}          & 39.1 \\ \hline
Ours       & {74.9} & {82.3} & {47.3} & {62.1} & {67.7} & {55.5} & {57.8} & {45.6} & {52.8} & \textbf{43.1} & 56.7          & \textbf{39.4} & \textbf{48.6} & \textbf{37.3} & 9.6           & 63.4 & 35.0 & 45.8          & \textbf{44.5} \\ \hline
~          & floormat      & clothes       & ceiling       & books         & fridge        & tv            & paper         & towel         & shower        & box           & board         & person        & nightstand    & toilet        & sink          & lamp          & bathhub       & bag           & mean          \\ \hline
\cite{song2015sun}     & 0.0           & 1.2           & 27.9          & 4.1           & 7.0           & 1.6           & 1.5           & 1.9           & 0.0           & 0.6           & 7.4           & 0.0           & 1.1           & 8.9           & 14.0          & 0.9           & 0.6           & 0.9           & 8.3           \\ \hline
\cite{song2015sun}   & 0.0           & 0.3           & 9.7           & 0.6           & 0.0           & 0.9           & 0.0           & 0.1           & 0.0           & 1.0           & 2.7           & 0.3           & 2.6           & 2.3           & 1.1           & 0.7           & 0.0           & 0.4           & 5.3           \\ \hline
\cite{song2015sun}   & 0.0           & 1.4           & 35.8          & 6.1           & 9.5           & 0.7           & 1.4           & 0.2           & 0.0           & 0.6           & 7.6           & 0.7           & 1.7           & 12.0          & 15.2          & 0.9           & 1.1           & 0.6           & 9.0           \\ \hline
\cite{liu2011sift}& 0.0           & 1.3           & 39.1          & 5.9           & 7.1           & 1.4           & 1.5           & 2.2           & 0.0           & 0.7           & 10.4          & 0.0           & 1.5           & 12.3          & 14.8          & 1.3           & 0.9           & 1.1           & 9.3           \\ \hline
\cite{liu2011sift}& 0.0           & 0.6           & 13.9          & 0.5           & 0.0           & 0.9           & 0.4           & 0.3           & 0.0           & 0.7           & 3.5           & 0.3           & 1.5           & 2.6           & 1.2           & 0.8           & 0.0           & 0.5           & 6.0           \\ \hline
\cite{liu2011sift}& 0.0           & 1.6           & 49.2          & 8.7           & 10.1          & 0.6           & 1.4           & 0.2           & 0.0           & 0.8           & 8.6           & 0.8           & 1.8           & 14.9          & 16.8          & 1.2           & 1.1           & 1.3           & 10.1          \\ \hline
\cite{ren2012rgb}& \textbf{5.6}  & 27.0          & \textbf{84.5} & 35.7          & 24.2          & 36.5          & 26.8          & 19.2          & \textbf{9.0}  & 11.7          & 51.4 & 35.7          & 25.0          & 64.1          & 53.0          & 44.2          & 47.0          & 18.6          & 36.3          \\ \hline
\cite{kendall2015bayesian}& 0.1           & 24.4 & 84.0          & \textbf{48.7} & 21.3 & 49.5 & 30.6 & 18.8 & 0.1             & 24.1 & \textbf{56.8}          & 17.9 & \textbf{42.9} & \textbf{73.0} & 66.2 & 48.8 & 45.1 & \textbf{24.1} & 45.9 \\ \hline
Ours       & 0.0           & \textbf{28.4} & 68.0          & 47.9 & \textbf{61.5} & \textbf{52.1} & \textbf{36.4} & \textbf{36.7} & 0             & \textbf{38.1} & 48.1          & \textbf{72.6} & 36.4 & 68.8 & \textbf{67.9} & \textbf{58.0} & \textbf{65.6} & 23.6 & \textbf{48.1} \\ \hline
\end{tabular}
}
\end{table}

\begin{figure}[t]
\centering
\subfigure[SUNRGBD]
{\includegraphics[width=1.5in]{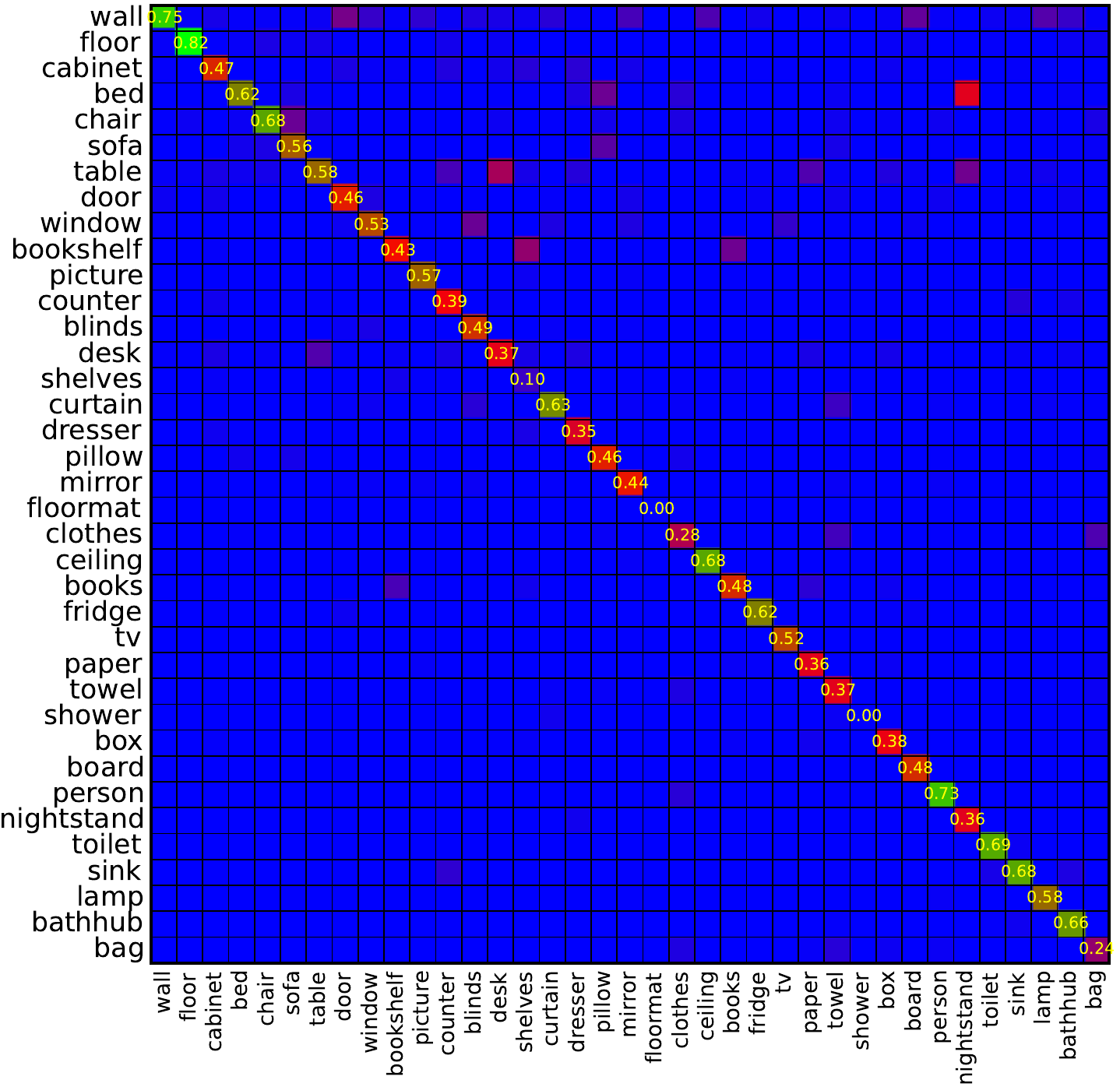}}\quad\quad\quad
\subfigure[NYUDv$2$]
{\includegraphics[width=1.5in]{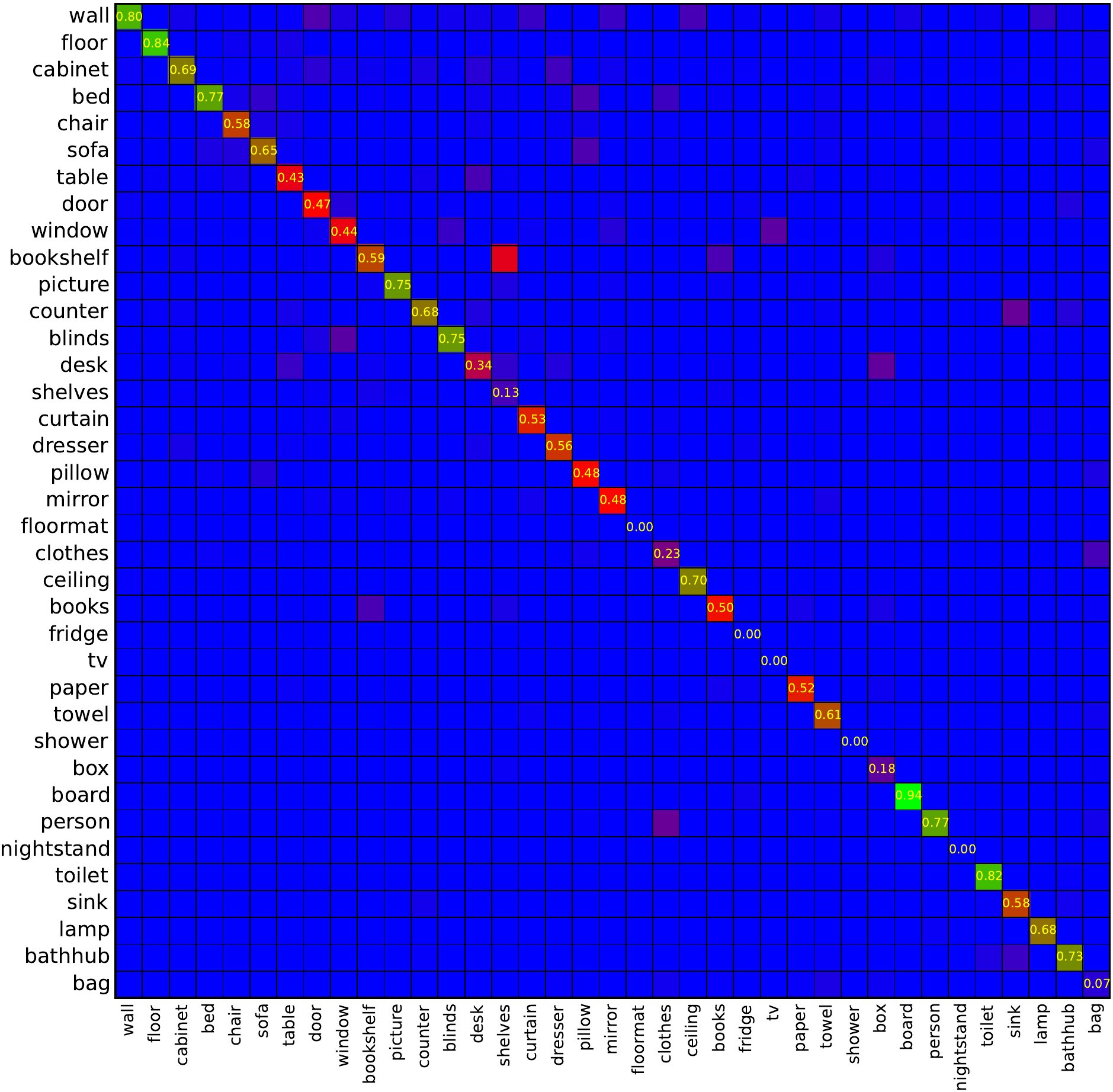}}
\caption{Confusion matrix for SUNRGBD and NYUDv$2$. Class-wise Jaccard Index is shown on the diagonal. Best viewed in color.}
\label{fig:confusion}
\end{figure}

{\bf{SUNRGBD dataset \cite{song2015sun}}}: The performance and comparison results on SUNRGBD are shown in Table \ref{table:sunrgbd}. Our proposed architecture can outperform existing techniques: $2.2\%$ higher than the performance reported in \cite{kendall2015bayesian}, $11.8\%$ higher than that in \cite{ren2012rgb}, $38\%$ higher than that in \cite{liu2011sift} and $39.1\%$ higher than that in \cite{song2015sun} in terms of $37$-class average Jaccard Index. Improvements can be observed in $15$ class-wise Jaccard Indices. For a better understanding, we also show the confusion matrix for this dataset in Fig. \ref{fig:confusion}(a).
It is worth mentioning that our proposed architecture and most previous methods achieve zero accuracy on two categories, i.e., floormat and shower, which mainly results from an imbalanced data distribution instead of the capacity of our model.

{\bf{NYUDv$2$ dataset:}} To further verify the effectiveness of our architecture and have more comparisons with existing state-of-the-art methods, we also conduct experiments on the NYUDv$2$ dataset. The results are presented in Table \ref{table:nyuv2}, where the $13$-class average Jaccard Index of our model is $20.3\%$ higher than that in \cite{couprie2014toward}. Class frequencies and the confusion matrix are also shown in Table \ref{table:nyuv2} and Fig. \ref{fig:confusion}(b) respectively. According to the reported results, our proposed architecture gains $5.6\%$ and $5.5\%$ improvement in average Jaccard Index over \cite{khan2015integrating} and FCN-$32$s~\cite{long2015fully} respectively. Considering the listed class frequencies, our proposed model significantly outperforms existing methods on high frequency categories and most low frequency categories, which primarily owes to the convolutional features of the RGB image and the fused global contexts of the complete RGB-D image. In terms of labeling categories with small and complex regions, e.g., pillows and chairs, our method also achieves a large improvement, which can be verified in the following visual comparisons.

\begin{table}[t]
\centering
\caption{Comparison of scene labeling on NYUDv$2$. We compare our proposed model with existing state-of-the-art methods, i.e., \protect\cite{silberman2012indoor}, \protect\cite{ren2012rgb}, \protect\cite{gupta2015indoor}, \protect\cite{wang2015unsupervised} and \protect\cite{khan2015integrating}. Class-wise Jaccard Index and average Jaccard Index of $37$ classes are presented. `Freq' stands for class frequency. Boldface numbers mean best performance.}
\label{table:nyuv2}
\resizebox{1.0\textwidth}{!}
{
\begin{tabular}{|l|l|l|l|l|l|l|l|l|l|l|l|l|l|l|l|l|l|l|l|}
\hline
~               & Wall          & floor         & cabinet       & bed           & chair         & sofa          & table         & door          & window        & bookshelf     & picture       & counter       & blinds        & desk          & shelves       & curtain       & dresser       & pillow        & mirror        \\ \hline
Freq       & 21.4          & 9.1           & 6.2           & 3.8           & 3.3           & 2.7           & 2.1           & 2.2           & 2.1           & 1.9           & 2.1           & 1.4           & 1.7           & 1.1           & 1.0           & 1.1           & 0.9           & 0.8           & 1.0           \\ \hline
\cite{silberman2012indoor} & 60.7          & 77.8          & 33.0          & 40.3          & 32.4          & 25.3          & 21.0          & 5.9           & 29.7          & 22.7          & 35.7          & 33.1          & 40.6          & 4.7           & 3.3           & 27.4          & 13.3          & 18.9          & 4.4           \\ \hline
\cite{ren2012rgb}      & 60.0          & 74.4          & 37.1          & 42.3          & 32.5          & 28.2          & 16.6          & 12.9          & 27.7          & 17.3          & 32.4          & 38.6          & 26.5          & 10.1          & 6.1           & 27.6          & 7.0           & 19.7          & 17.9          \\ \hline
\cite{gupta2015indoor}    & 67.4          & 80.5          & 41.4          & 56.4          & 40.4          & 44.8          & 30.0          & 12.1          & 34.1          & 20.5          & 38.7          & 50.7          & 44.7          & 10.1          & 1.6           & 26.3          & 21.6          & 31.3          & 14.6          \\ \hline
\cite{wang2015unsupervised} & 61.4          & 66.4          & 38.2          & 43.9          & 34.4          & 33.8          & 22.6          & 8.3          & 27.6          & 17.6          & 27.7          & 30.2          & 33.6          & 5.1          & 2.7           & 18.9          & 16.8          & 12.5          & 10.7          \\ \hline
\cite{khan2015integrating}      & 65.7          & 62.5          & 40.1          & 32.1          & 44.5          & 50.8          & 43.5          & \textbf{51.6} & \textbf{49.2} & 36.3          & 41.4          & 39.2          & 55.8          & \textbf{48.0} & \textbf{45.2} & 53.1          & 55.3          & \textbf{50.5} & 46.1          \\ \hline
Ours            & \textbf{79.6} & \textbf{83.5} & \textbf{69.3} & \textbf{77.0} & \textbf{58.3} & \textbf{64.9} & \textbf{42.6} & 47.0          & 43.6          & \textbf{59.5} & \textbf{74.5} & \textbf{68.2} & \textbf{74.6} & 33.6          & 13.1          & \textbf{53.2} & \textbf{56.5} & 48.0          & \textbf{47.7} \\ \hline
~               & floormat      & clothes       & ceiling       & books         & fridge        & tv            & paper         & towel         & shower        & box           & board         & person        & nightstand    & toilet        & sink          & lamp          & bathhub       & bag           & mean          \\ \hline
Freq       & 0.7           & 0.7           & 1.4           & 0.6           & 0.6           & 0.5           & 0.4           & 0.4           & 0.4           & 0.3           & 0.3           & 0.3           & 0.3           & 0.3           & 0.3           & 0.3           & 0.3           & 0.2           &               \\ \hline
\cite{silberman2012indoor} & 7.1           & 6.5           & 73.2          & 5.5           & 1.4           & 5.7           & 12.7          & 0.1           & 3.6           & 0.1           & 0.0           & 6.6           & 6.3           & 26.7          & 25.1          & 15.9          & 0.0           & 0.0           & 17.5          \\ \hline
\cite{ren2012rgb}       & 20.1          & 9.5           & 53.9          & 14.8          & 1.9           & 18.6          & 11.7          & 12.6          & 5.4           & 3.3           & 0.2           & 13.6          & 9.2           & 35.2          & 28.9          & 14.2          & 7.8           & 1.2           & 20.2          \\ \hline
\cite{gupta2015indoor}    & 28.2          & 8.0           & 61.8          & 5.8           & 14.5          & 14.4          & 14.1          & 19.8          & 6.0           & 1.1           & 12.9          & 1.5           & 15.7          & 52.5          & 47.9          & 31.2          & 29.4          & 0.2           & 30.0          \\ \hline
\cite{wang2015unsupervised} & 13.8          & 2.7          & 46.1          & 3.6          & 2.9          & 3.2          & 2.6          & 6.2          & 6.1          & 0.8          & 28.2          & 5          & 6.9          & 32          & 20.9           & 5.4          & 16.2          & 0.2          & 29.2          \\ \hline
\cite{khan2015integrating}      & \textbf{54.1} & \textbf{35.4} & 50.6          & 39.1          & \textbf{53.6} & \textbf{50.1} & 35.4          & 39.9          & \textbf{41.8} & \textbf{36.3} & 60.6          & 35.6          & \textbf{32.5} & 31.8          & 22.5          & 26.3          & 38.5          & \textbf{37.3} & 43.9          \\ \hline
Ours            & 0.0           & 22.7          & \textbf{70.2} & \textbf{49.7} & 0.0           & 0.0           & \textbf{52.1} & \textbf{60.6} & 0             & 17.6          & \textbf{93.9} & \textbf{77.0} & 0             & \textbf{81.8} & \textbf{58.4} & \textbf{67.6} & \textbf{72.6} & 7.5           & \textbf{49.4} \\ \hline
\end{tabular}
}
\end{table}

{\bf{SUN$3$D dataset:}} Table \ref{table.sun3d} gives comparison results on the $1539$ test images in the SUN$3$D dataset. For fair comparison, the $12$-class average Jaccard Index is used in the comparison with the state-of-the-art results recently reported in \cite{khan2015integrating}. Note that the $12$-class accuracy of our network is calculated through the model previously trained for $37$ classes. Our model substantially outperforms the one from \cite{khan2015integrating} on large planar regions such as those labeled as floors and ceilings. This also results from the incorporated convolutional features and the fused global contexts.
These comparison results further confirm the power and generalization capability of our LSTM-based model.
\begin{table}[t]
\centering
\caption{Comparison of class-wise Jaccard Index and $12$-class average Jaccard Index on SUN$3$D.}
\label{table.sun3d}
\begin{tabular}{l|l|l|l|l|l|l|l|l|l|l|l|l|l}
\hline
     & wall        & floor       & bed         & chair       & table       & counter     & curtain     & ceiling     & tv          & toilet      & bathtub     & bag         & Mean          \\ \hline
\cite{khan2015integrating}     & \textbf{73}          & 35          & \textbf{71} & 35          & 30          & \textbf{52} & 68          & 27          & 56          & 23          & 49          & \textbf{29} & 45.7         \\ \hline
Ours & \textbf{73} & \textbf{86} & 32          & \textbf{65} & \textbf{57} & 22          & \textbf{76} & \textbf{69} & \textbf{75} & \textbf{62} & \textbf{62} & 23          & \textbf{58.5} \\ \hline
\end{tabular}
\end{table}
\begin{table}[t]
\centering
\caption{Ablation Study}
\label{table.ablation}
\begin{tabular}{l|c}
\hline
\multicolumn{1}{c|}{Model}      & \multicolumn{1}{c}{Mean Accuracy} \\ \hline
Without RGB path, using Deeplab+Renet for depth path & $15.8\%$                            \\ \hline
Without depth path                  & $43.7\%$                            \\ \hline
Without multi-scale RGB feature concatenation         & $42.1\%$                            \\ \hline
Without cross-layer integration of RGB convolutional features & $15.2\%$                            \\ \hline
Without memorized fusion layer  & $44.7\%$                            \\ \hline
Without memorized context layers & $45.7\%$                            \\ \hline
Without any memorized (context or fusion) layers     & $45.0\%$                            \\ \hline
\end{tabular}
\end{table}

\begin{figure}[t]
\centering
\subfigure[]{
\begin{minipage}[]{0.095\textwidth}
\includegraphics[width=1\textwidth]{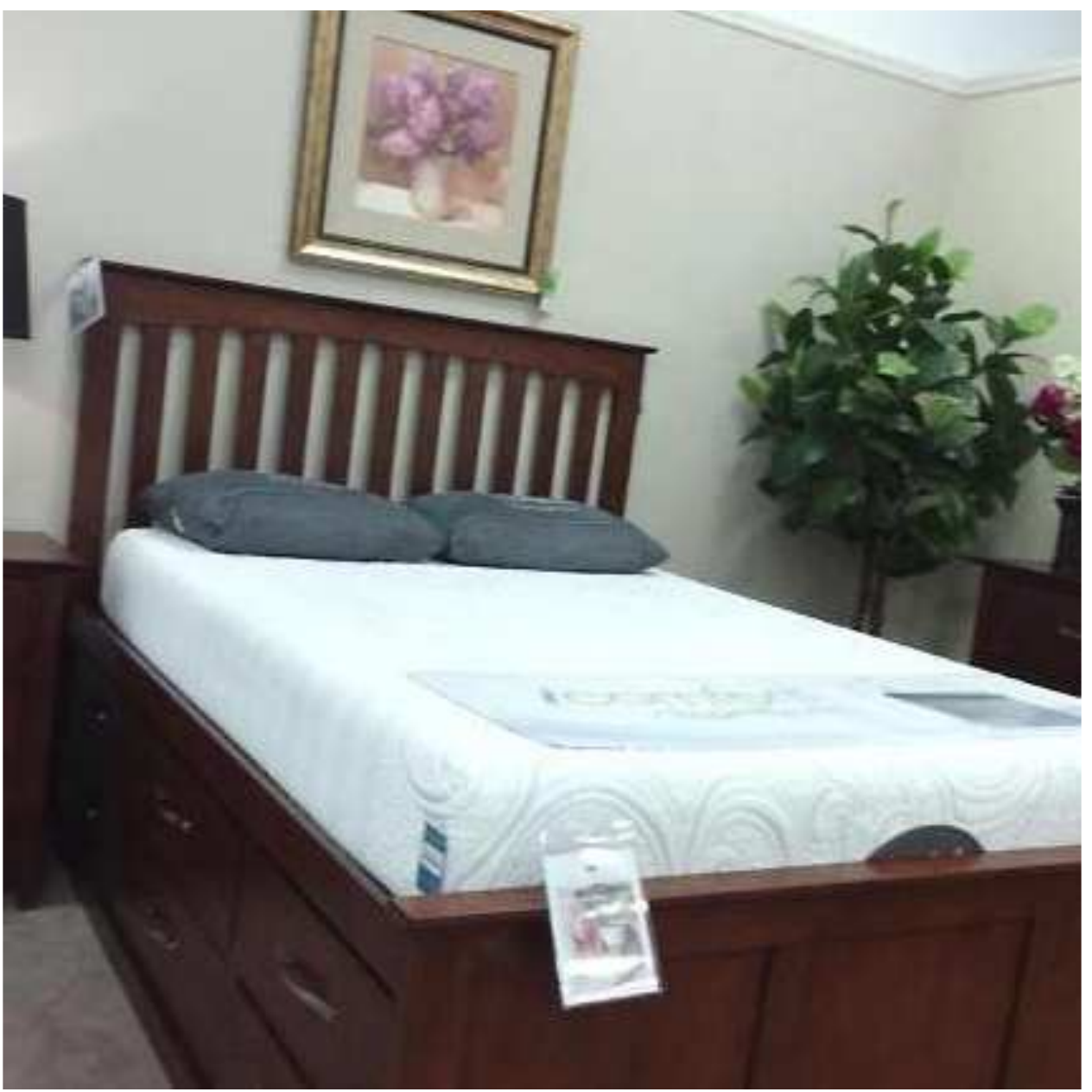} \\
\includegraphics[width=1\textwidth]{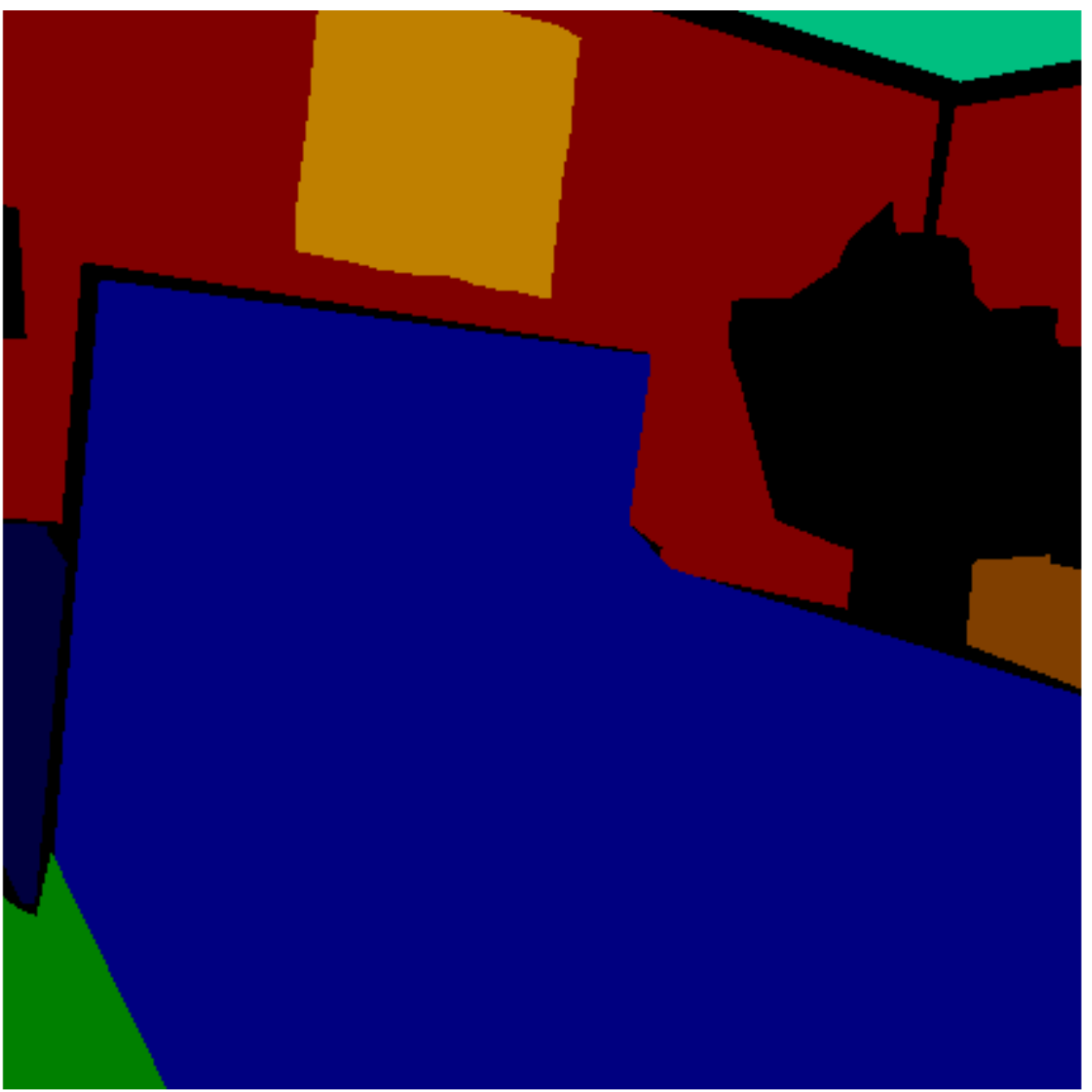} \\
\includegraphics[width=\textwidth]{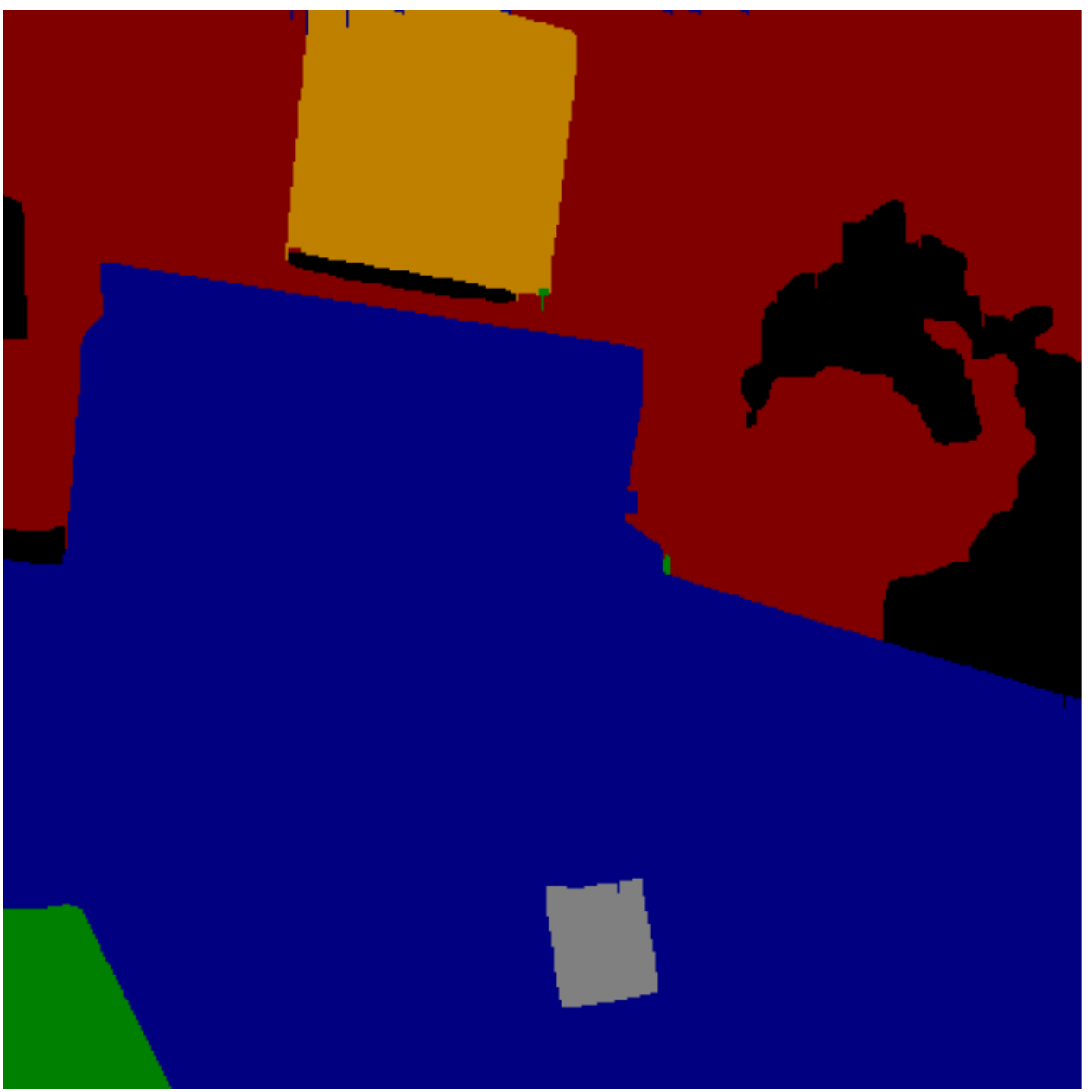}
\end{minipage}
}
\subfigure[]{
\begin{minipage}[]{0.095\textwidth}
\includegraphics[width=1\textwidth]{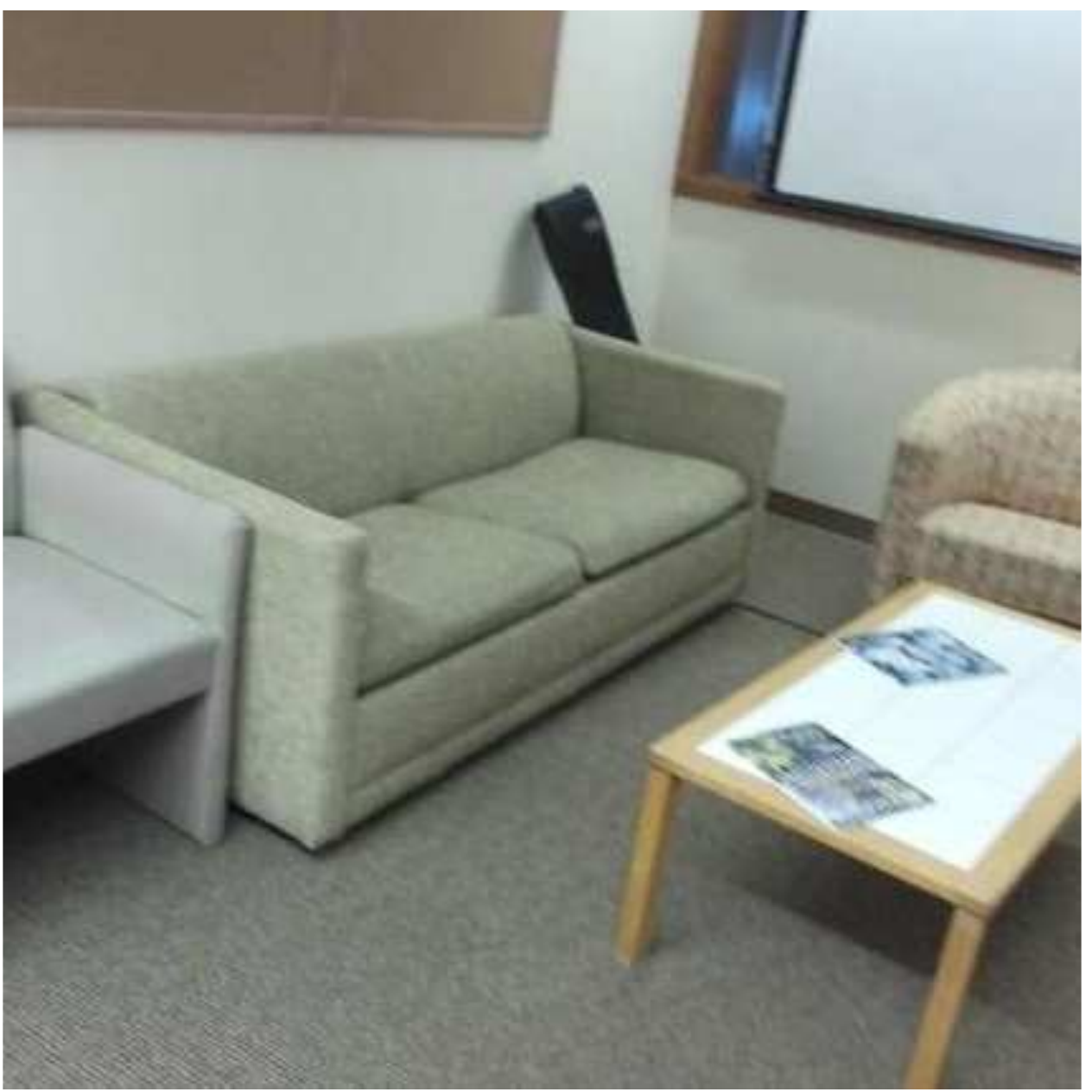} \\
\includegraphics[width=1\textwidth]{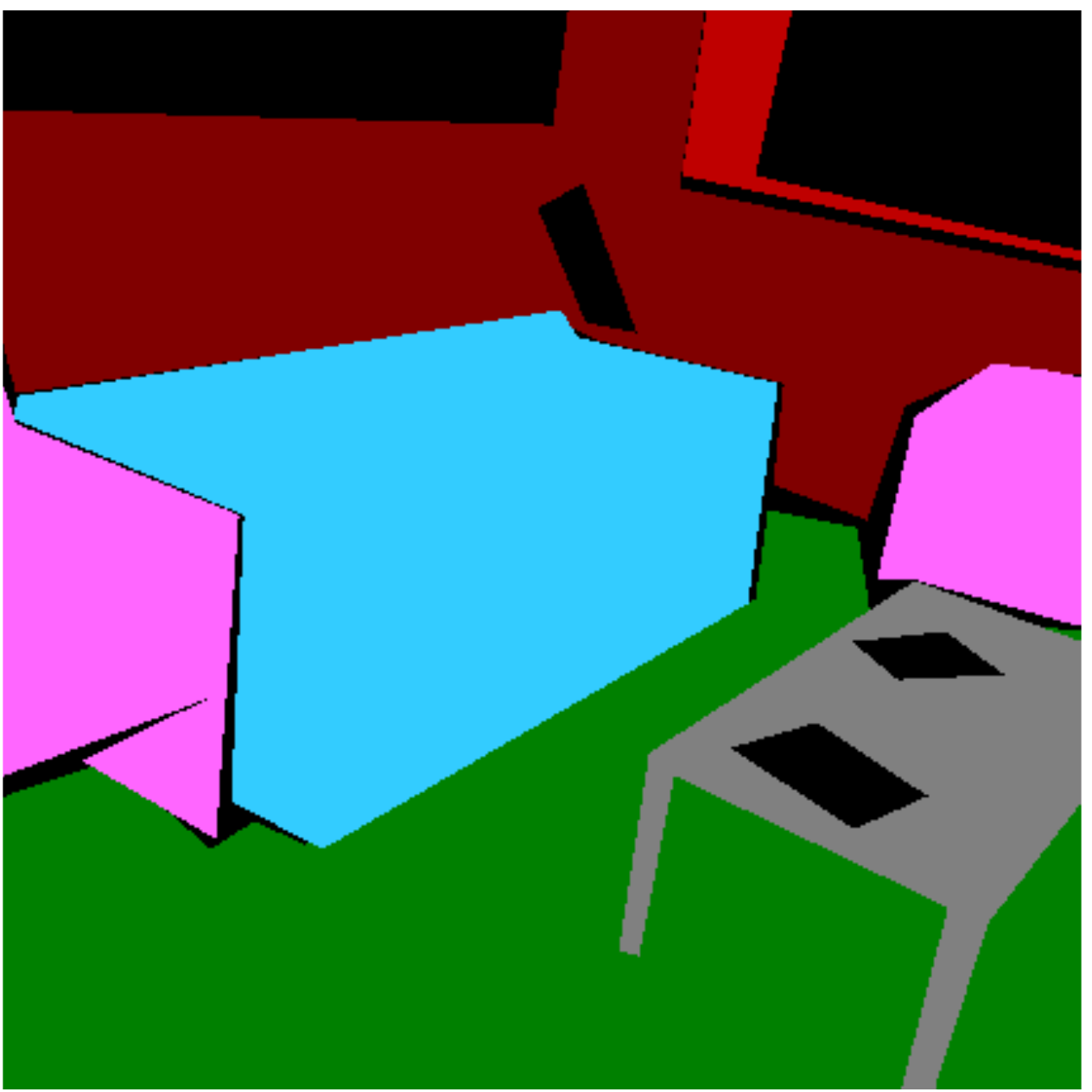} \\
\includegraphics[width=\textwidth]{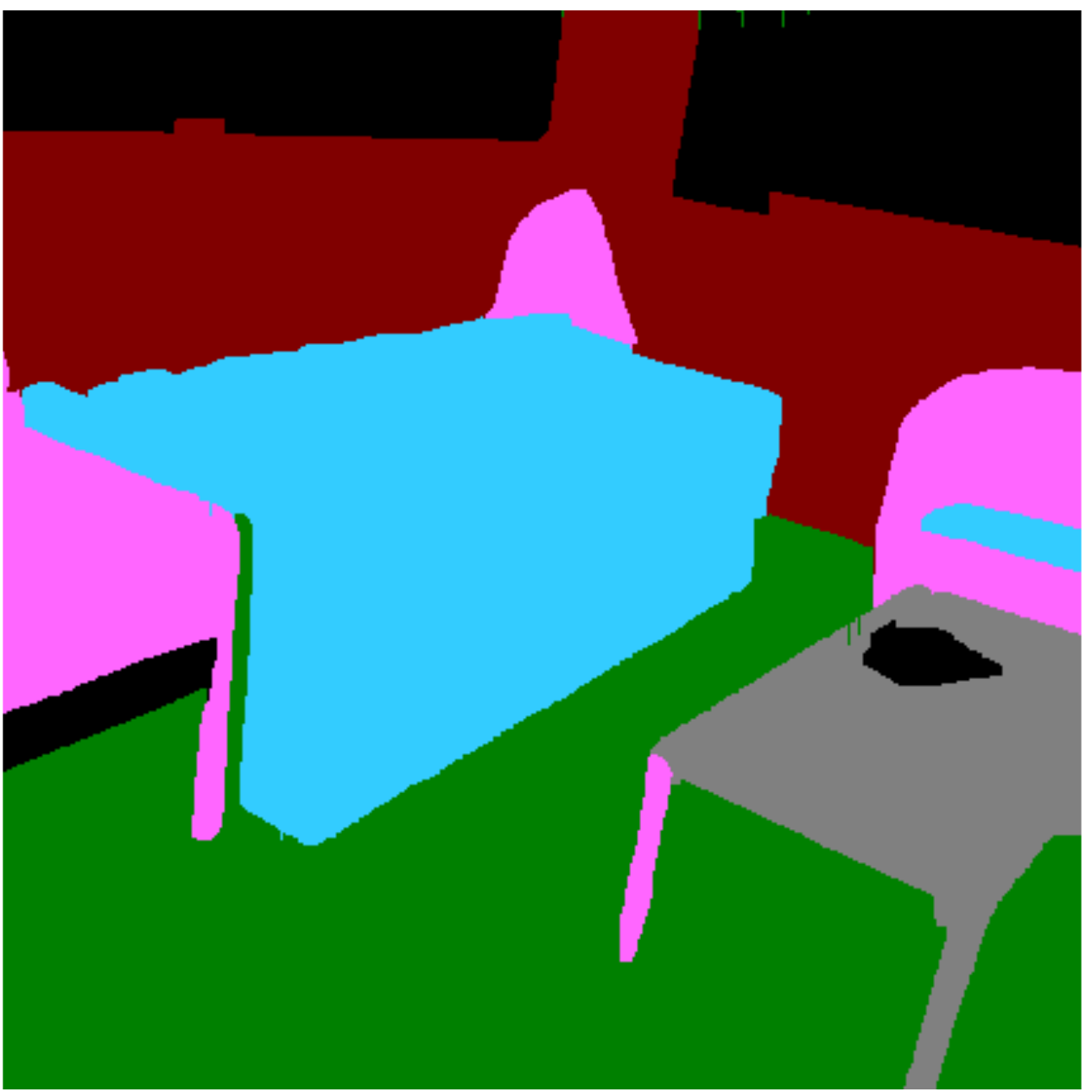}
\end{minipage}
}
\subfigure[]{
\begin{minipage}[]{0.095\textwidth}
\includegraphics[width=1\textwidth]{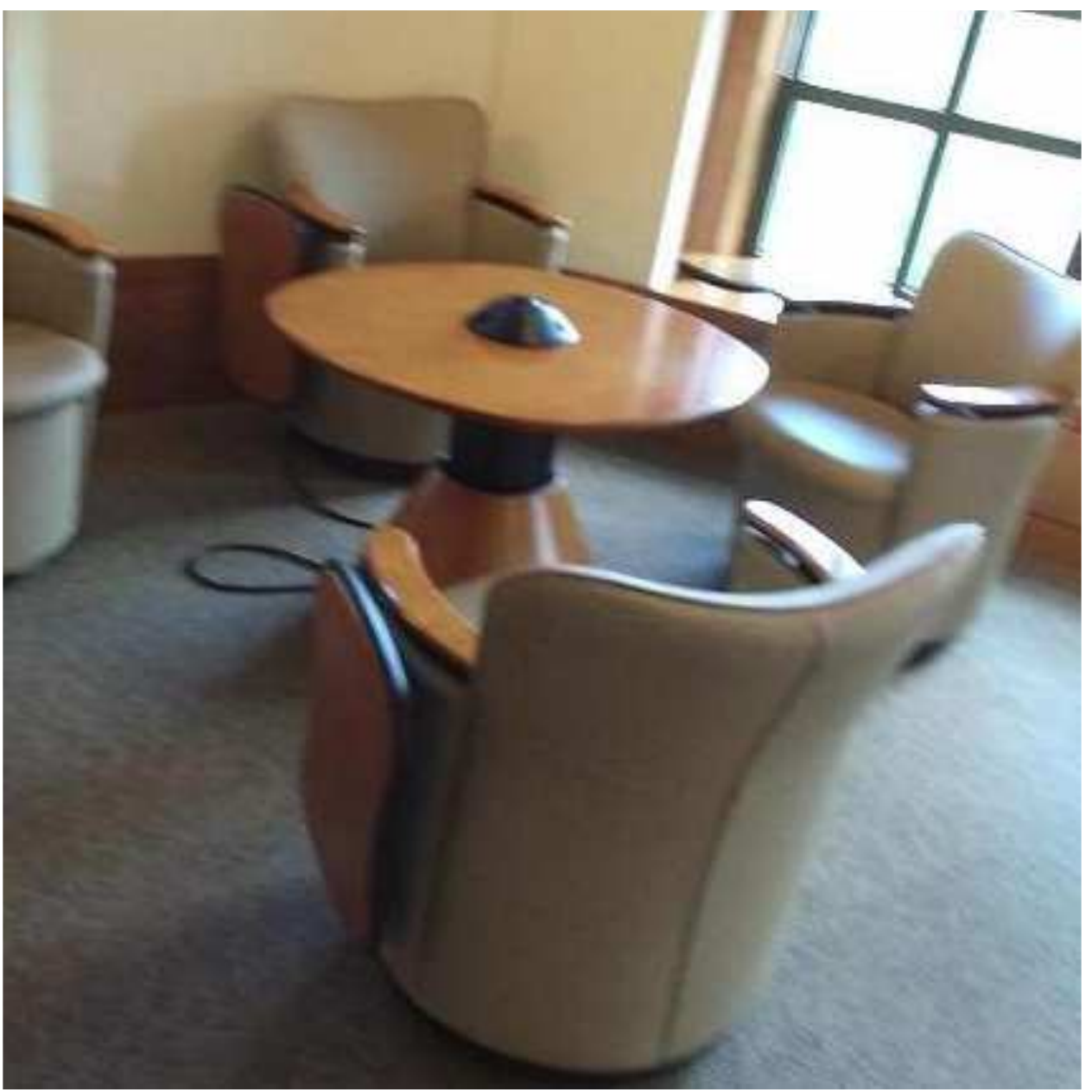} \\
\includegraphics[width=1\textwidth]{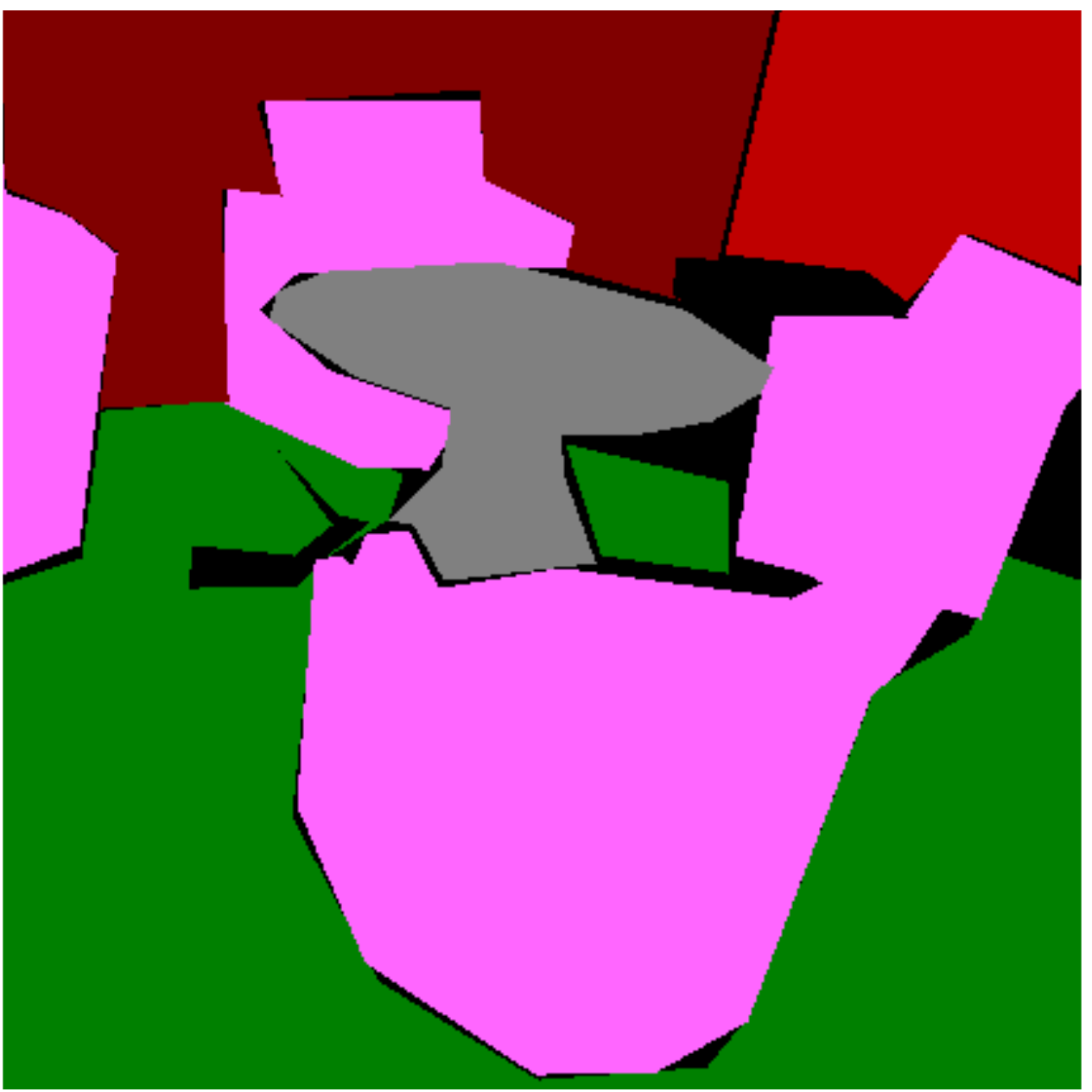} \\
\includegraphics[width=\textwidth]{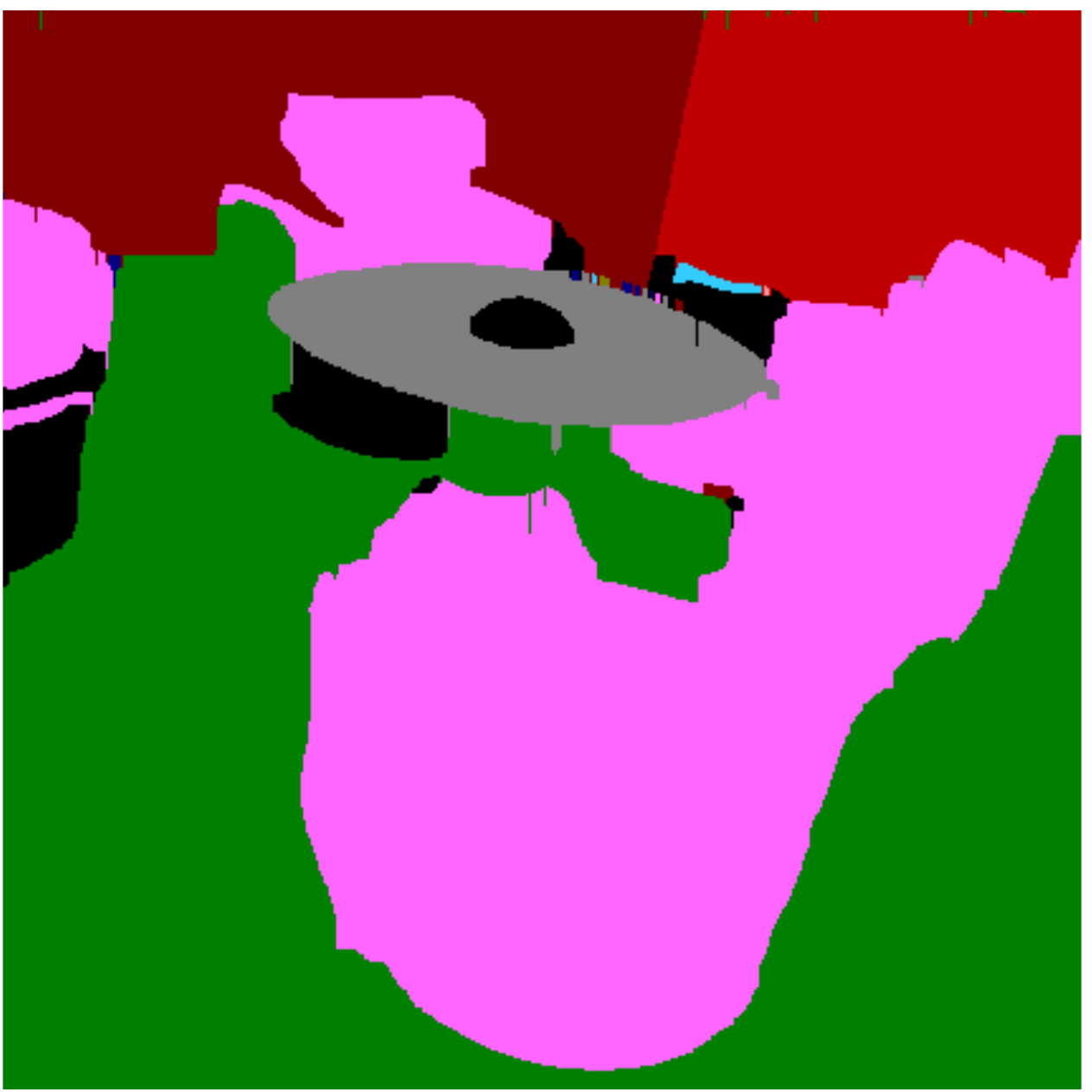}
\end{minipage}
}
\subfigure[]{
\begin{minipage}[]{0.095\textwidth}
\includegraphics[width=1\textwidth]{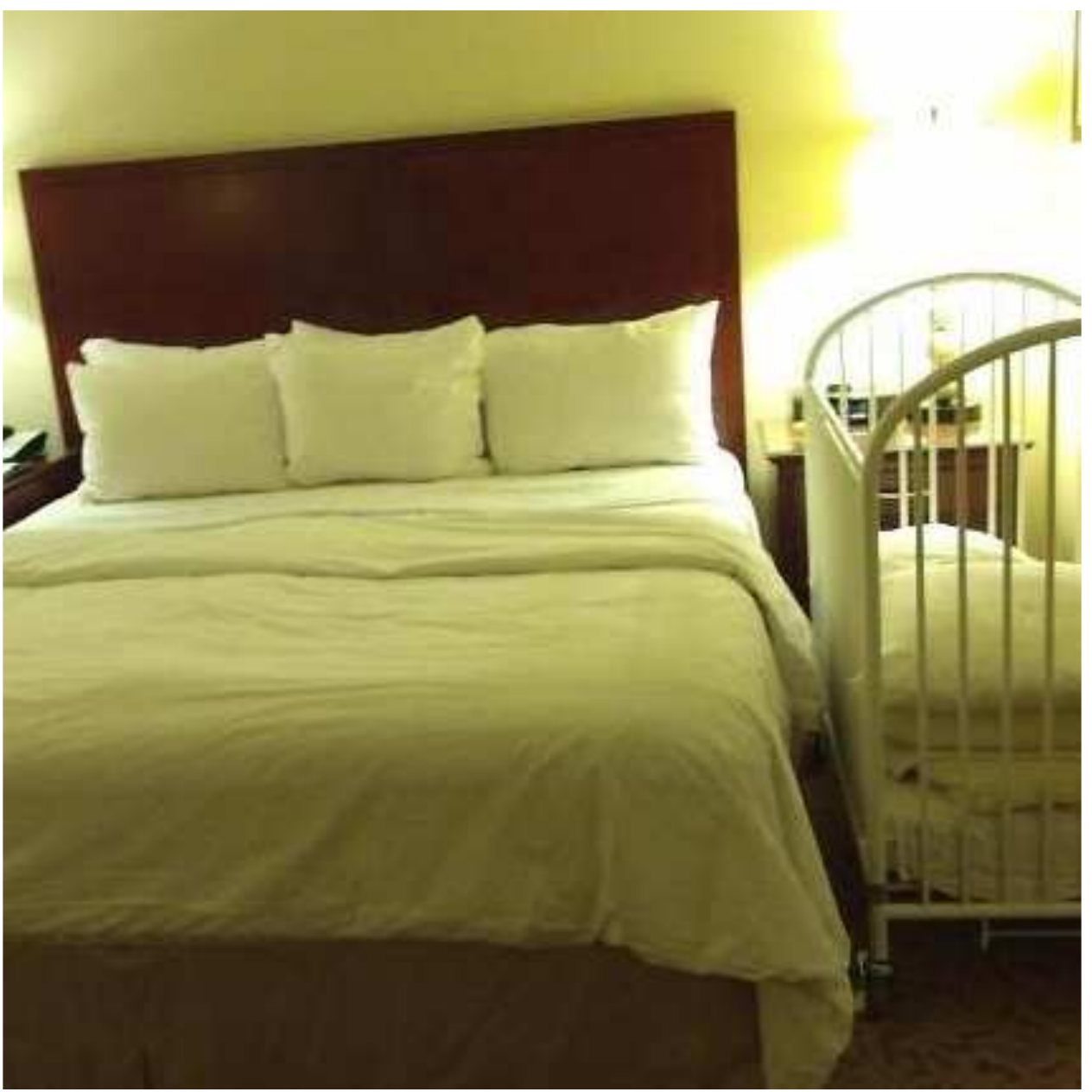} \\
\includegraphics[width=1\textwidth]{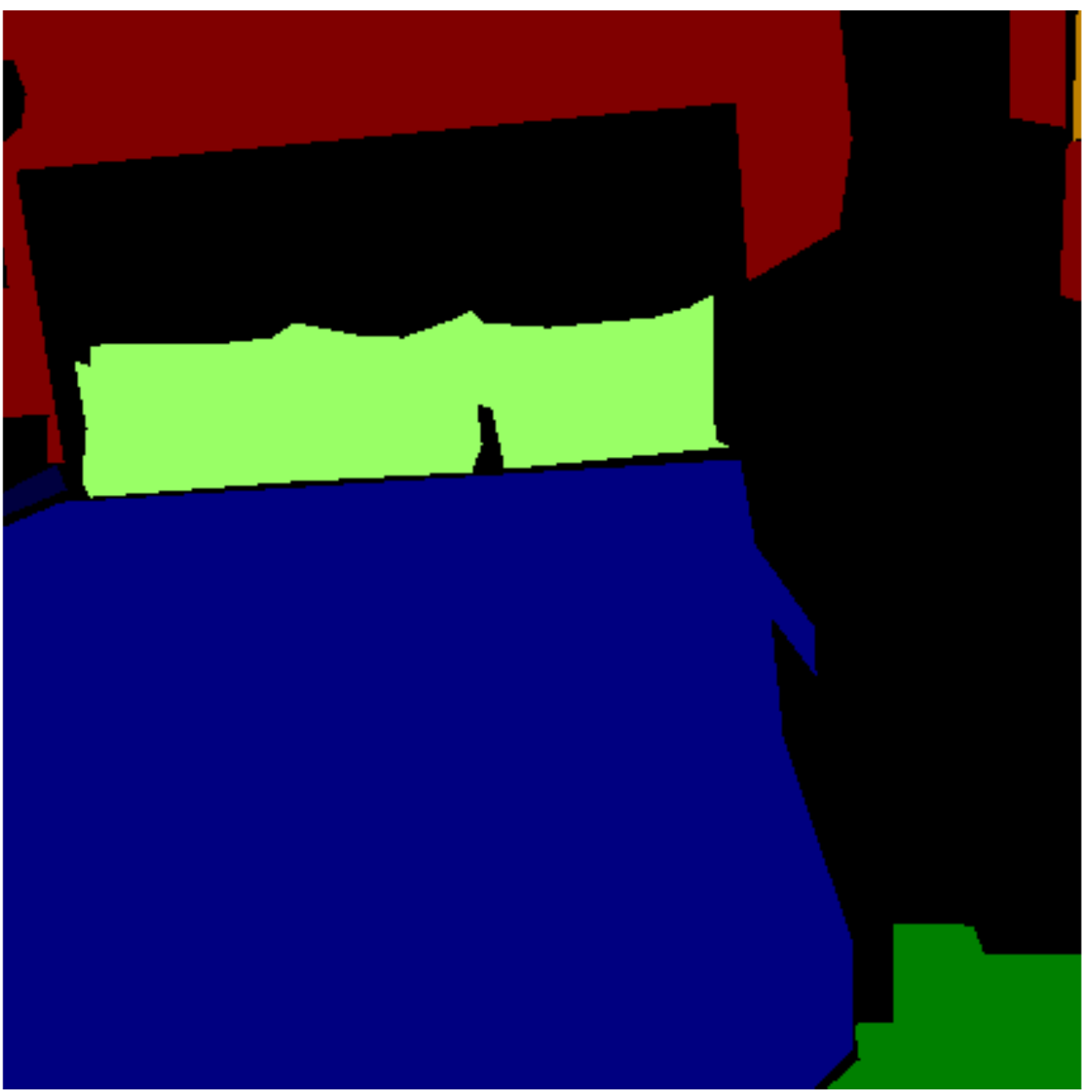} \\
\includegraphics[width=\textwidth]{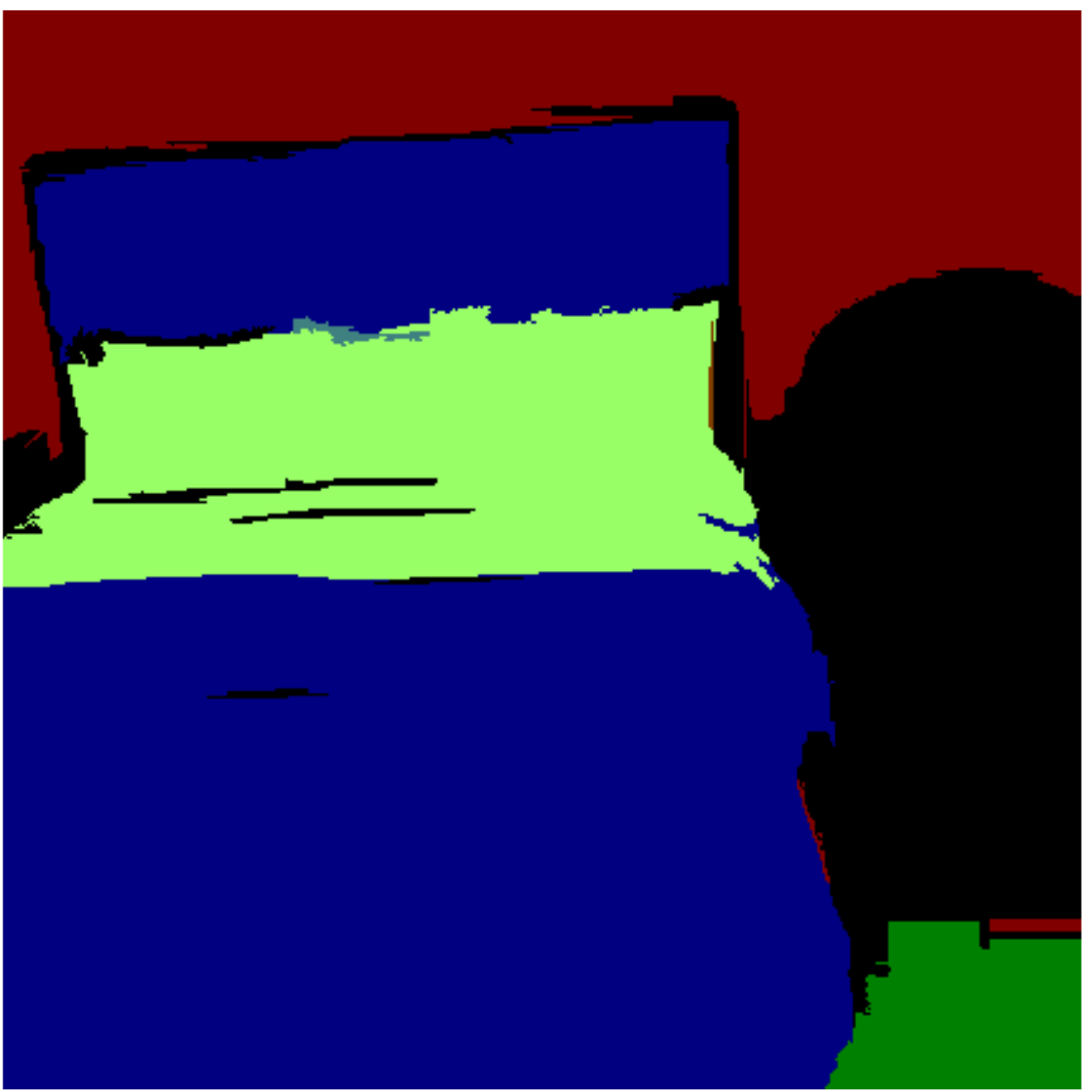}
\end{minipage}
}
\subfigure[]{
\begin{minipage}[]{0.095\textwidth}
\includegraphics[width=1\textwidth]{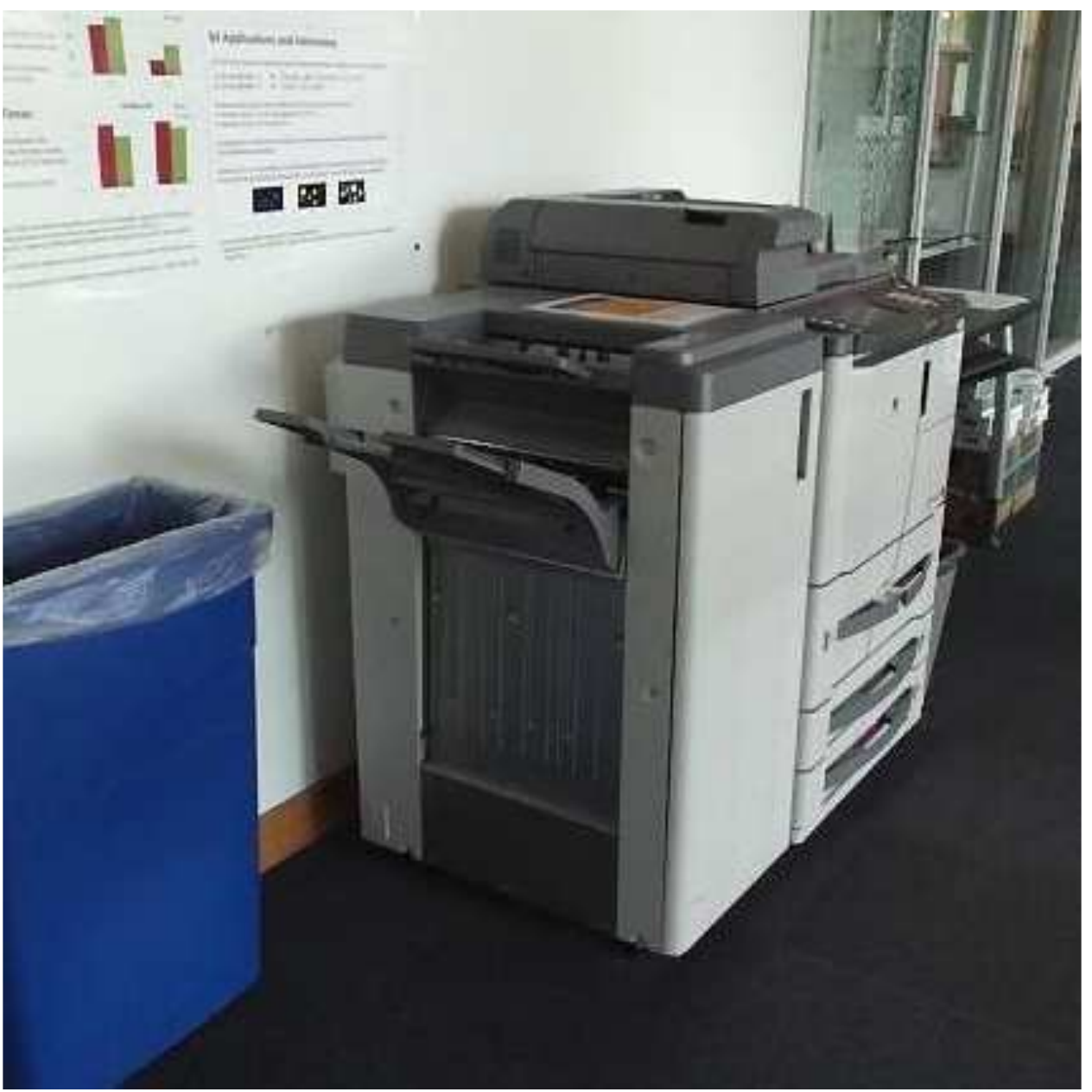} \\
\includegraphics[width=1\textwidth]{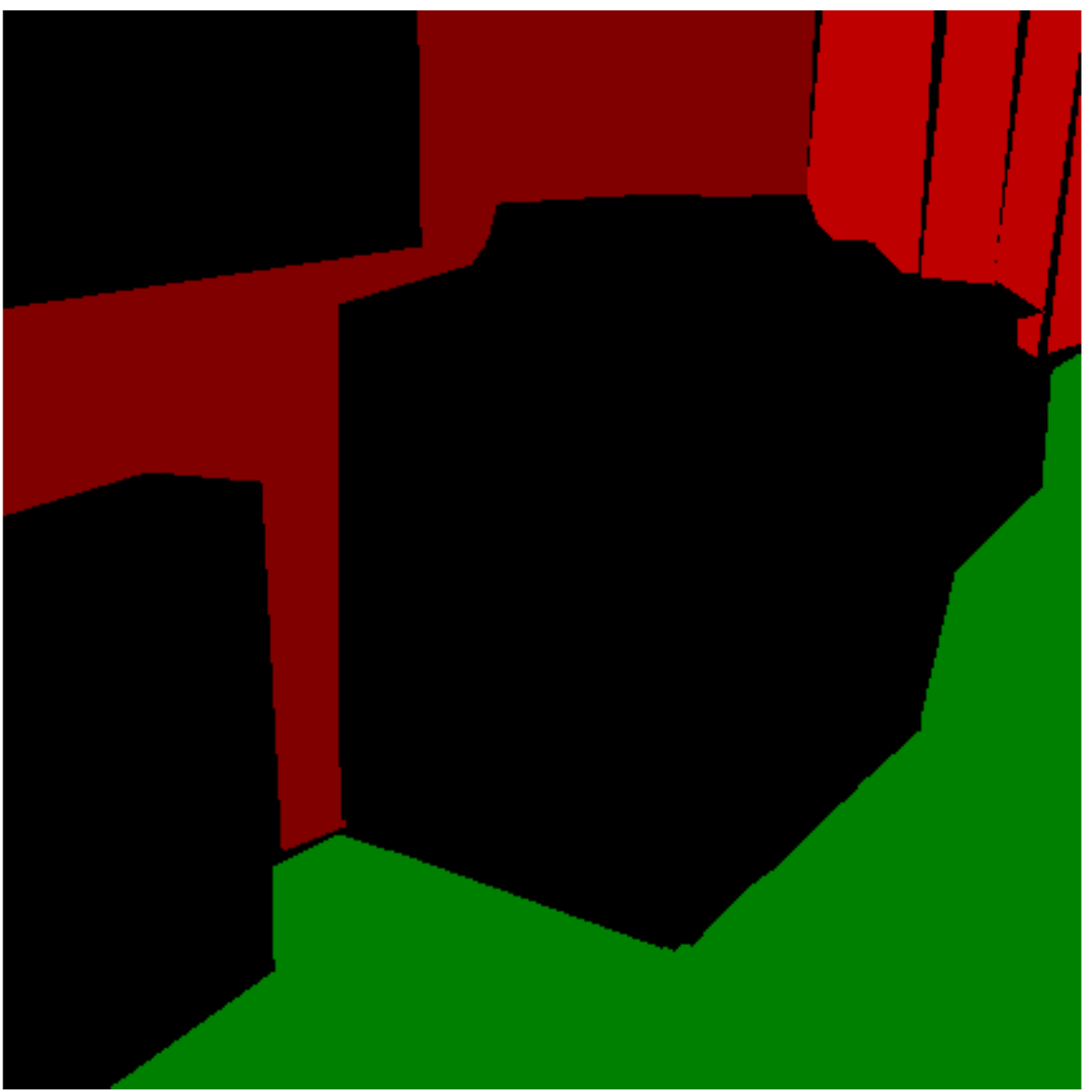} \\
\includegraphics[width=\textwidth]{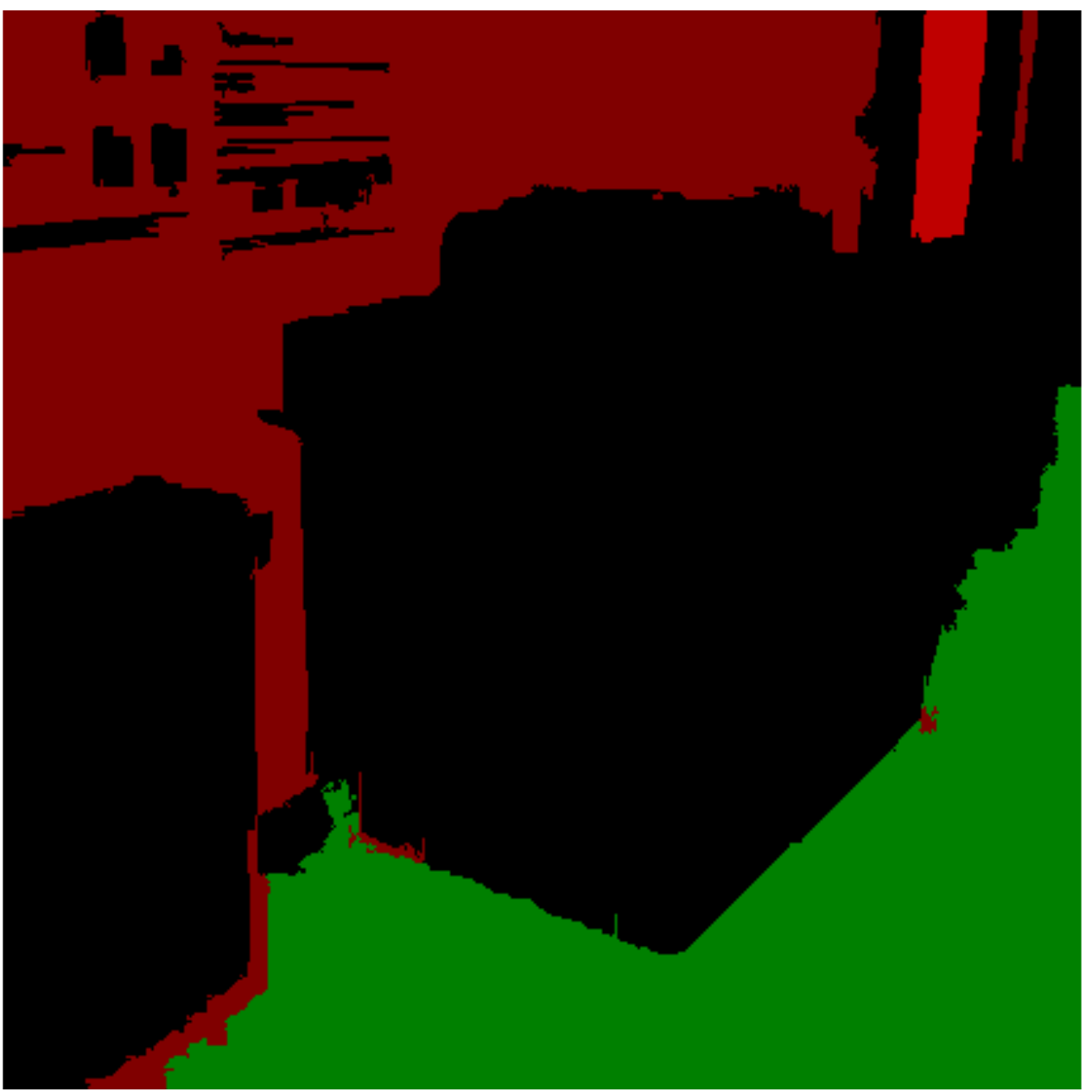}
\end{minipage}
}
\subfigure[]{
\begin{minipage}[]{0.095\textwidth}
\includegraphics[width=1\textwidth]{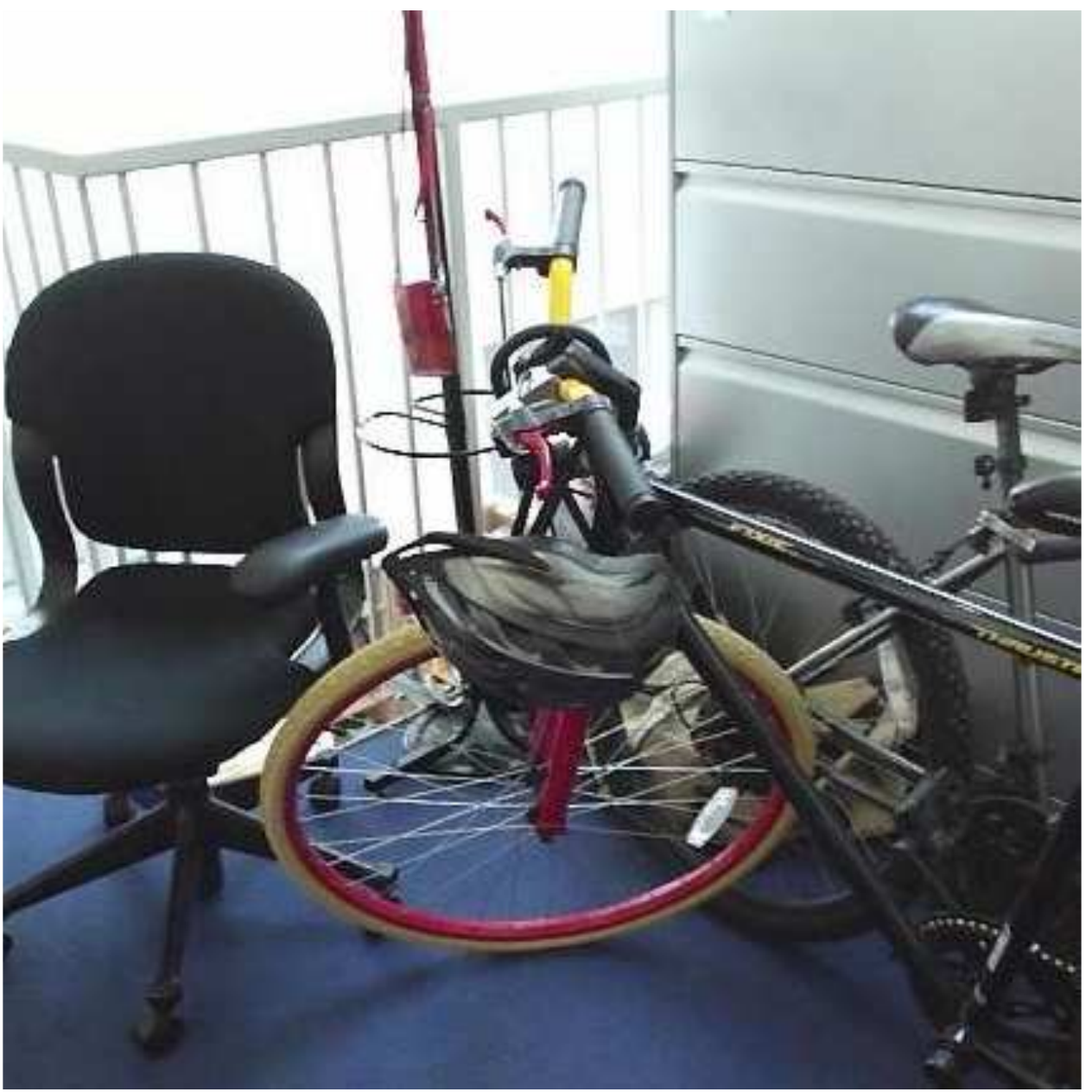} \\
\includegraphics[width=1\textwidth]{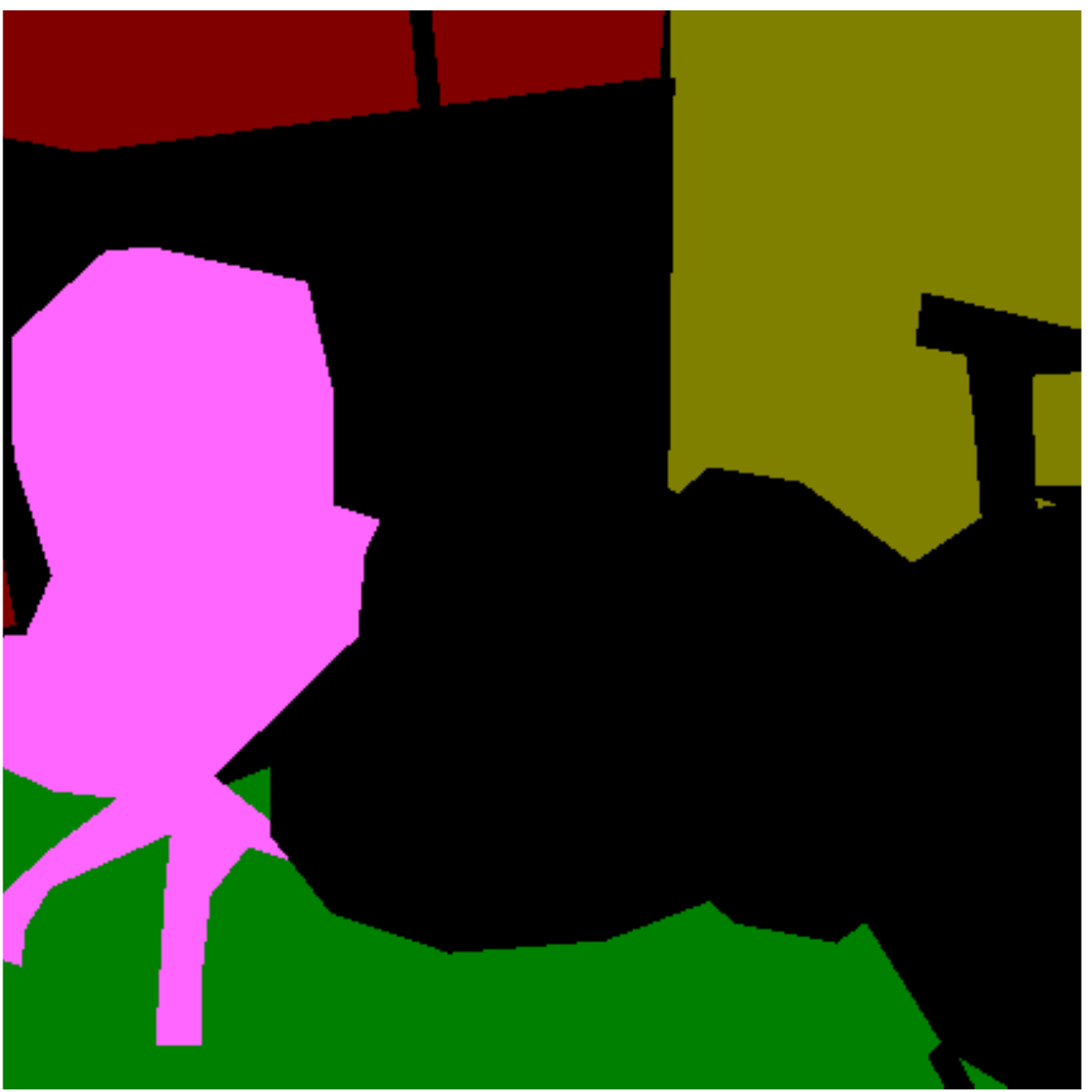} \\
\includegraphics[width=\textwidth]{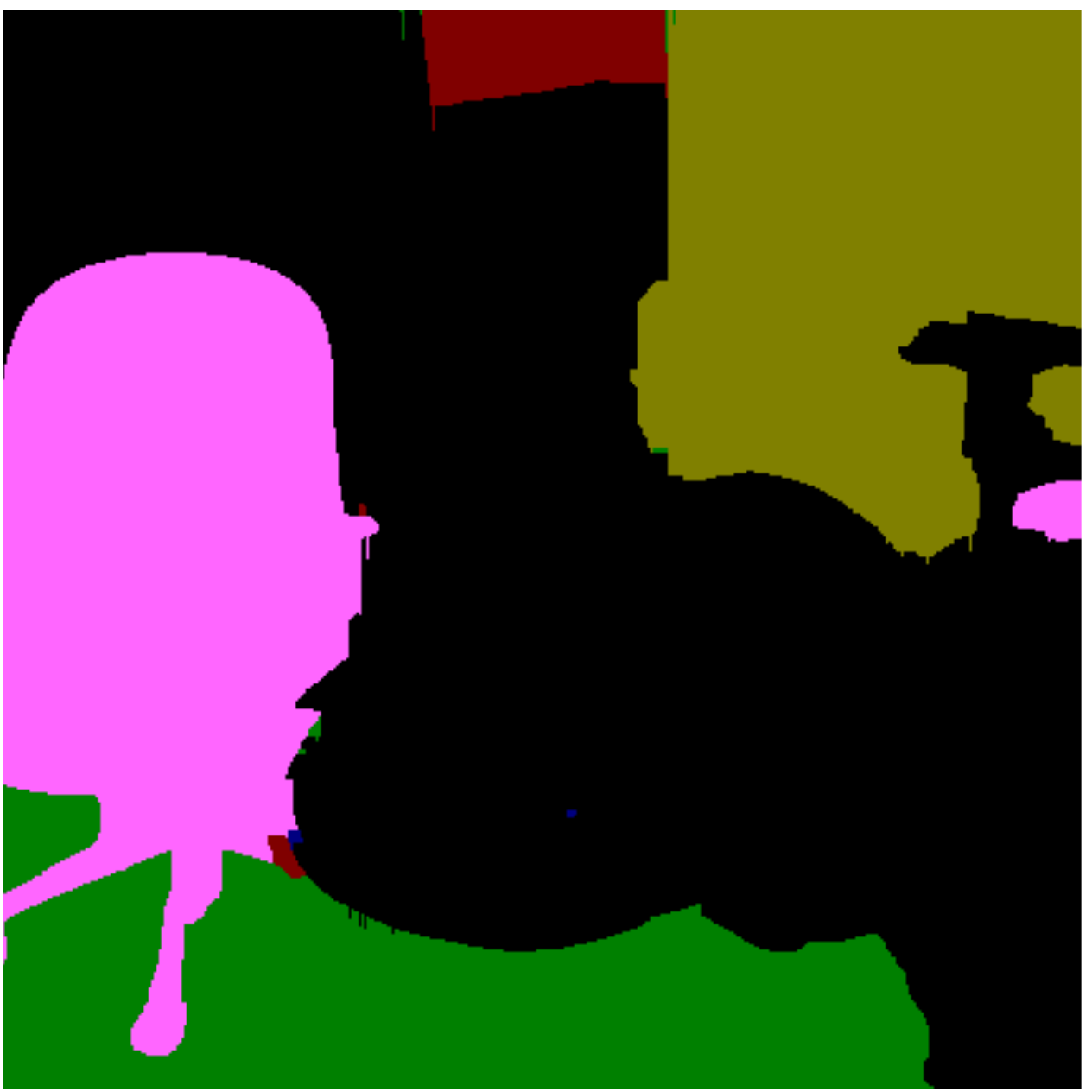}
\end{minipage}
}
\subfigure[]{
\begin{minipage}[]{0.095\textwidth}
\includegraphics[width=1\textwidth]{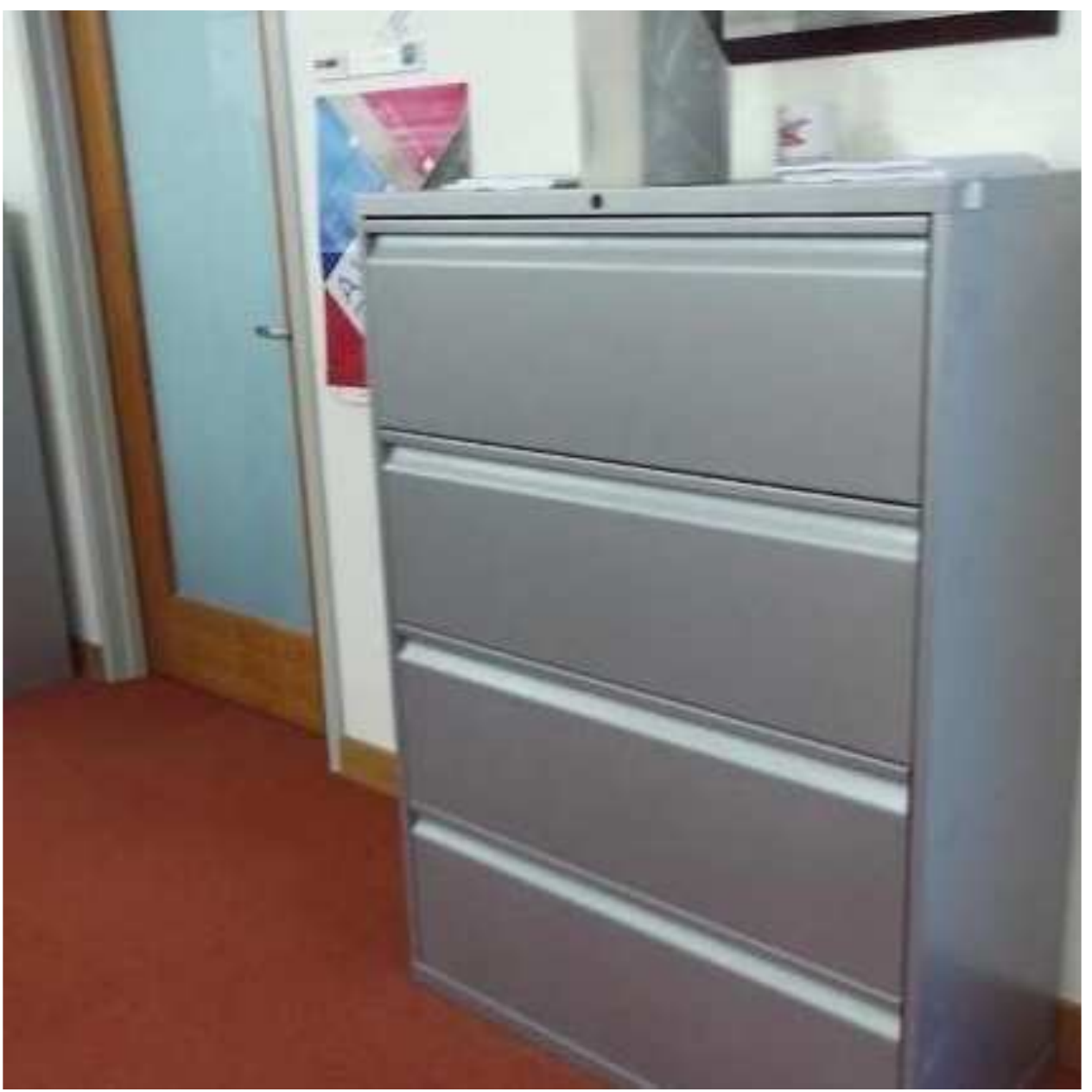} \\
\includegraphics[width=1\textwidth]{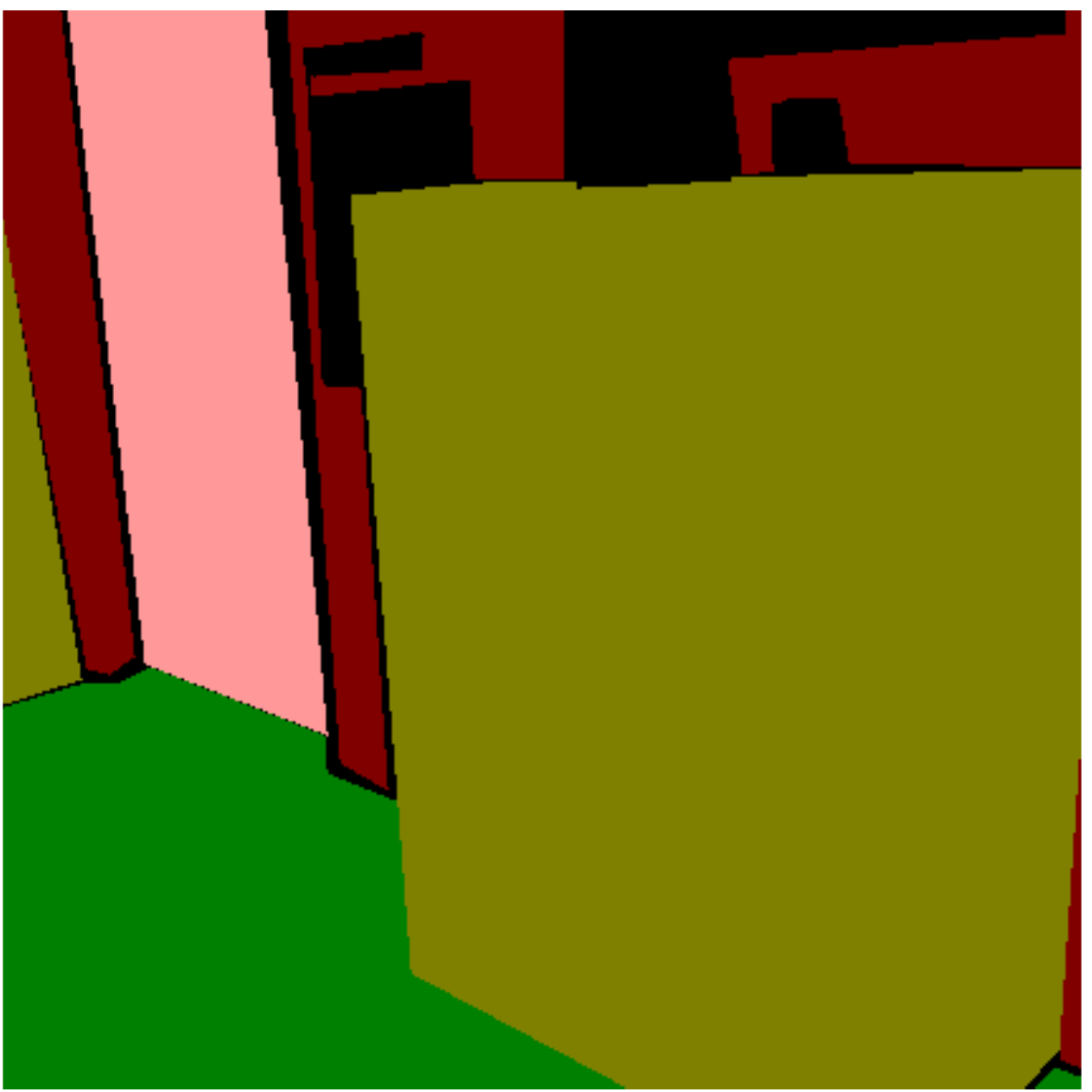} \\
\includegraphics[width=\textwidth]{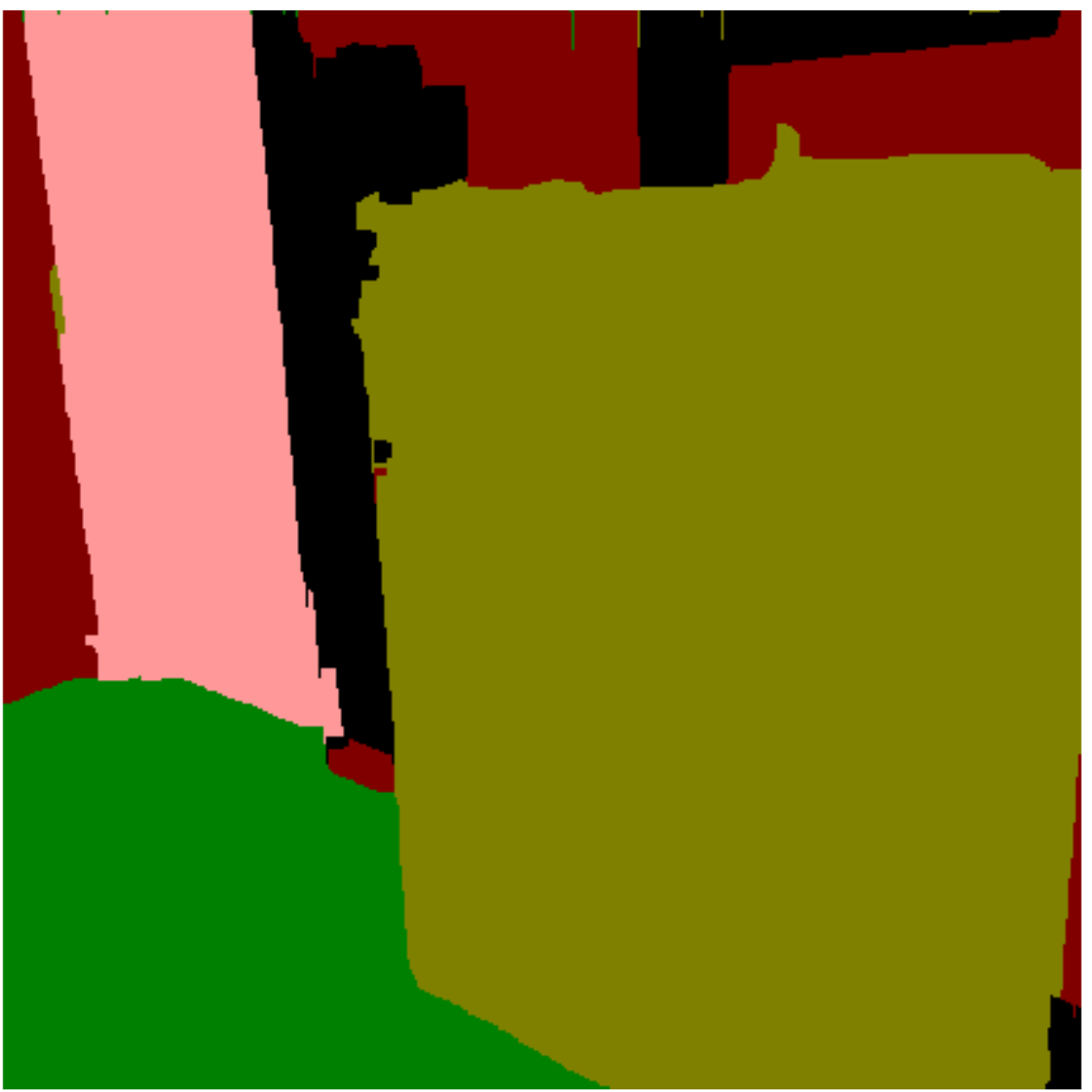}
\end{minipage}
}
\subfigure[]{
\begin{minipage}[]{0.095\textwidth}
\includegraphics[width=1\textwidth]{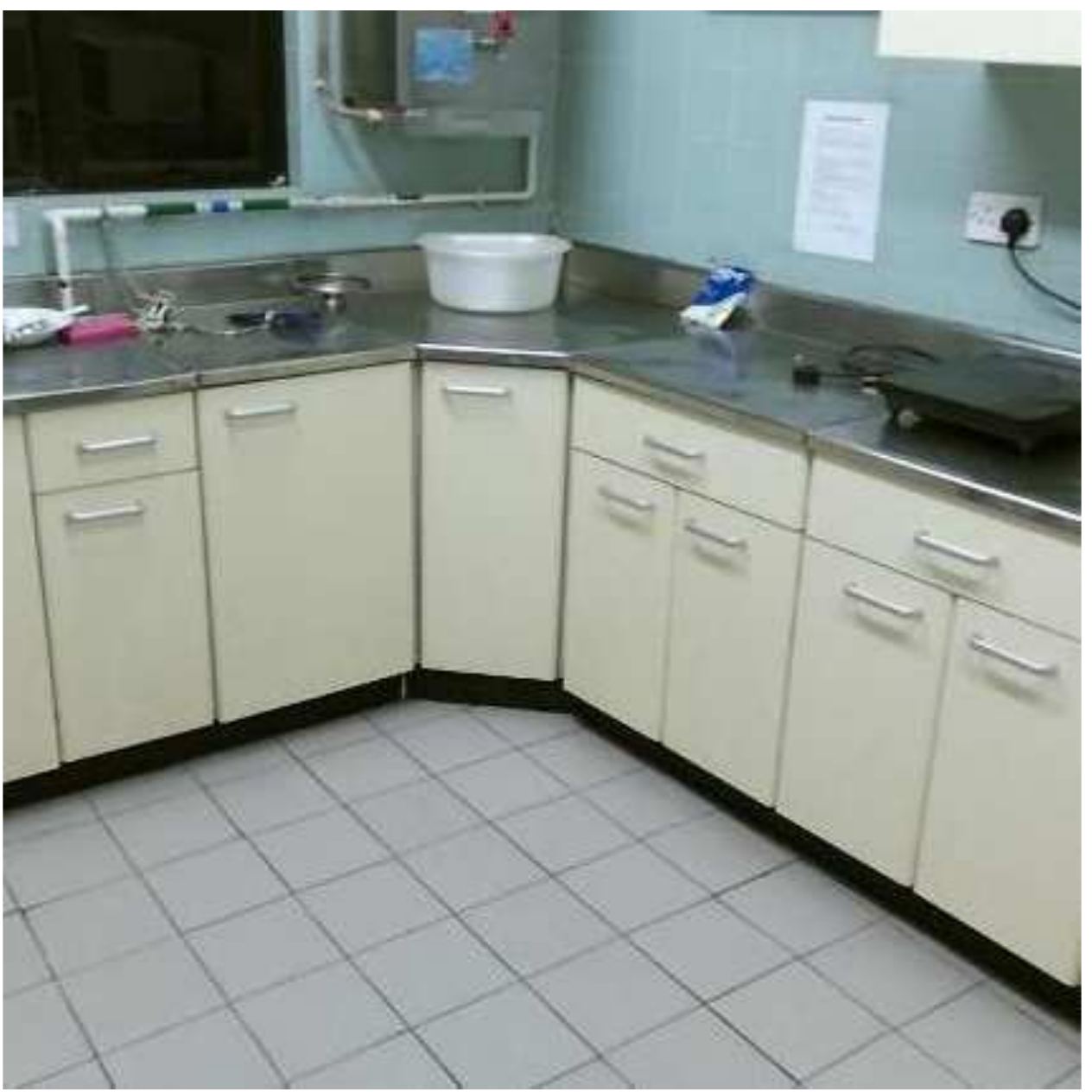} \\
\includegraphics[width=1\textwidth]{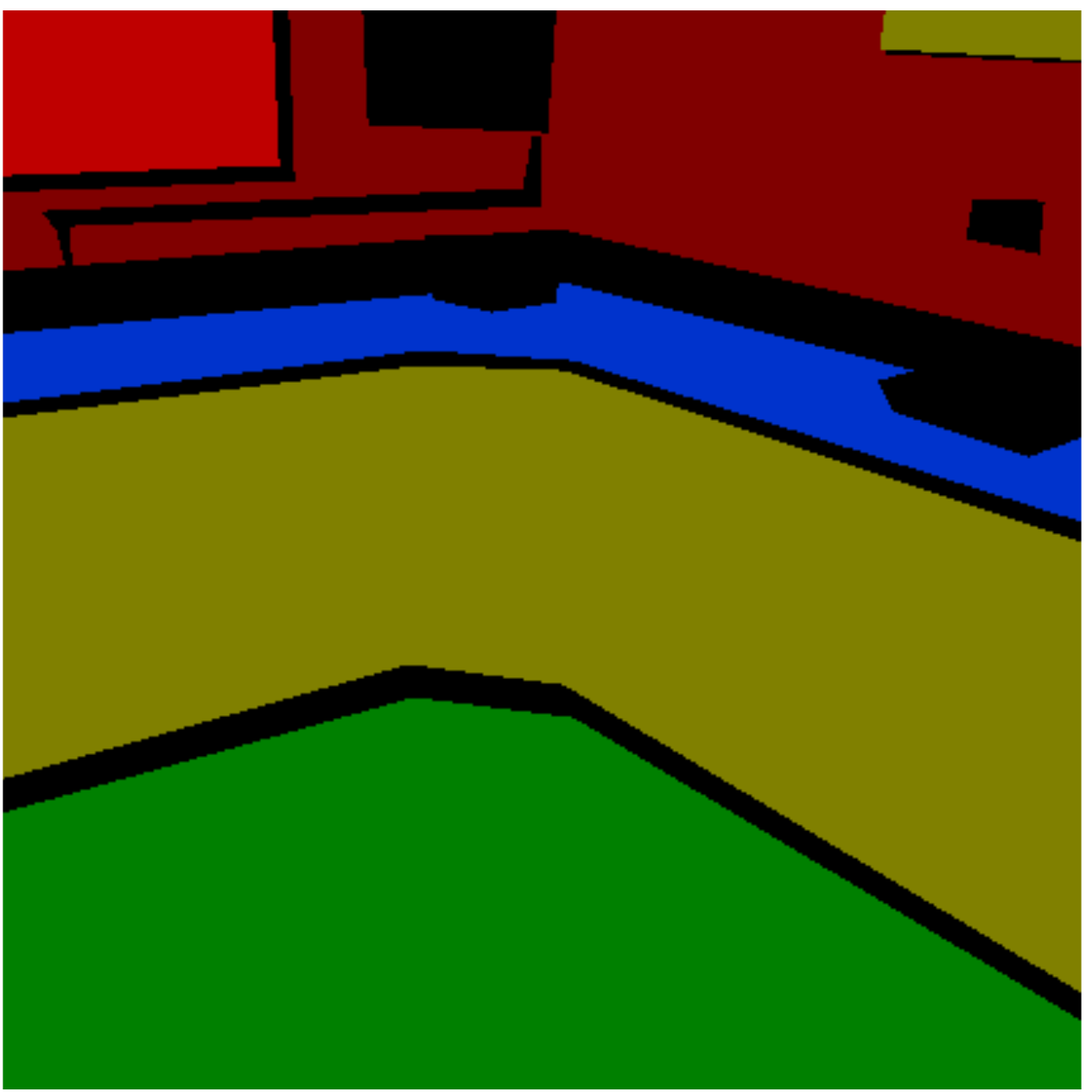} \\
\includegraphics[width=\textwidth]{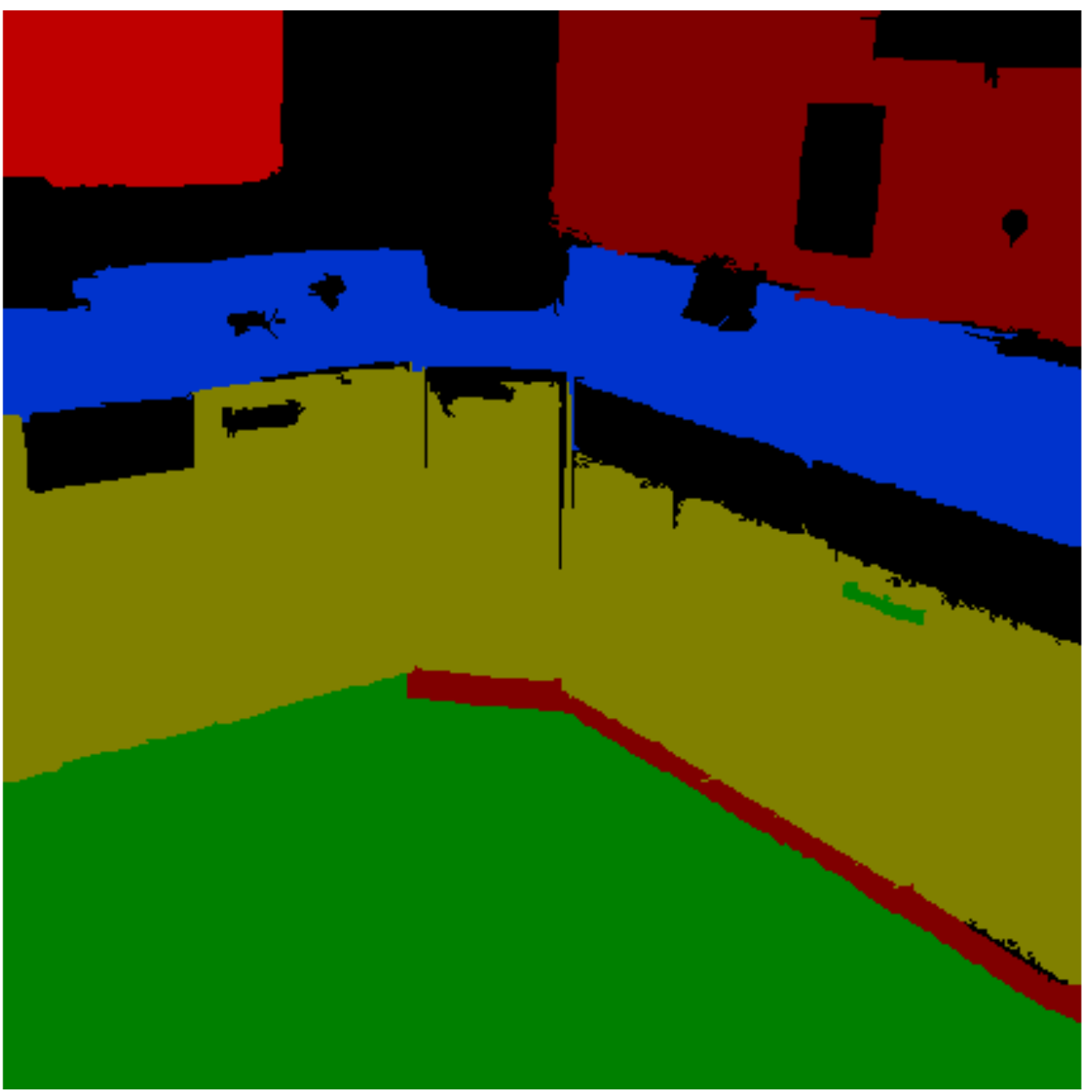}
\end{minipage}
}
\subfigure[]{
\begin{minipage}[]{0.095\textwidth}
\includegraphics[width=1\textwidth]{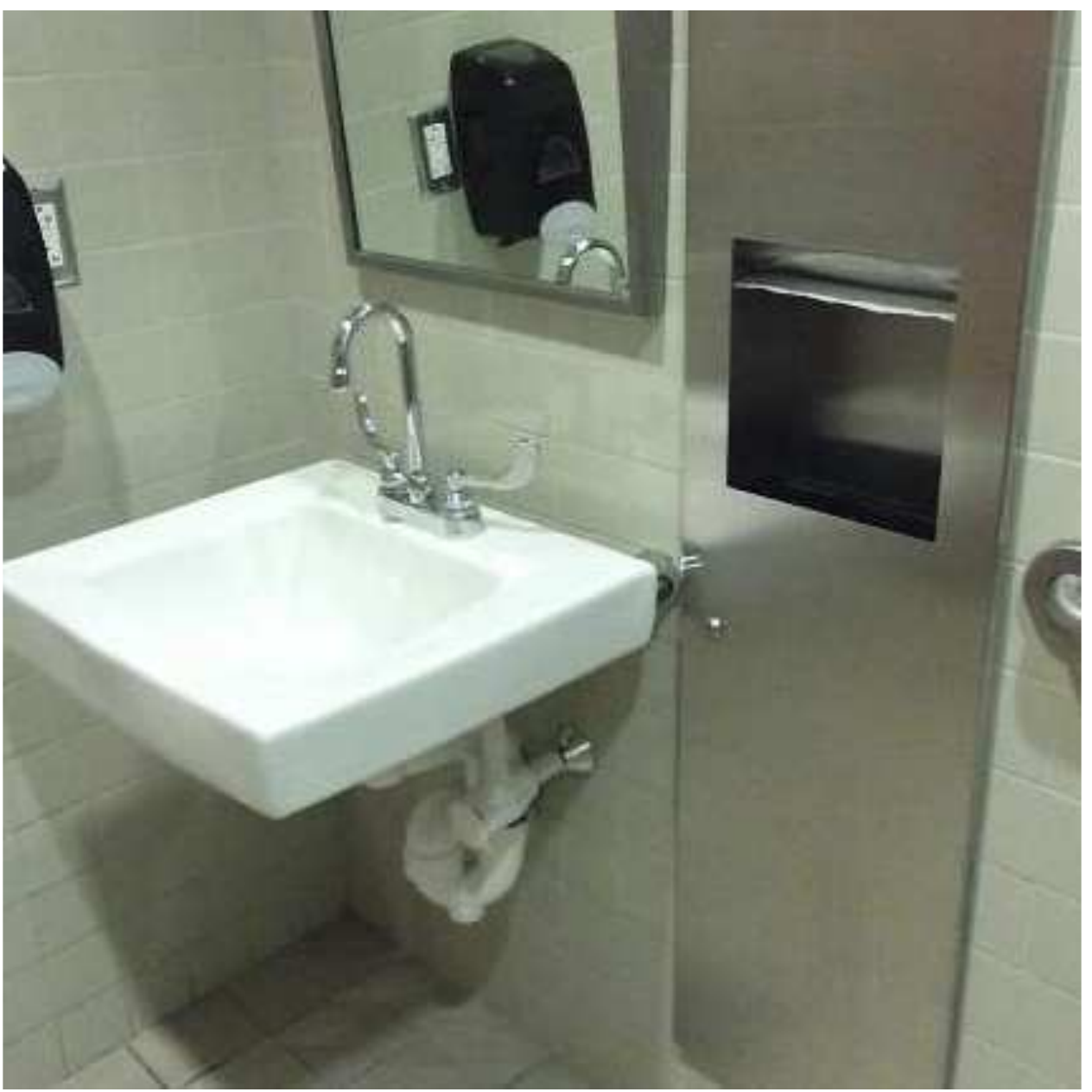} \\
\includegraphics[width=1\textwidth]{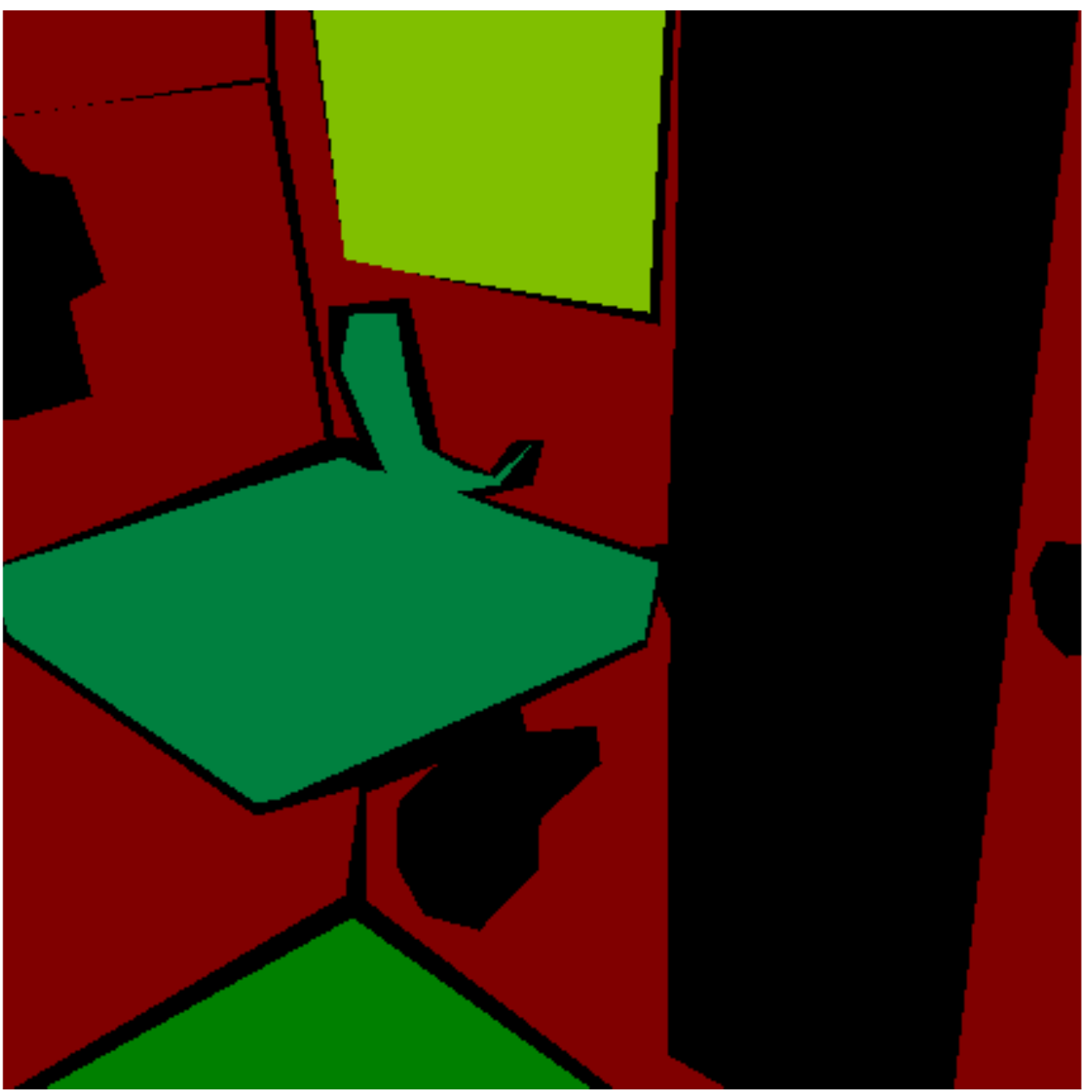} \\
\includegraphics[width=\textwidth]{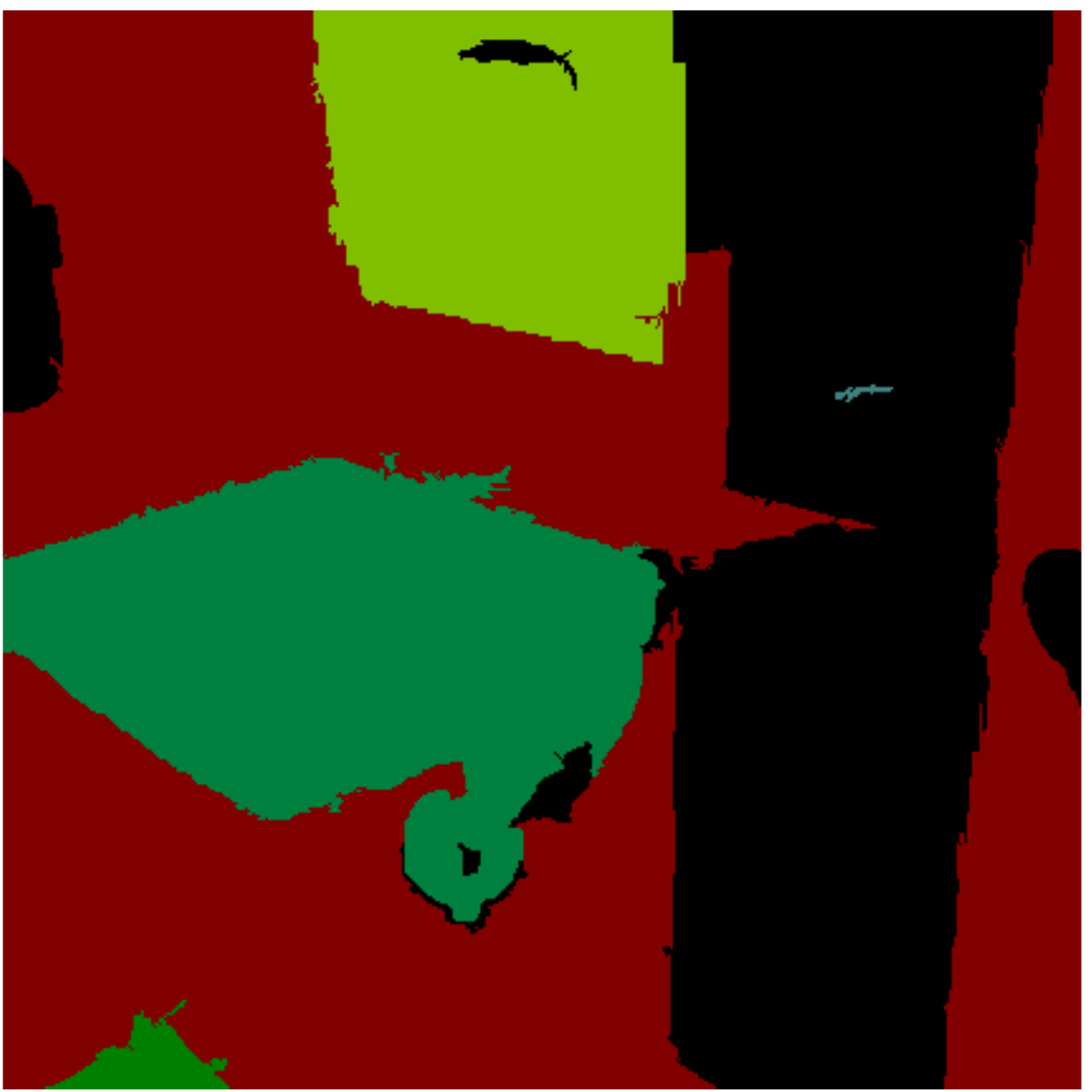}
\end{minipage}
}
\subfigure[]{
\begin{minipage}[]{0.095\textwidth}
\includegraphics[width=1\textwidth]{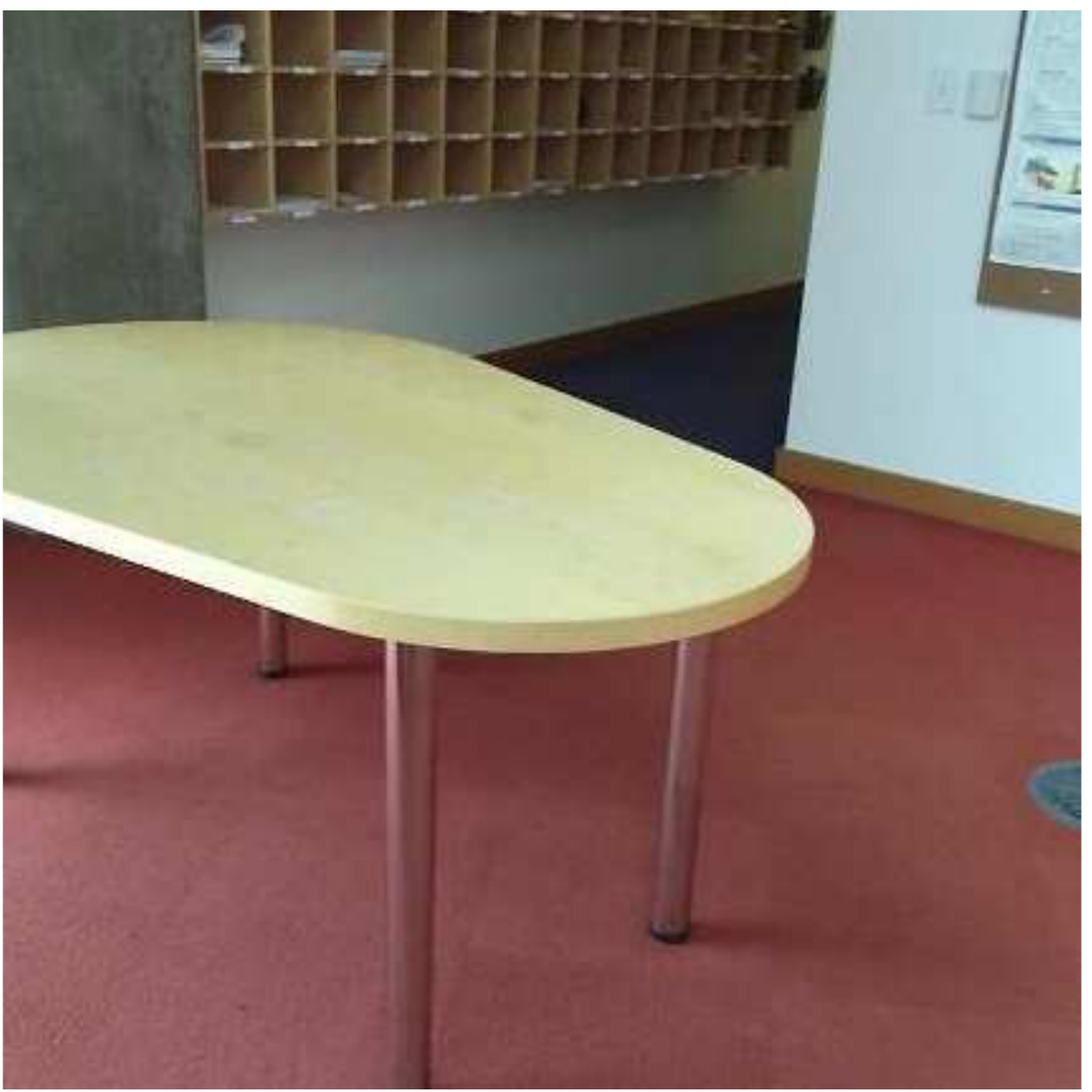} \\
\includegraphics[width=1\textwidth]{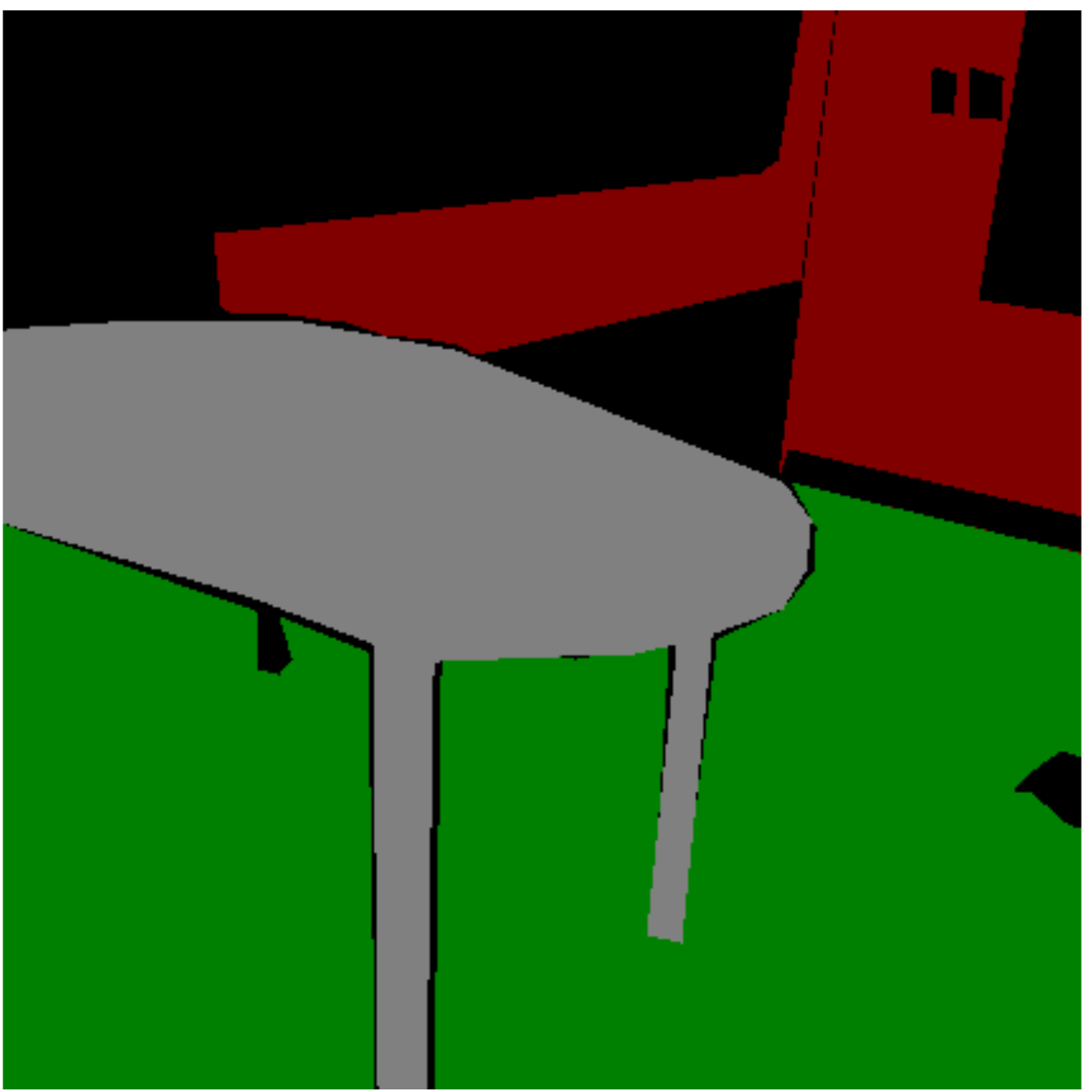} \\
\includegraphics[width=\textwidth]{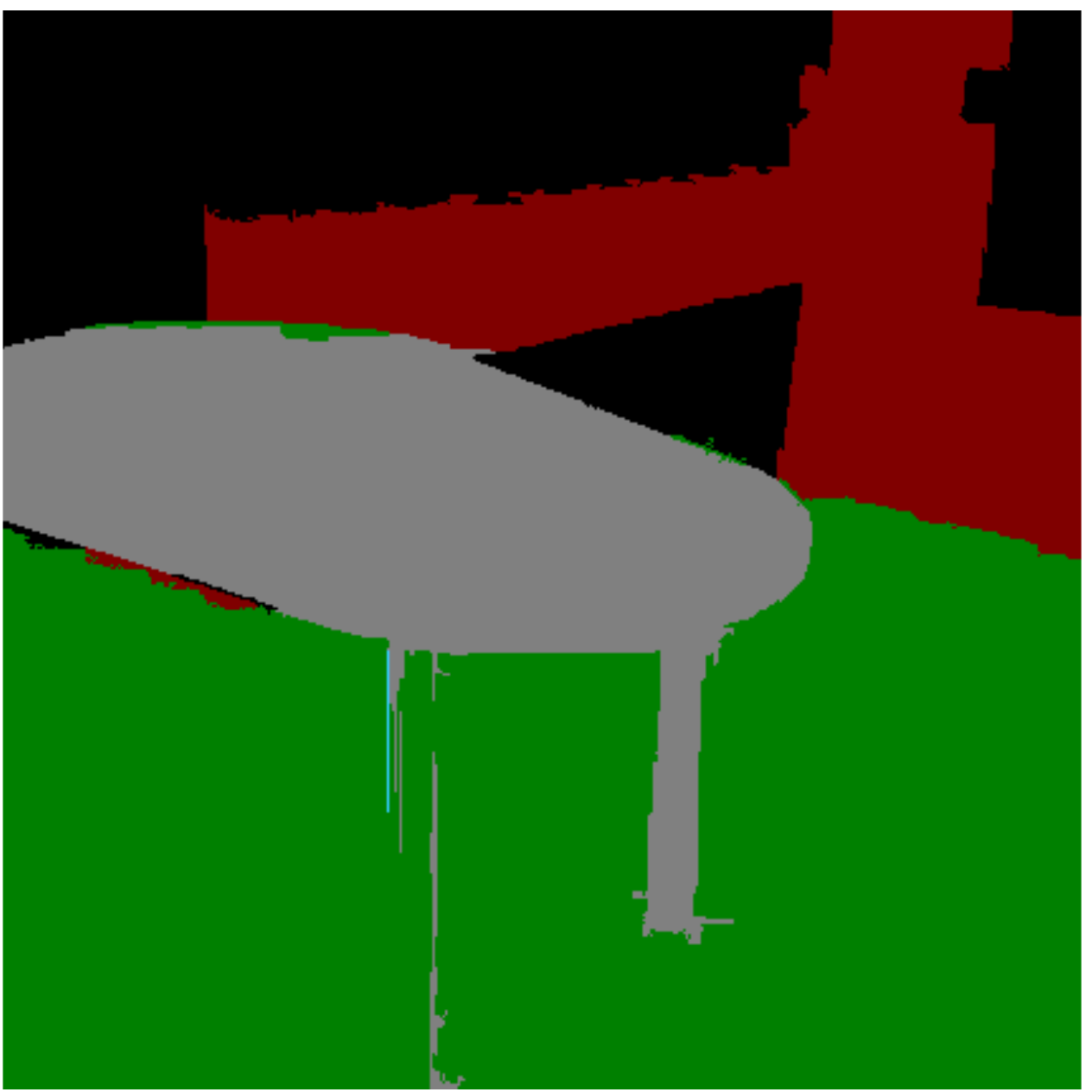}
\end{minipage}
}
\subfigure[]{
\begin{minipage}[]{0.095\textwidth}
\includegraphics[width=1\textwidth]{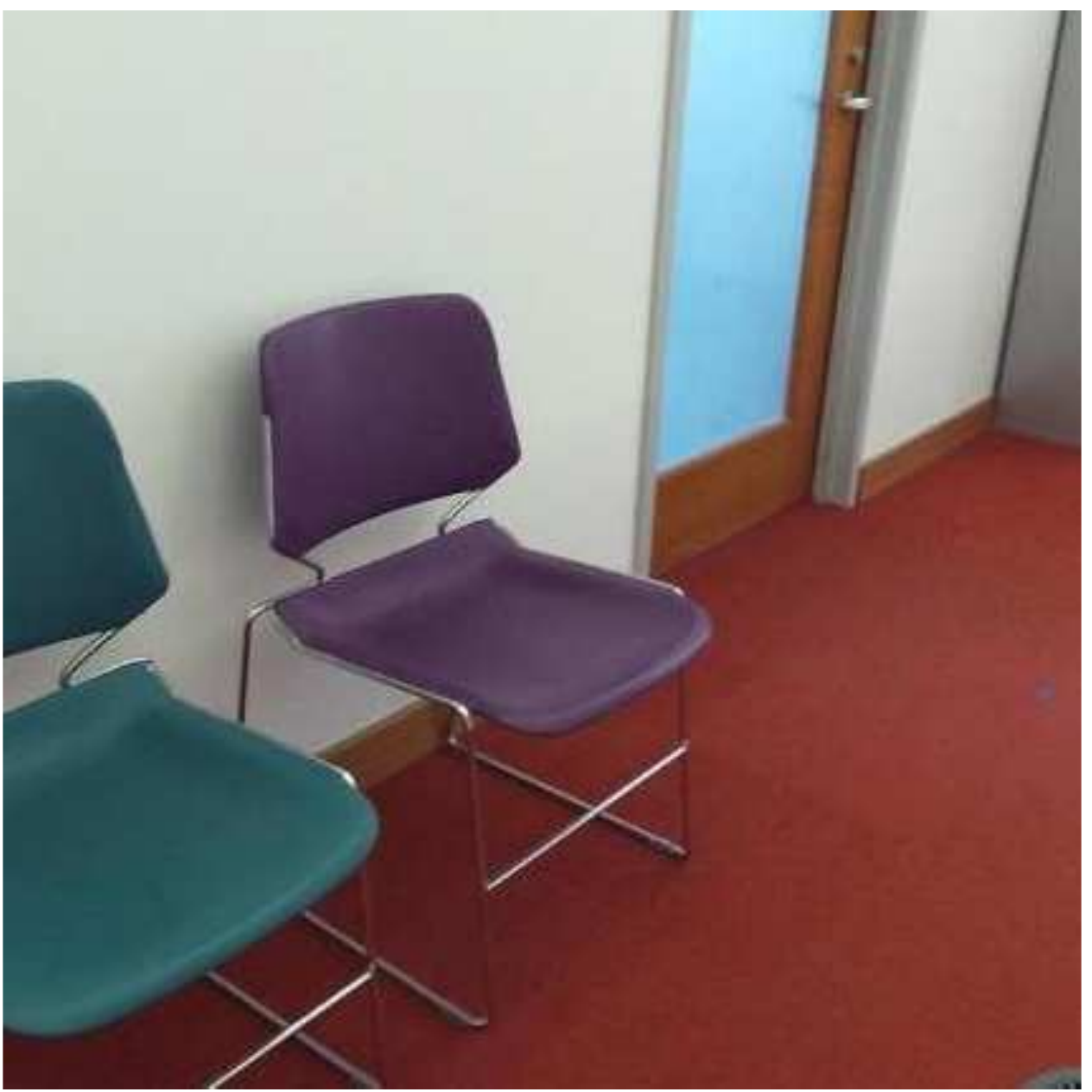} \\
\includegraphics[width=1\textwidth]{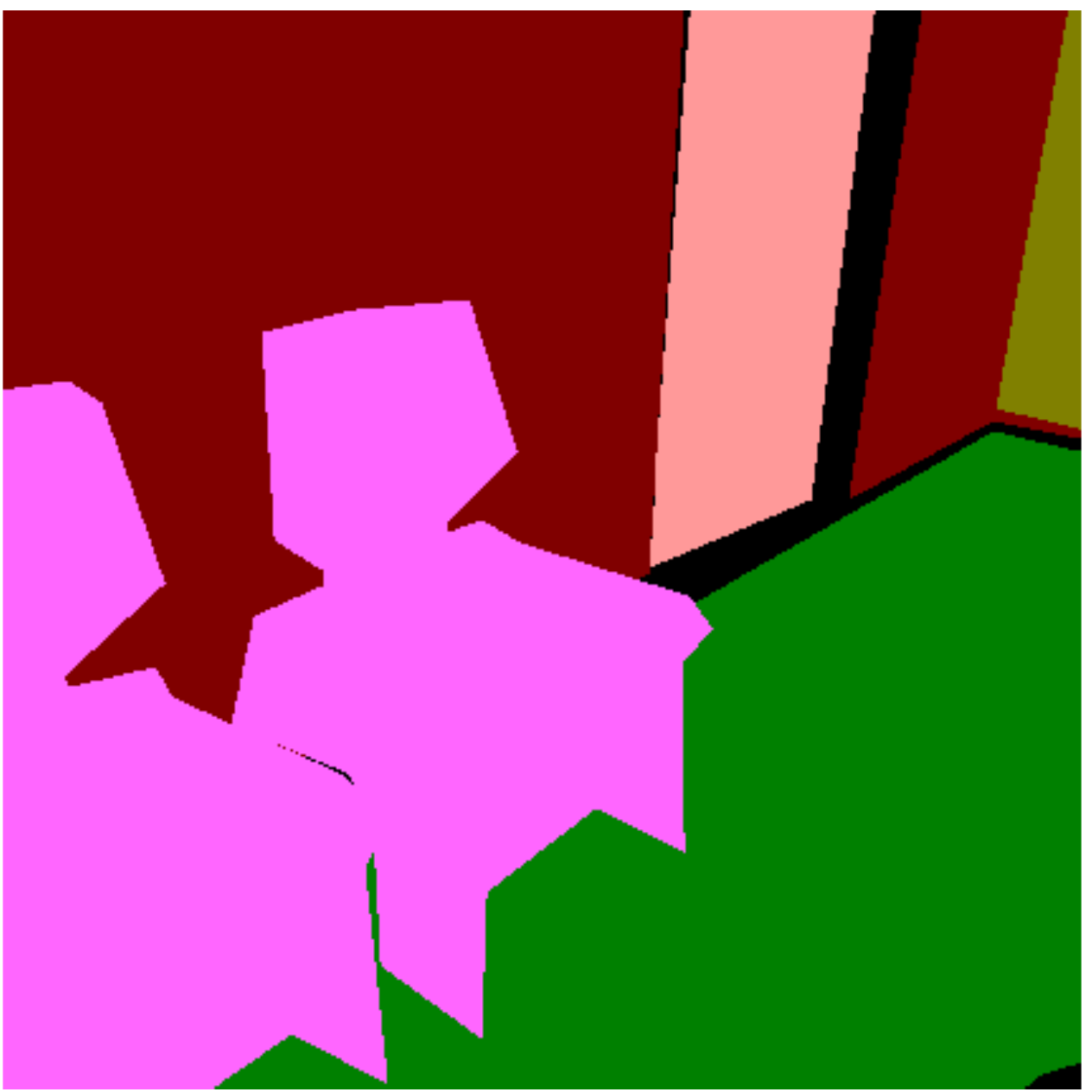} \\
\includegraphics[width=\textwidth]{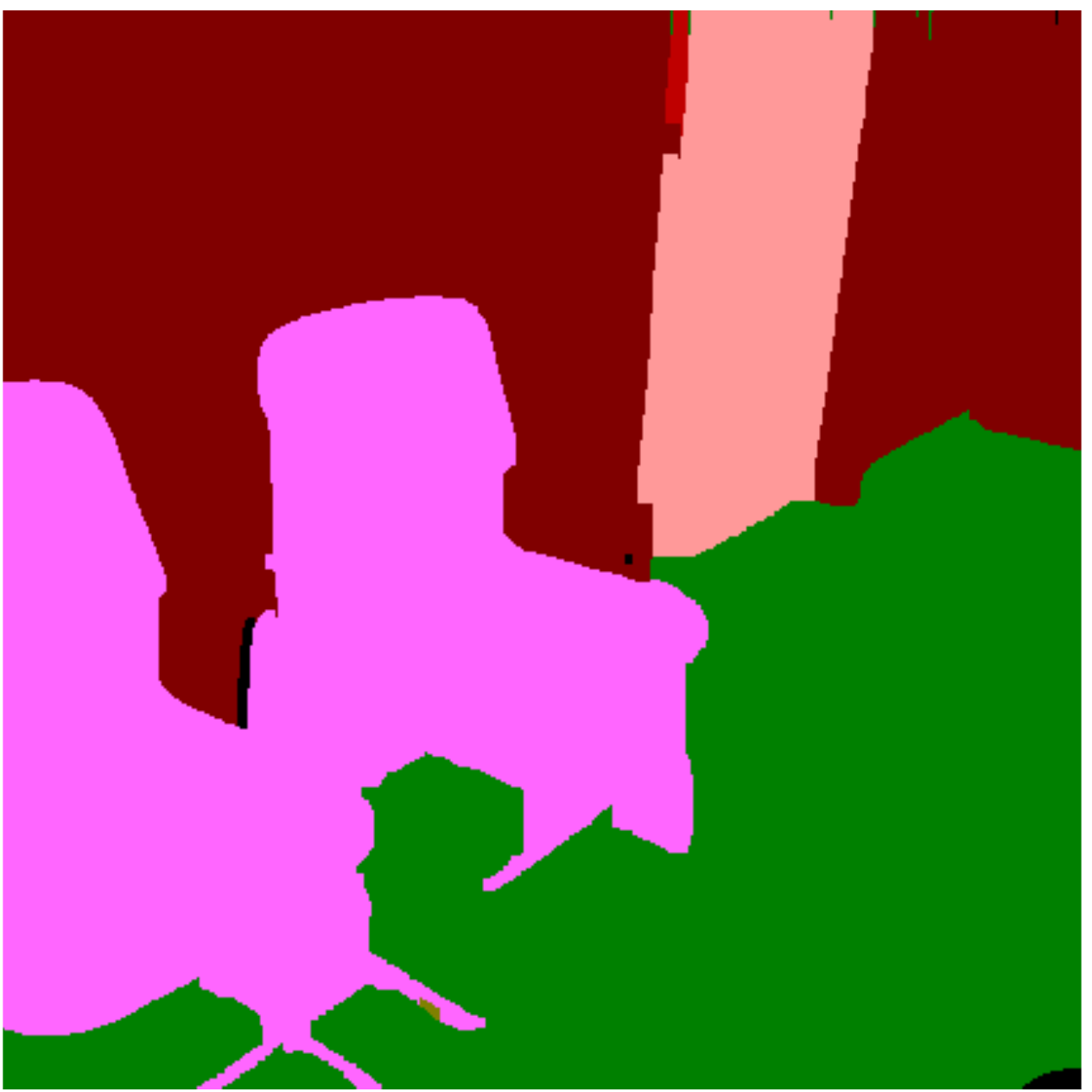}
\end{minipage}
}
\subfigure[]{
\begin{minipage}[]{0.095\textwidth}
\includegraphics[width=1\textwidth]{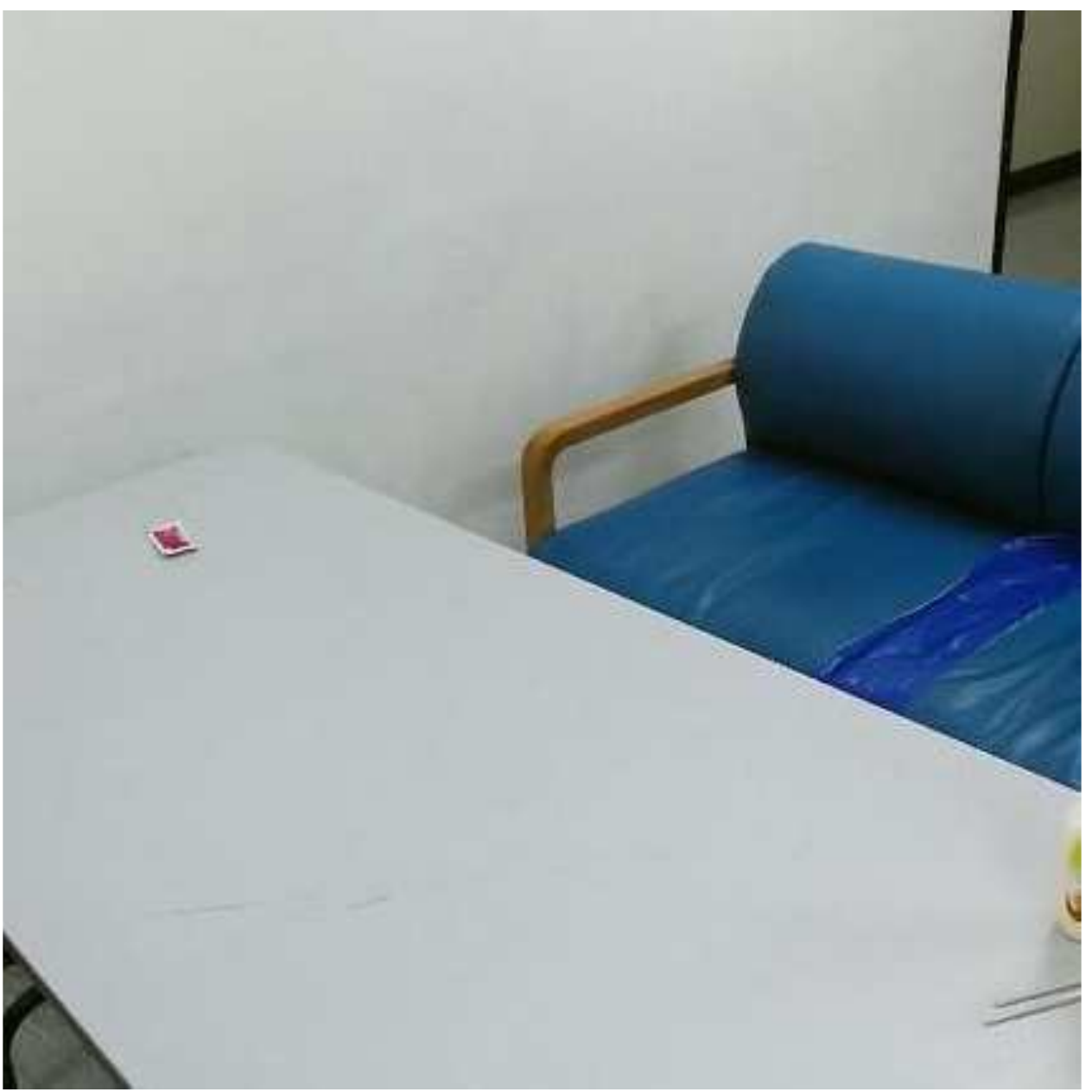} \\
\includegraphics[width=1\textwidth]{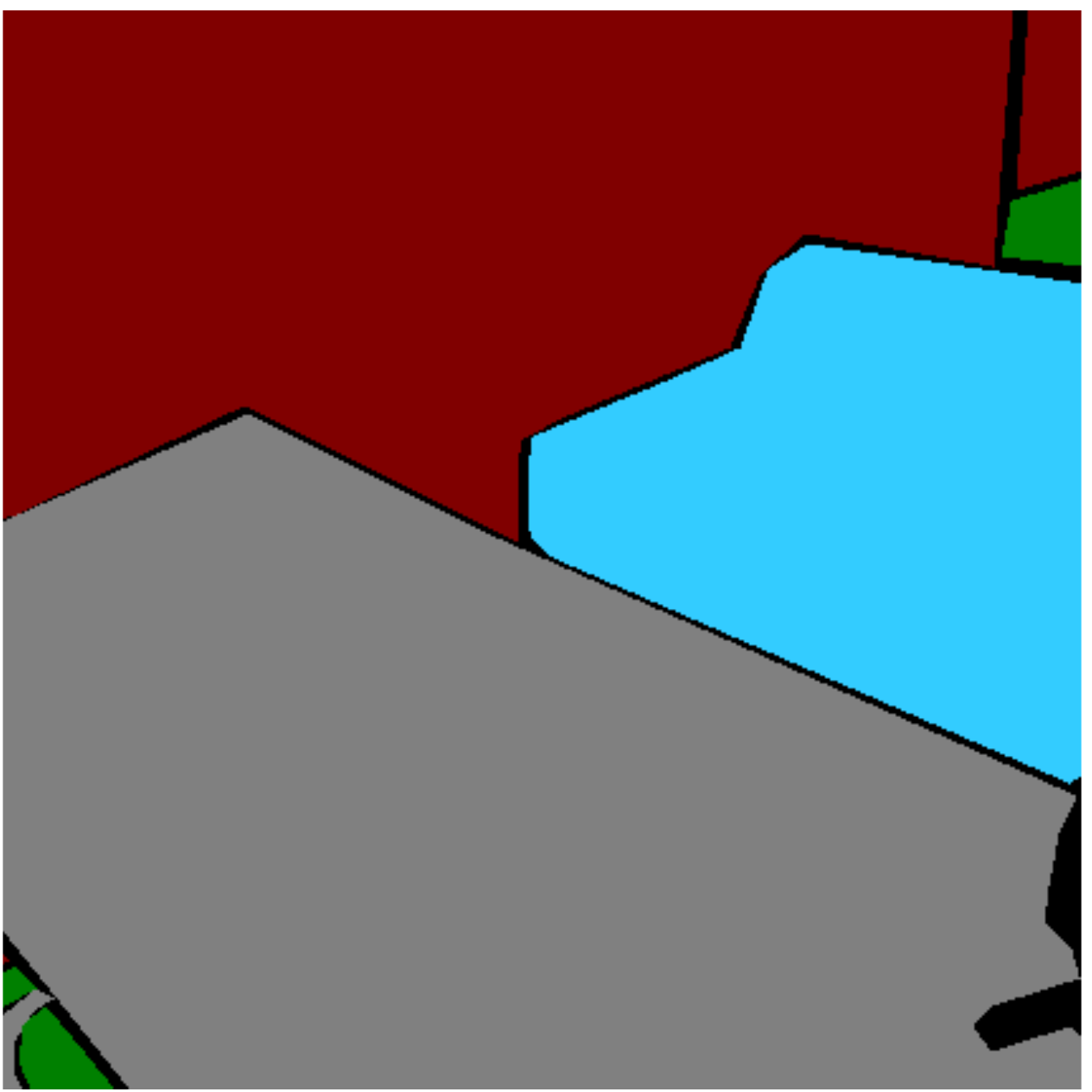} \\
\includegraphics[width=\textwidth]{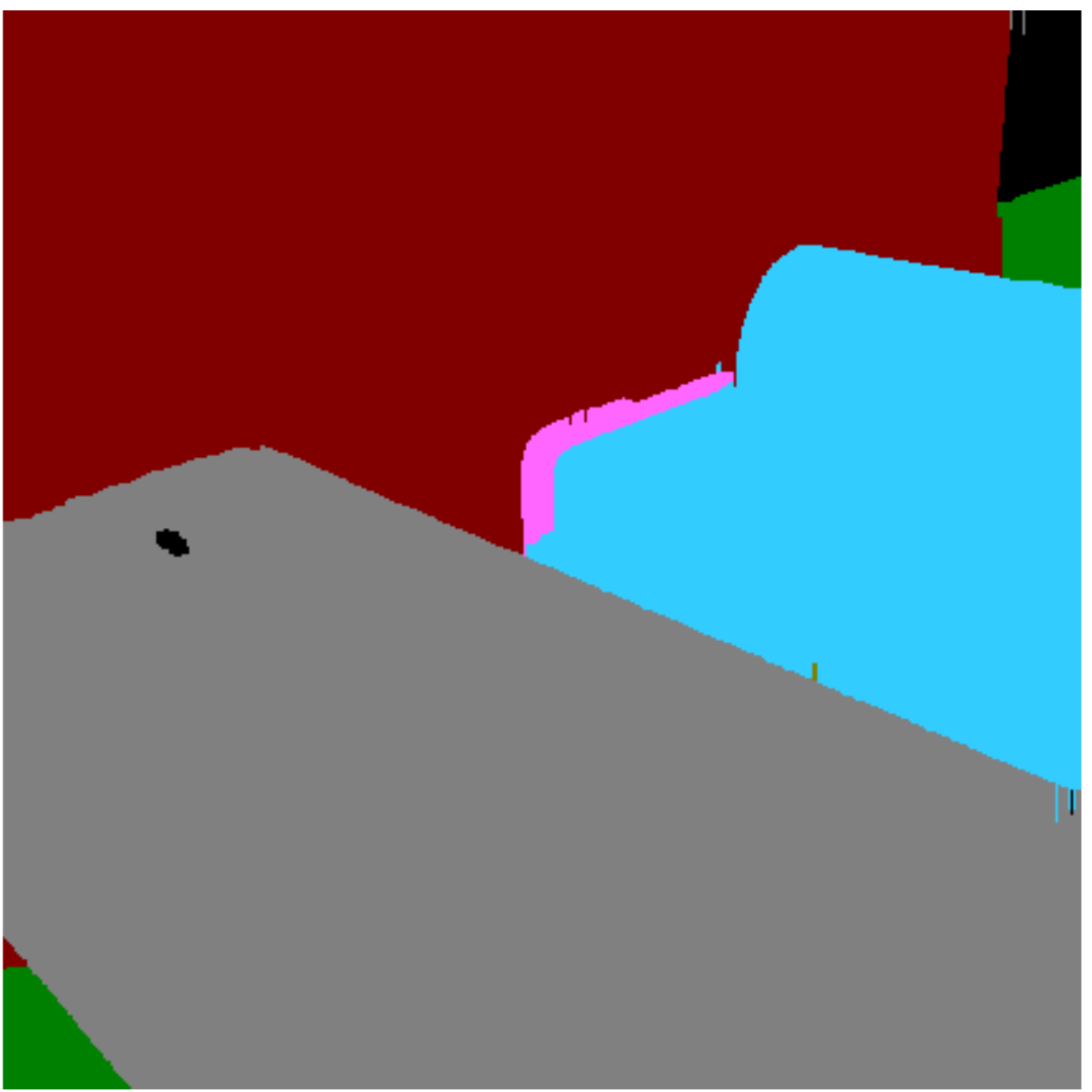}
\end{minipage}
}
\subfigure[]{
\begin{minipage}[]{0.095\textwidth}
\includegraphics[width=1\textwidth]{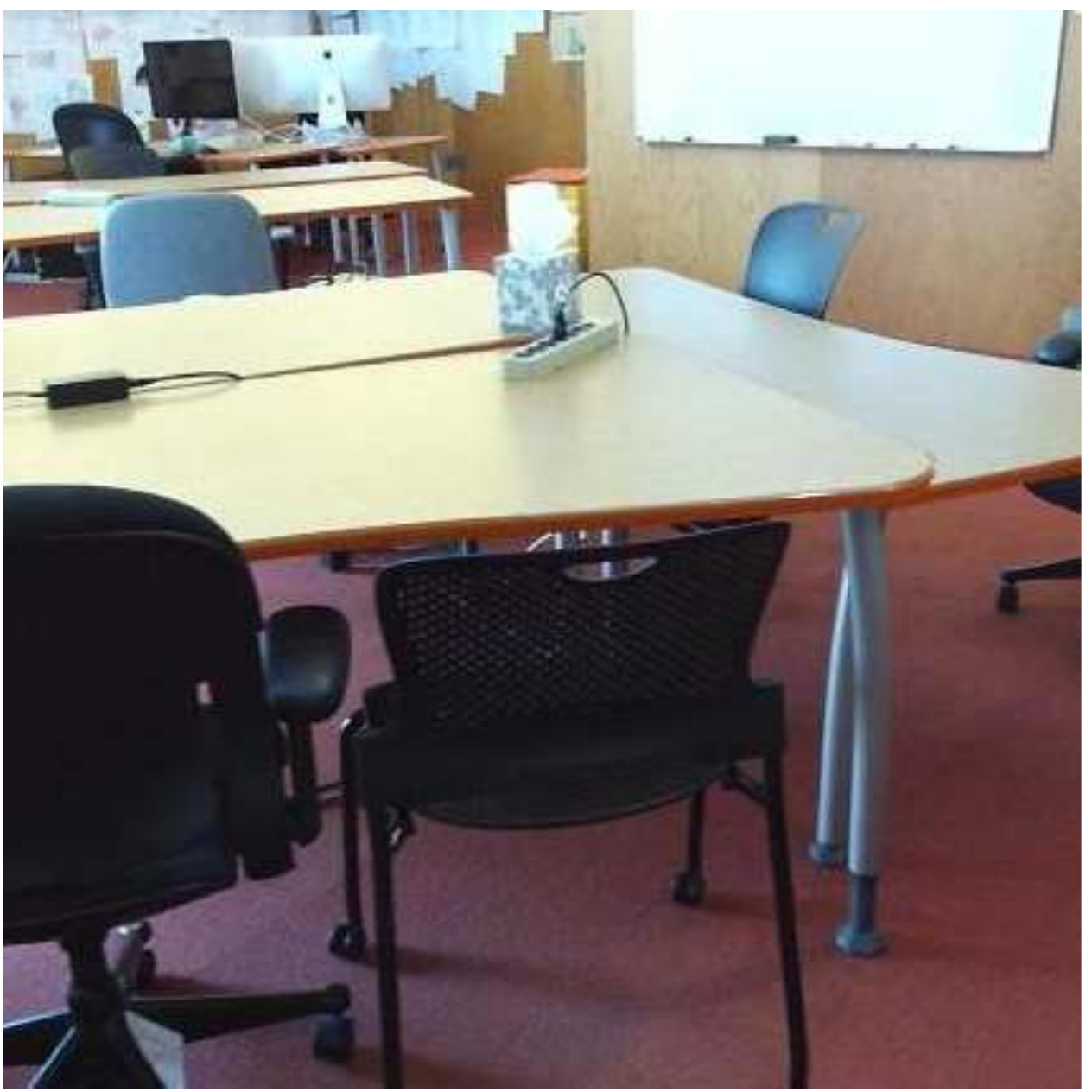} \\
\includegraphics[width=1\textwidth]{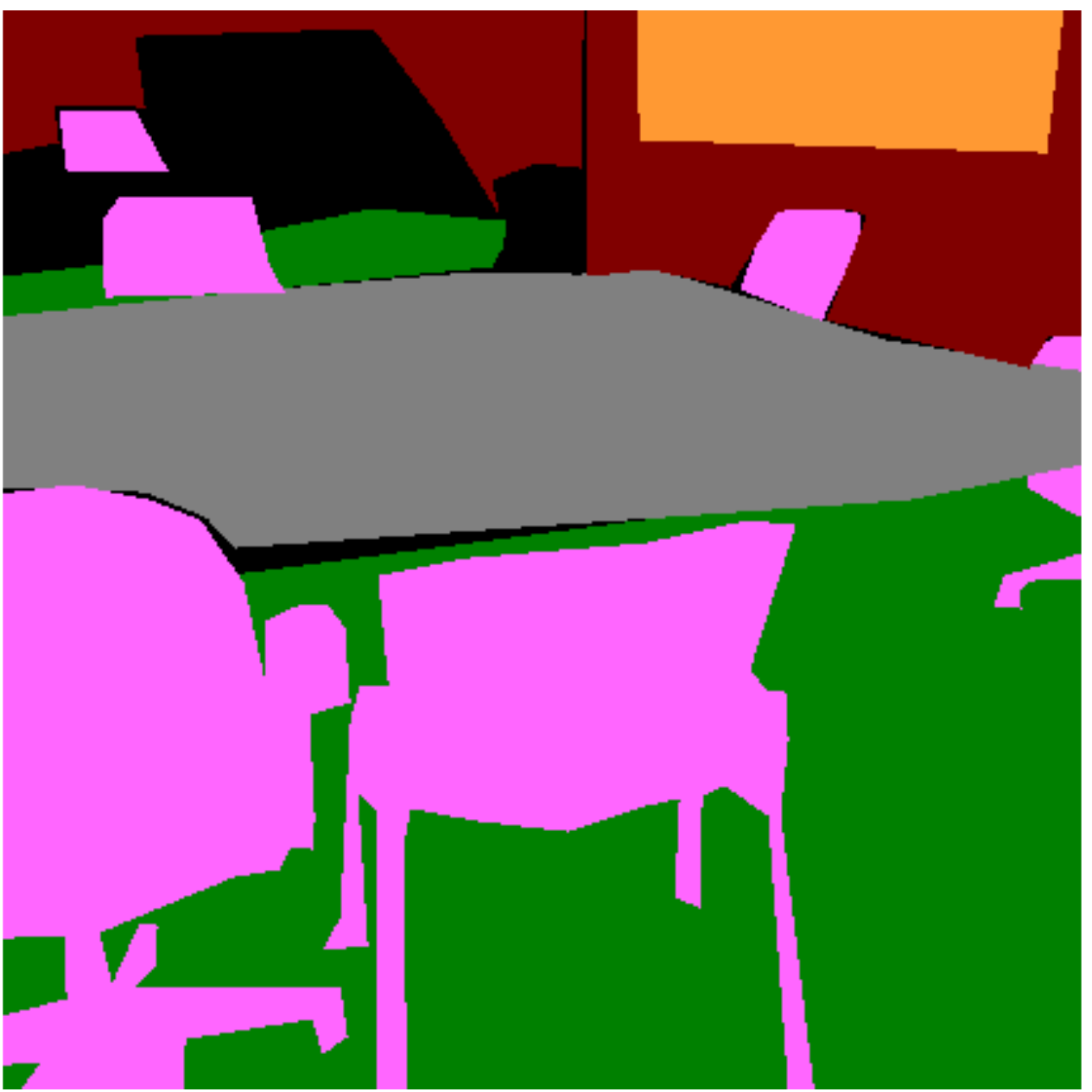} \\
\includegraphics[width=\textwidth]{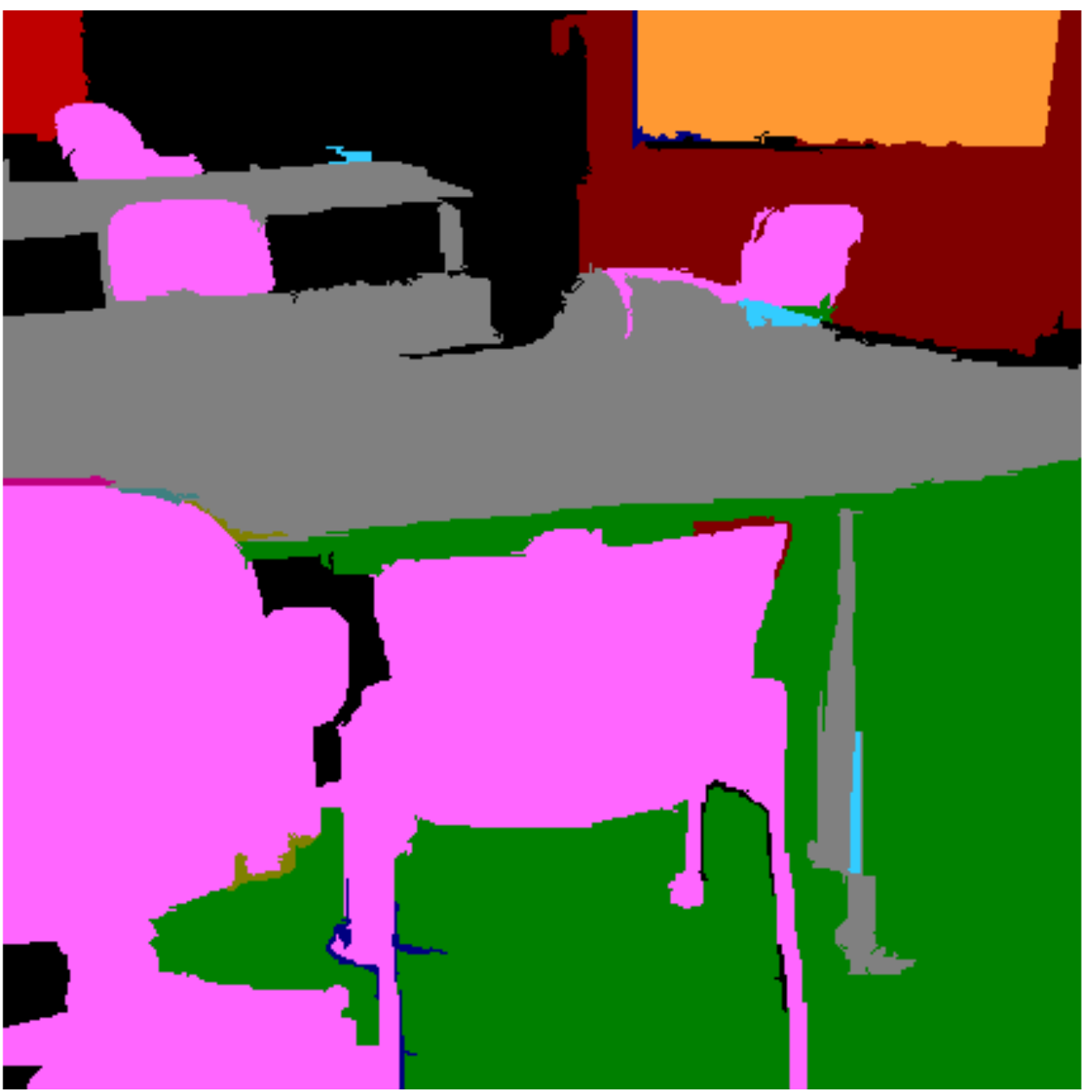}
\end{minipage}
}
\subfigure[]{
\begin{minipage}[]{0.095\textwidth}
\includegraphics[width=1\textwidth]{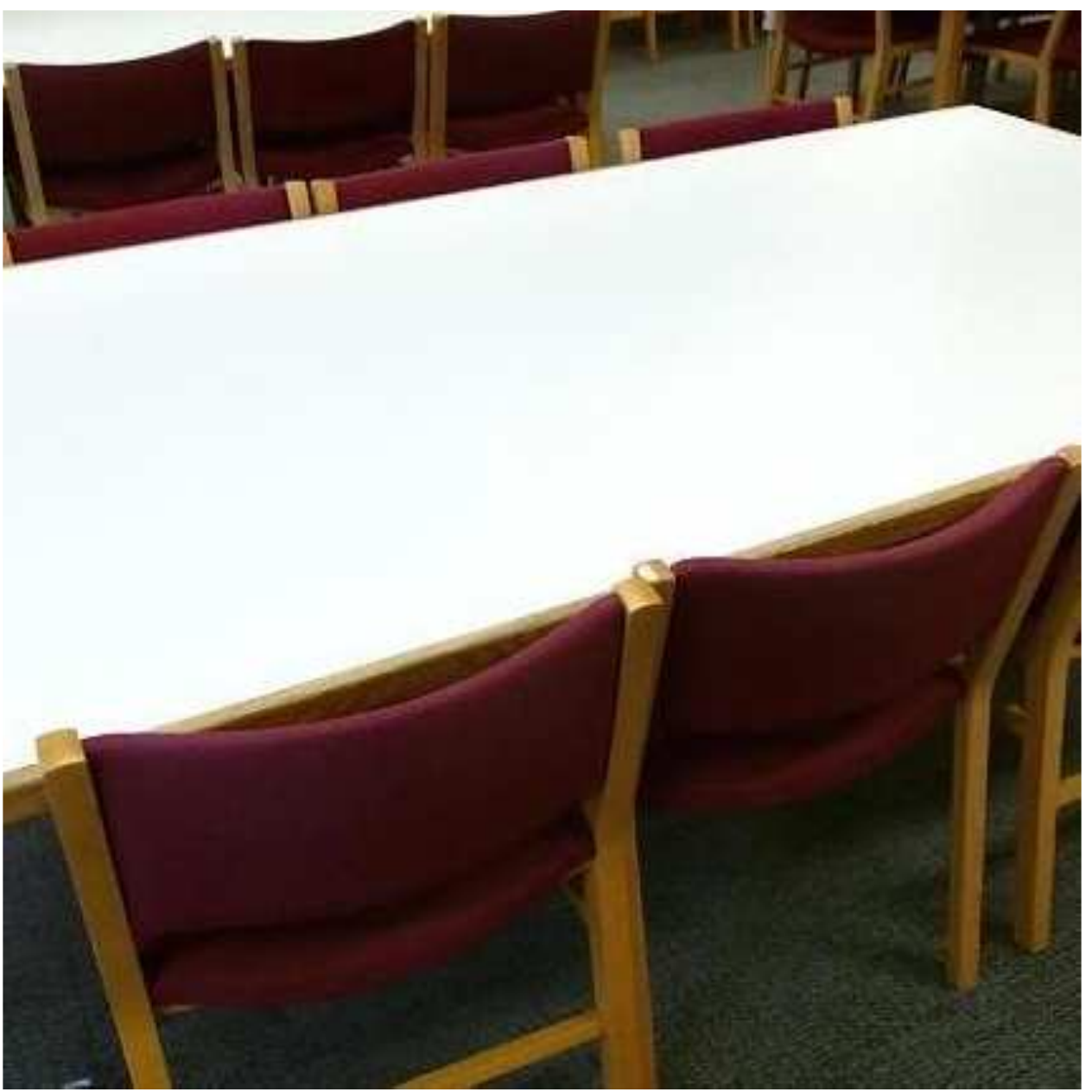} \\
\includegraphics[width=1\textwidth]{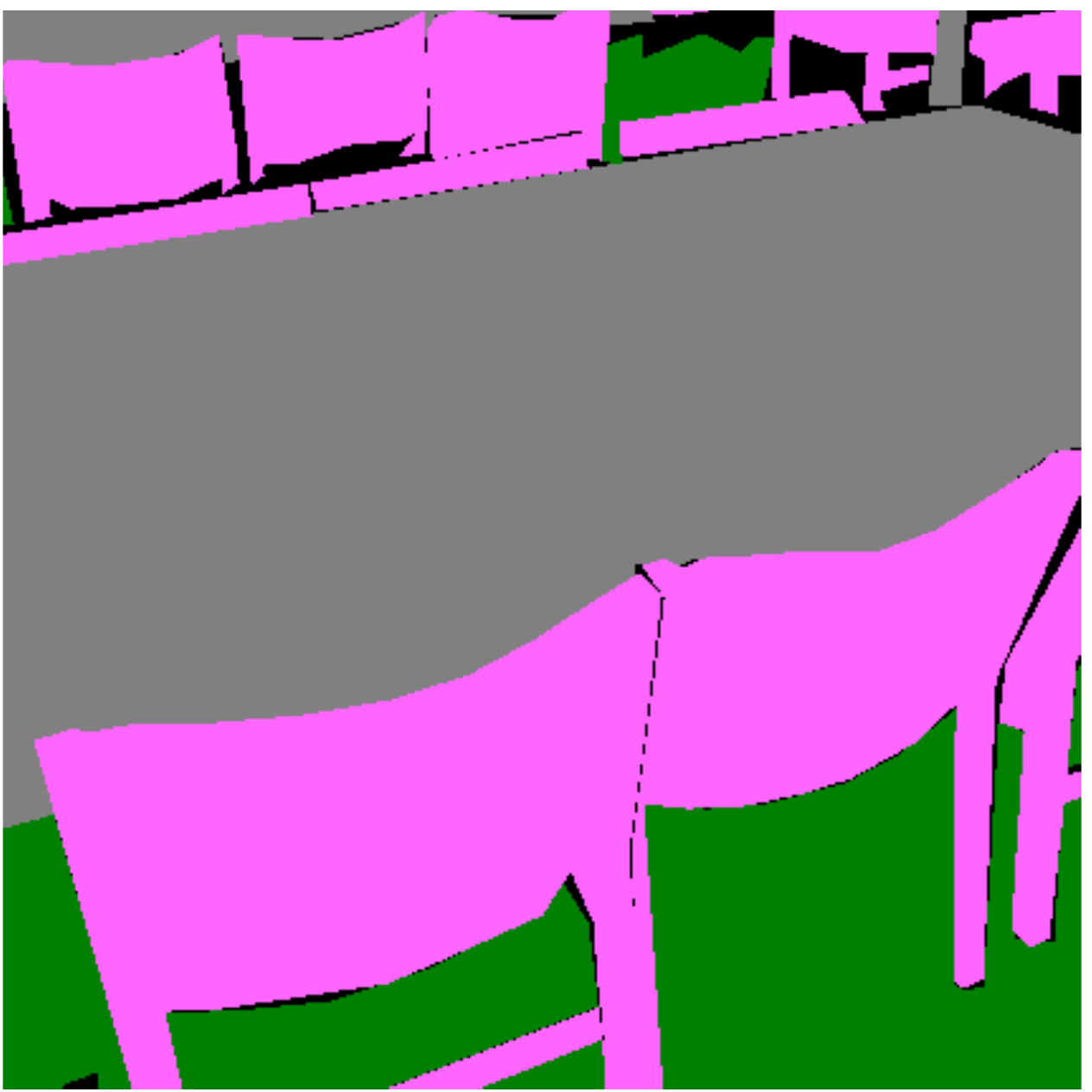} \\
\includegraphics[width=\textwidth]{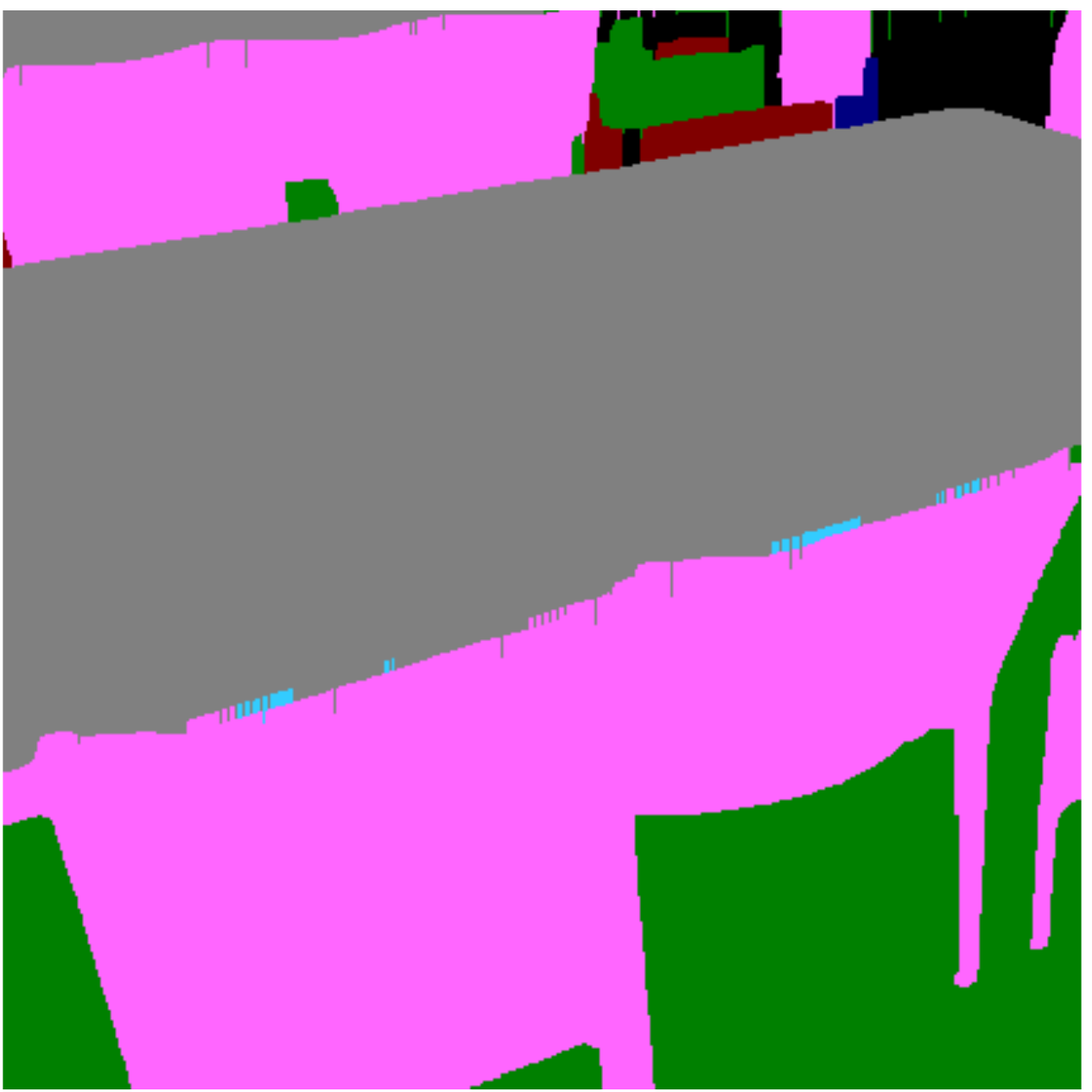}
\end{minipage}
}
\subfigure[]{
\begin{minipage}[]{0.095\textwidth}
\includegraphics[width=1\textwidth]{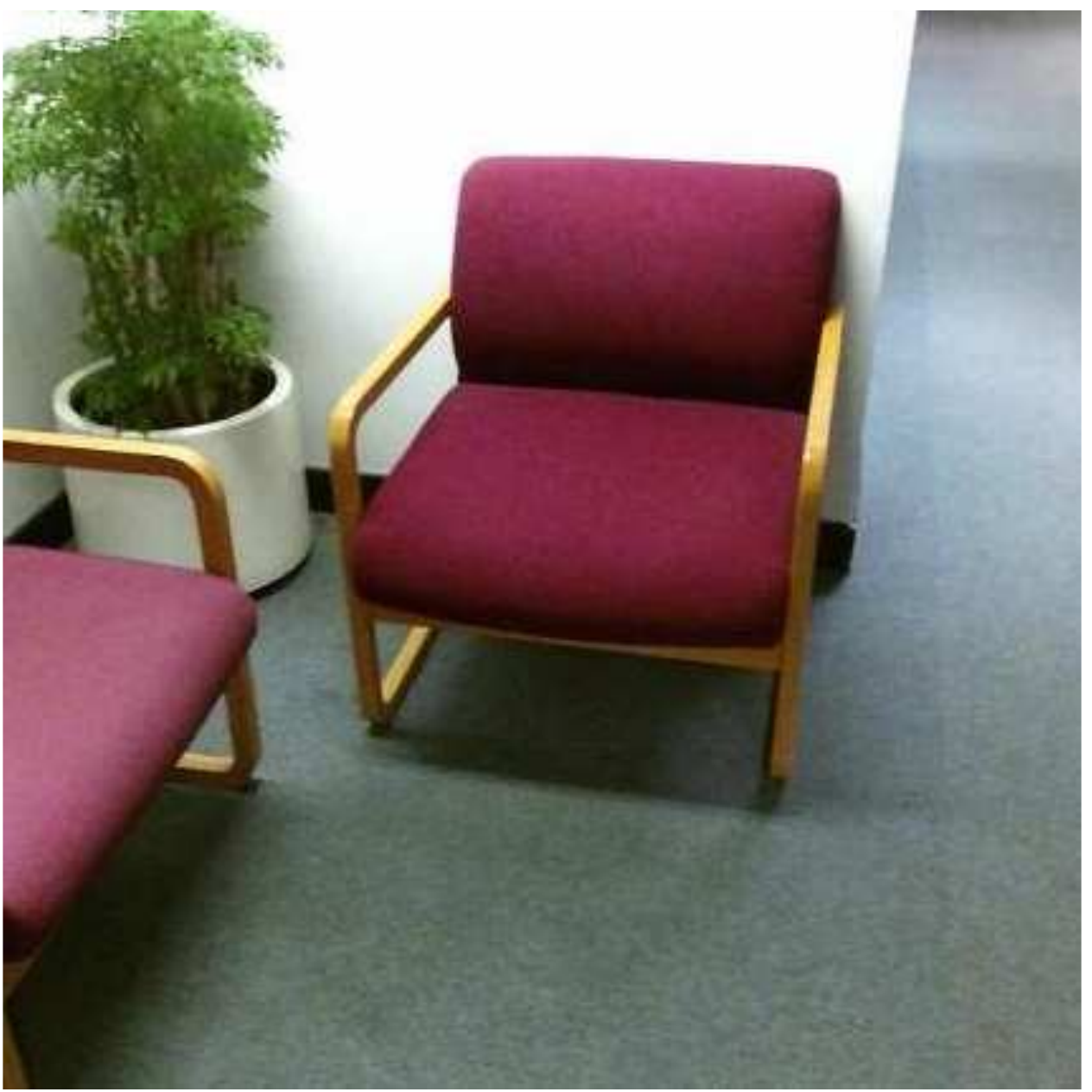} \\
\includegraphics[width=1\textwidth]{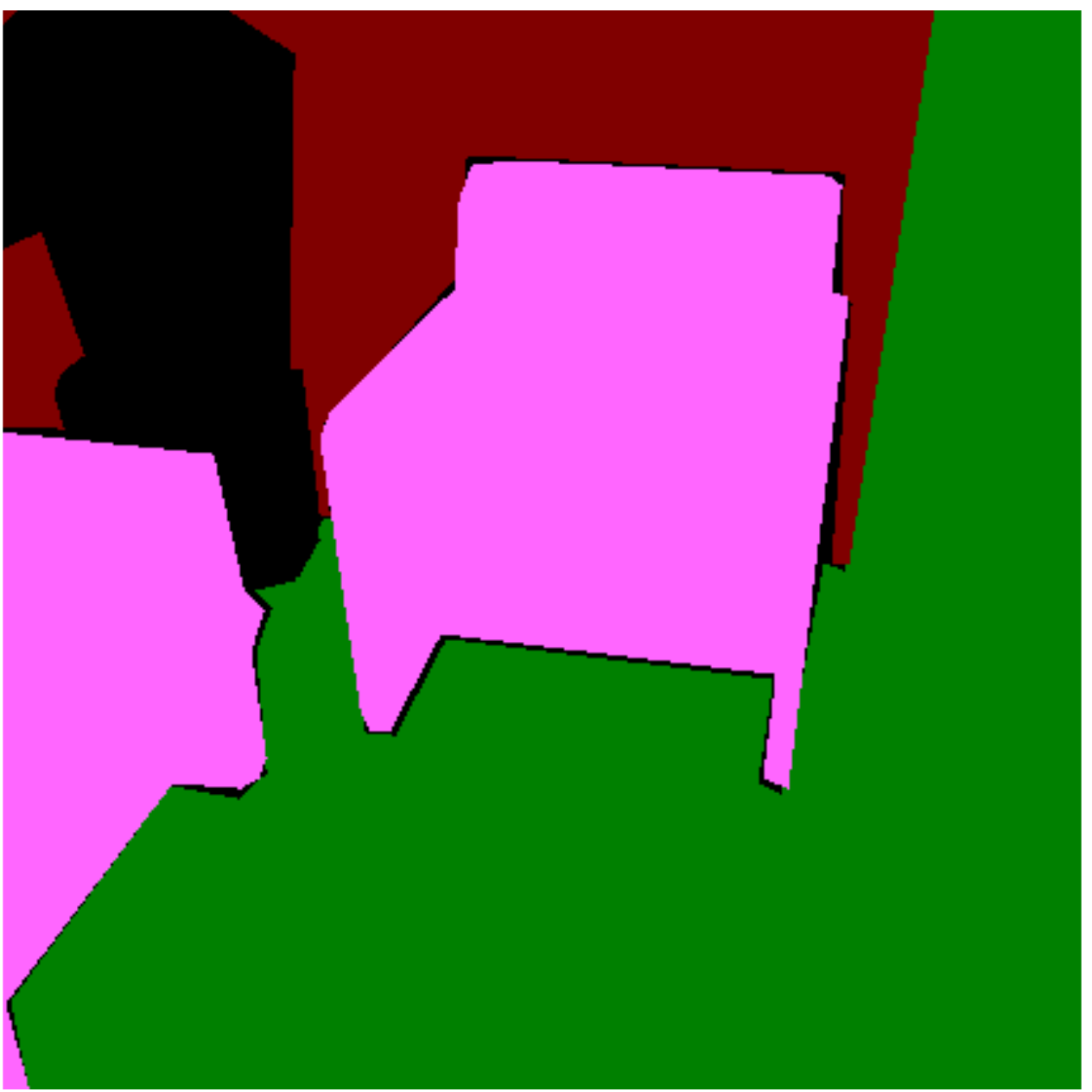} \\
\includegraphics[width=\textwidth]{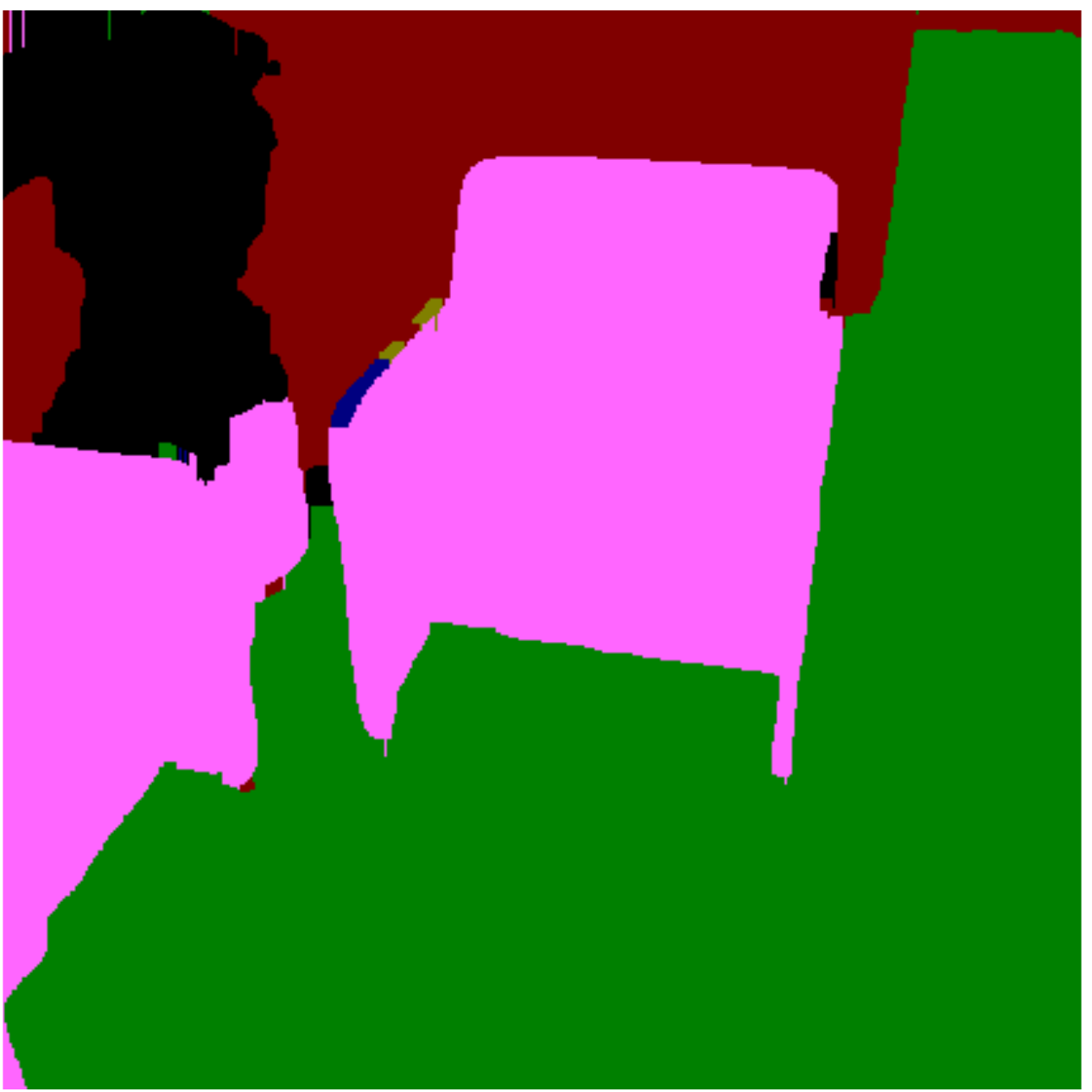}
\end{minipage}
}
\subfigure[]{
\begin{minipage}[]{0.095\textwidth}
\includegraphics[width=0.9\textwidth]{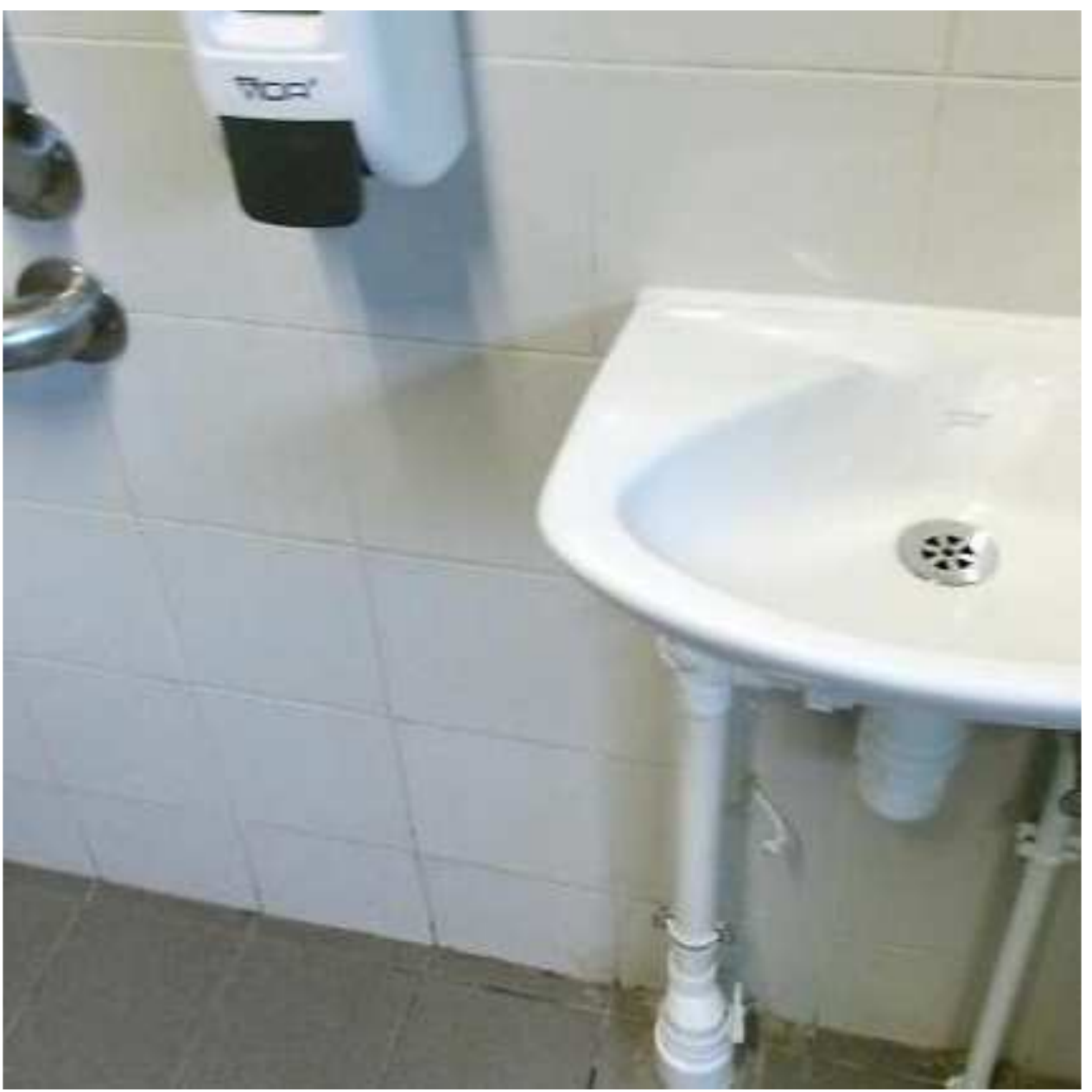} \\
\includegraphics[width=0.9\textwidth]{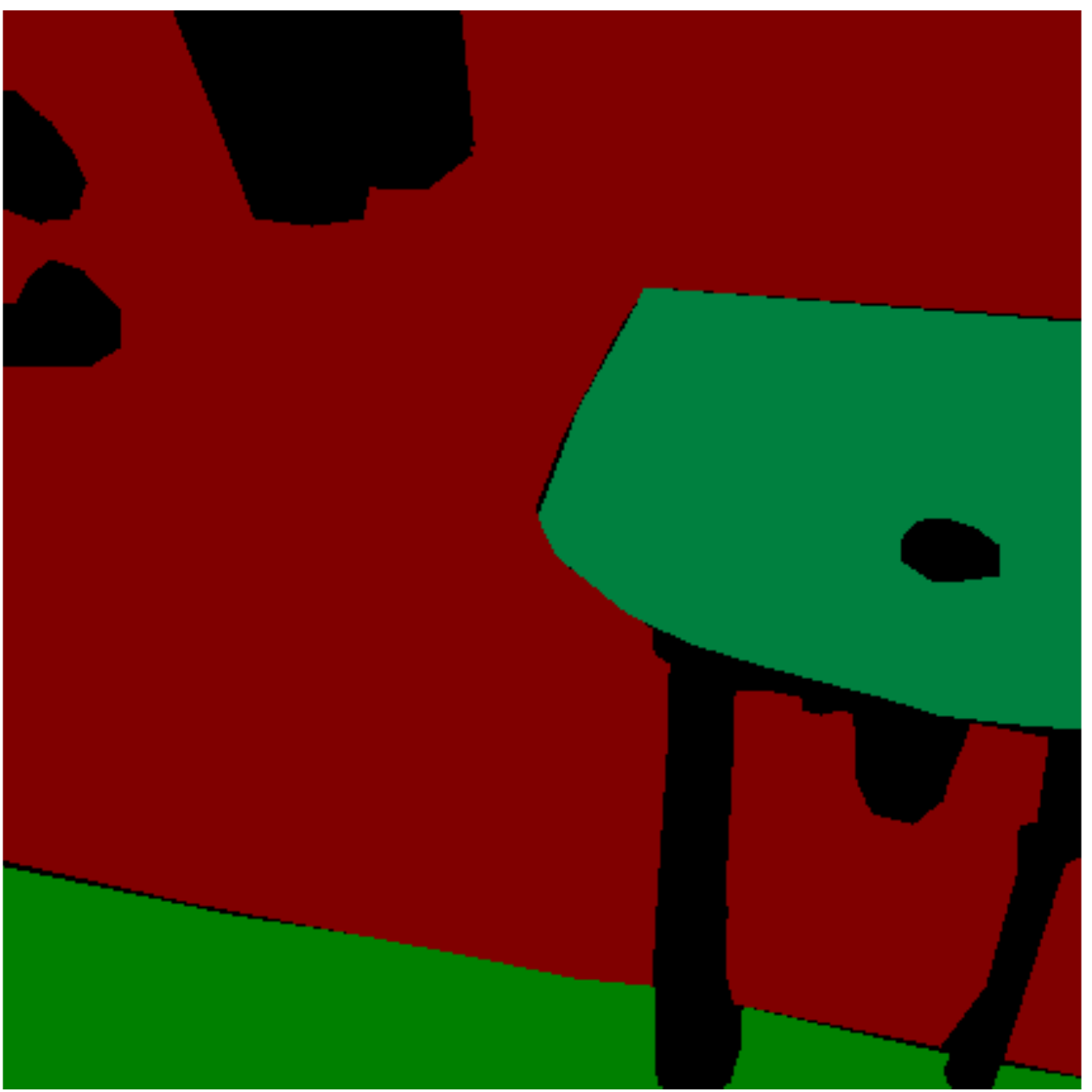} \\
\includegraphics[width=0.9\textwidth]{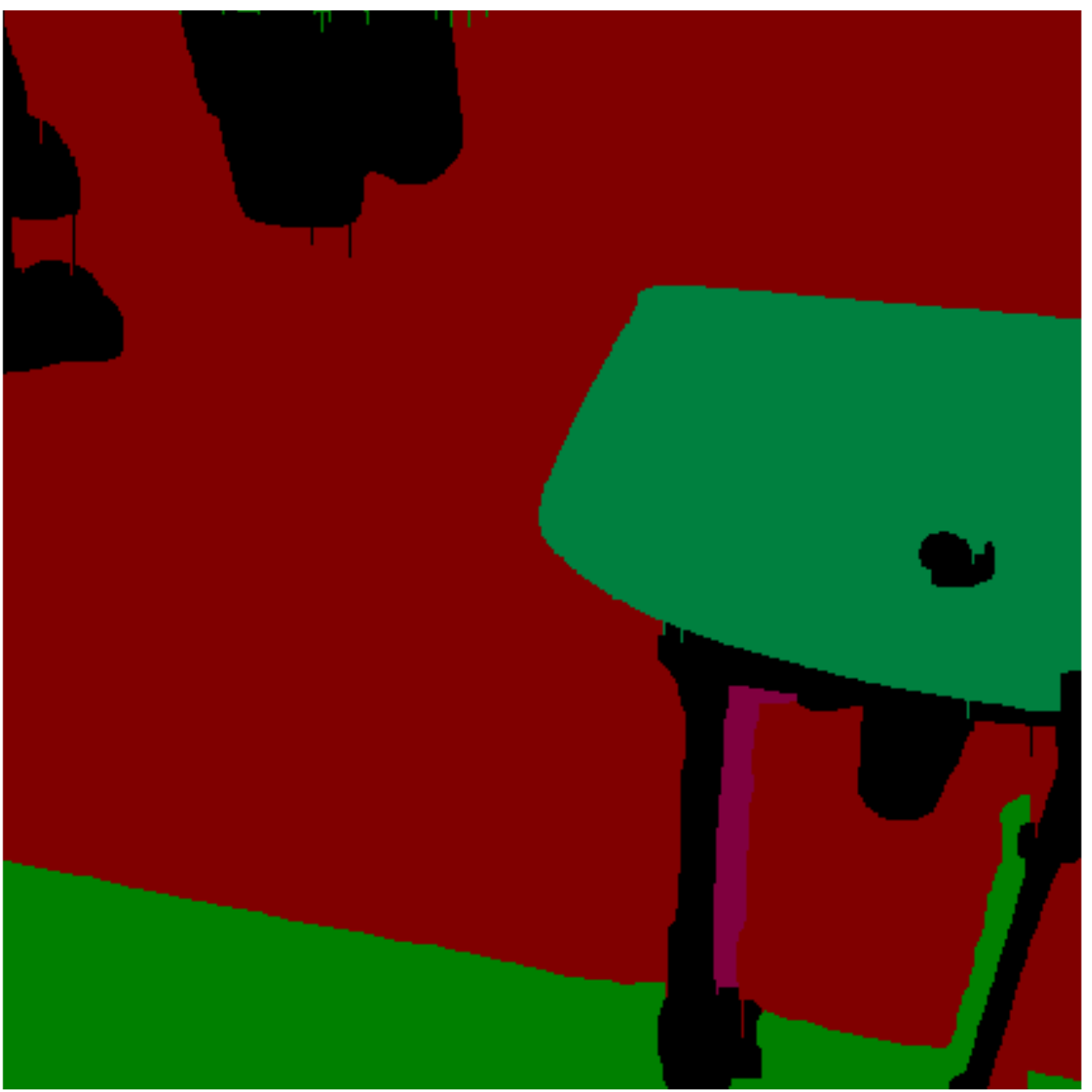}
\end{minipage}
}
\subfigure[legend of semantic labels]{
\begin{minipage}[]{0.9\textwidth}
\includegraphics[width=1\textwidth]{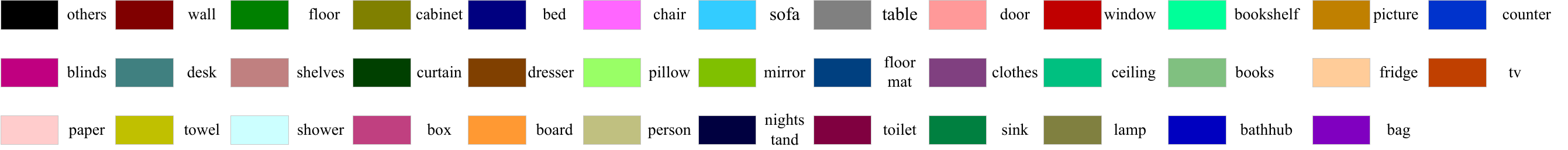} \\
\end{minipage}
}
\caption{Examples of semantic labeling results on the SUNRGBD dataset. The top row shows the input RGB images, the bottom row shows scene labeling obtained with our model and the middle row has the ground truth. Semantic labels and their corresponding colors are shown at the bottom.}
\label{fig:visulization}
\end{figure}

\begin{figure}[t]
\centering
\subfigure[]{
\begin{minipage}[b]{0.1\textwidth}
\includegraphics[width=0.9\textwidth]{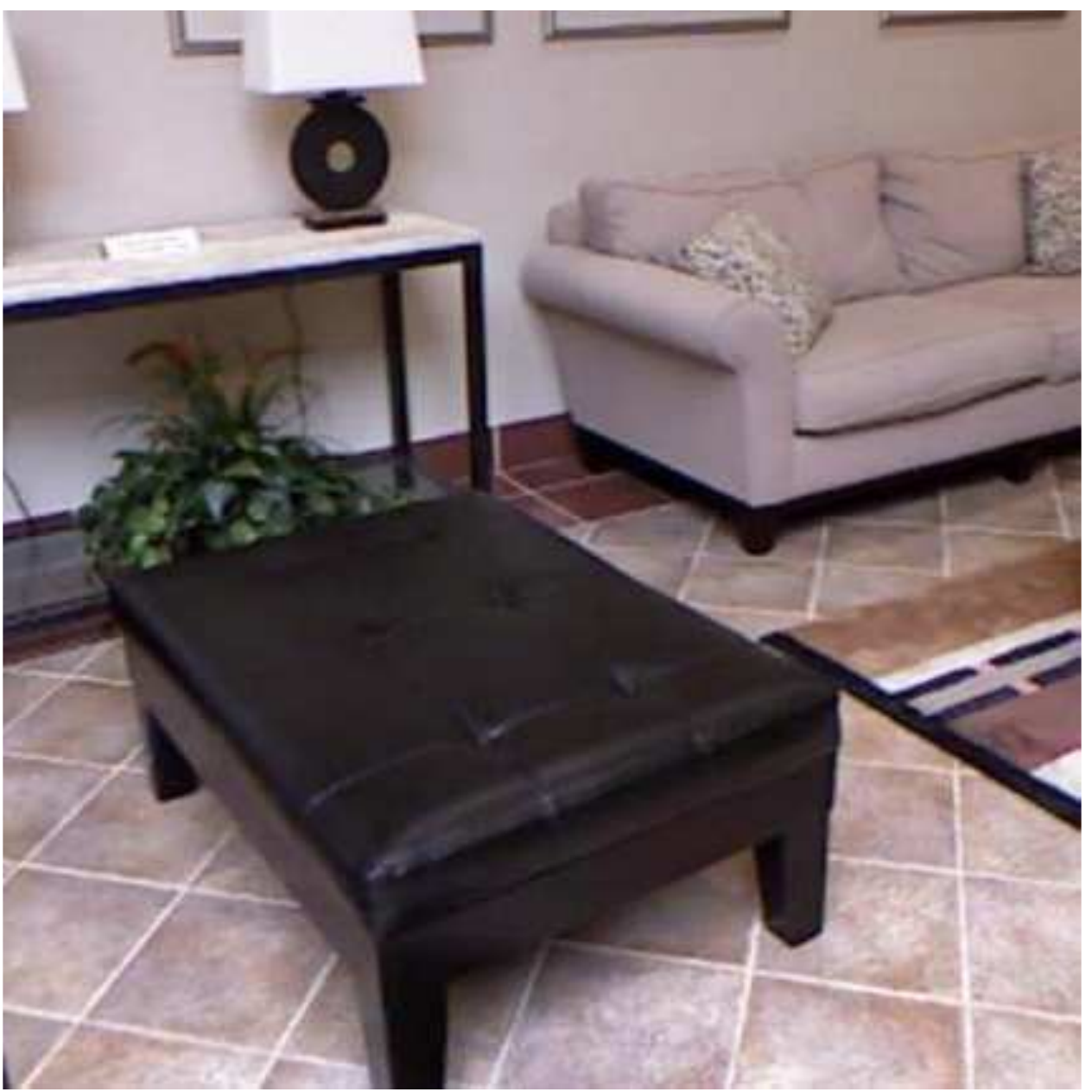} \\
\includegraphics[width=0.9\textwidth]{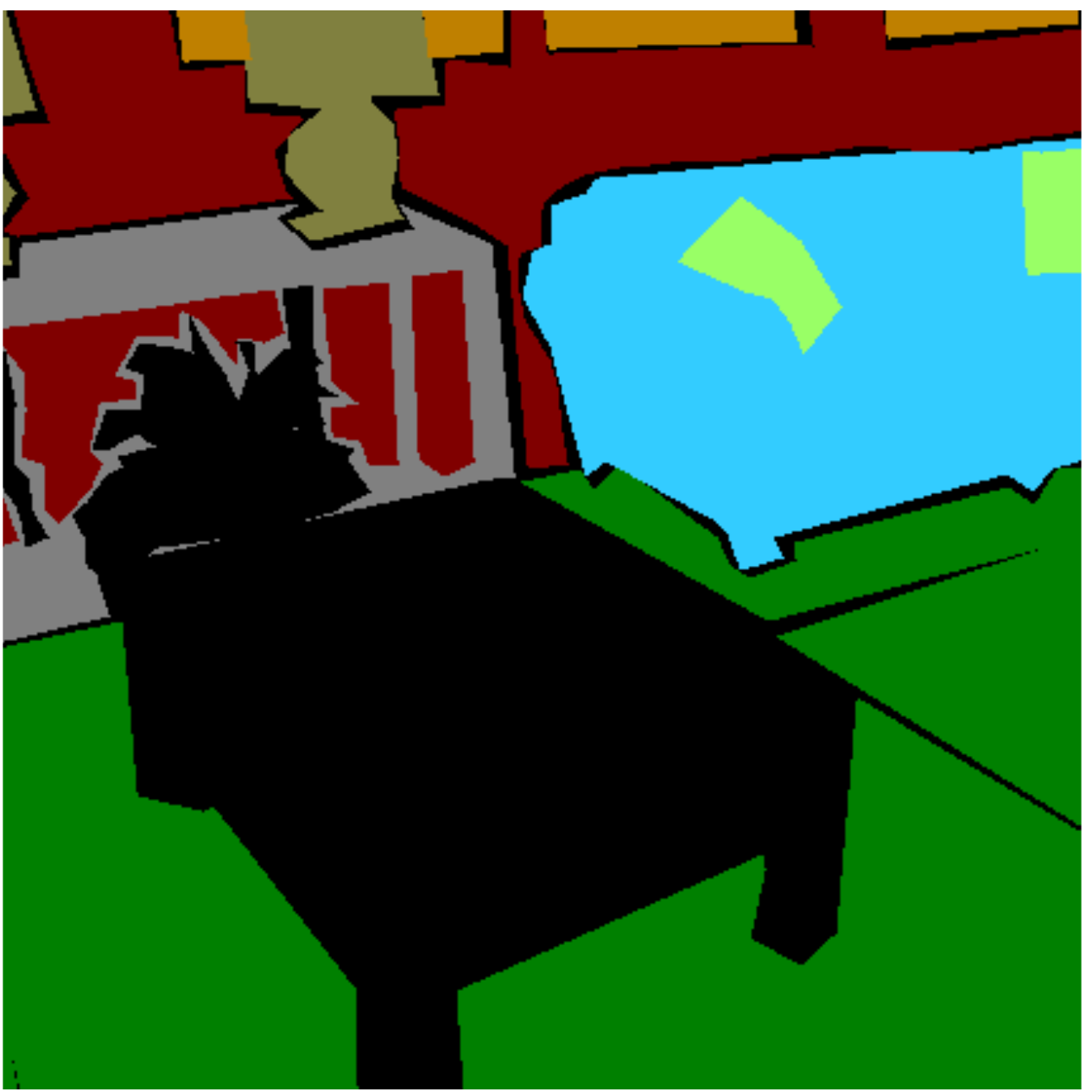} \\
\includegraphics[width=0.9\textwidth]{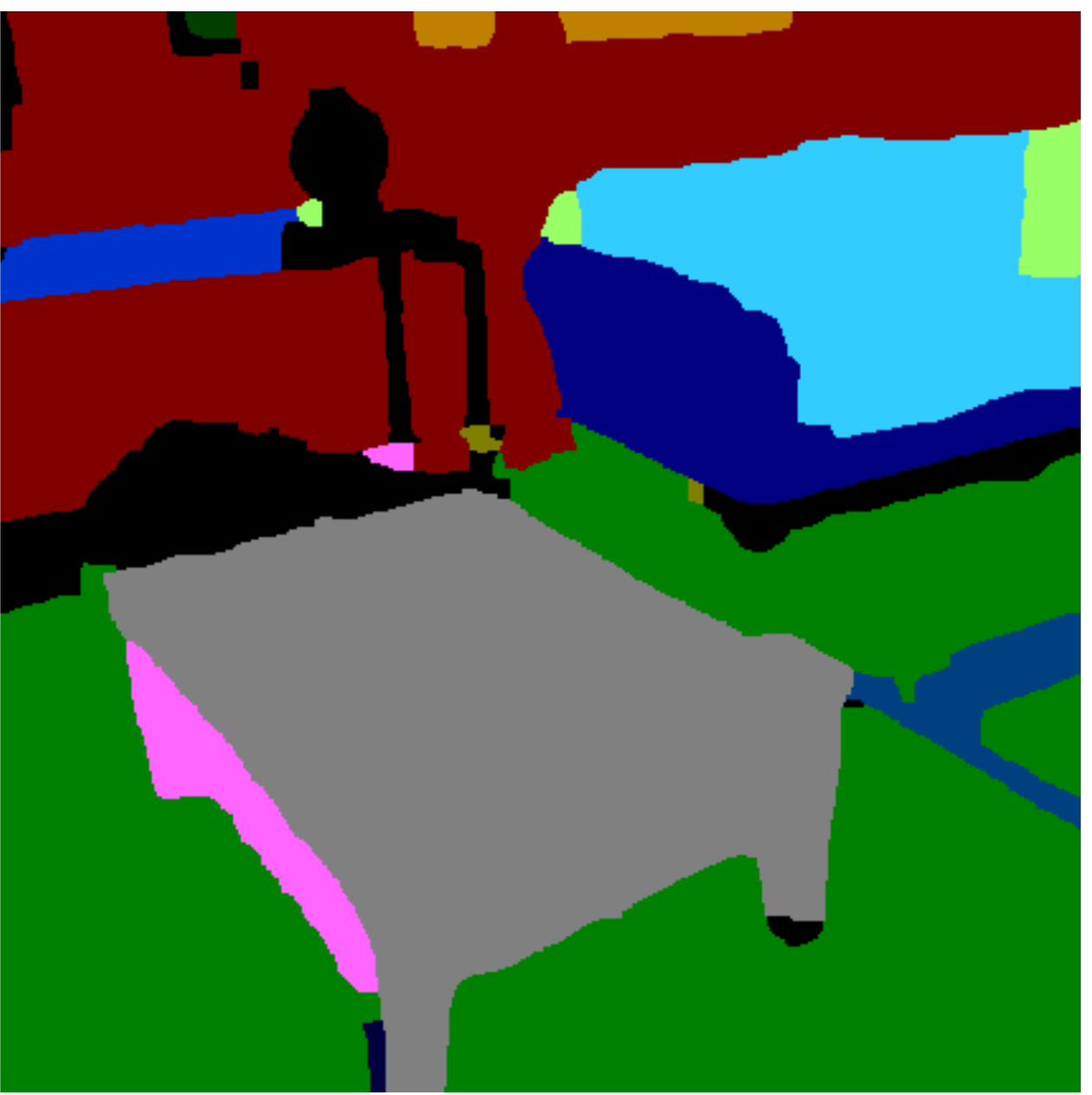} \\
\includegraphics[width=\textwidth]{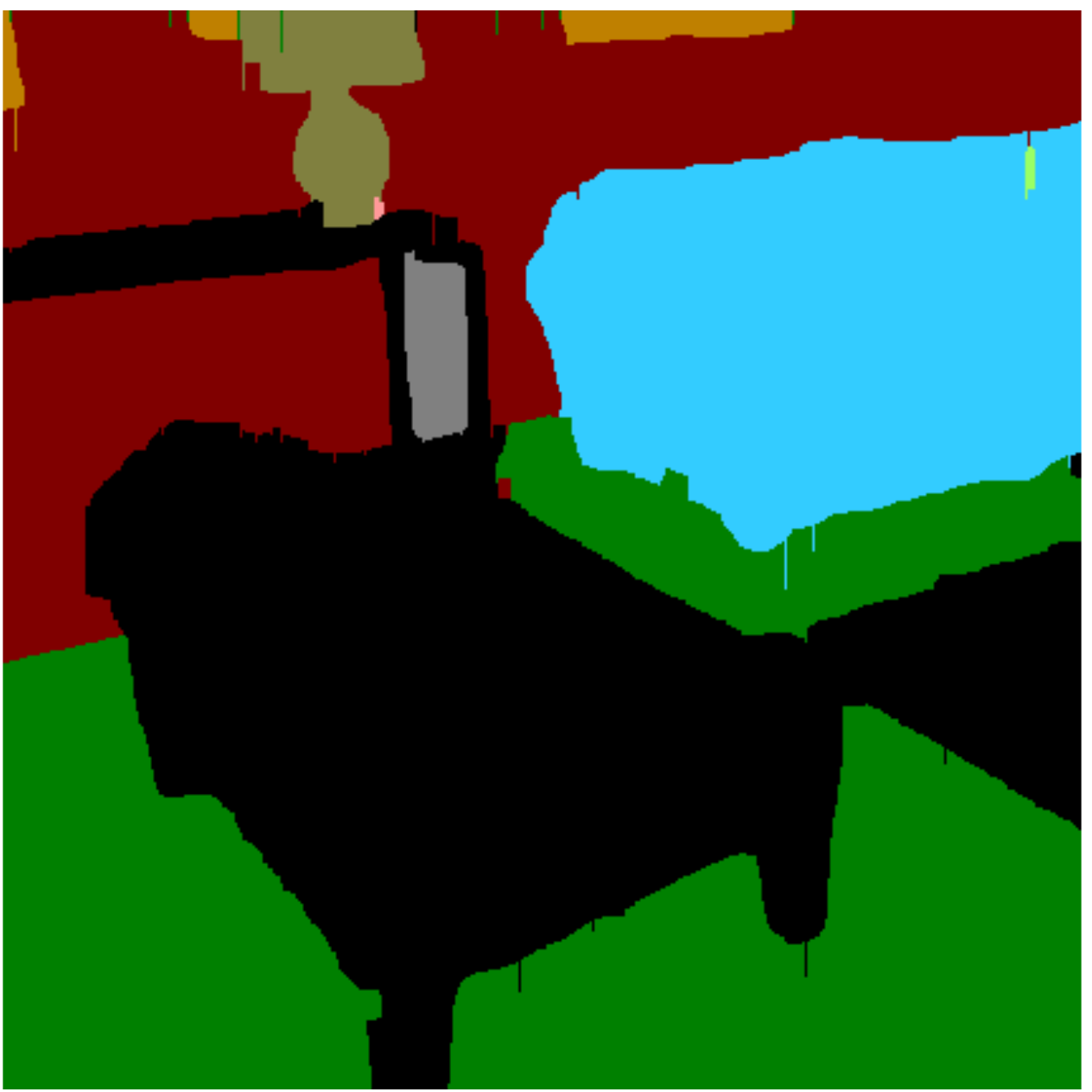}
\end{minipage}
}
\subfigure[]{
\begin{minipage}[b]{0.1\textwidth}
\includegraphics[width=0.9\textwidth]{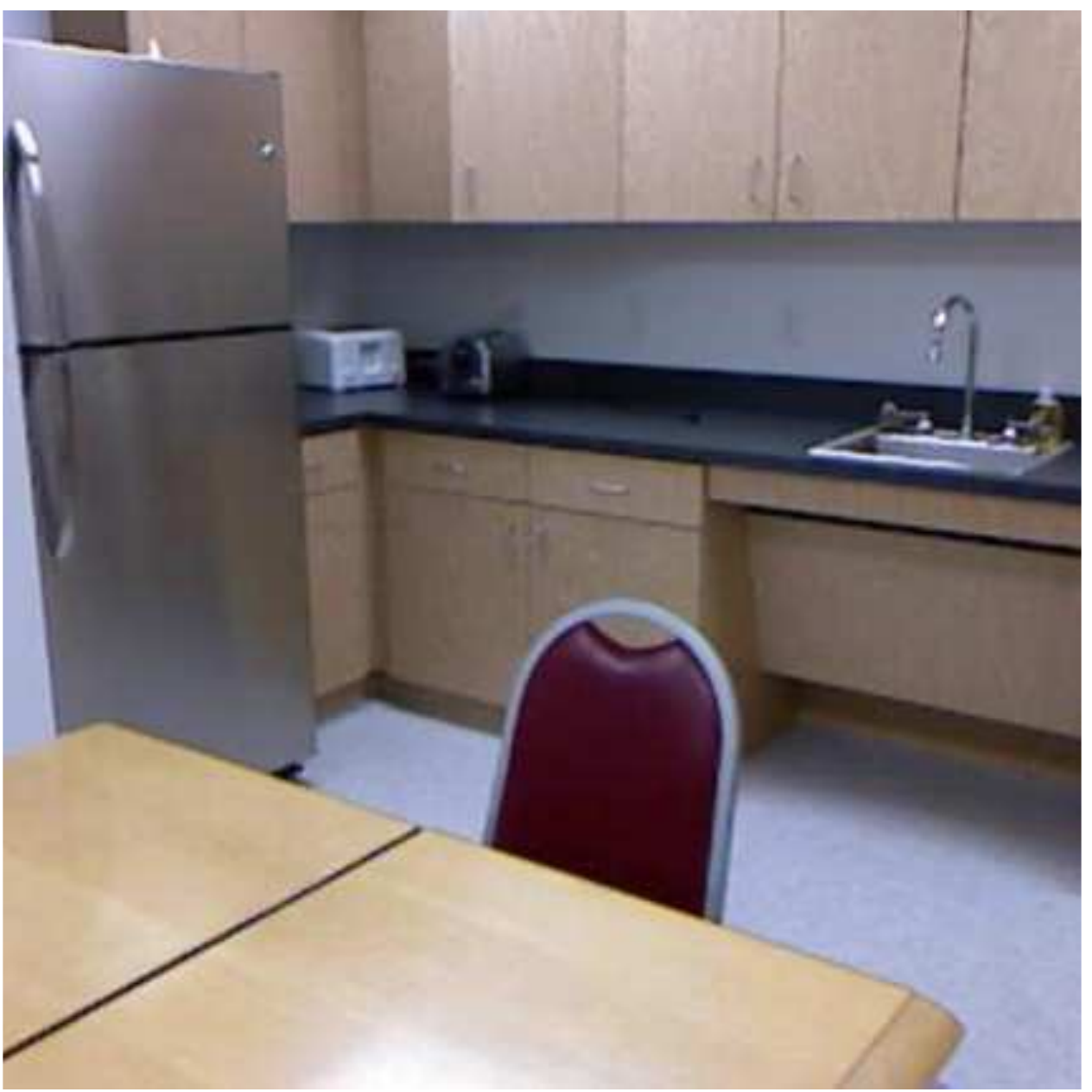} \\
\includegraphics[width=0.9\textwidth]{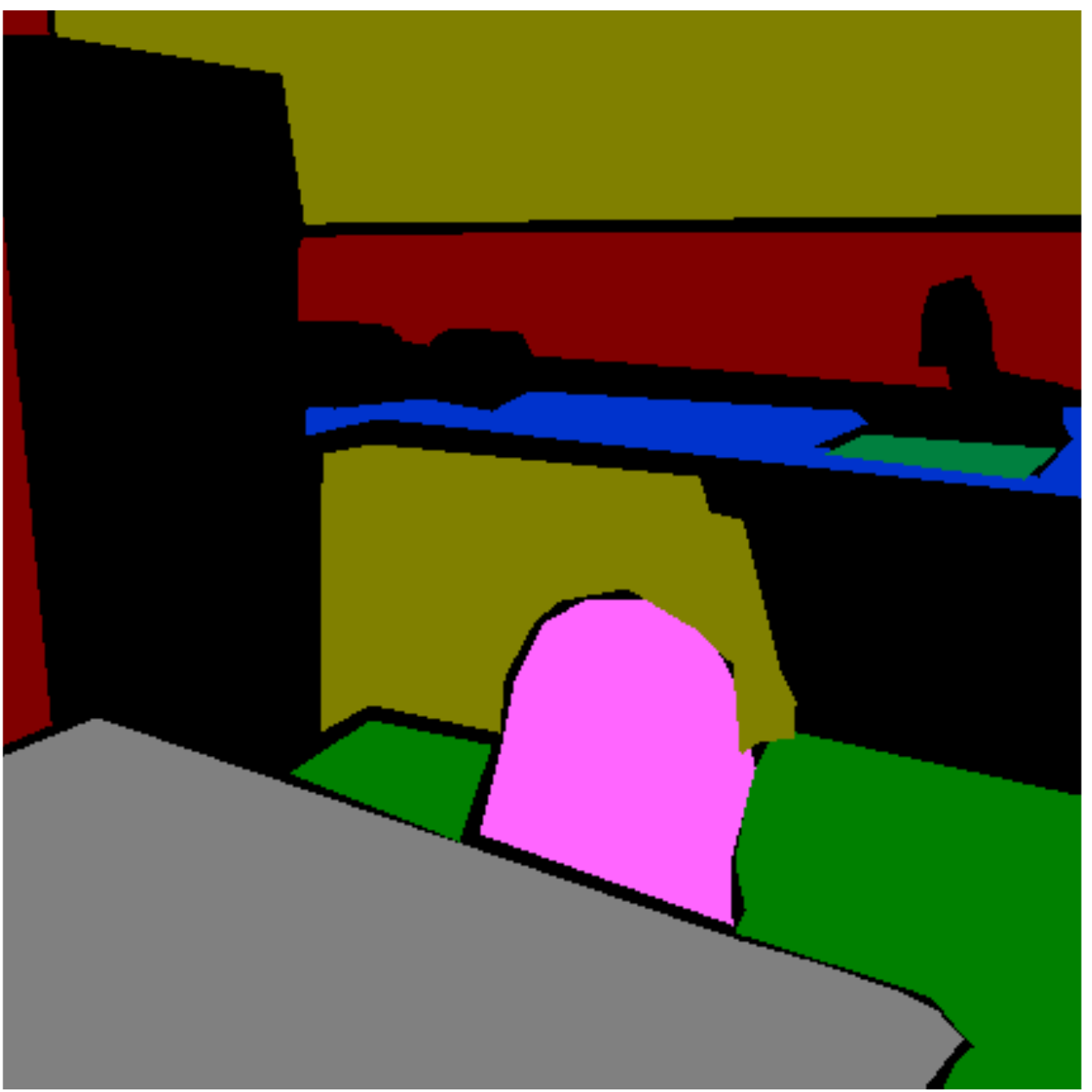} \\
\includegraphics[width=0.9\textwidth]{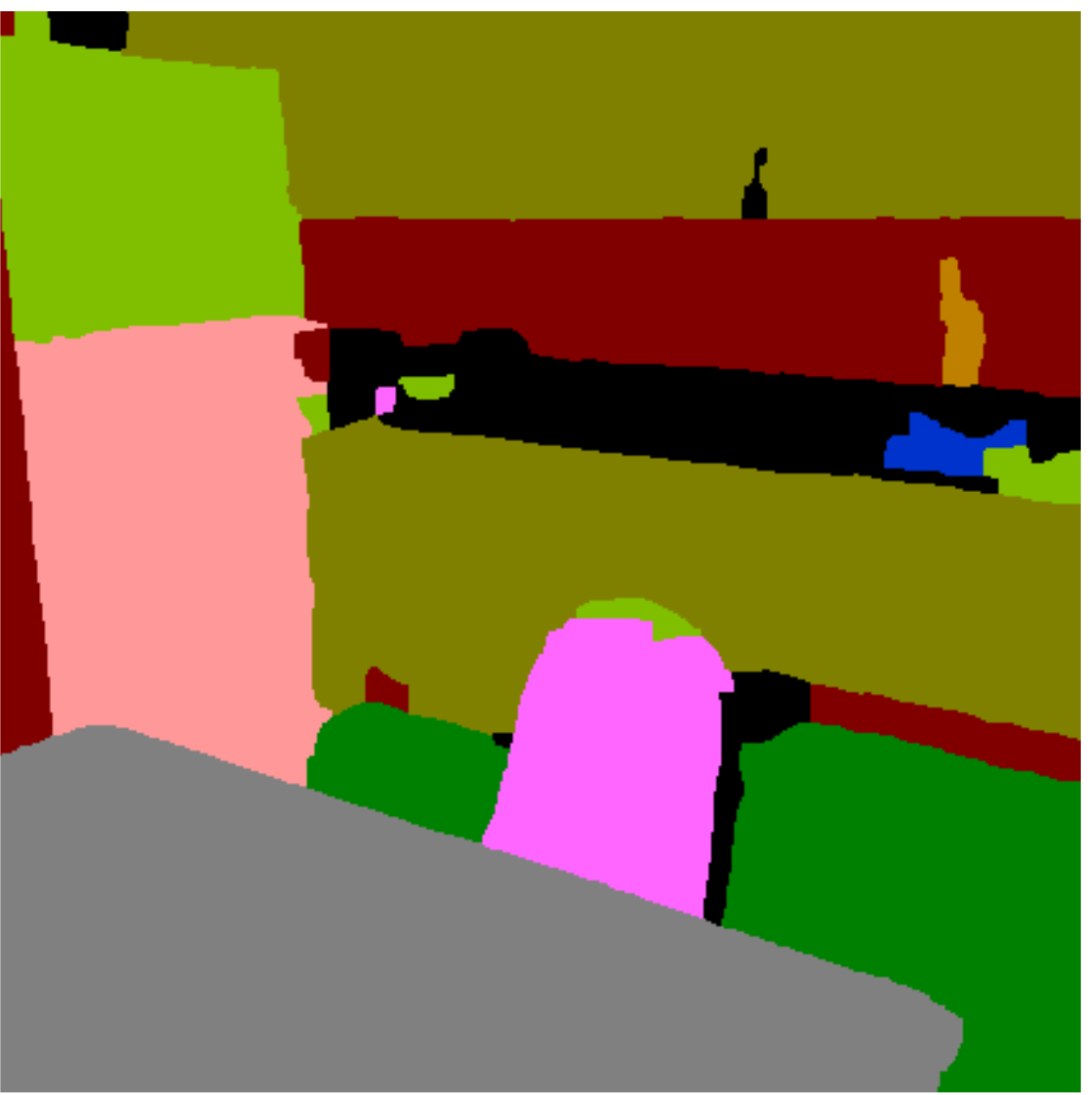} \\
\includegraphics[width=\textwidth]{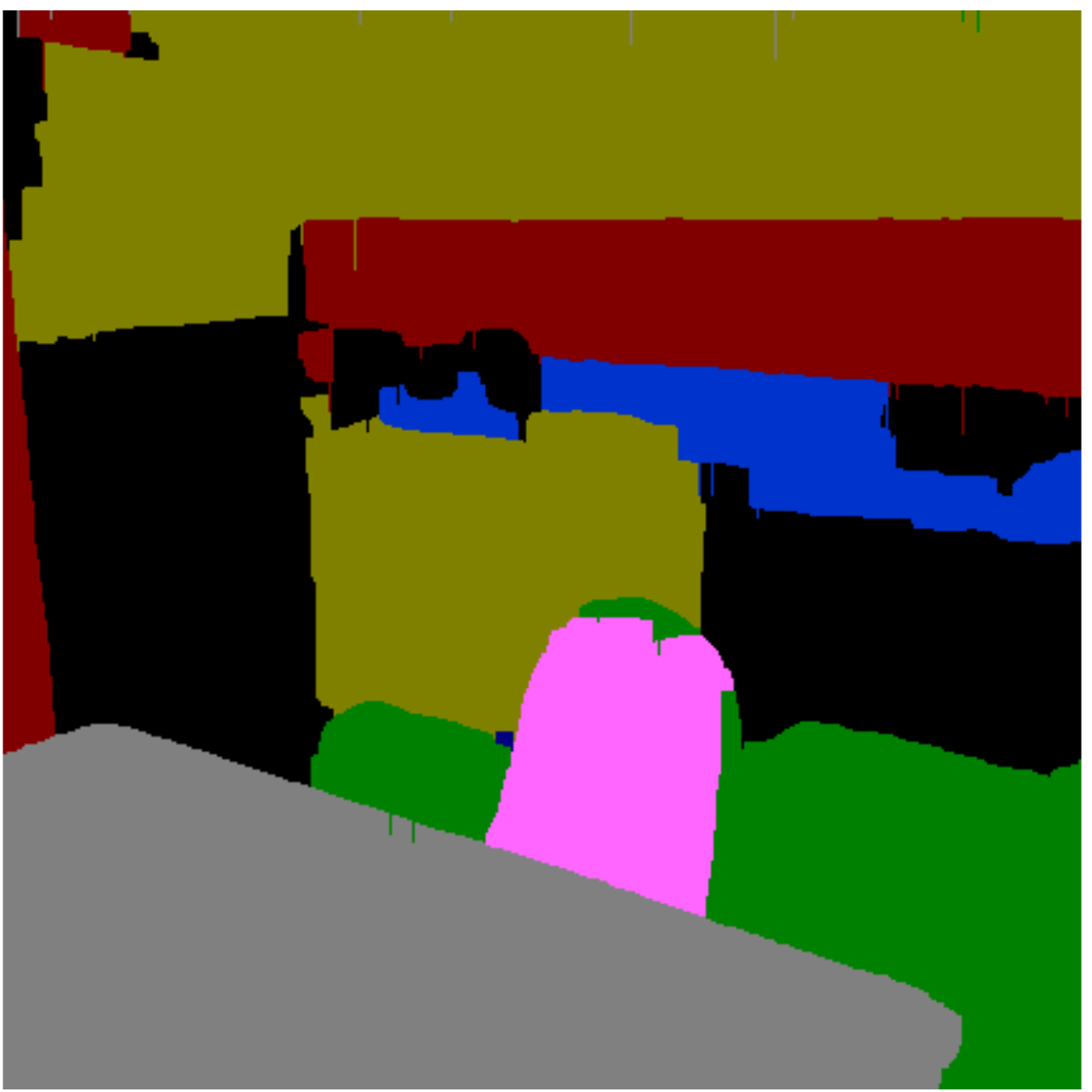}
\end{minipage}
}
\subfigure[]{
\begin{minipage}[b]{0.1\textwidth}
\includegraphics[width=0.9\textwidth]{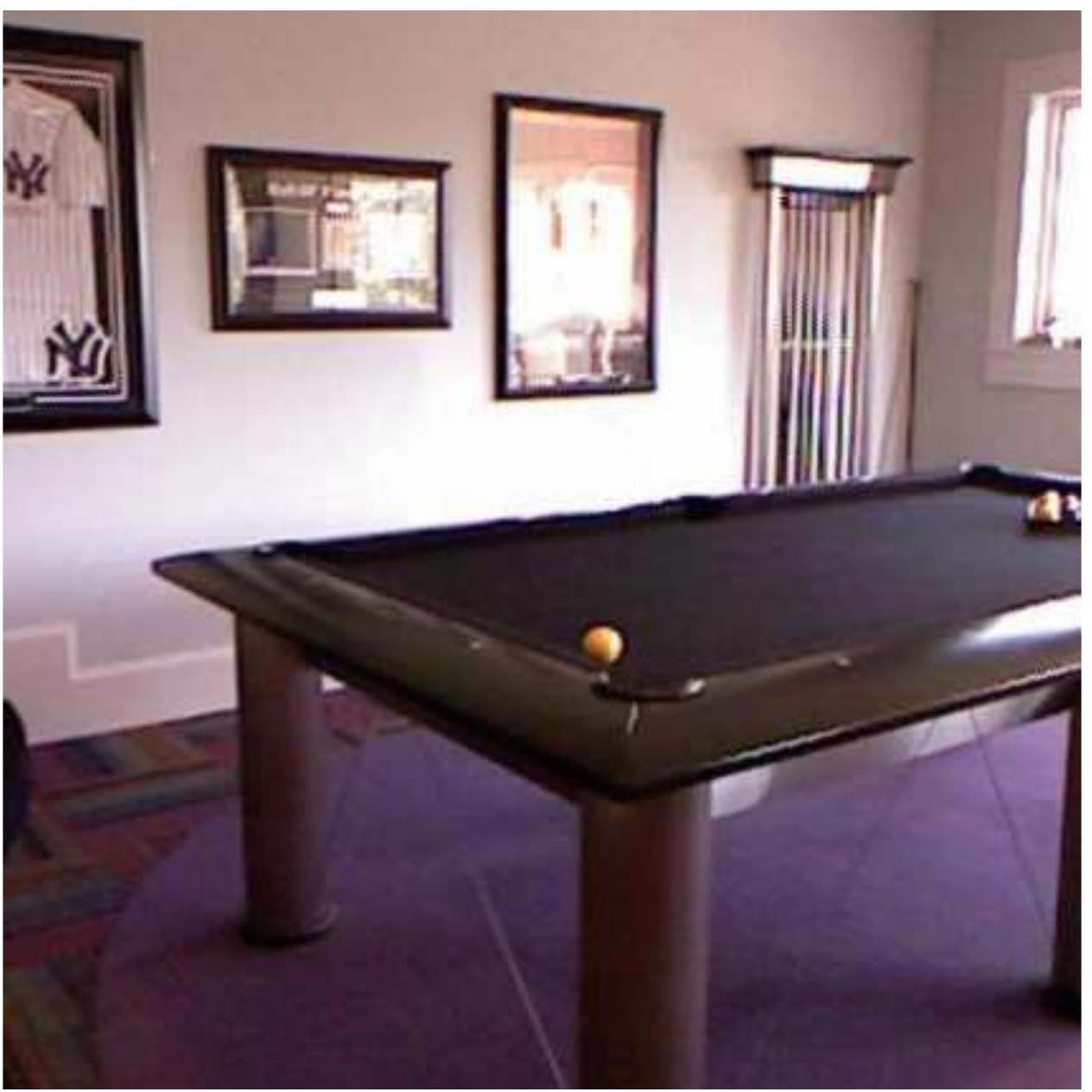} \\
\includegraphics[width=0.9\textwidth]{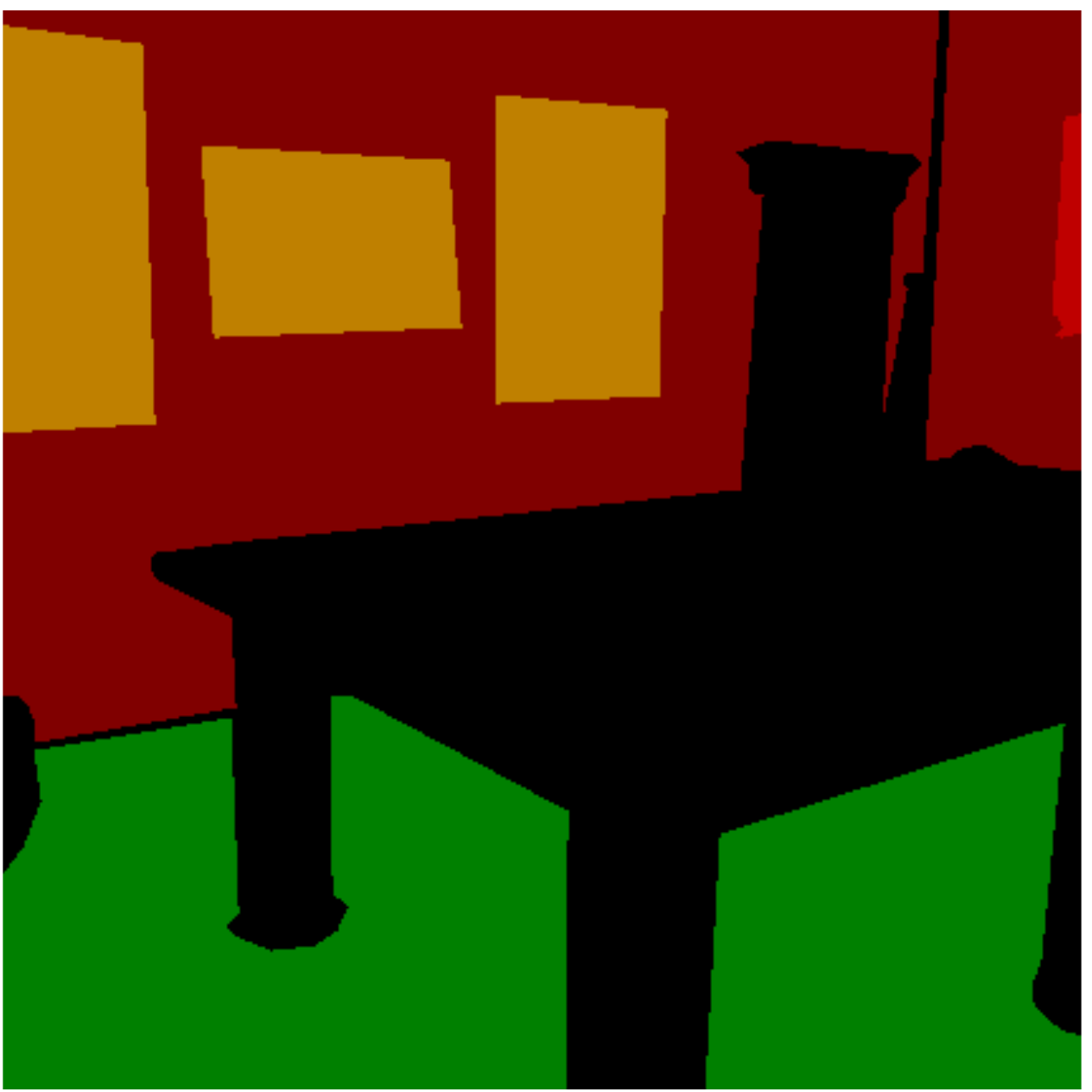} \\
\includegraphics[width=0.9\textwidth]{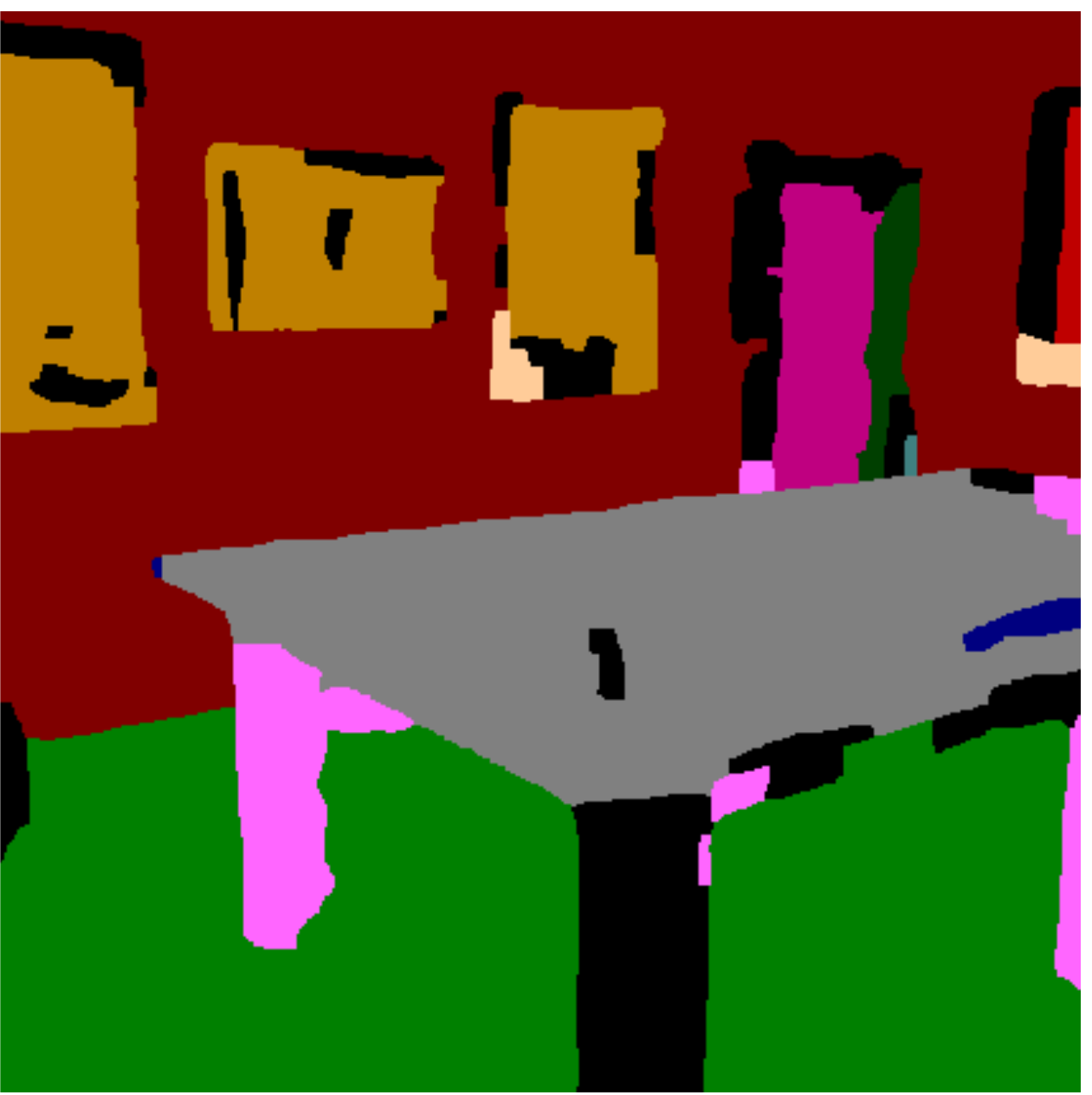} \\
\includegraphics[width=0.9\textwidth]{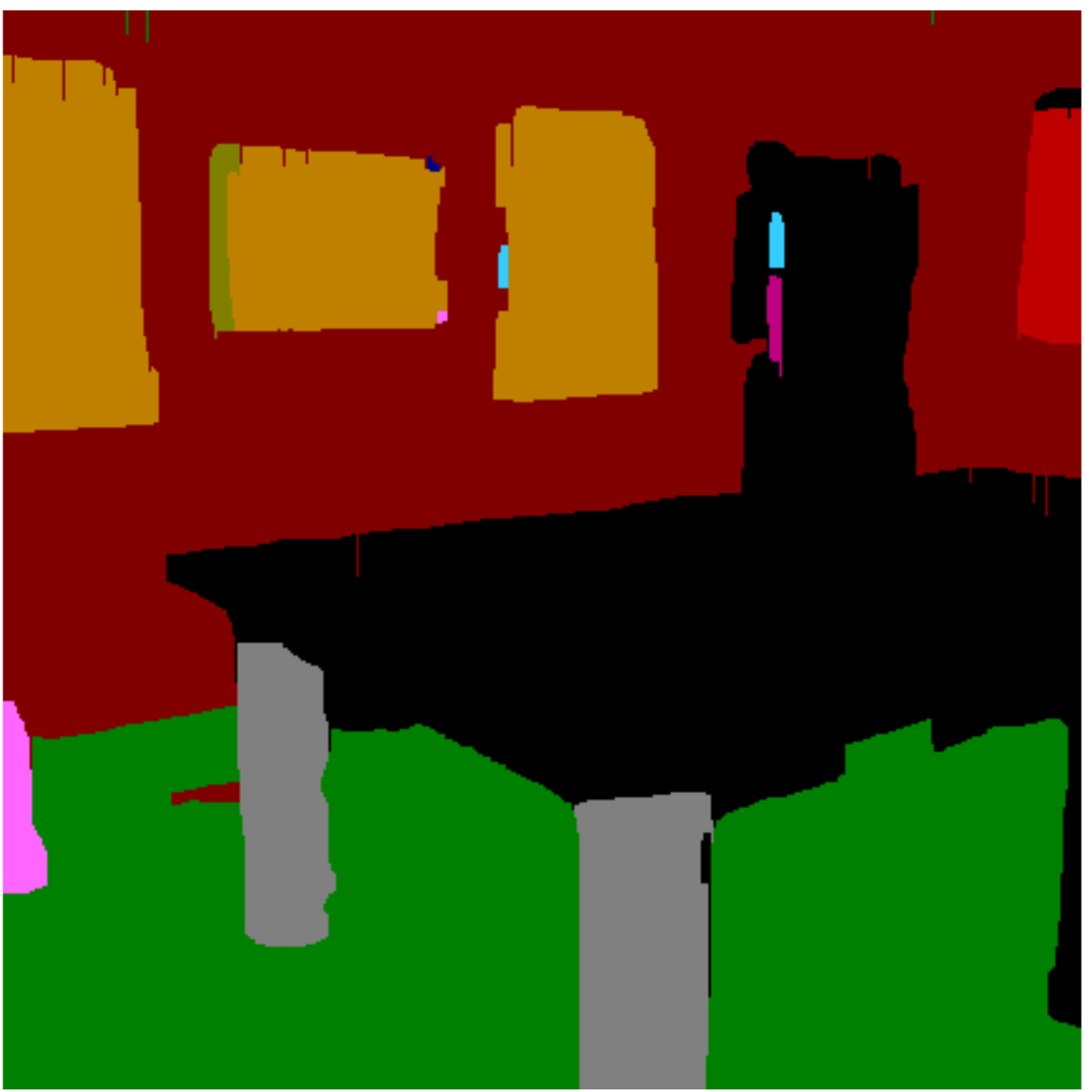}
\end{minipage}
}
\subfigure[]{
\begin{minipage}[b]{0.1\textwidth}
\includegraphics[width=0.9\textwidth]{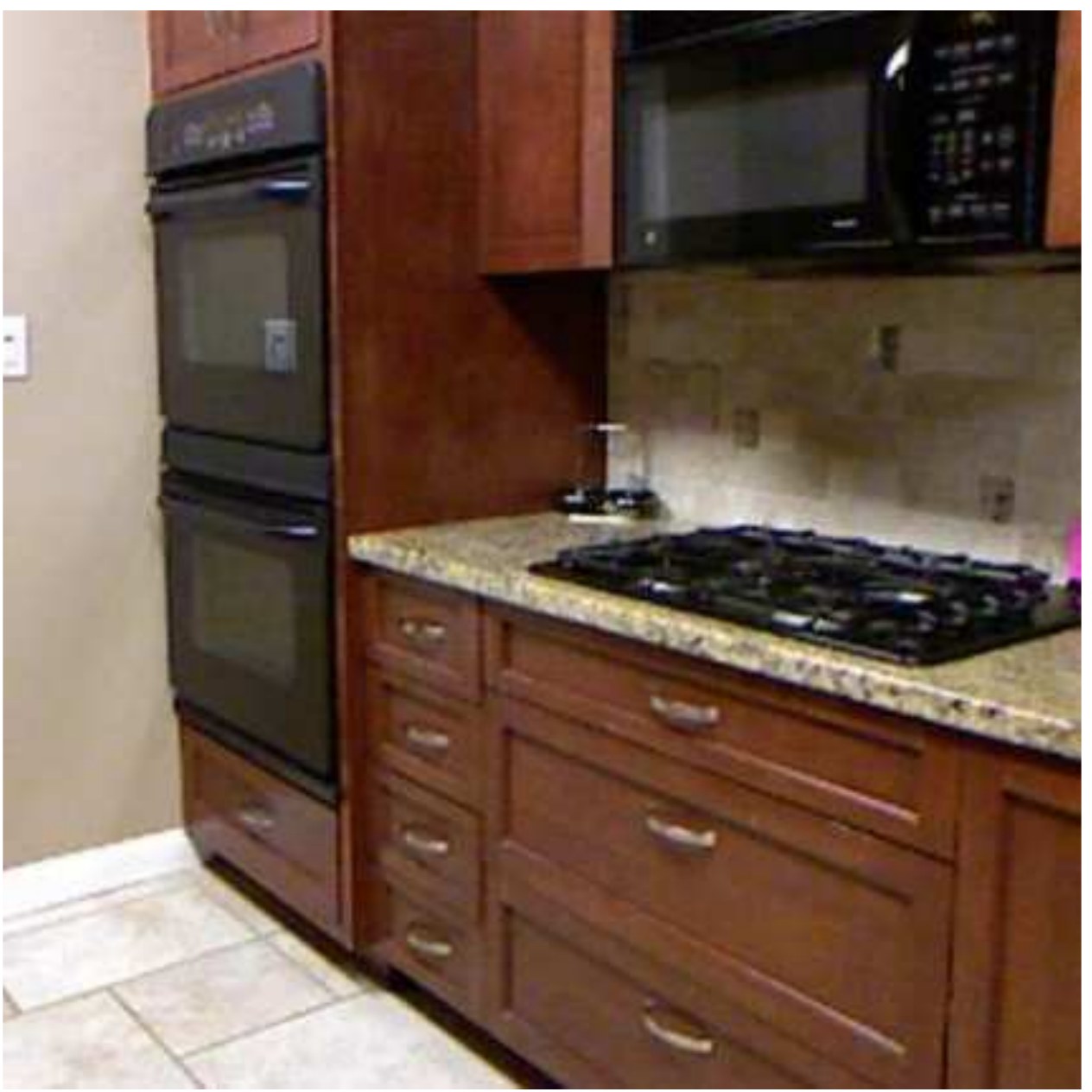} \\
\includegraphics[width=0.9\textwidth]{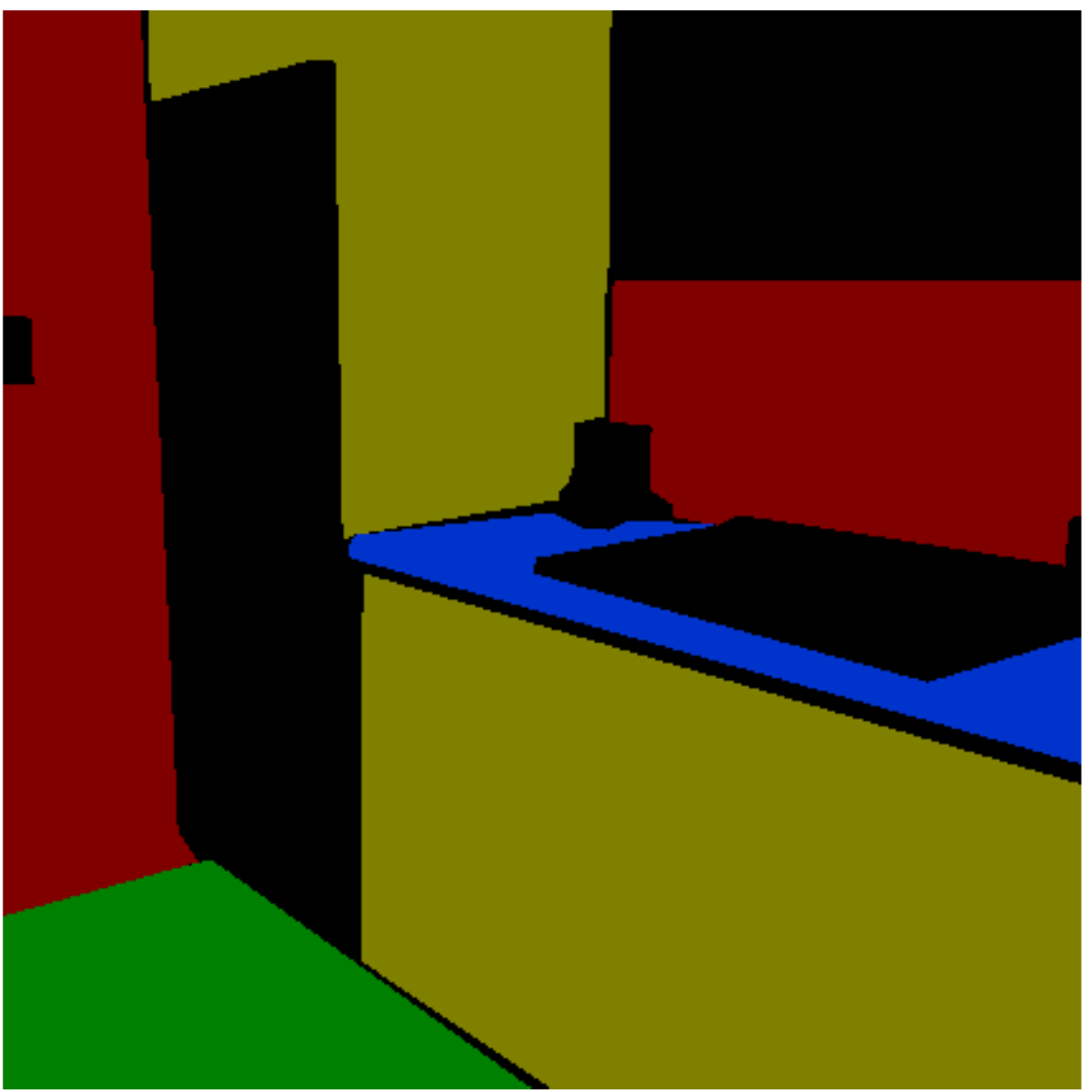} \\
\includegraphics[width=0.9\textwidth]{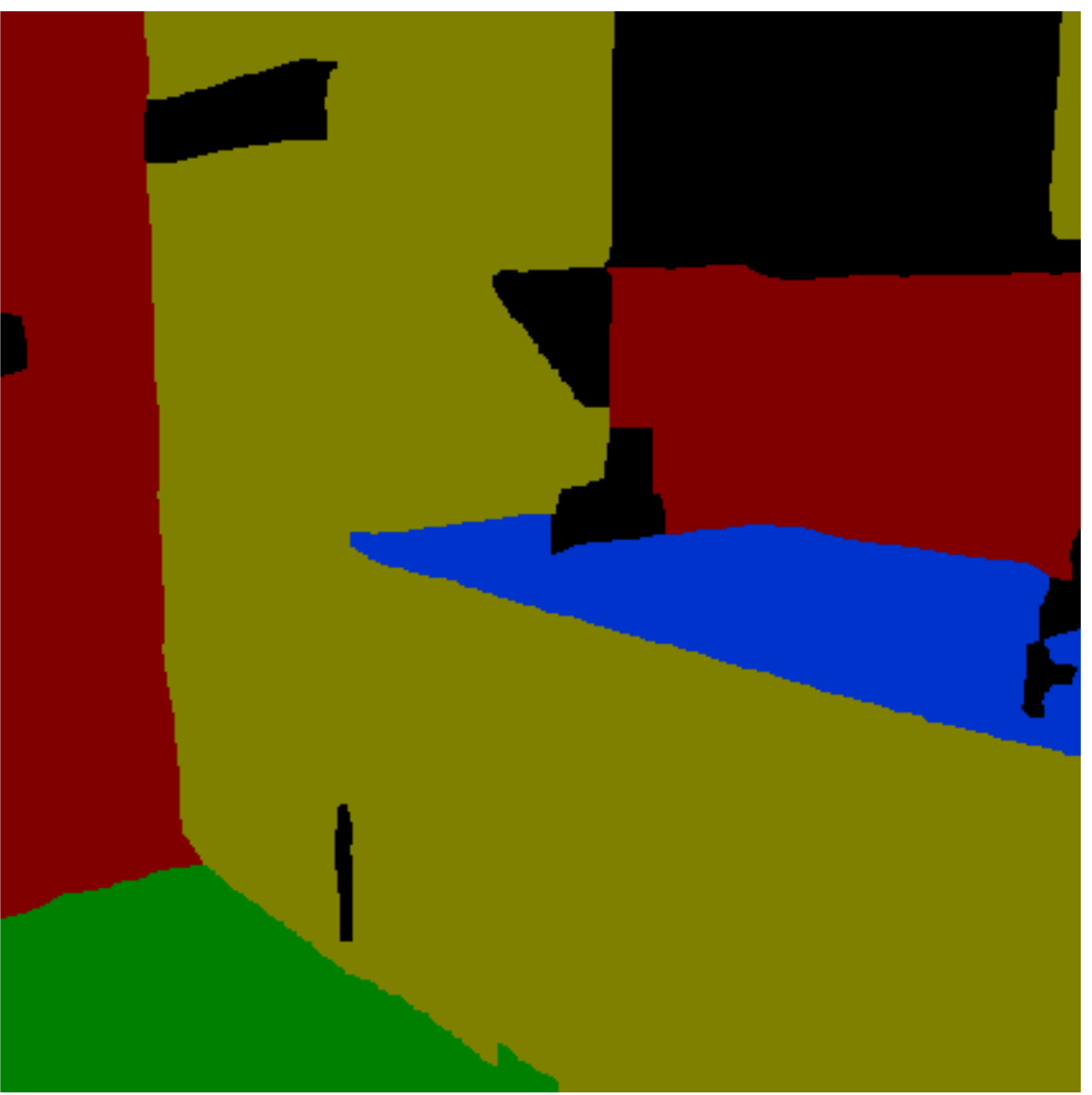} \\
\includegraphics[width=0.9\textwidth]{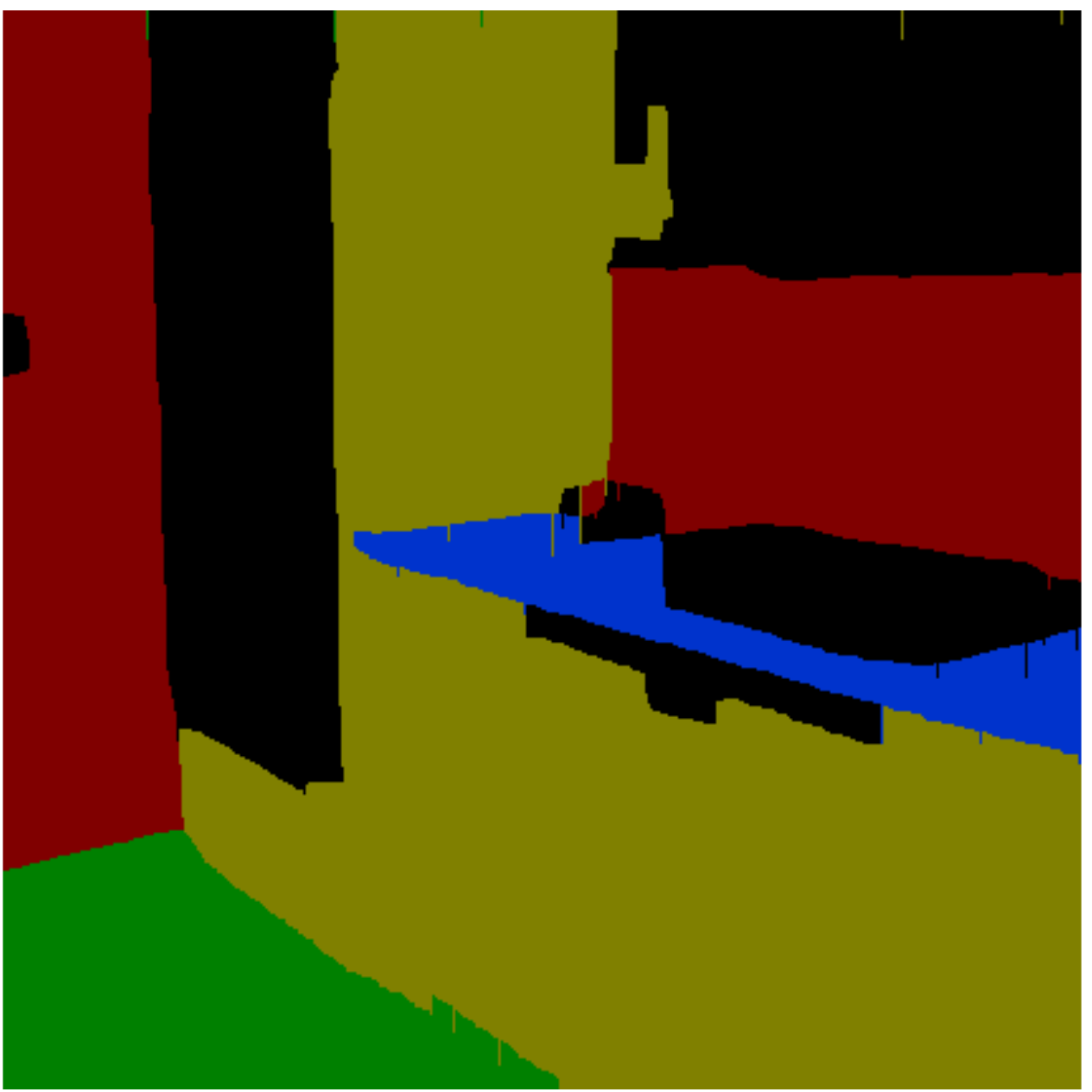}
\end{minipage}
}
\subfigure[]{
\begin{minipage}[b]{0.1\textwidth}
\includegraphics[width=0.9\textwidth]{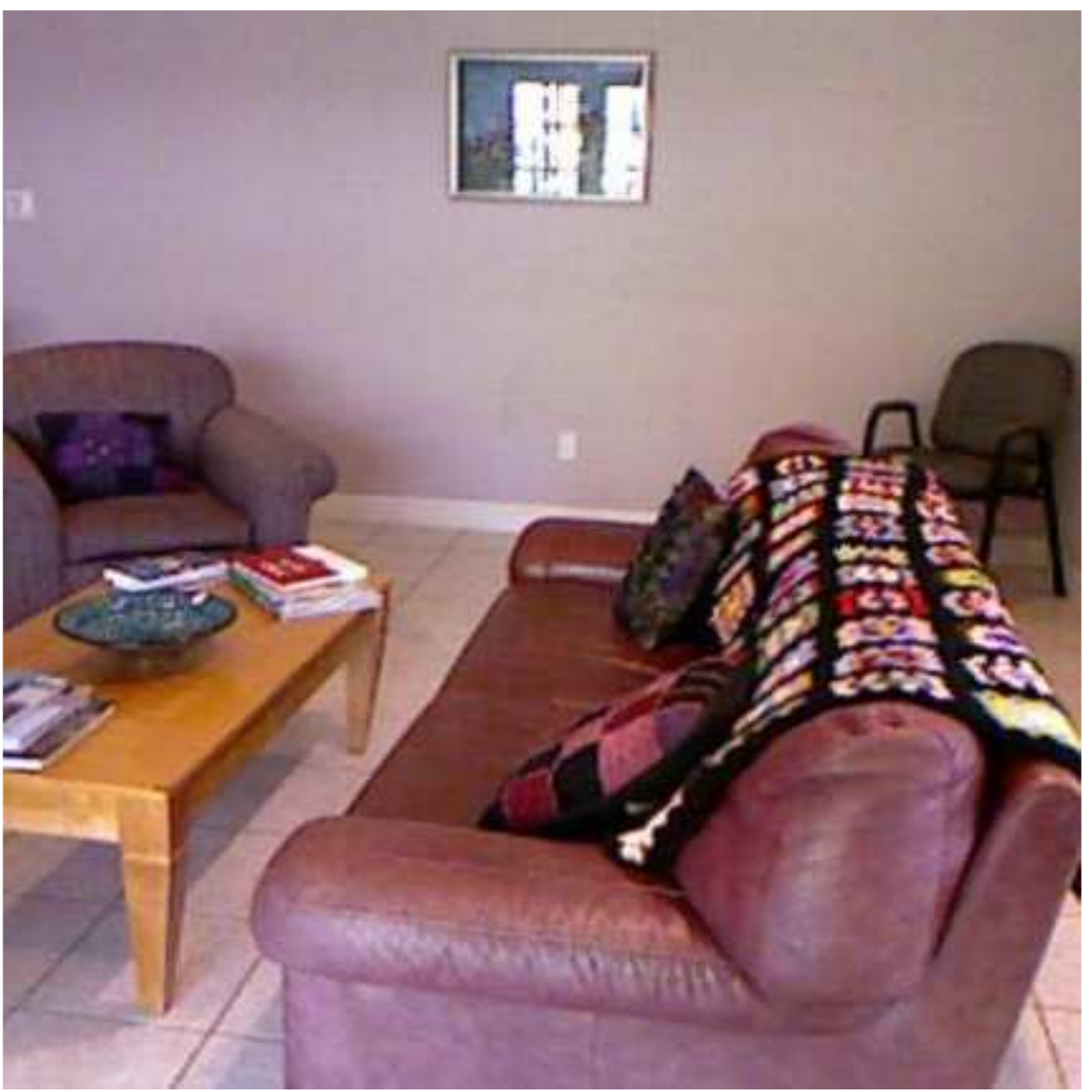} \\
\includegraphics[width=0.9\textwidth]{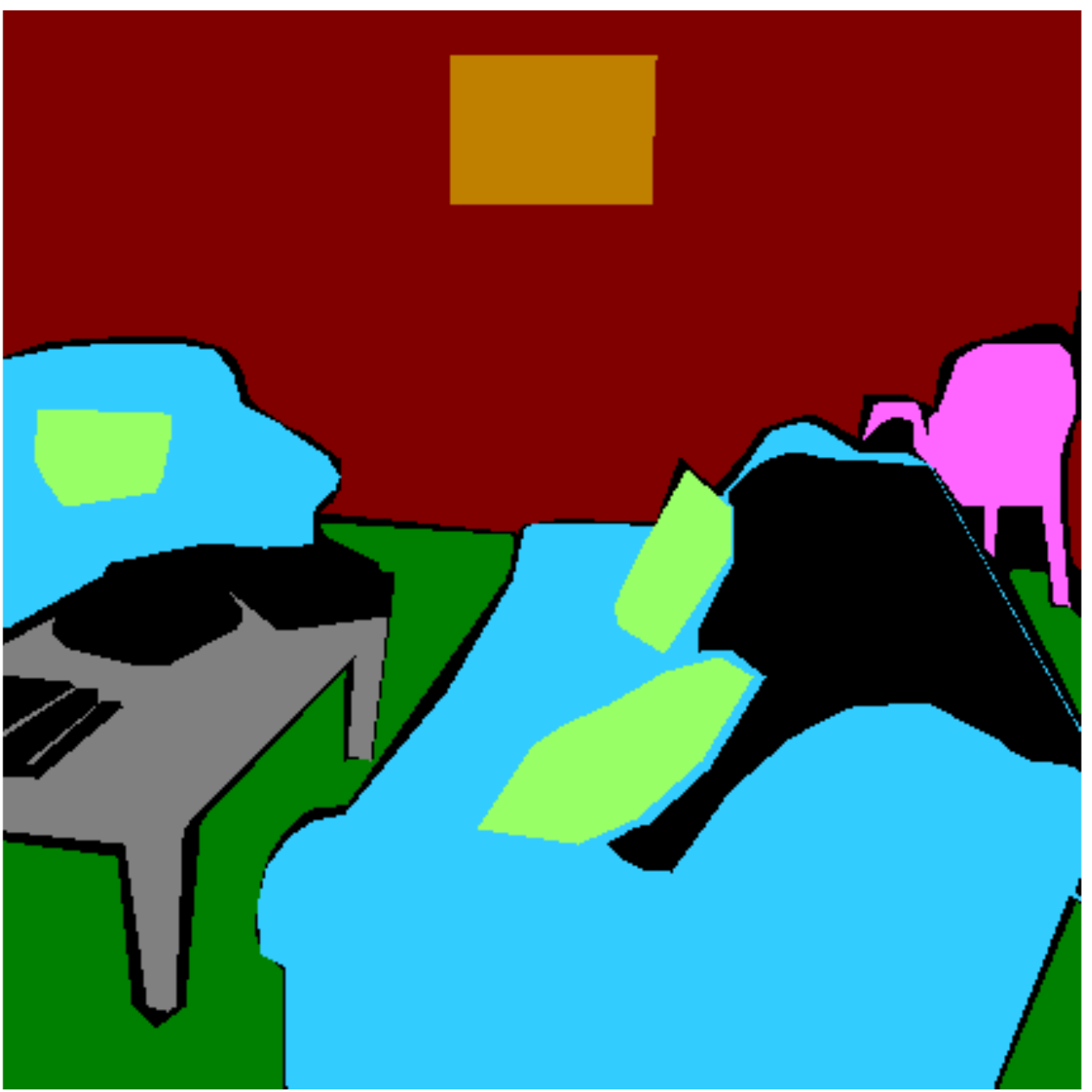} \\
\includegraphics[width=0.9\textwidth]{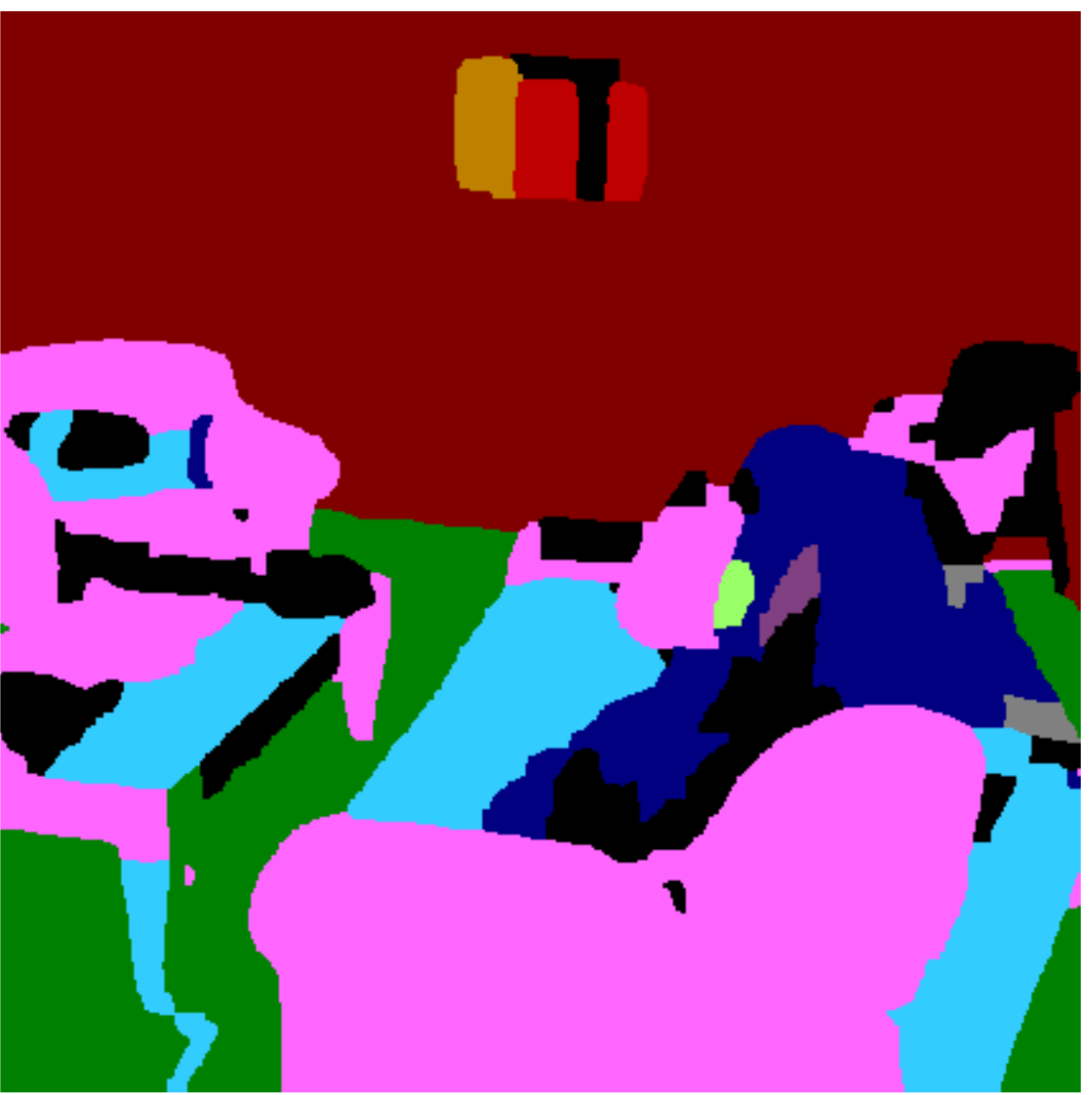} \\
\includegraphics[width=0.9\textwidth]{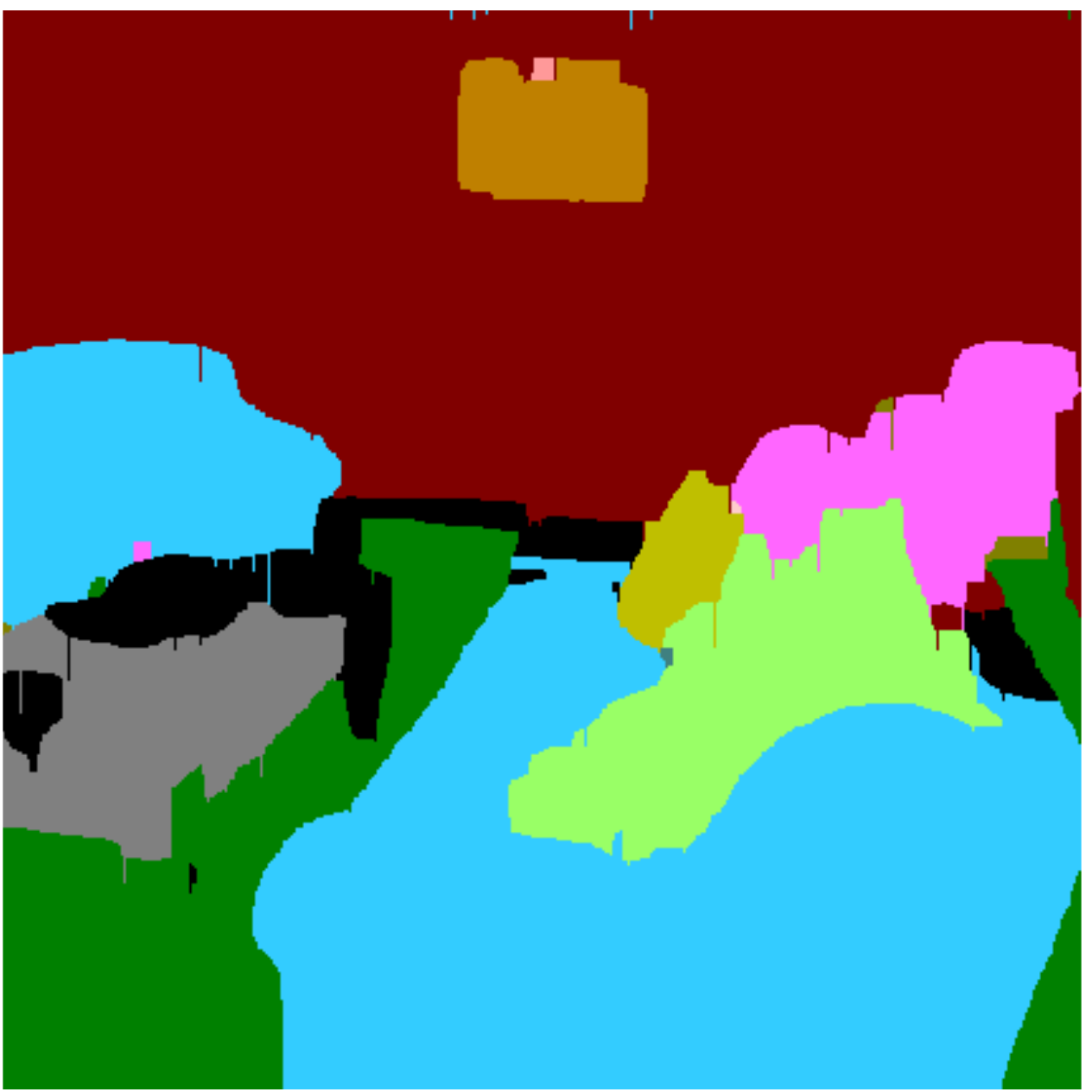}
\end{minipage}
}
\subfigure[]{
\begin{minipage}[b]{0.1\textwidth}
\includegraphics[width=0.9\textwidth]{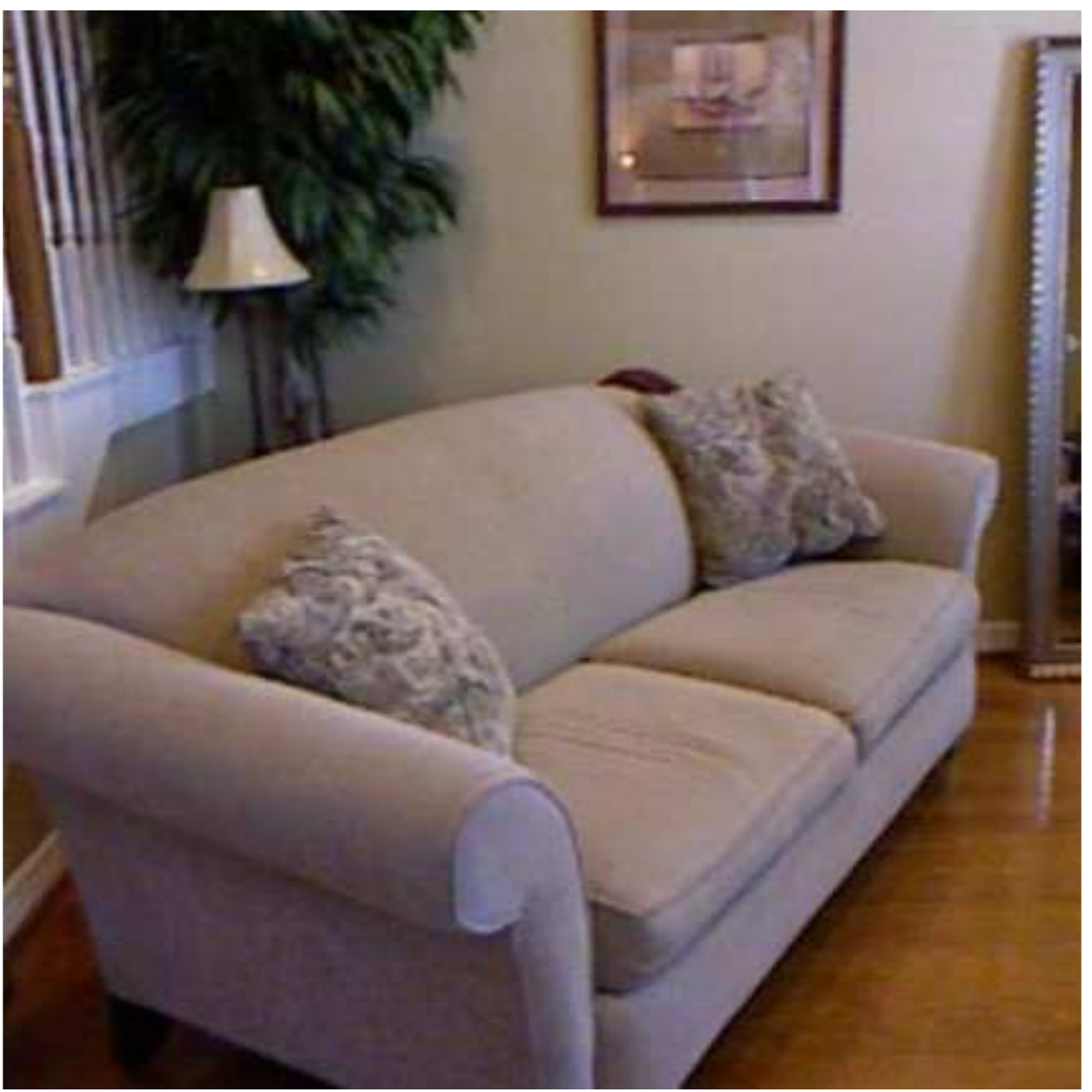} \\
\includegraphics[width=0.9\textwidth]{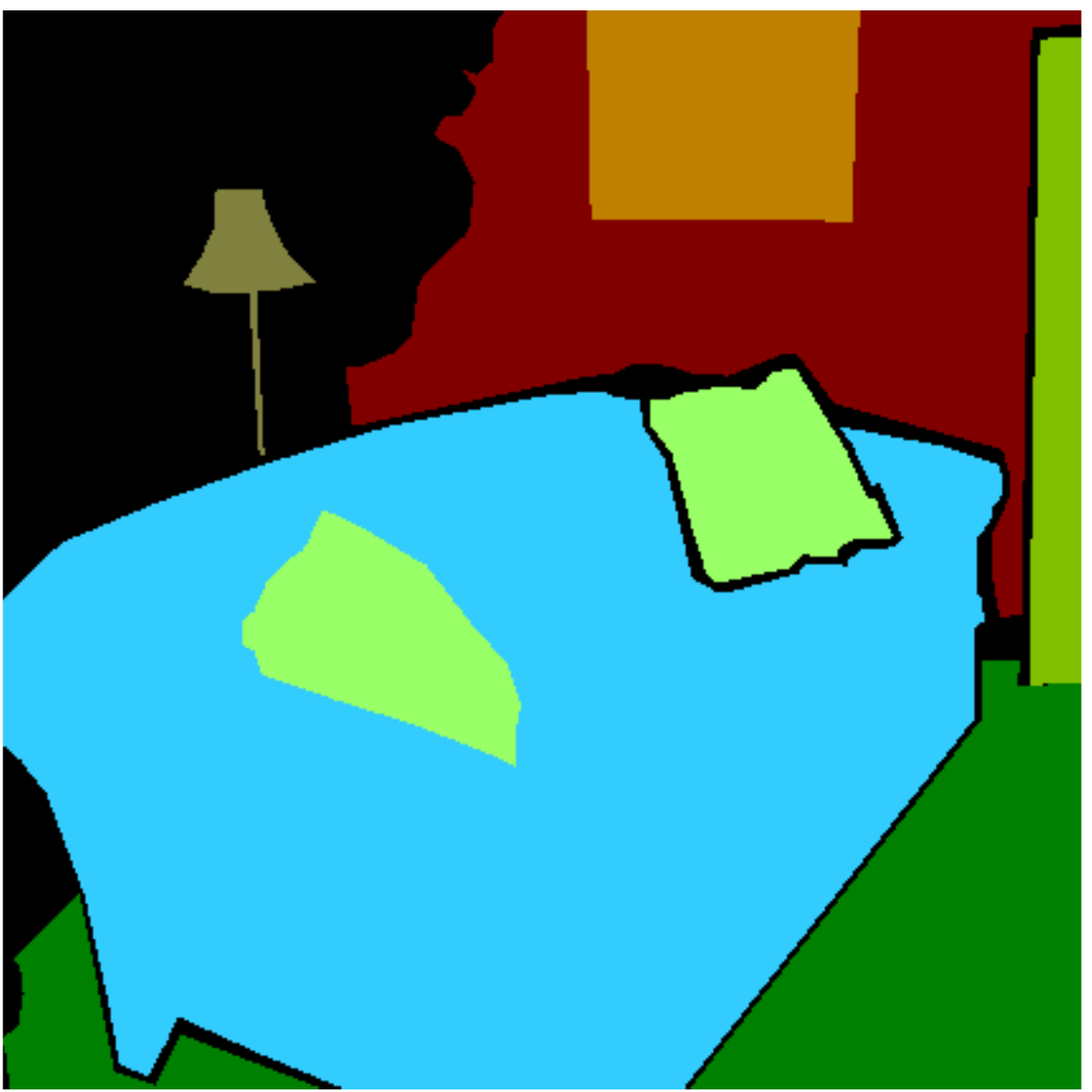} \\
\includegraphics[width=0.9\textwidth]{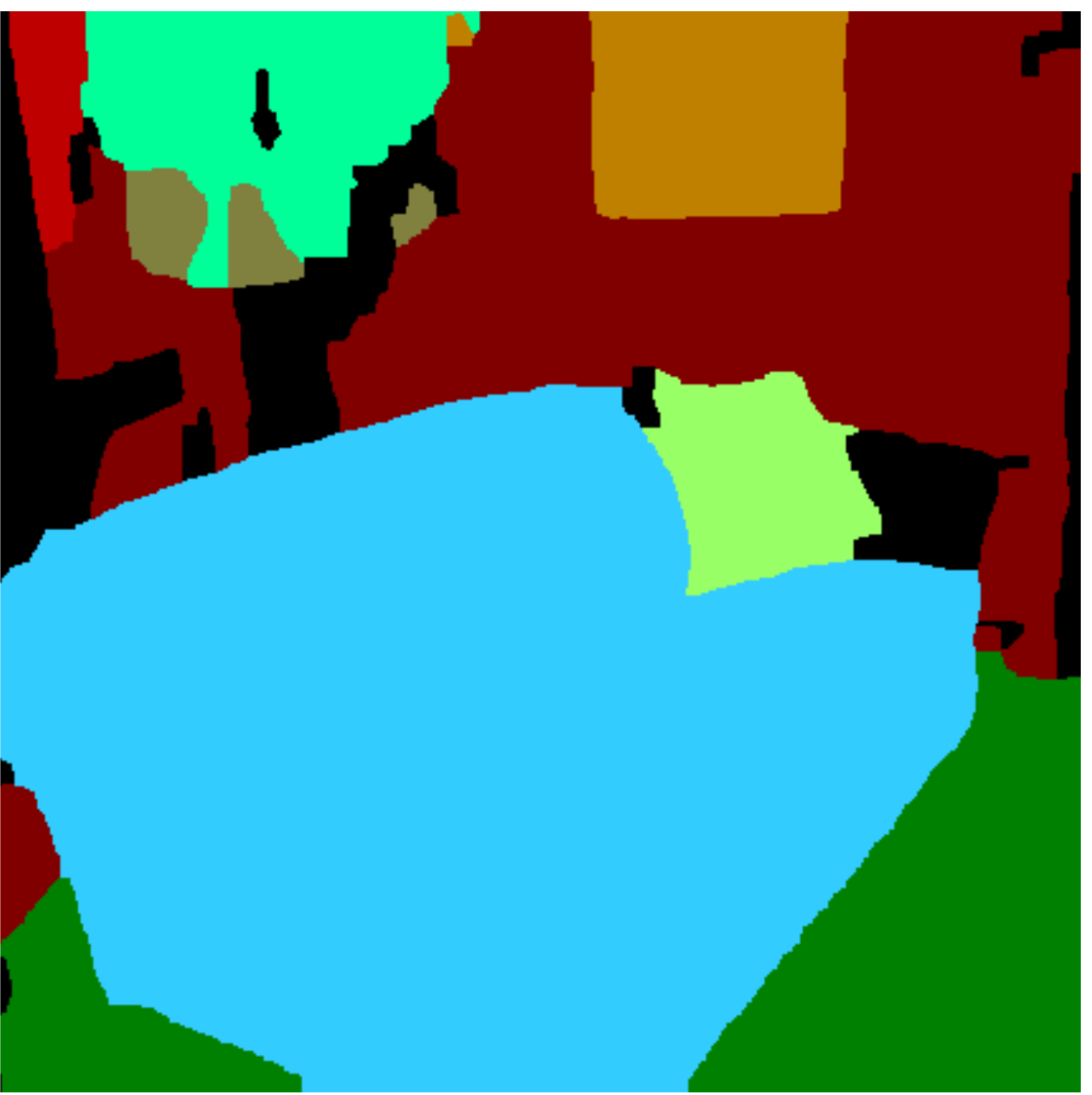} \\
\includegraphics[width=\textwidth]{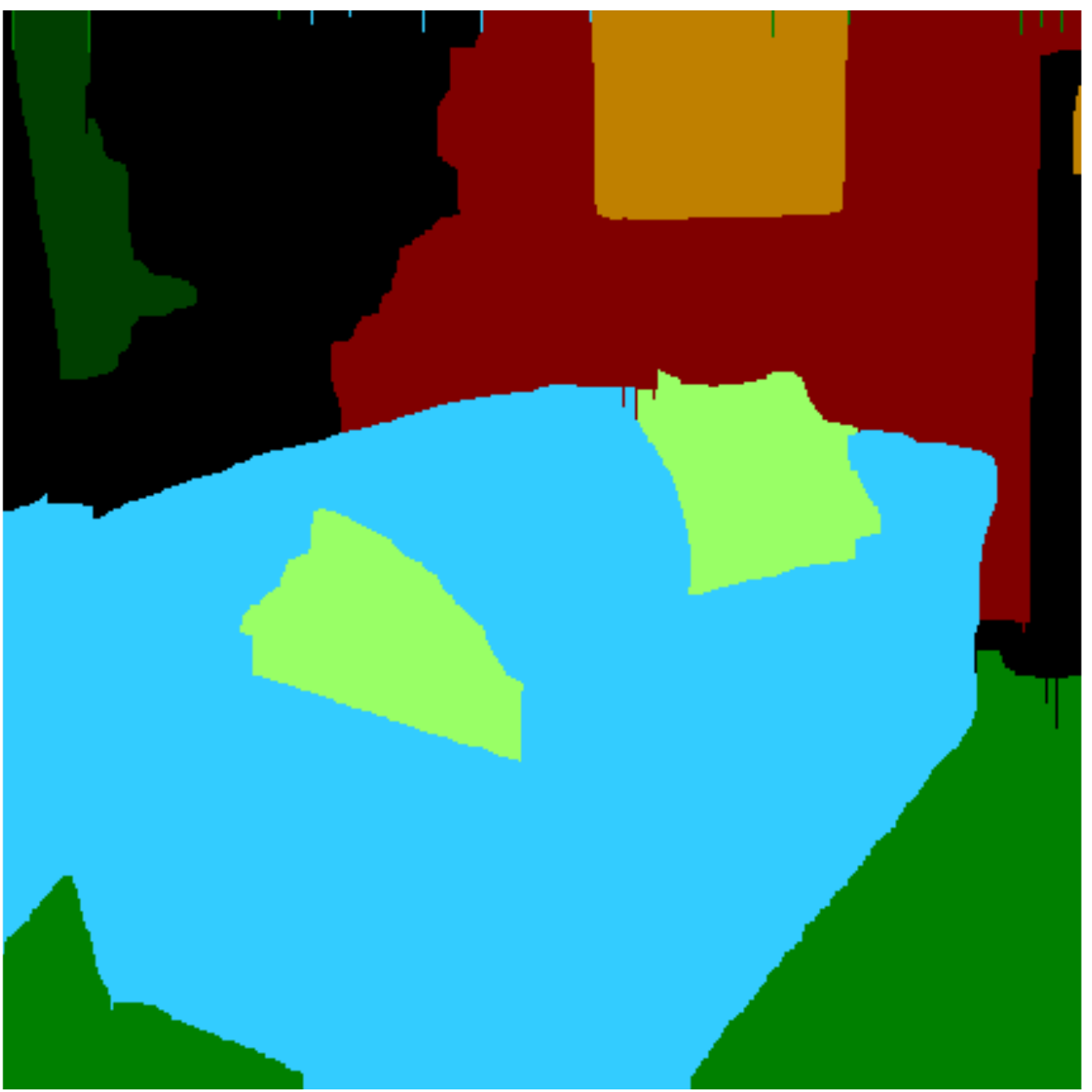}
\end{minipage}
}
\subfigure[]{
\begin{minipage}[b]{0.1\textwidth}
\includegraphics[width=0.9\textwidth]{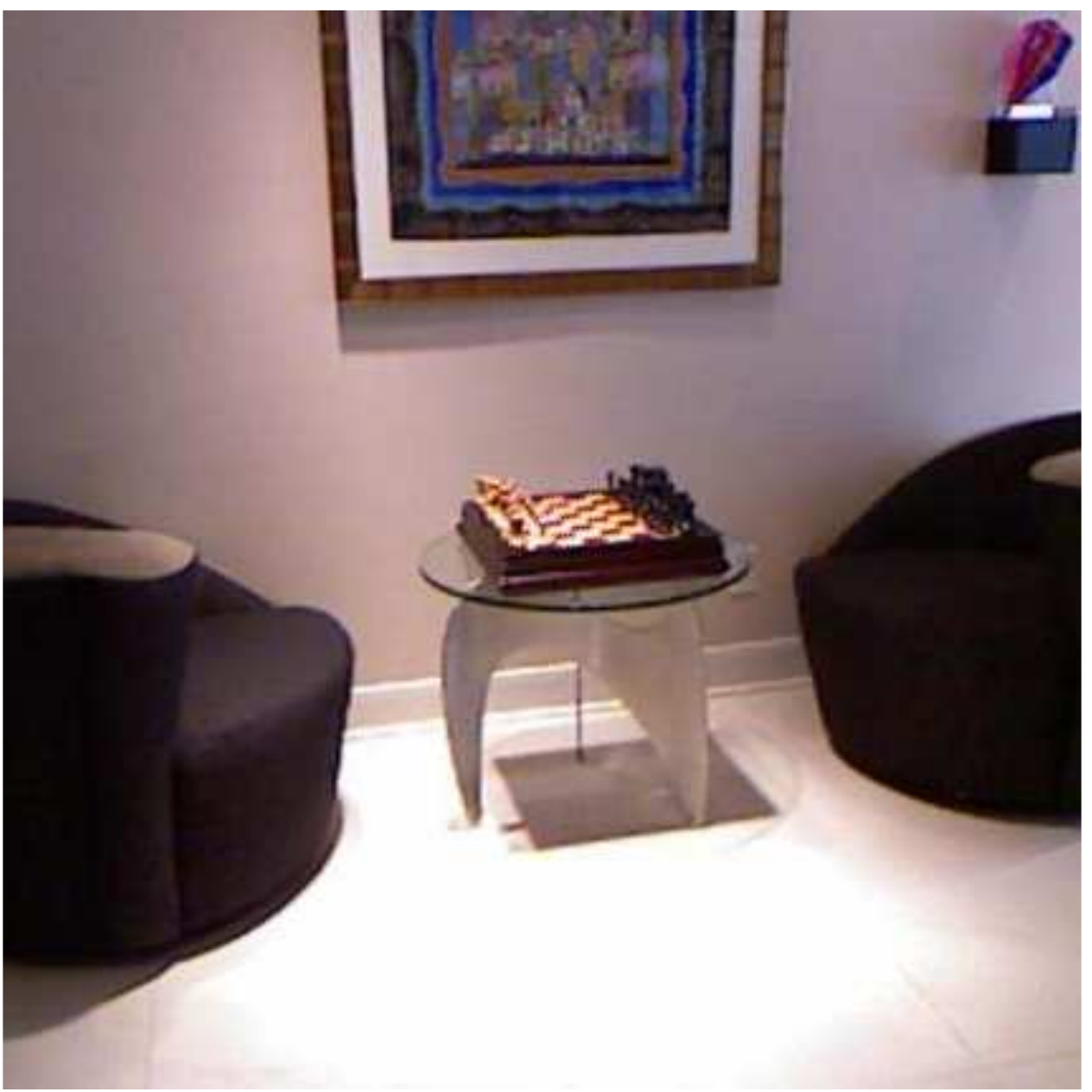} \\
\includegraphics[width=0.9\textwidth]{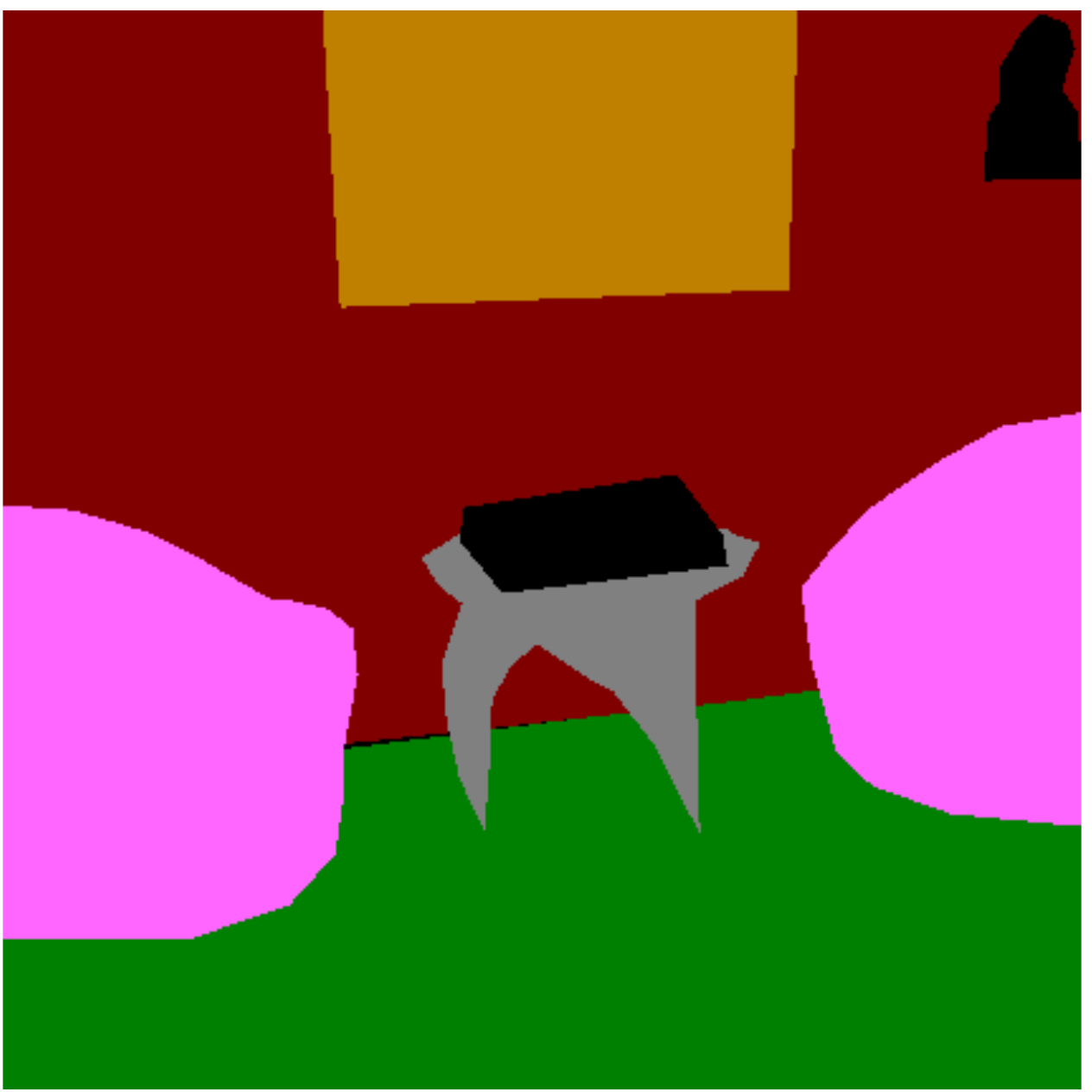} \\
\includegraphics[width=0.9\textwidth]{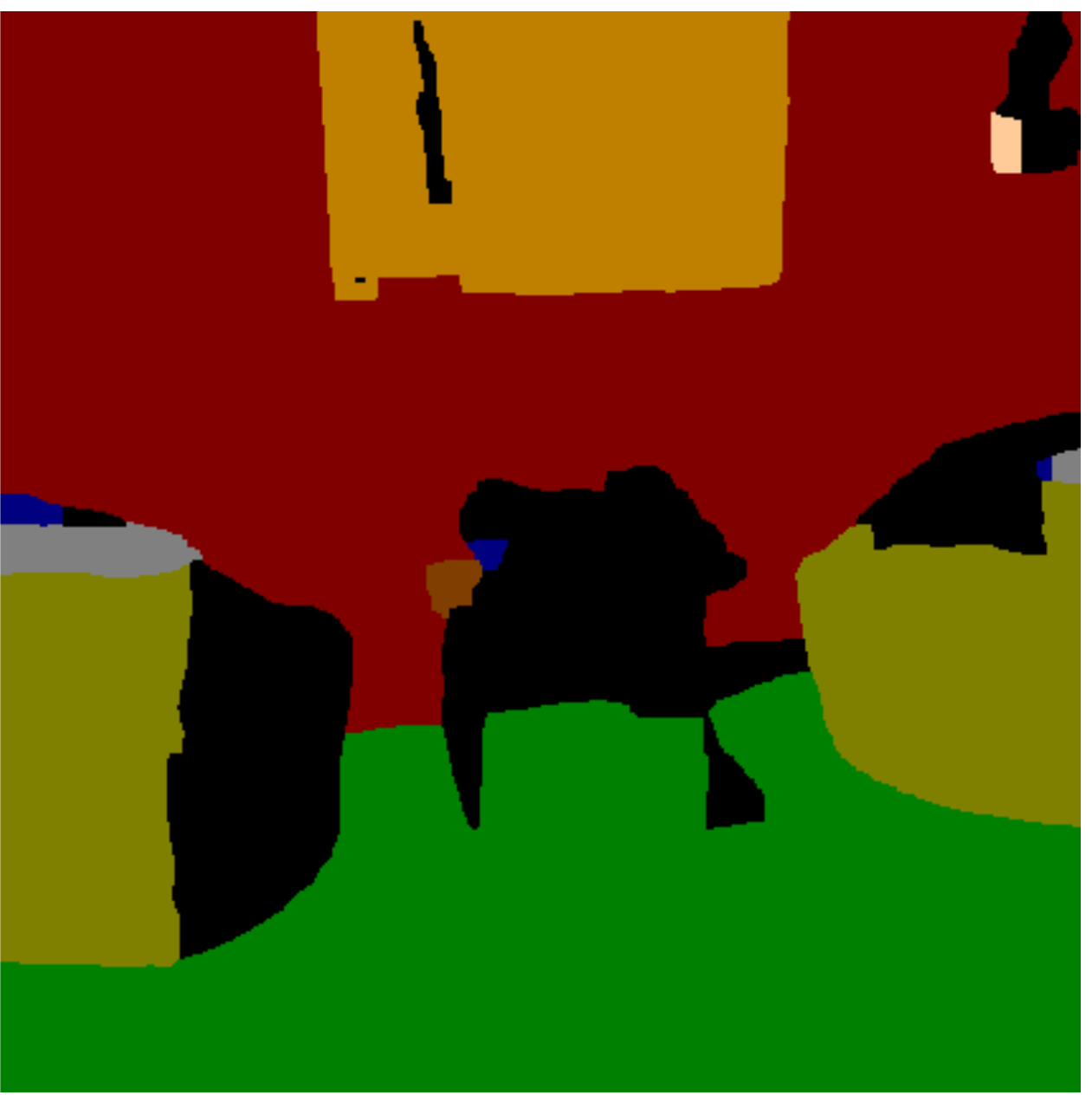} \\
\includegraphics[width=0.9\textwidth]{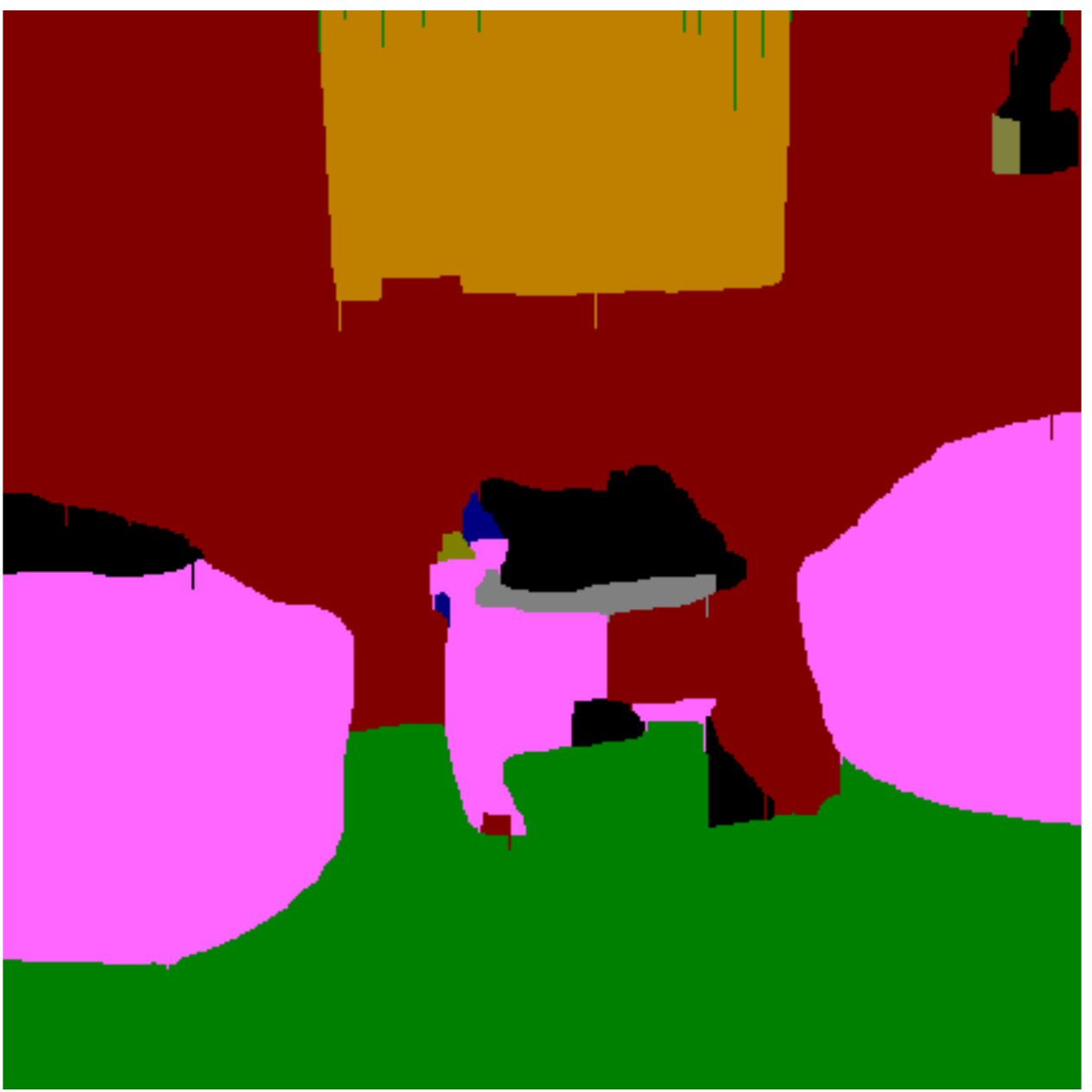}
\end{minipage}
}
\subfigure[]{
\begin{minipage}[b]{0.1\textwidth}
\includegraphics[width=0.9\textwidth]{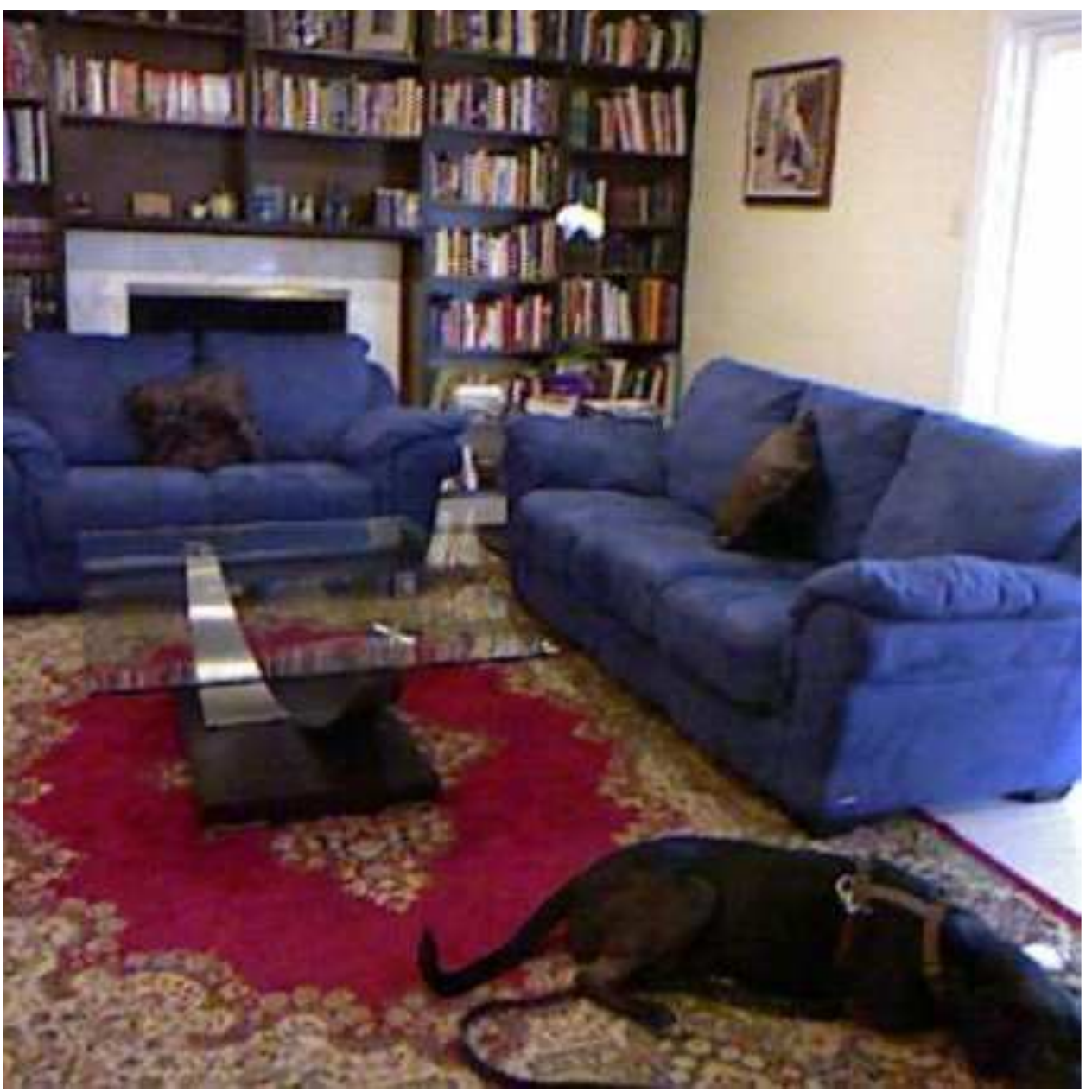} \\
\includegraphics[width=0.9\textwidth]{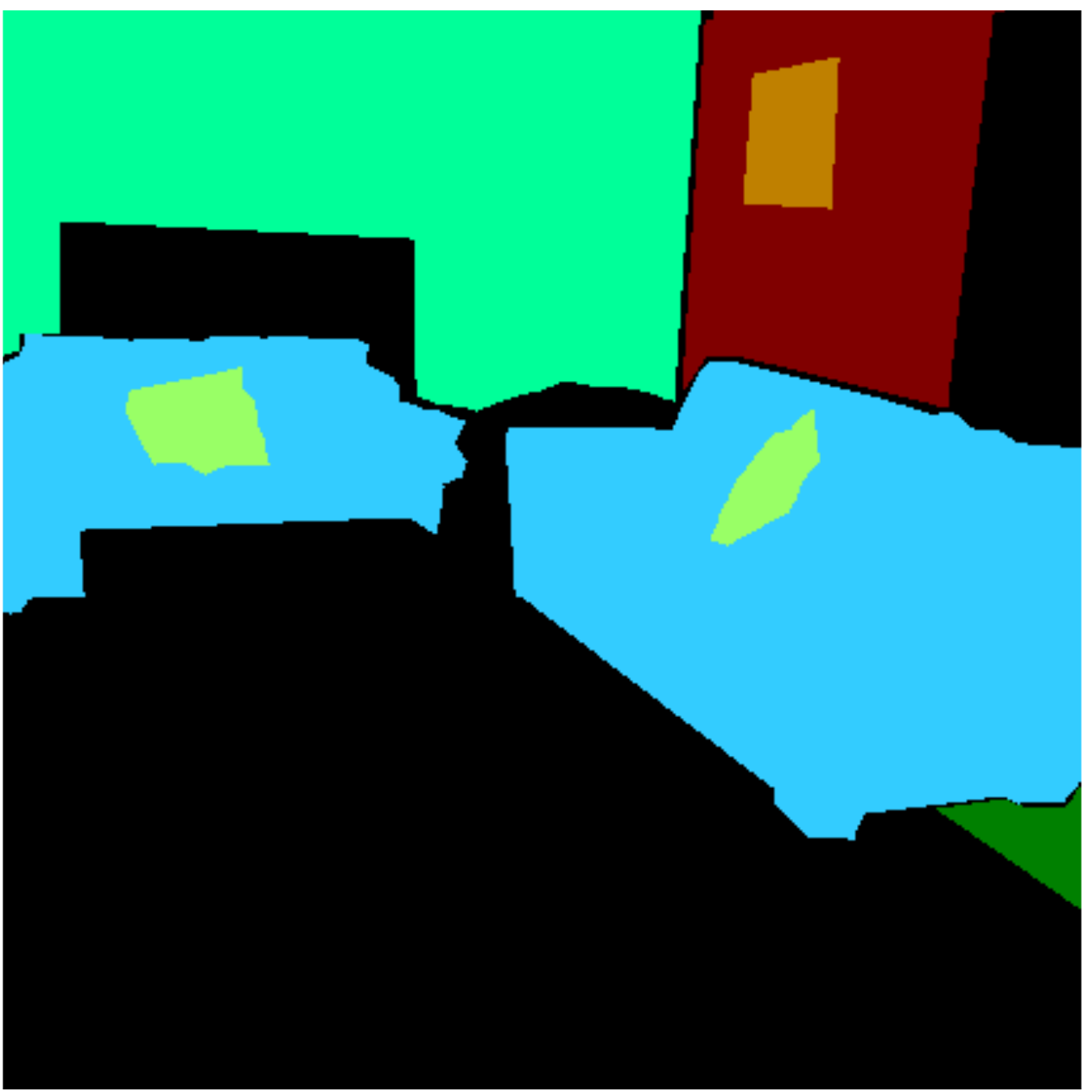} \\
\includegraphics[width=0.9\textwidth]{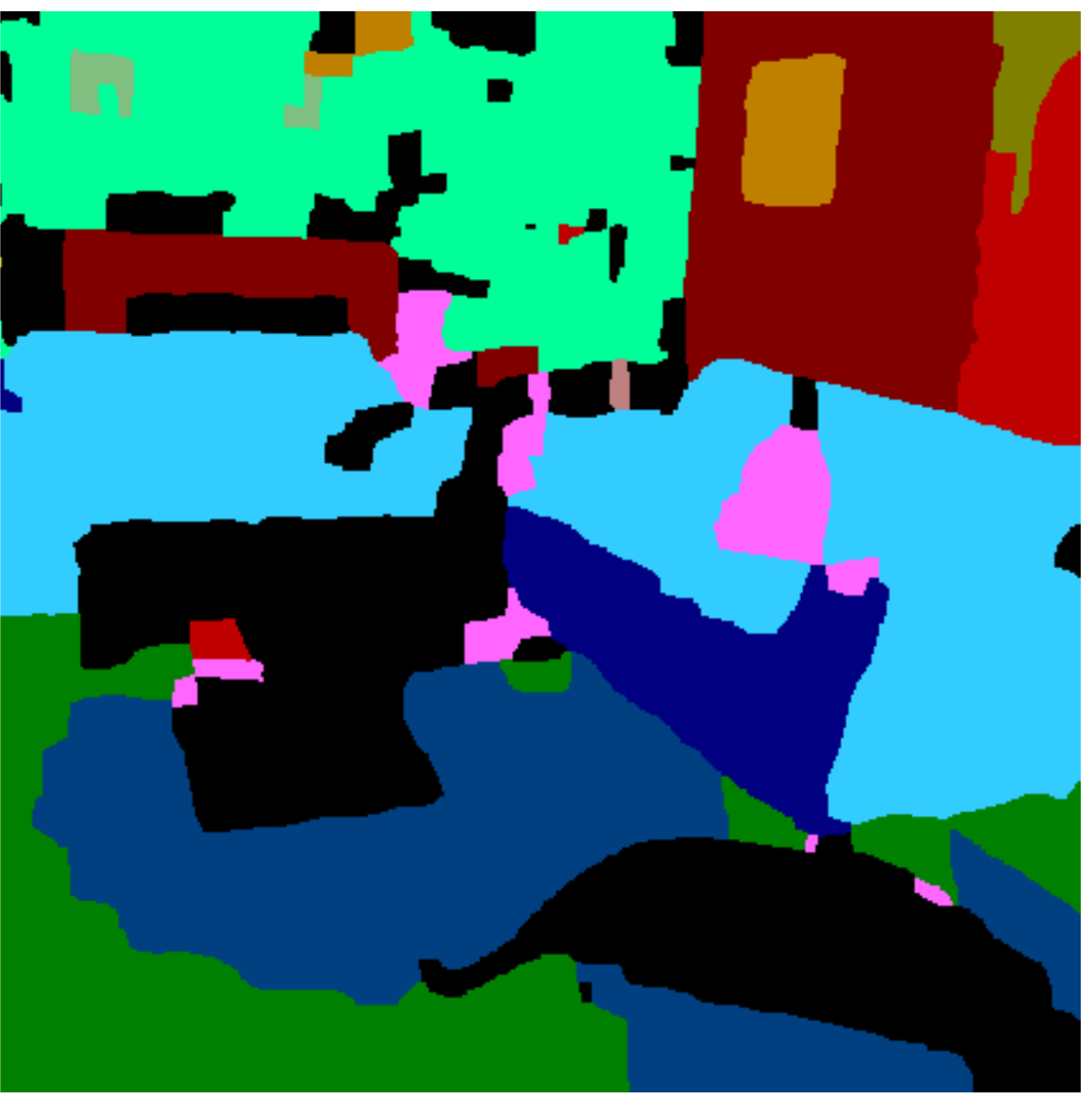} \\
\includegraphics[width=0.9\textwidth]{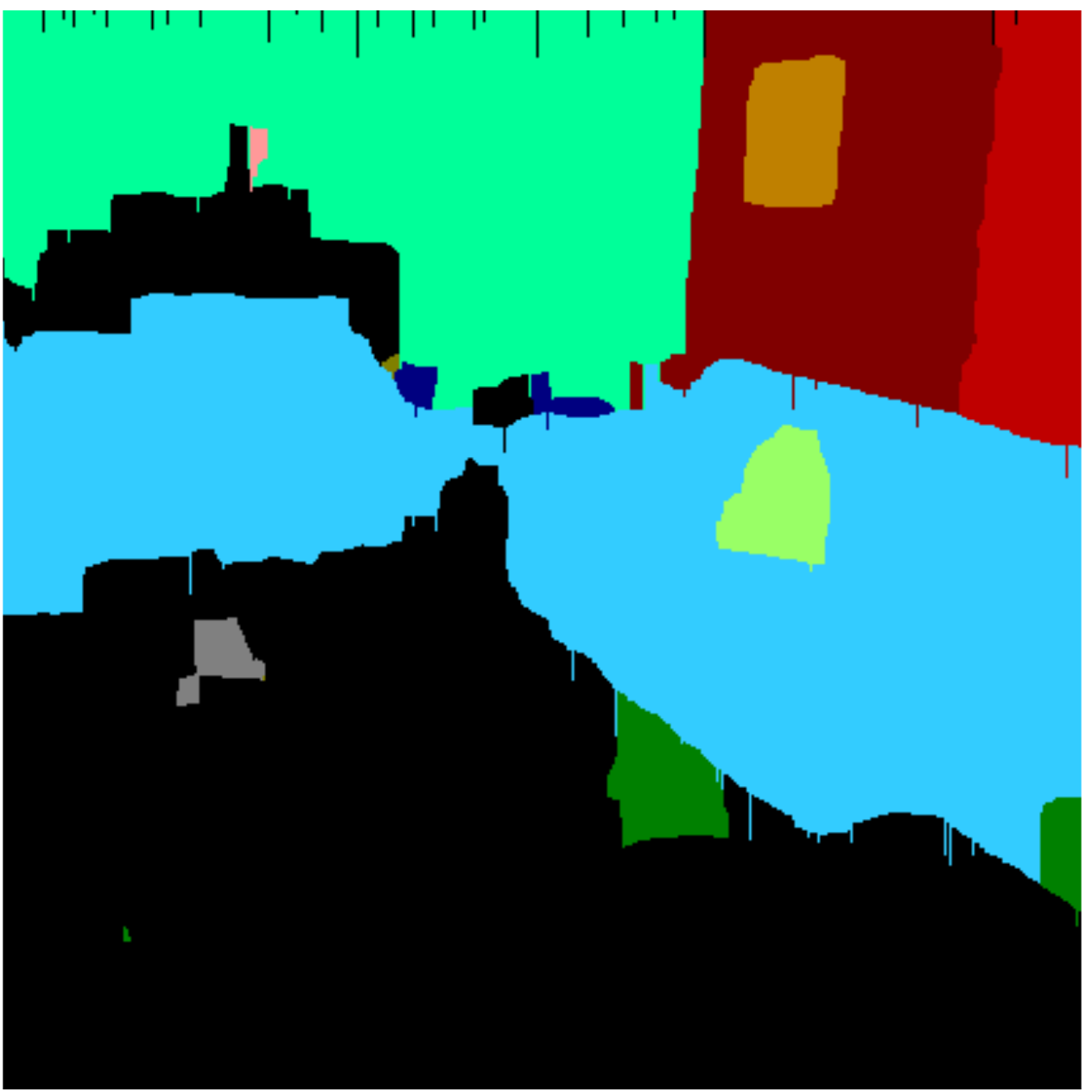}
\end{minipage}
}
\caption{Visual comparison of scene labeling results on the NYUDv2 dataset. The first and second rows show the input RGB images and their corresponding groundtruth labeling. The third row shows the results from \protect\cite{gupta2015indoor} and the last row shows the results from our model.}
\label{fig:comparison}
\end{figure}

\begin{figure}[t]
\centering
\subfigure[]{
\begin{minipage}[b]{0.1\textwidth}
\includegraphics[width=0.9\textwidth]{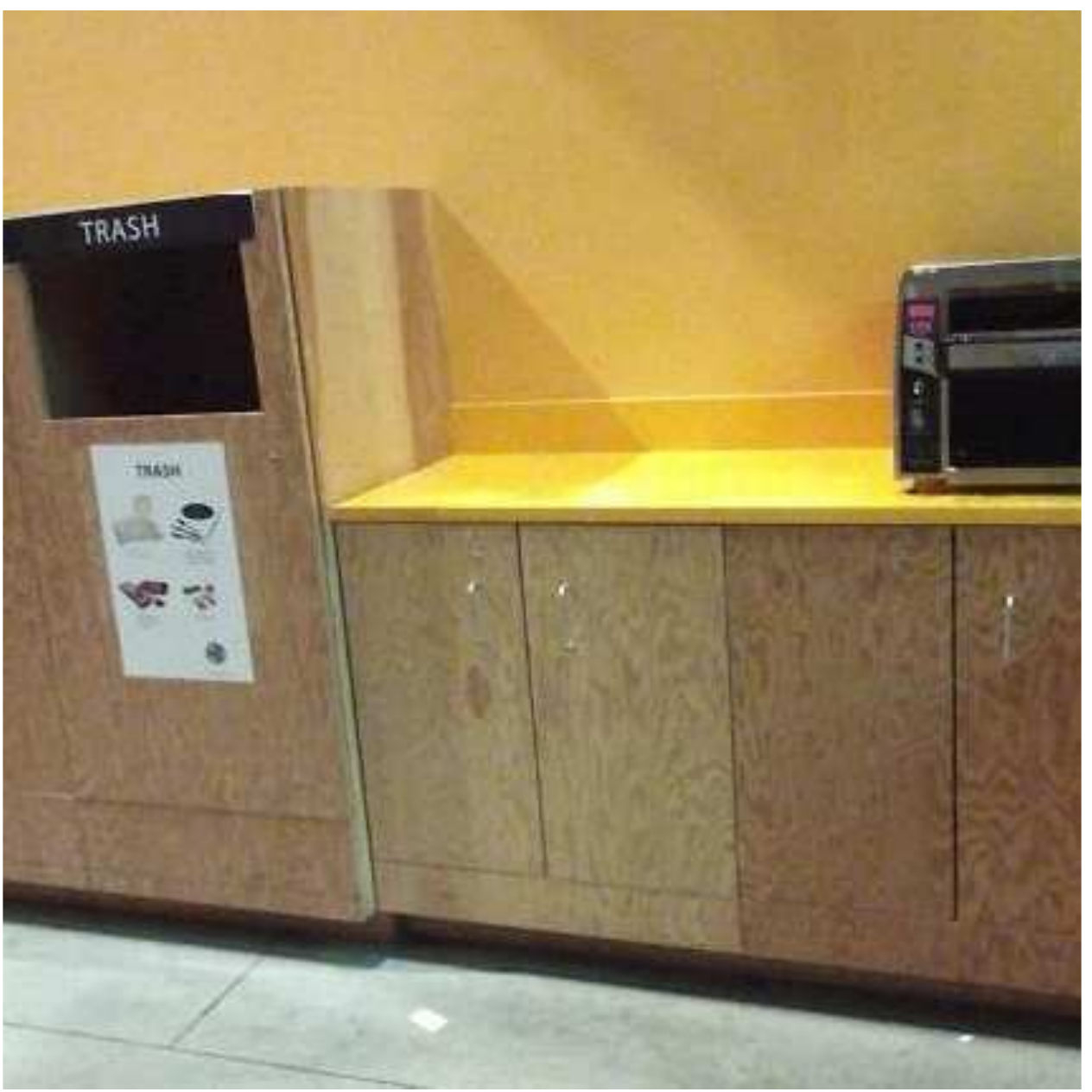} \\
\includegraphics[width=0.9\textwidth]{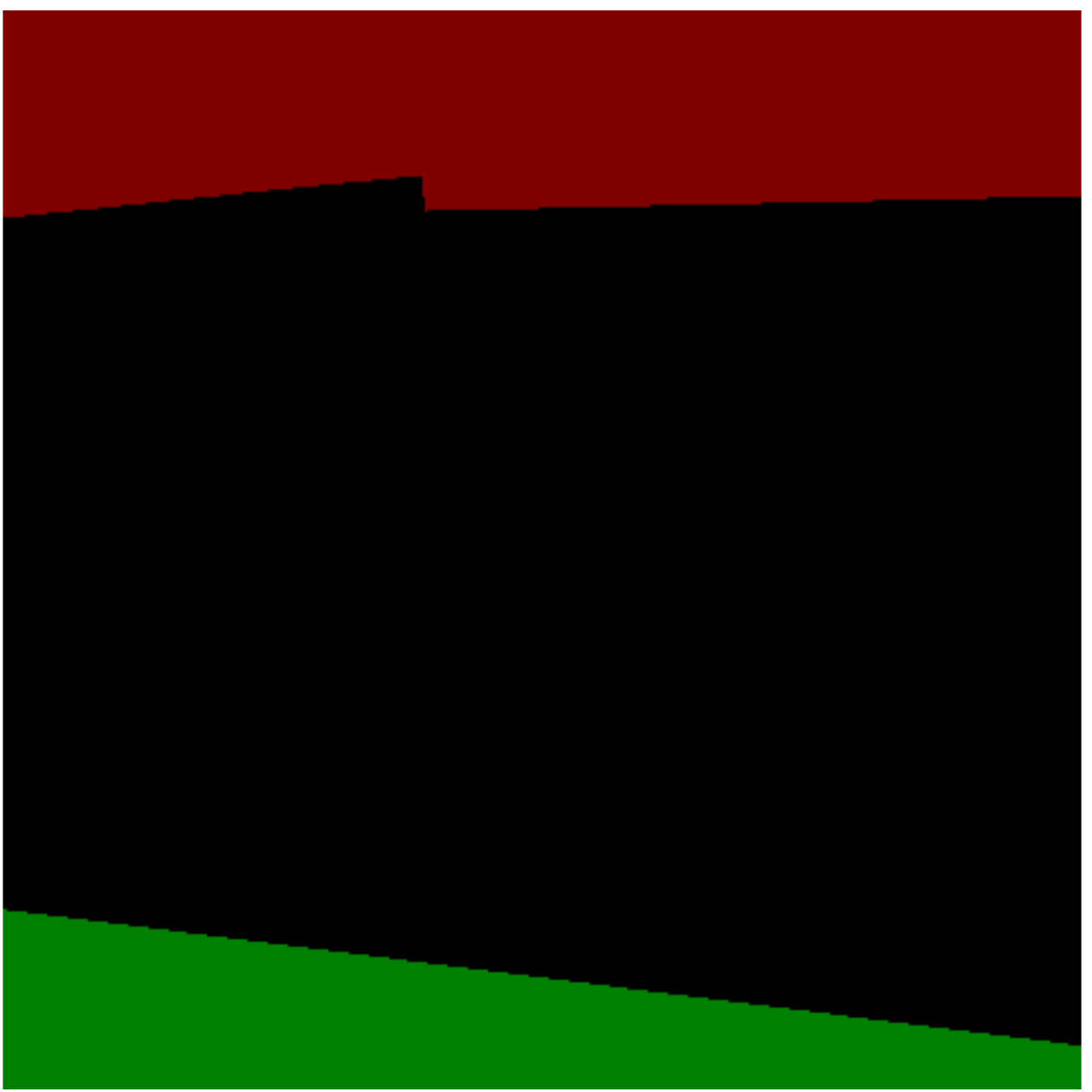} \\
\includegraphics[width=0.9\textwidth]{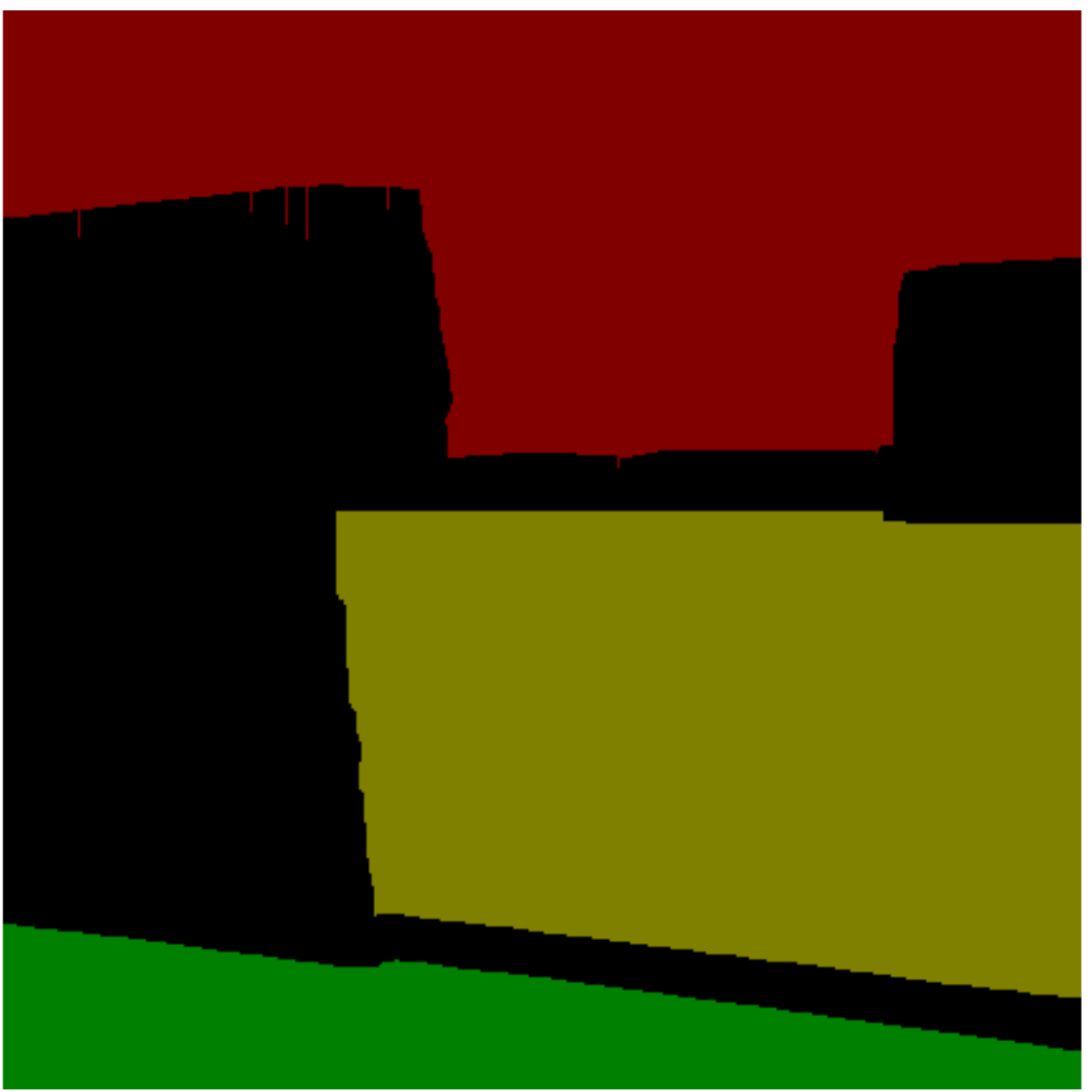}
\end{minipage}
}
\subfigure[]{
\begin{minipage}[b]{0.1\textwidth}
\includegraphics[width=0.9\textwidth]{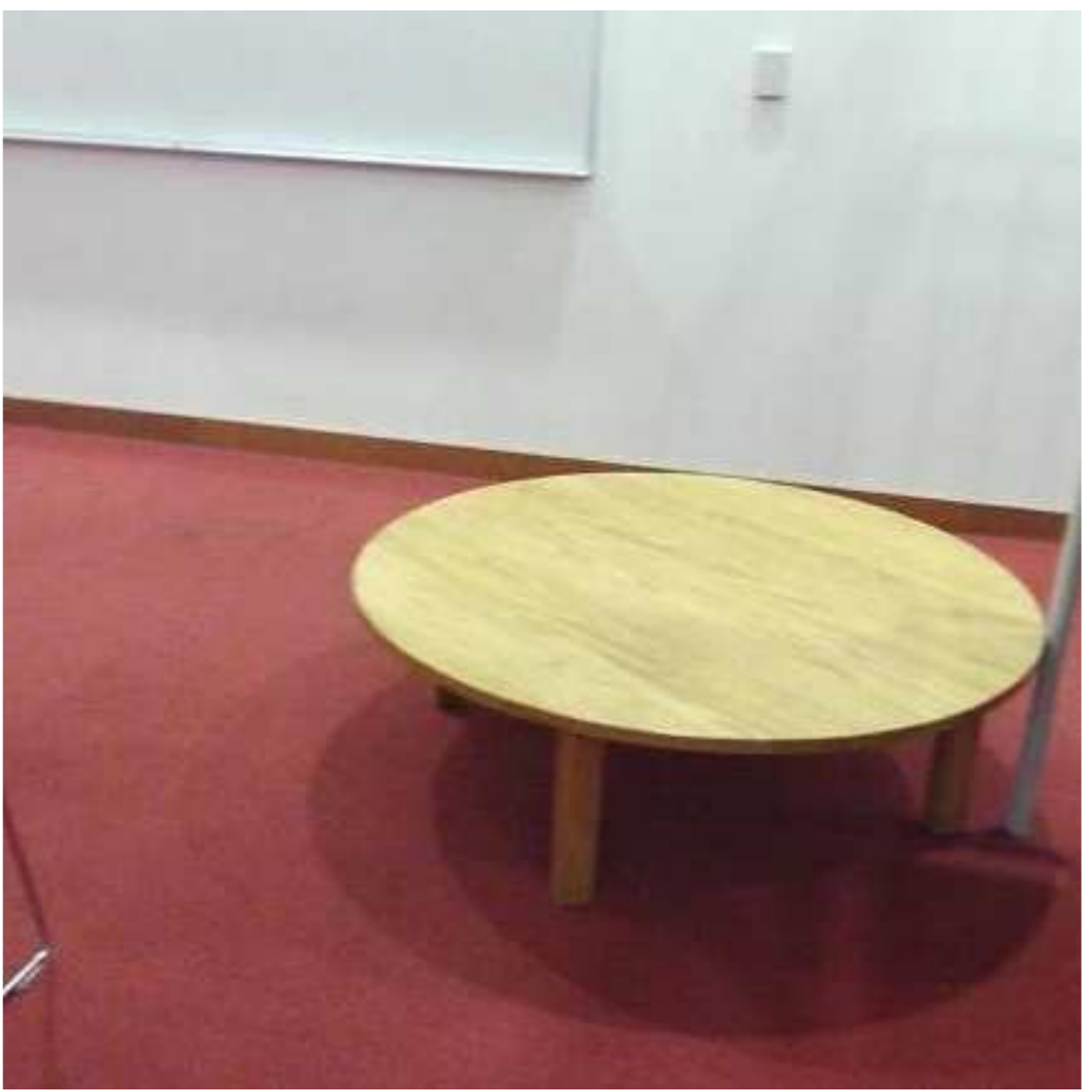} \\
\includegraphics[width=0.9\textwidth]{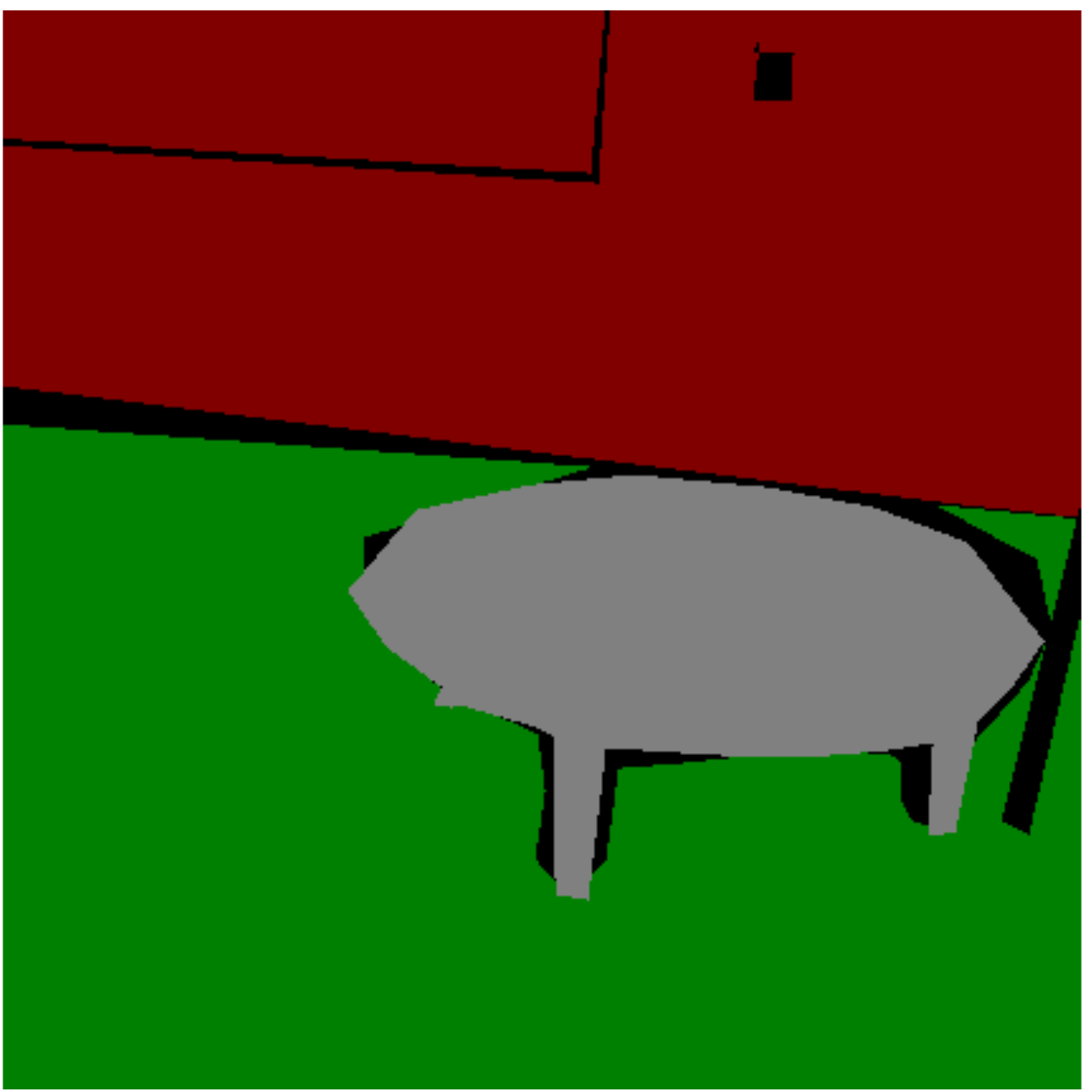} \\
\includegraphics[width=0.9\textwidth]{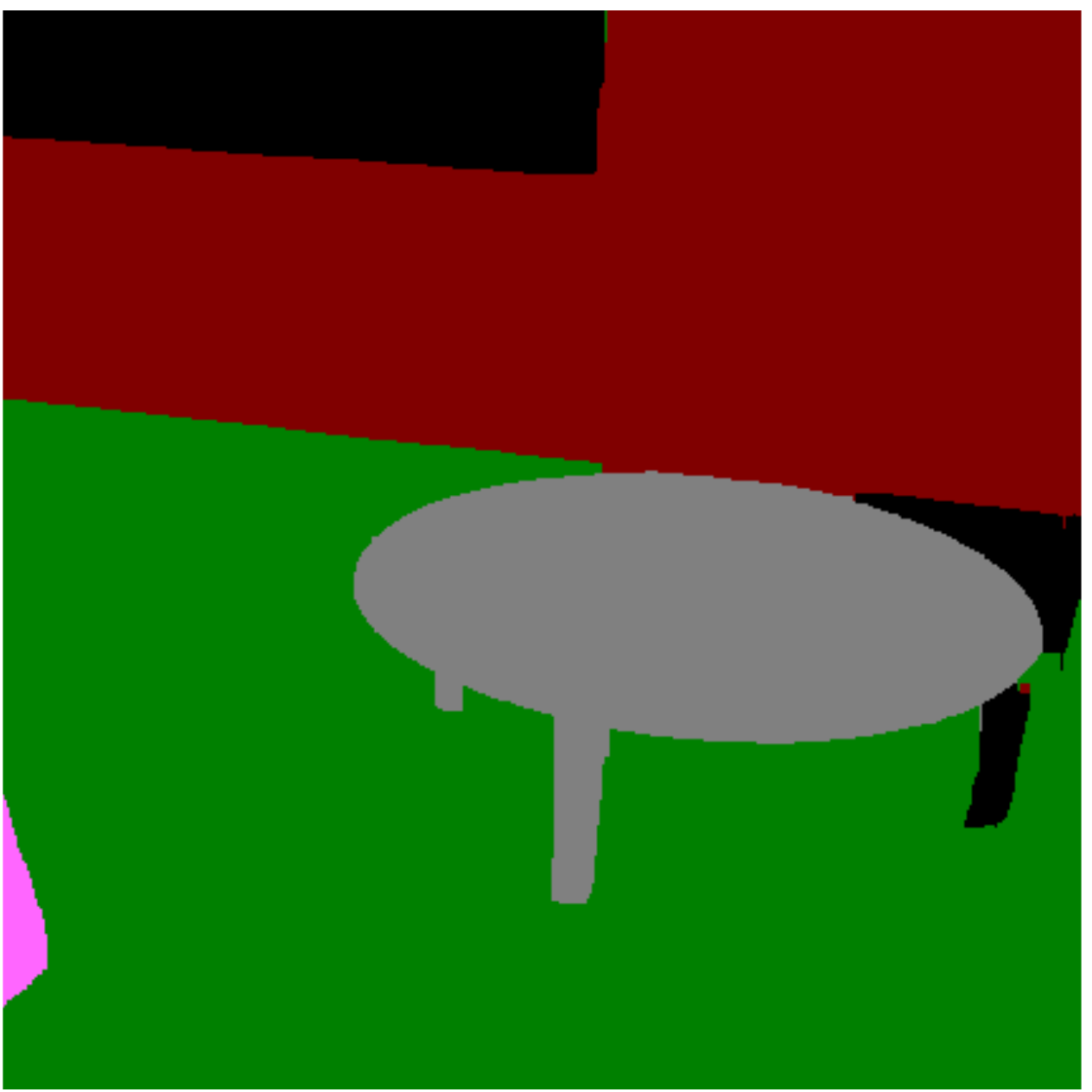}
\end{minipage}
}
\subfigure[]{
\begin{minipage}[b]{0.1\textwidth}
\includegraphics[width=0.9\textwidth]{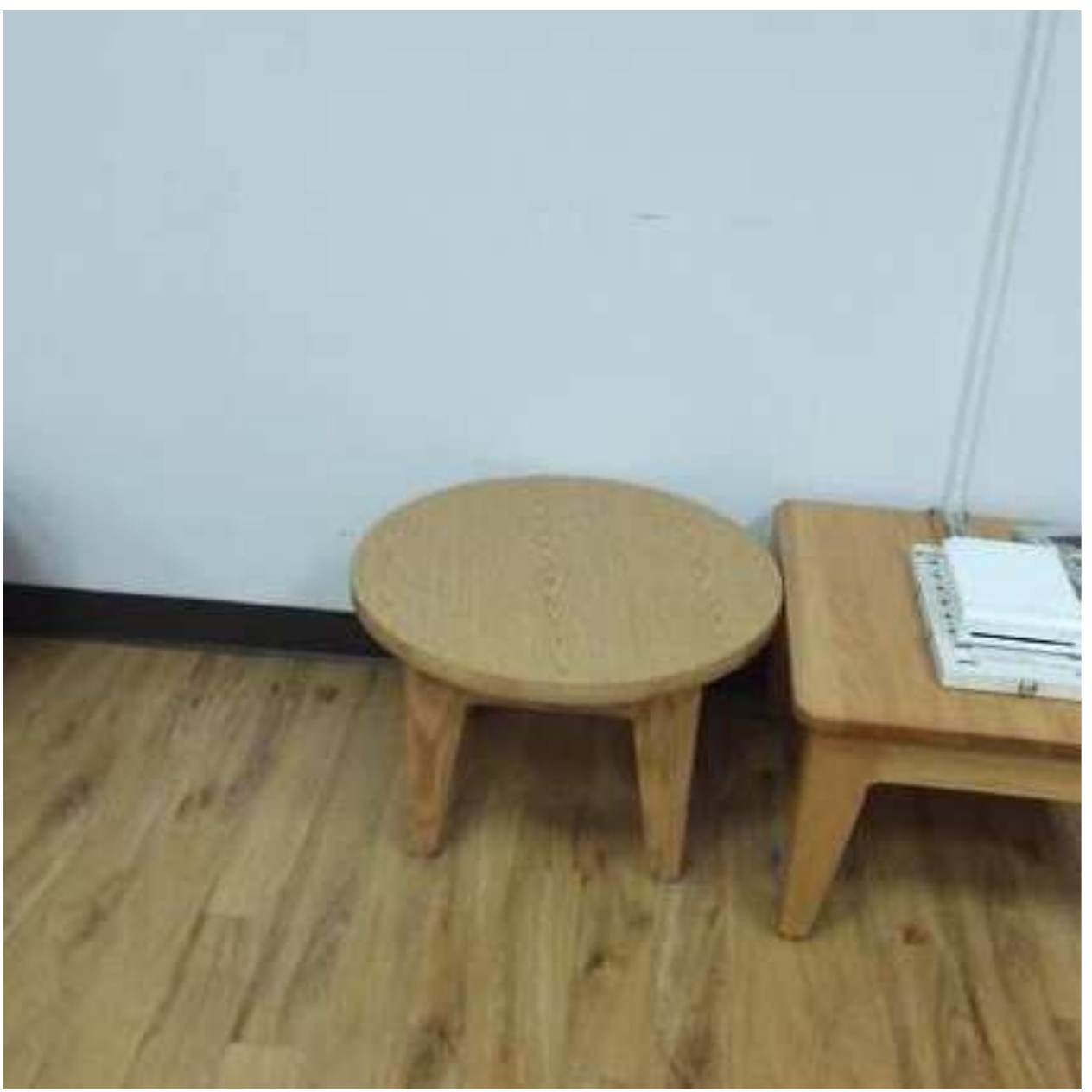} \\
\includegraphics[width=0.9\textwidth]{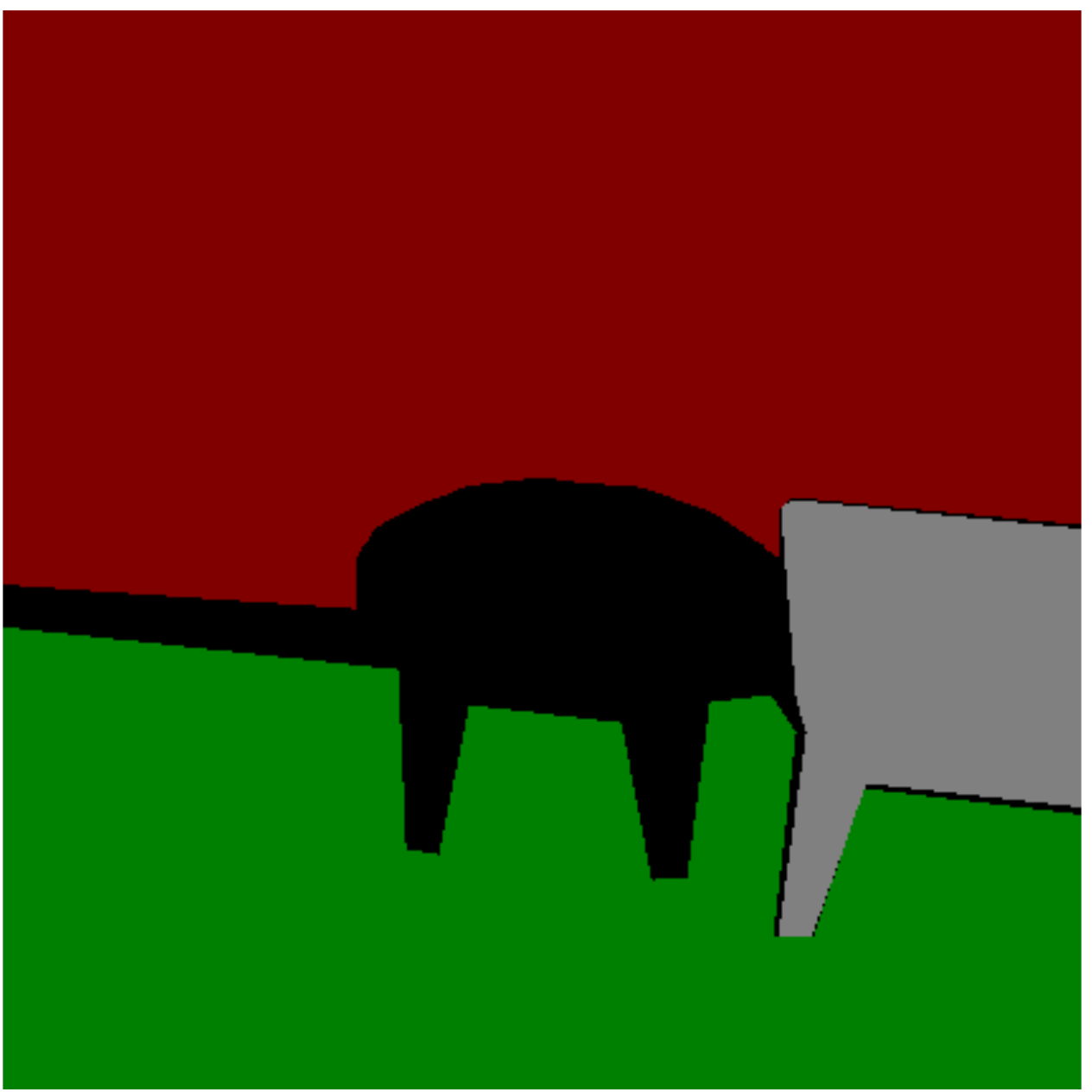} \\
\includegraphics[width=0.9\textwidth]{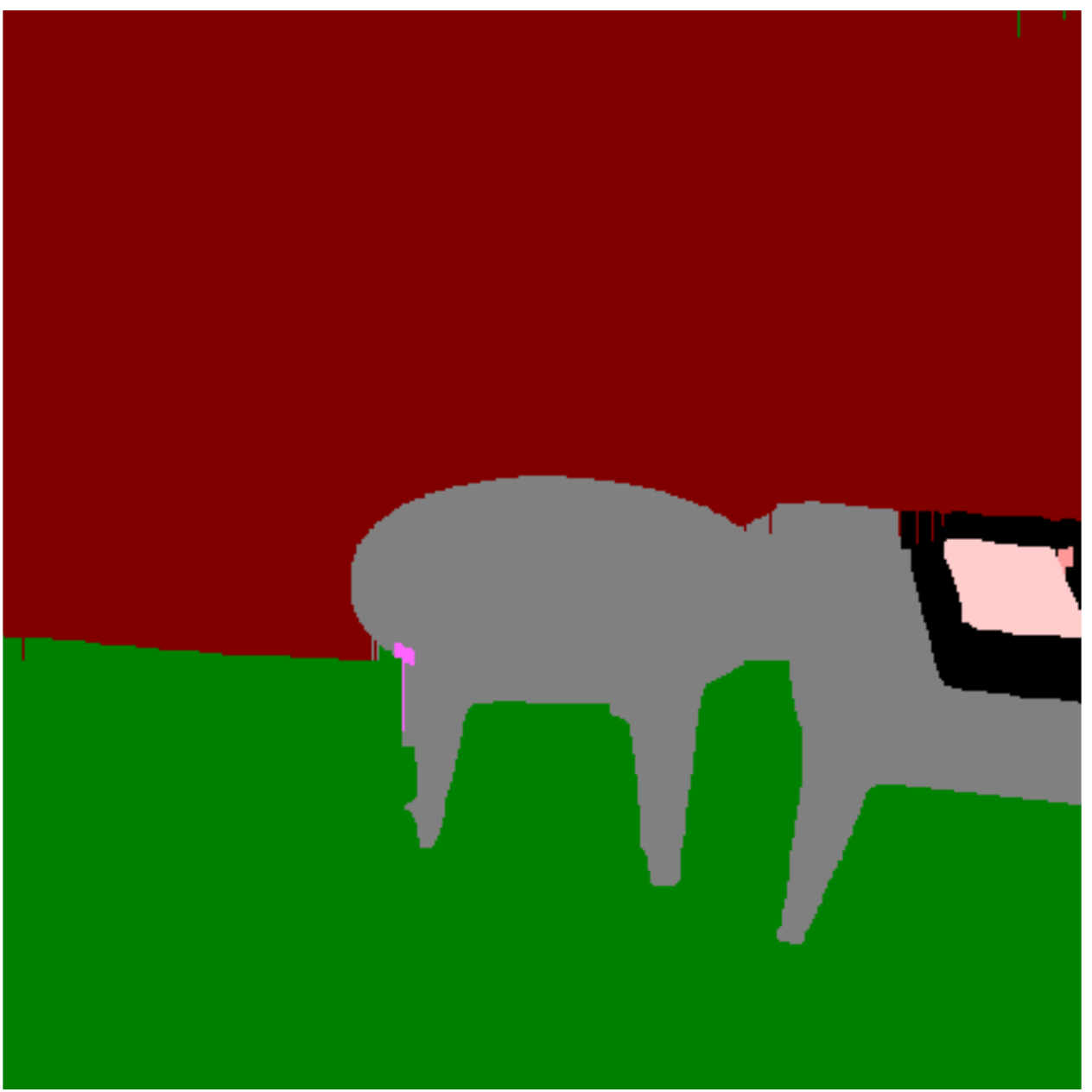}
\end{minipage}
}
\subfigure[]{
\begin{minipage}[b]{0.1\textwidth}
\includegraphics[width=0.9\textwidth]{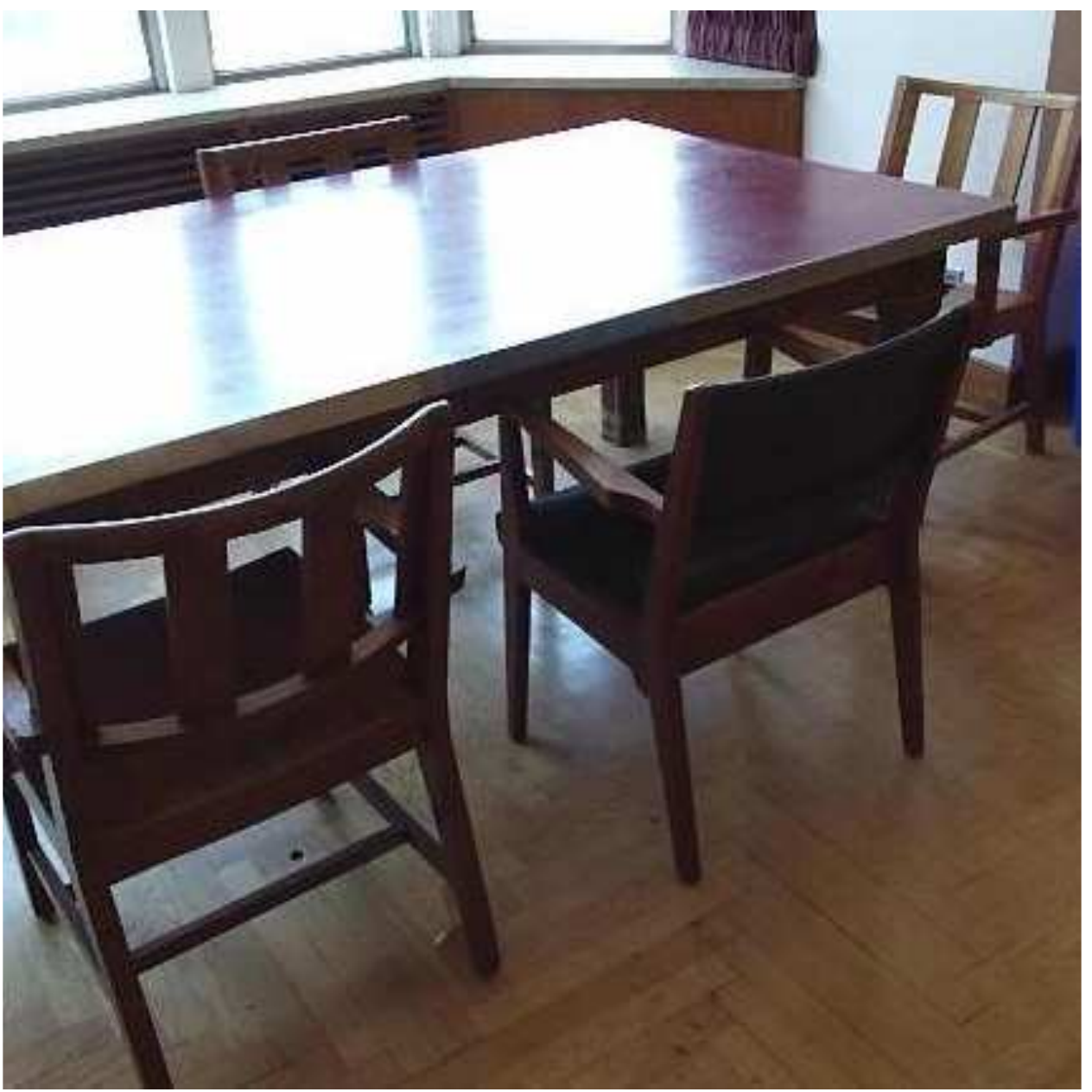} \\
\includegraphics[width=0.9\textwidth]{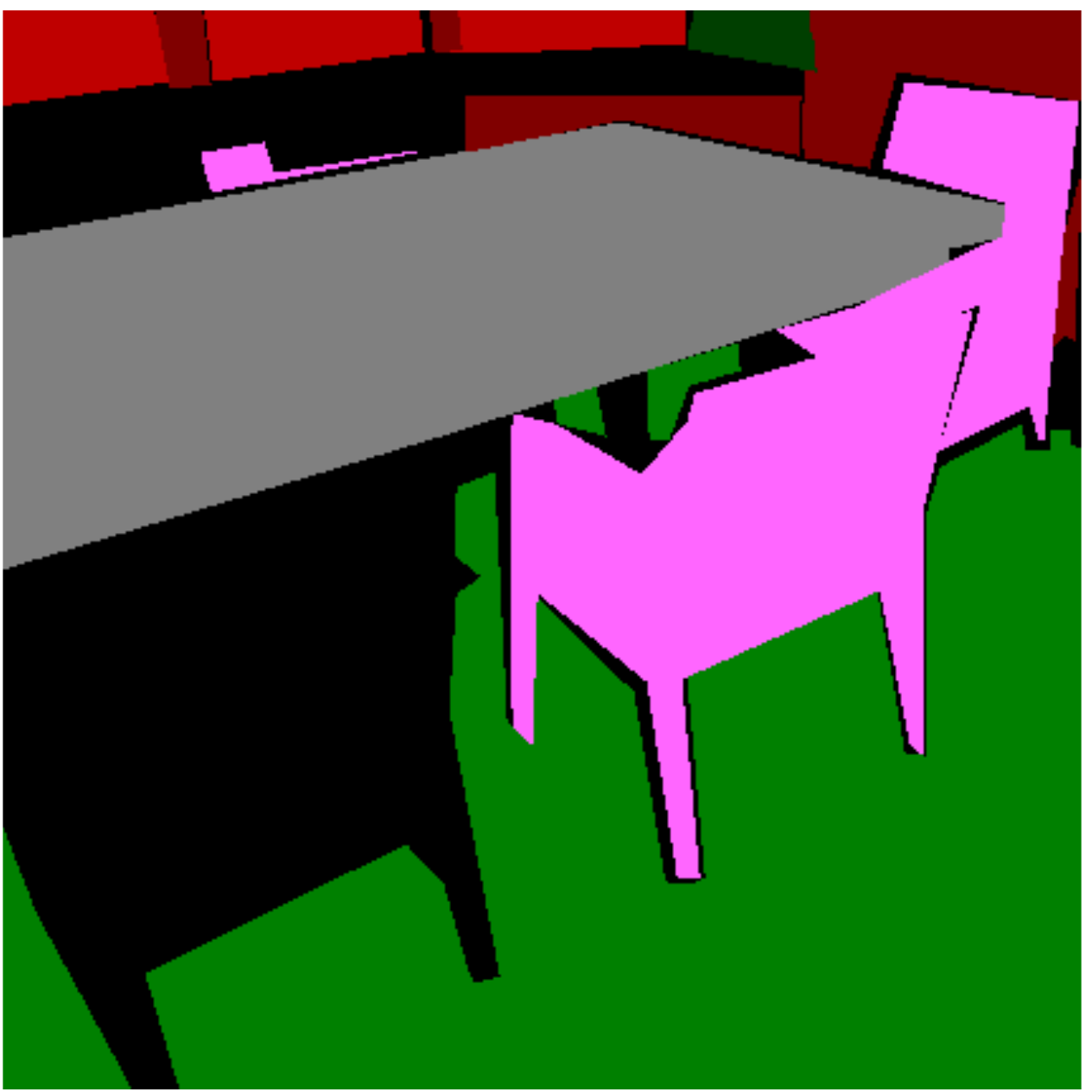} \\
\includegraphics[width=0.9\textwidth]{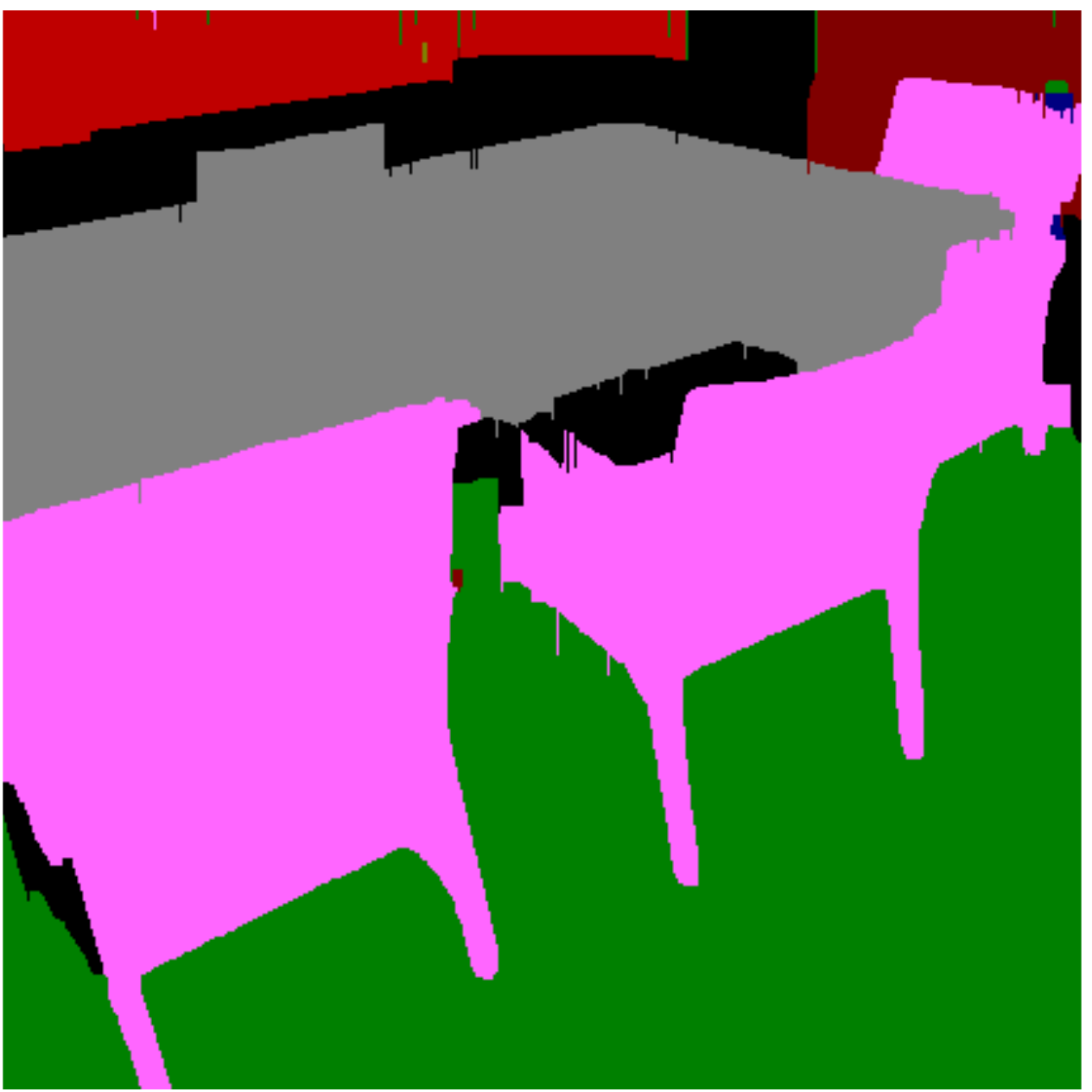}
\end{minipage}
}
\subfigure[]{
\begin{minipage}[b]{0.1\textwidth}
\includegraphics[width=0.9\textwidth]{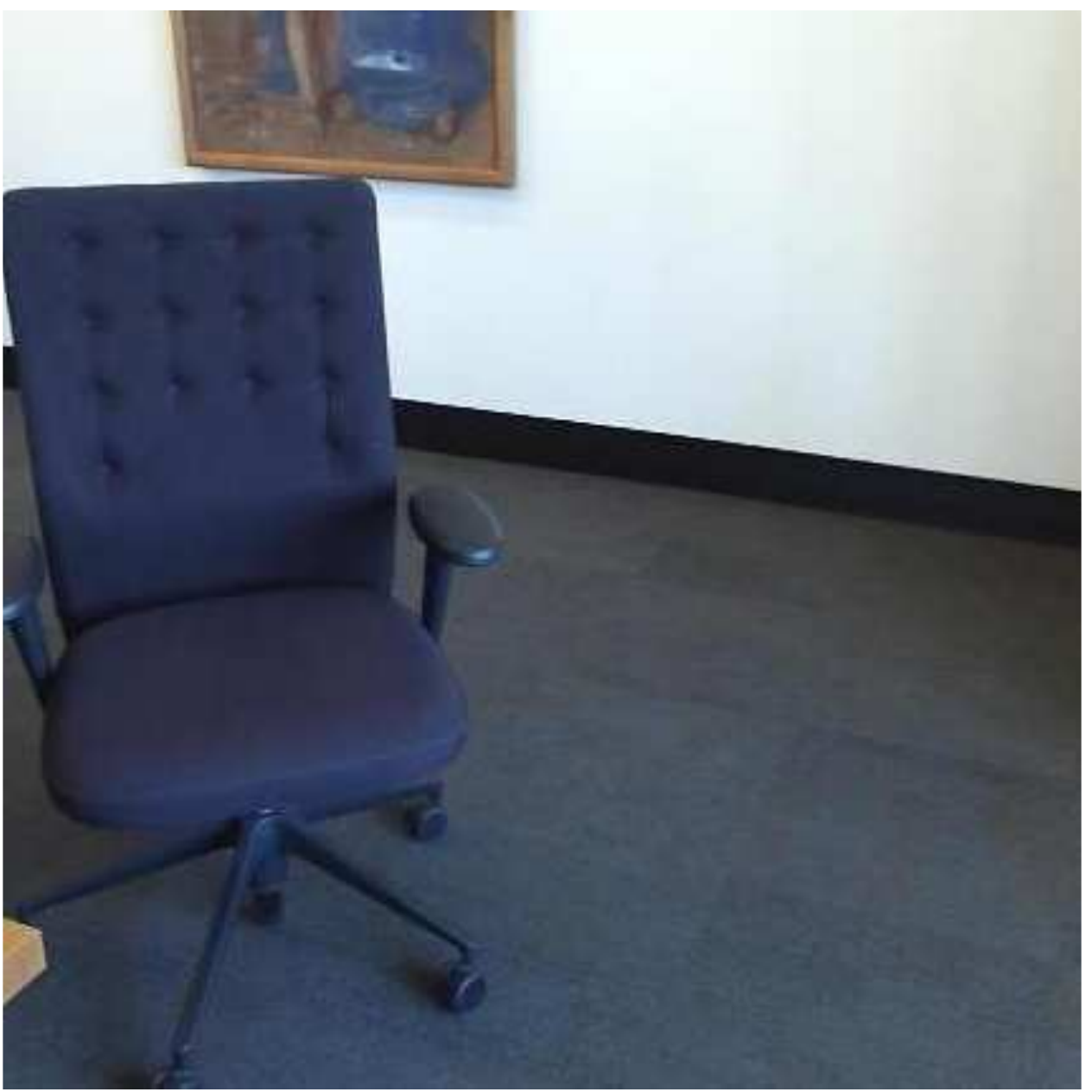} \\
\includegraphics[width=0.9\textwidth]{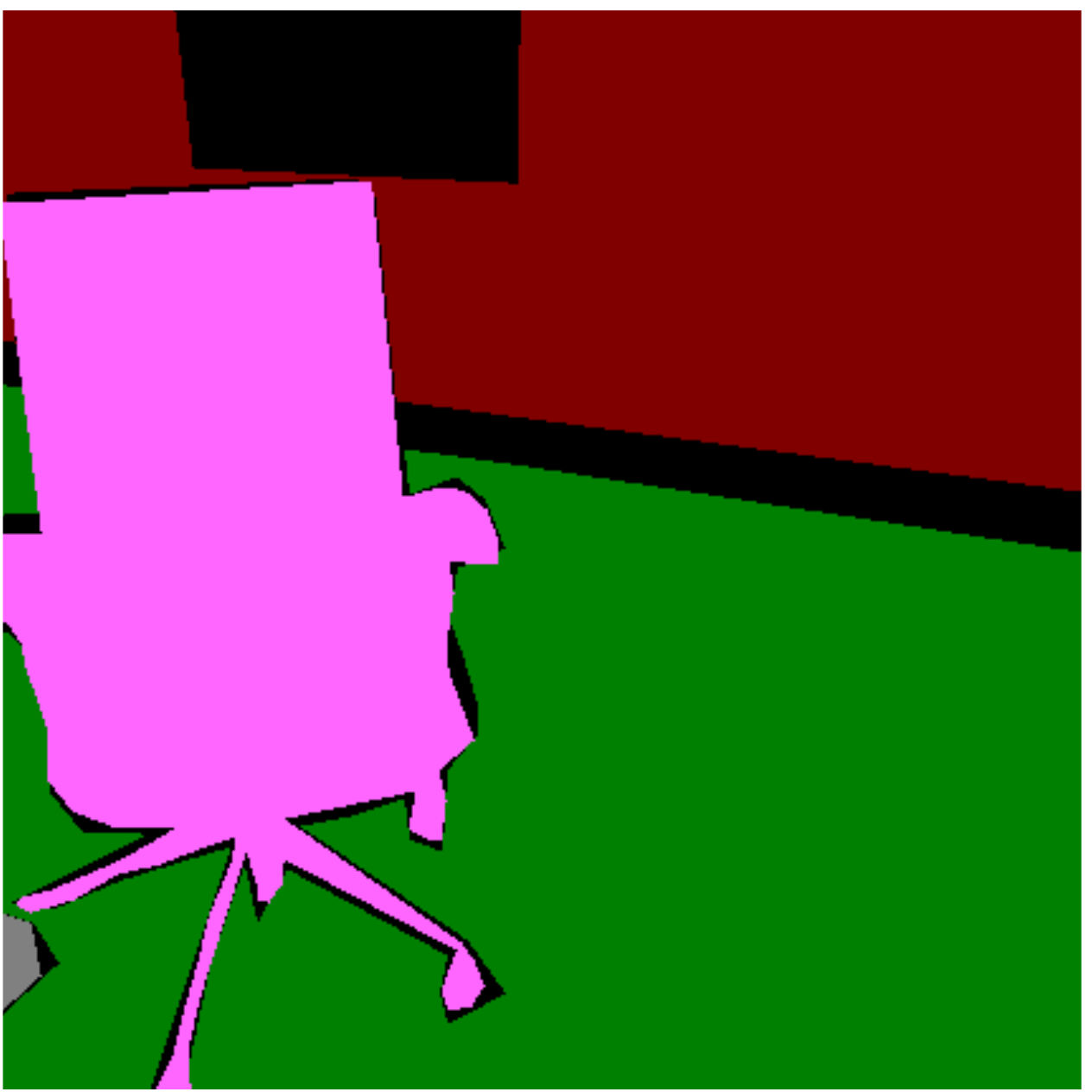} \\
\includegraphics[width=0.9\textwidth]{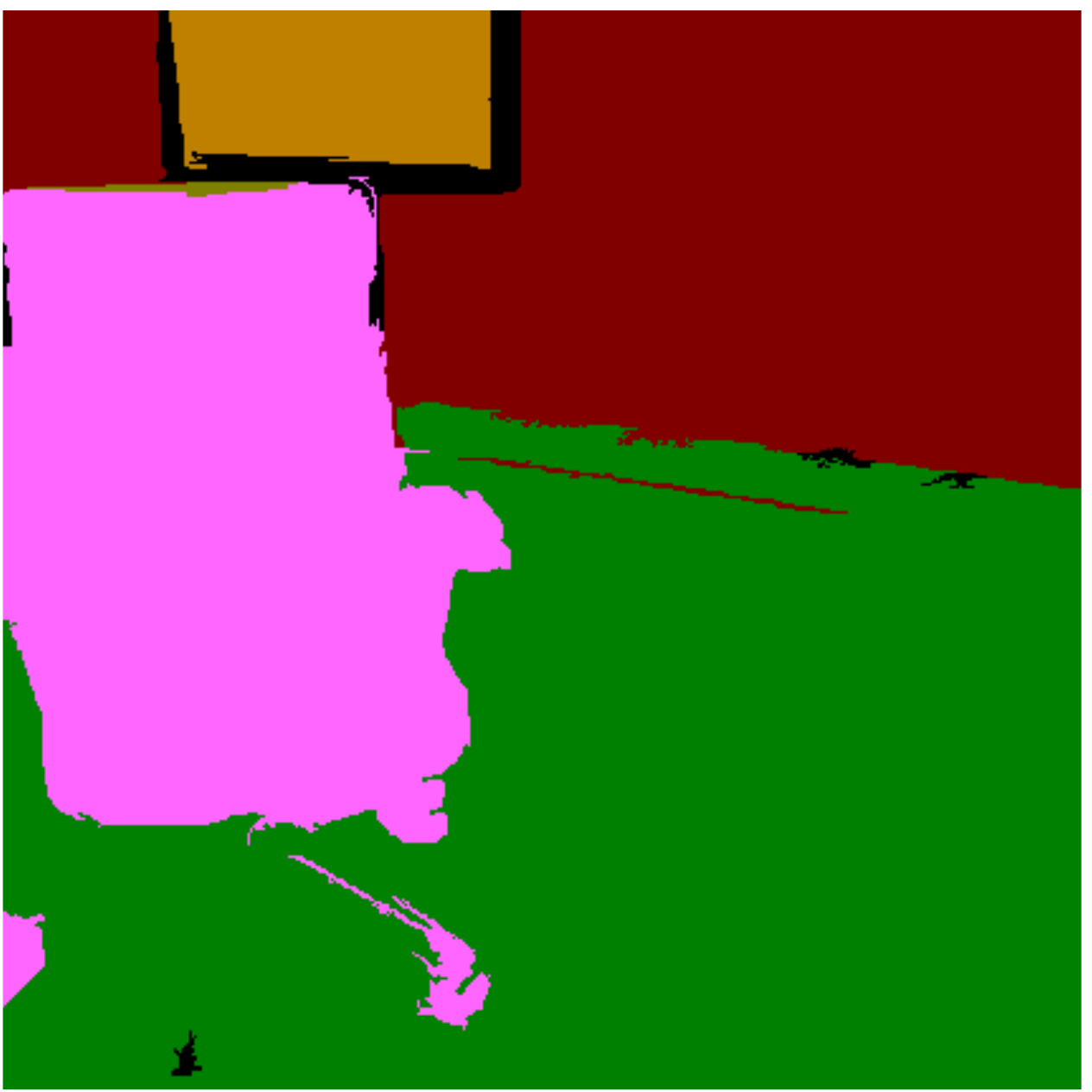}
\end{minipage}
}
\subfigure[]{
\begin{minipage}[b]{0.1\textwidth}
\includegraphics[width=0.9\textwidth]{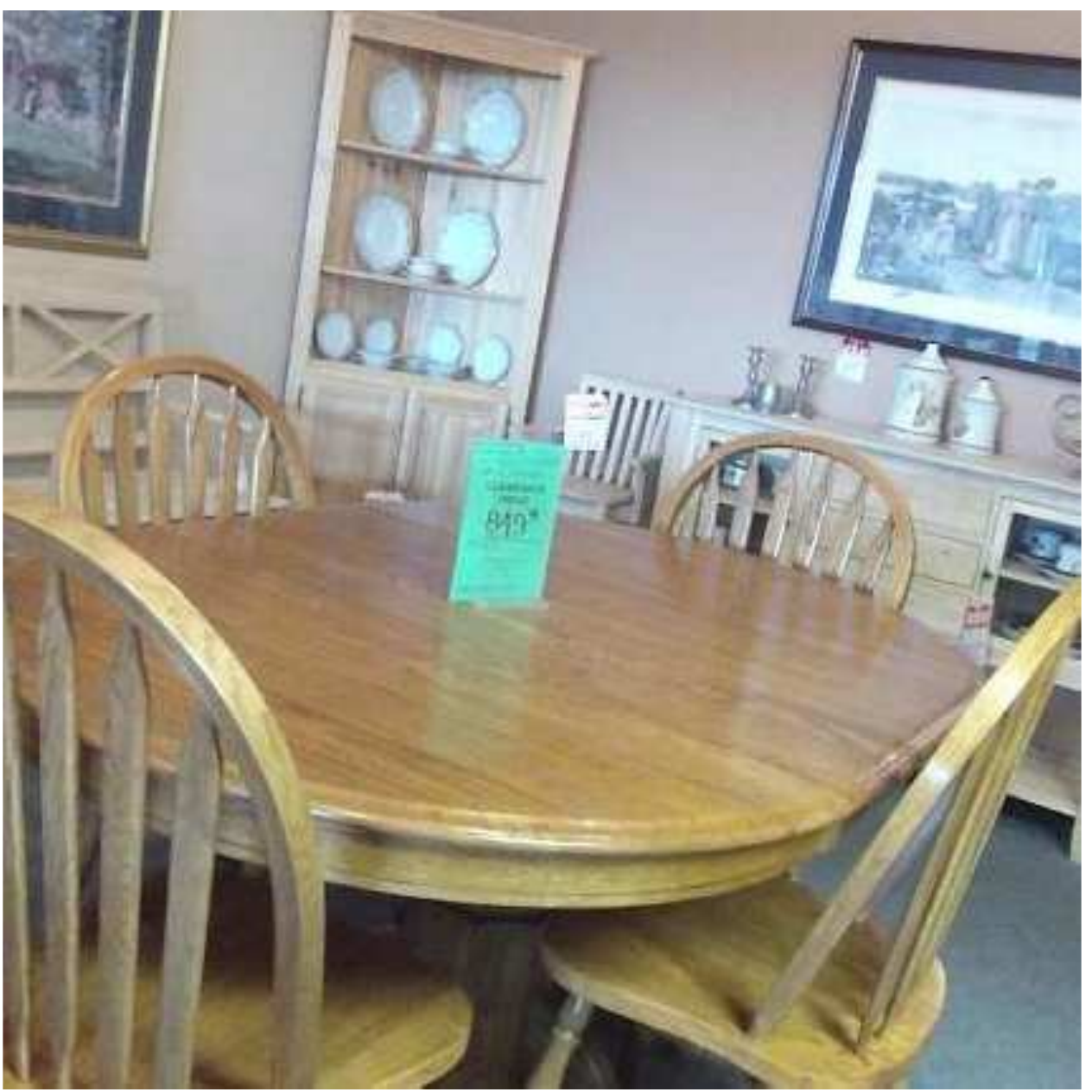} \\
\includegraphics[width=0.9\textwidth]{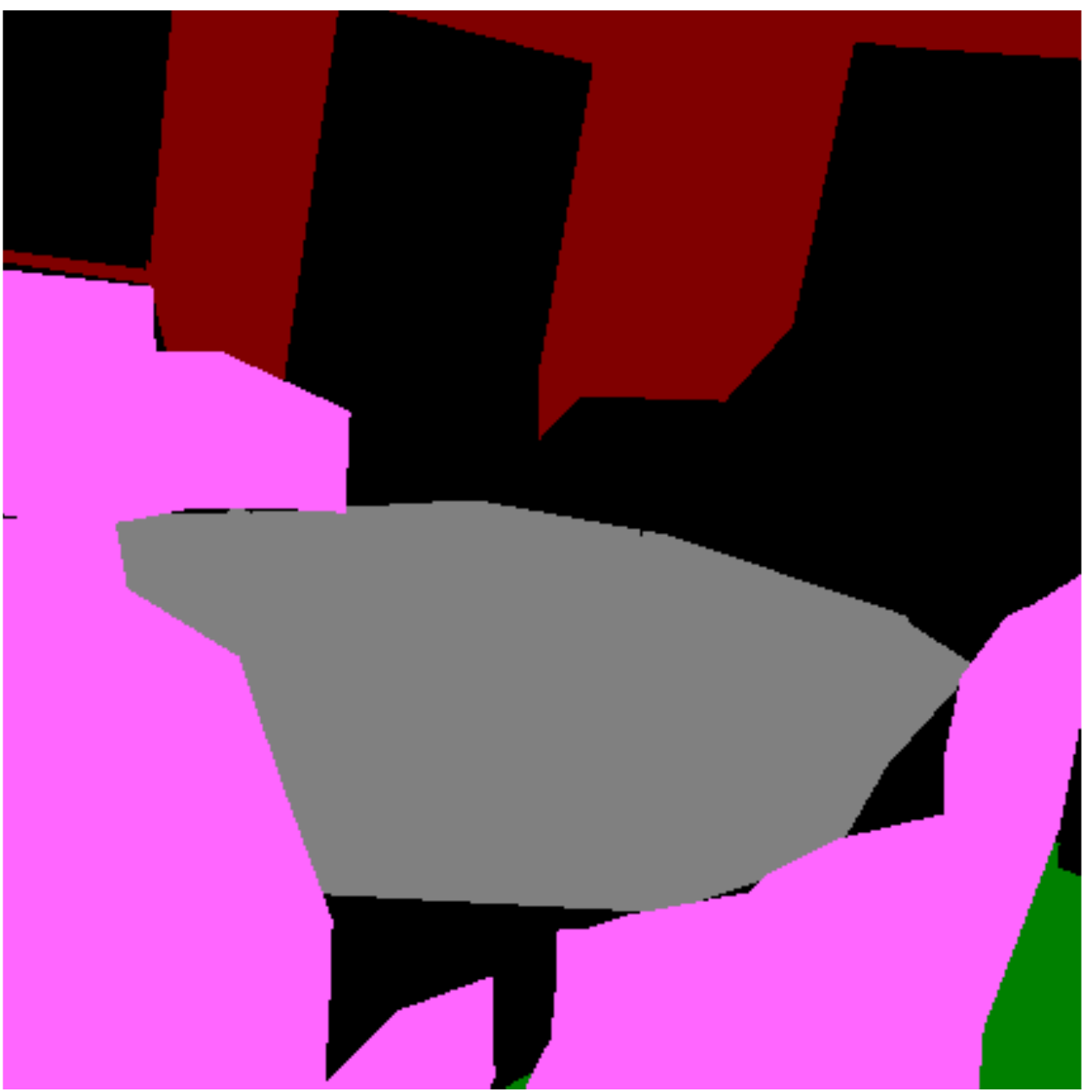} \\
\includegraphics[width=0.9\textwidth]{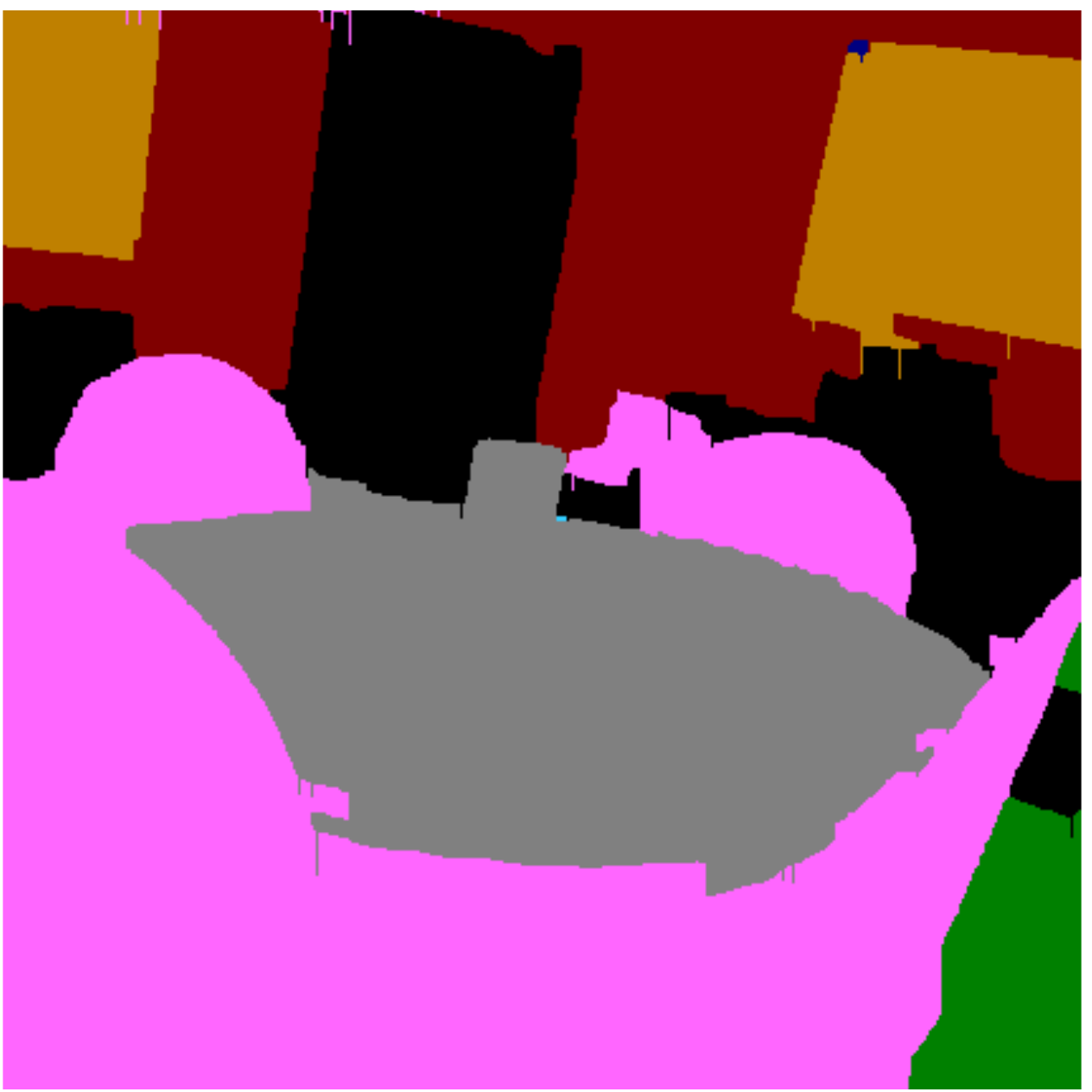}
\end{minipage}
}
\subfigure[]{
\begin{minipage}[b]{0.1\textwidth}
\includegraphics[width=0.9\textwidth]{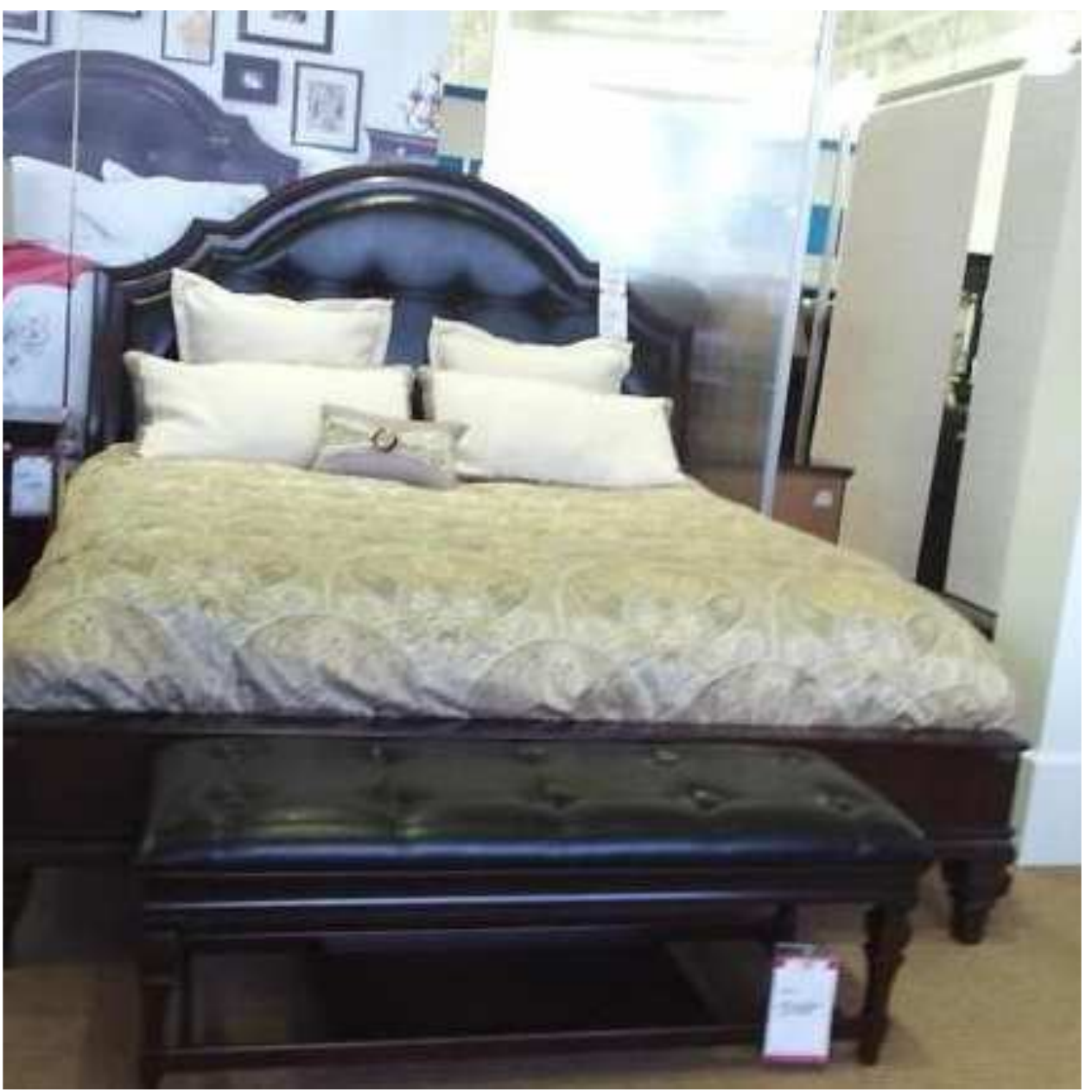} \\
\includegraphics[width=0.9\textwidth]{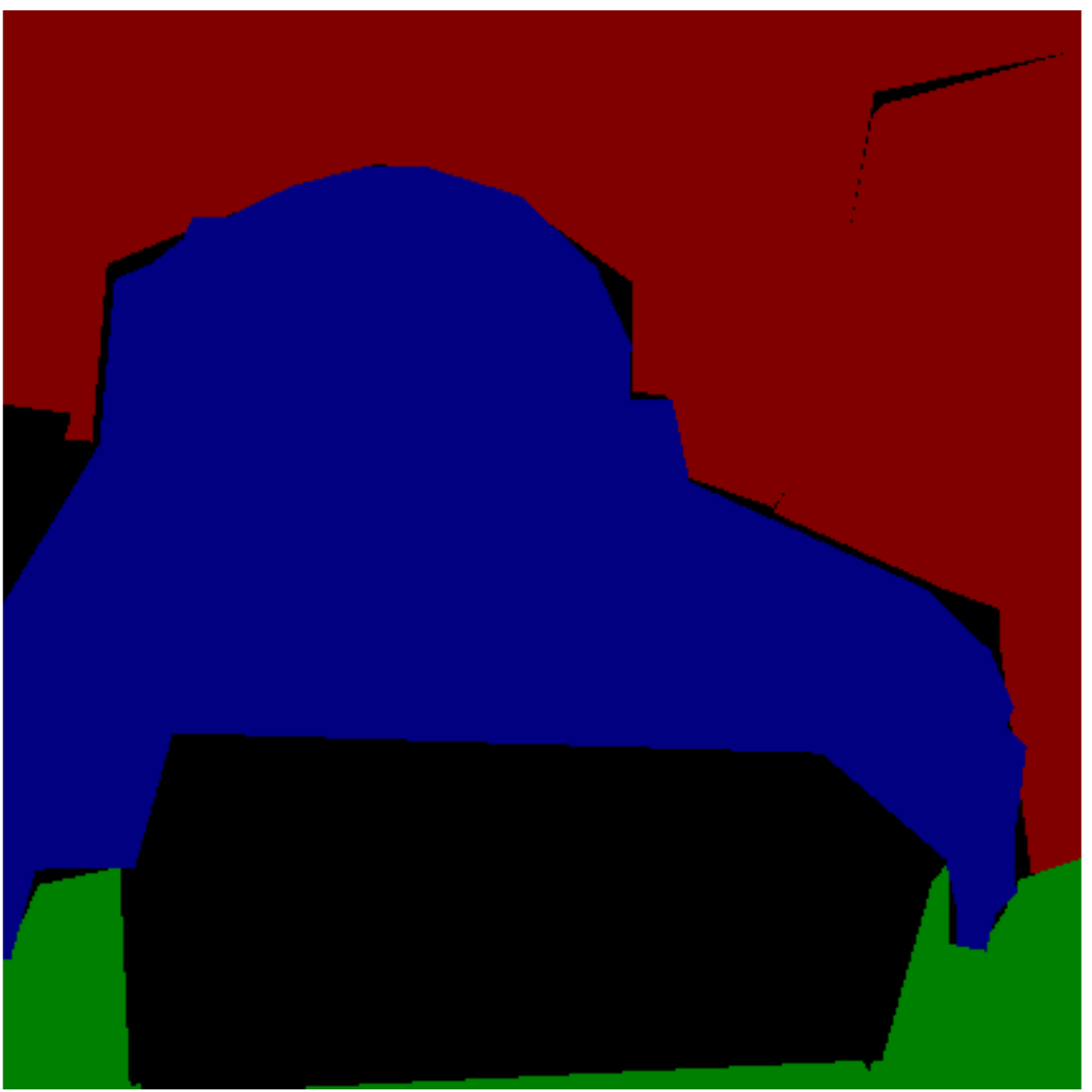} \\
\includegraphics[width=0.9\textwidth]{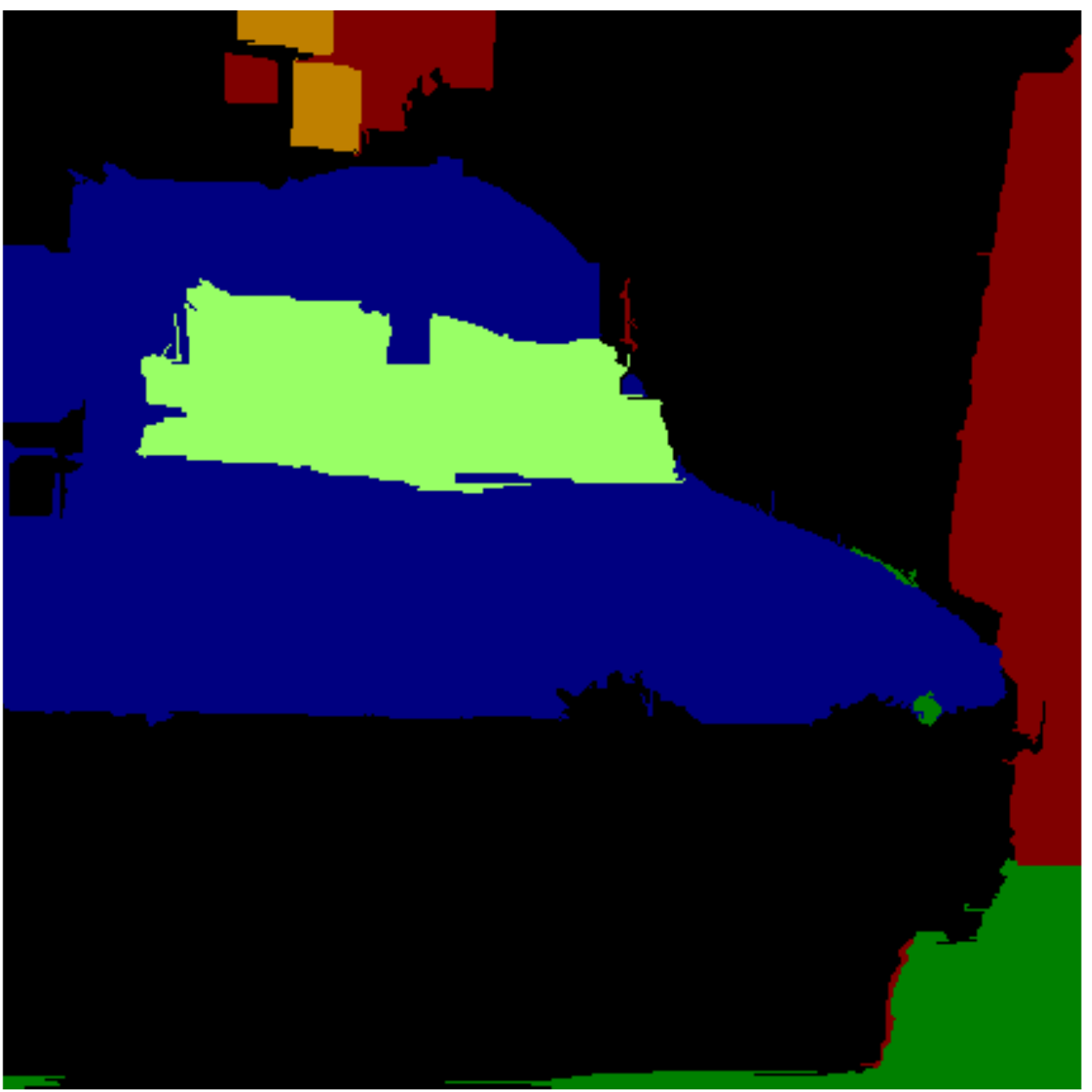}
\end{minipage}
}
\subfigure[]{
\begin{minipage}[b]{0.1\textwidth}
\includegraphics[width=0.9\textwidth]{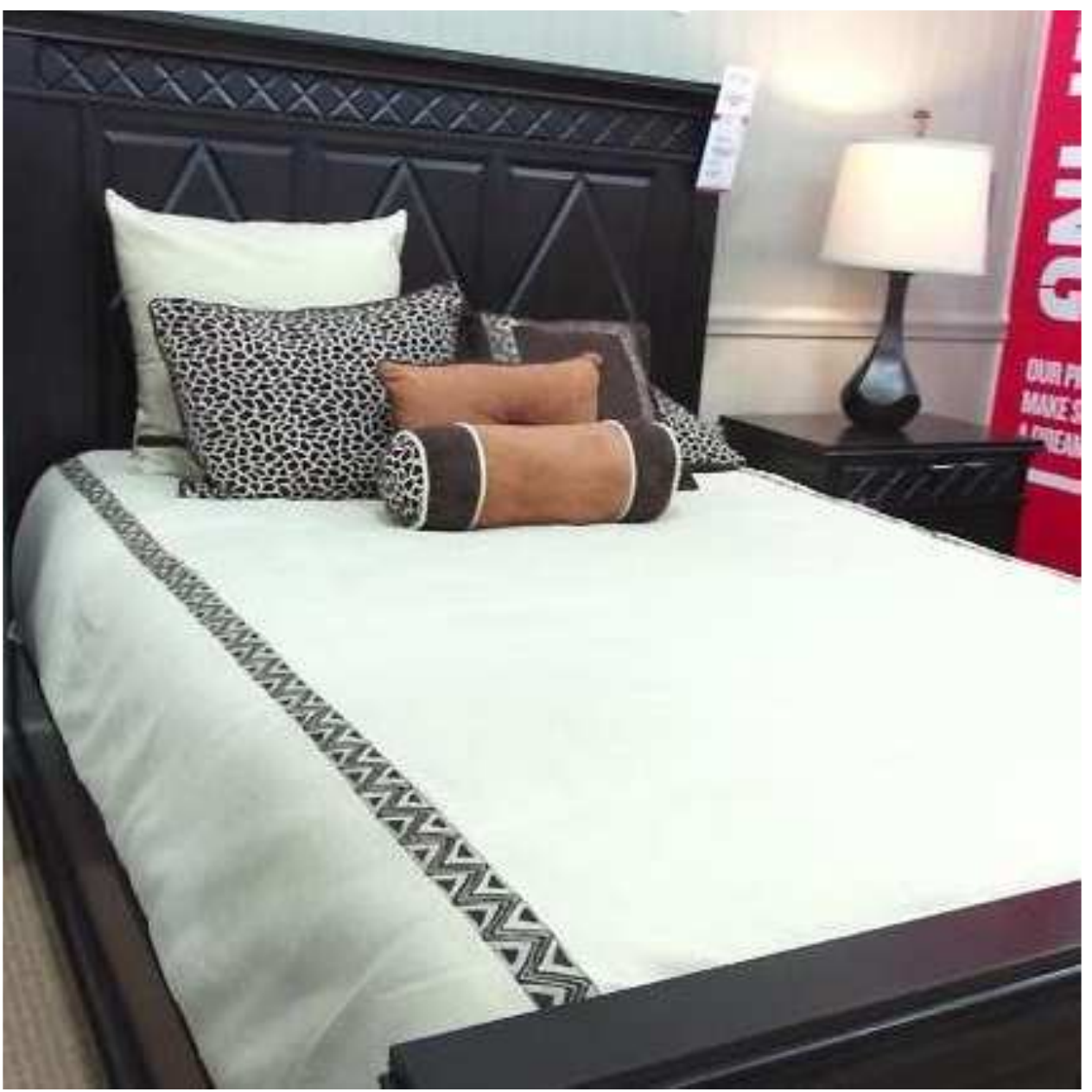} \\
\includegraphics[width=0.9\textwidth]{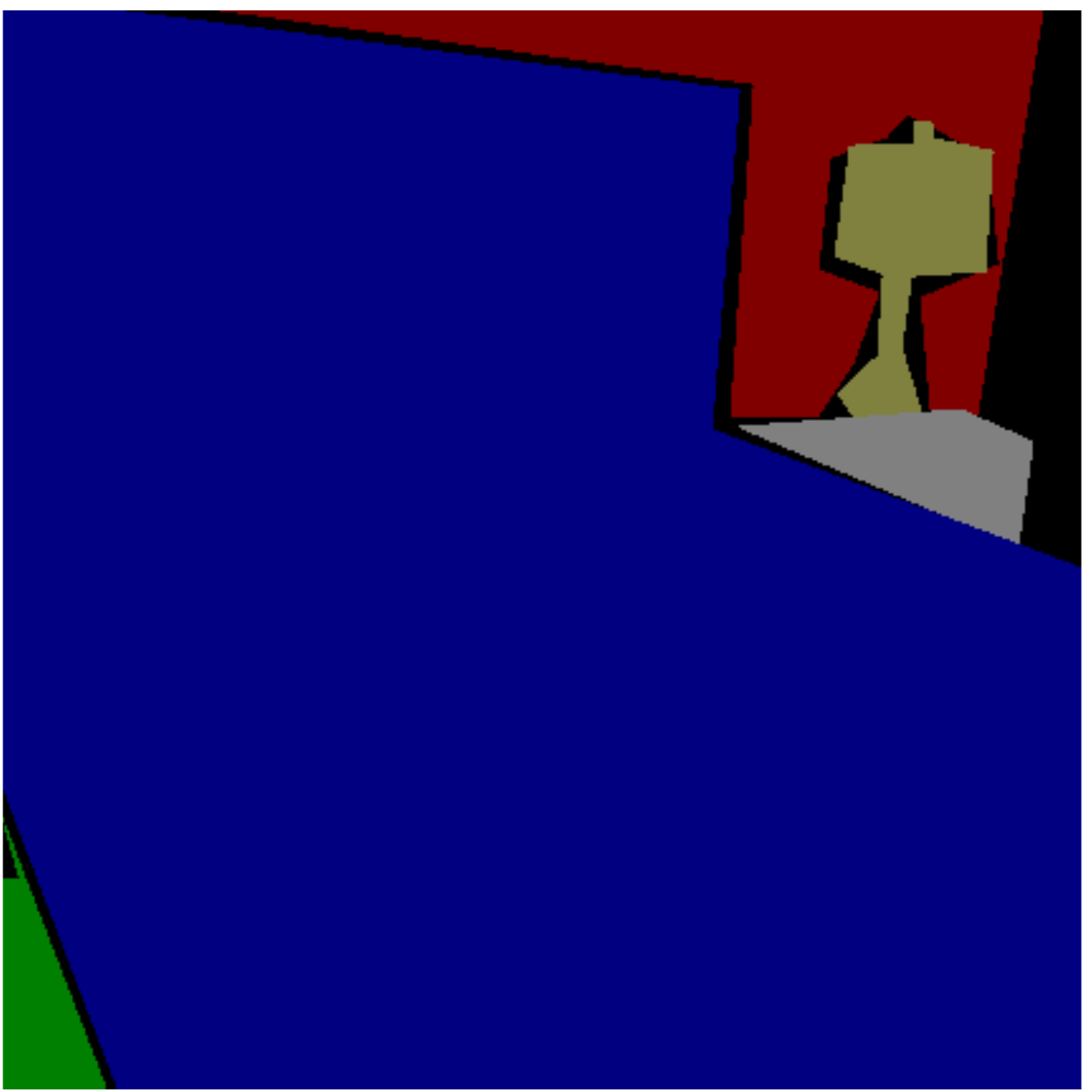} \\
\includegraphics[width=0.9\textwidth]{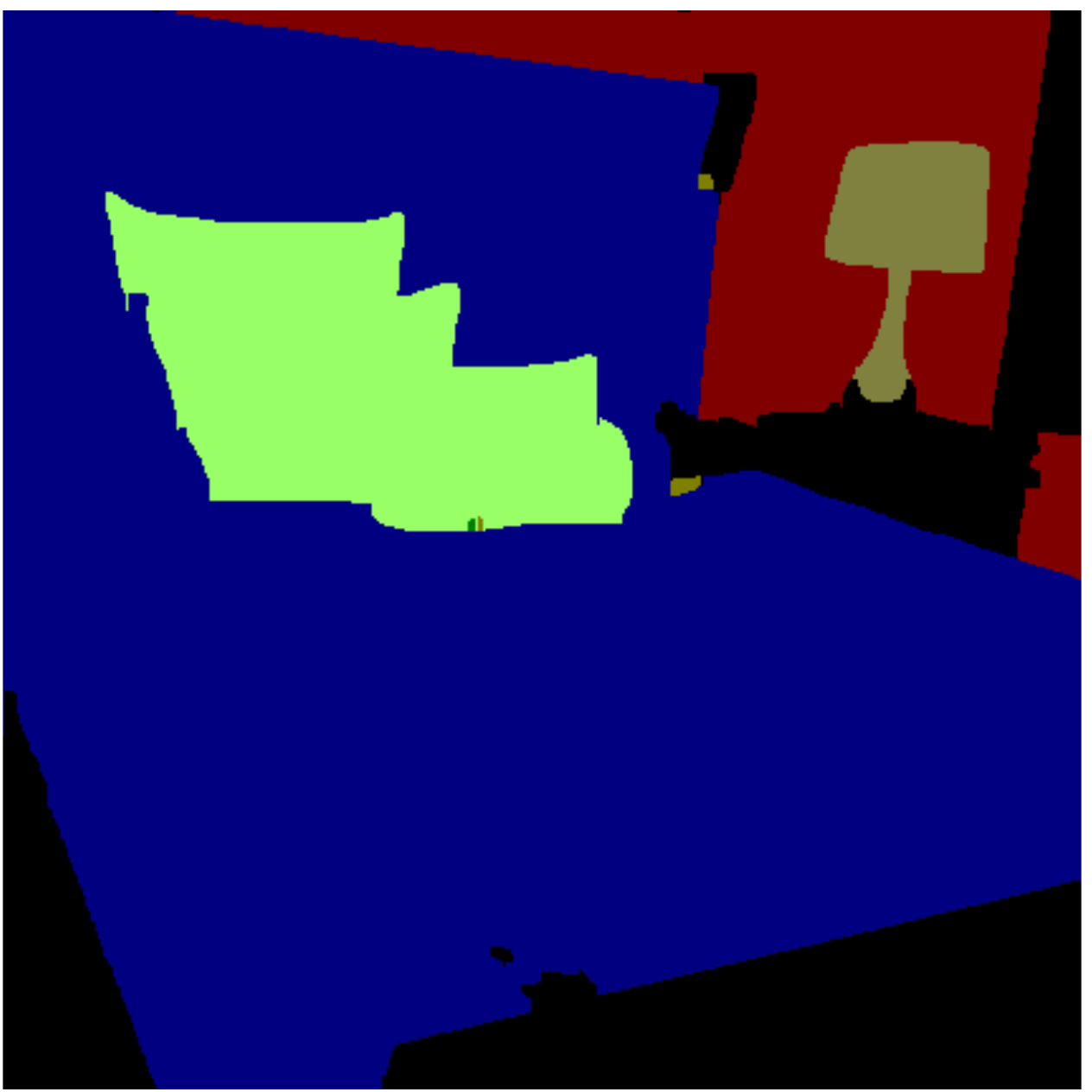}
\end{minipage}
}
\subfigure[]{
\begin{minipage}[b]{0.1\textwidth}
\includegraphics[width=0.9\textwidth]{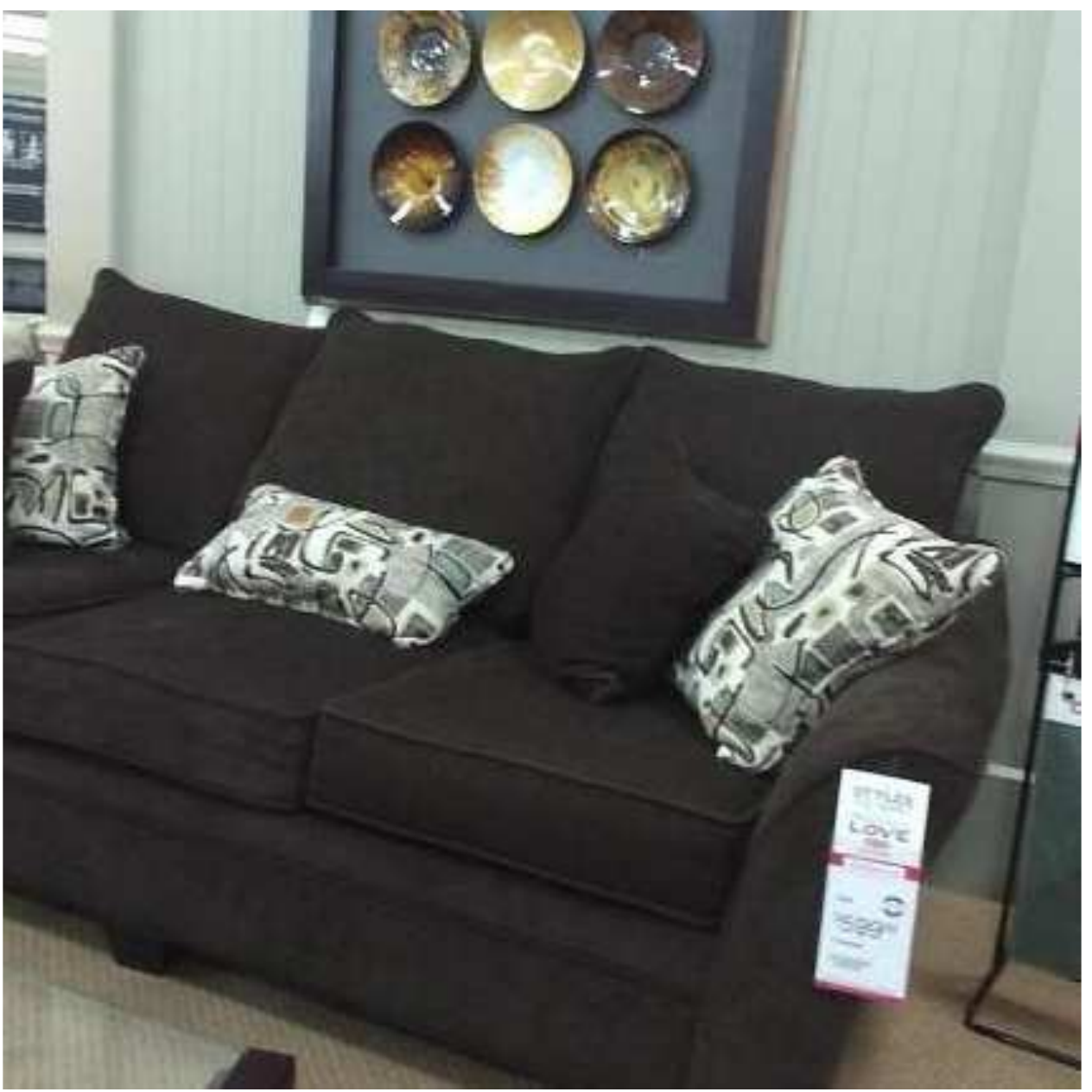} \\
\includegraphics[width=0.9\textwidth]{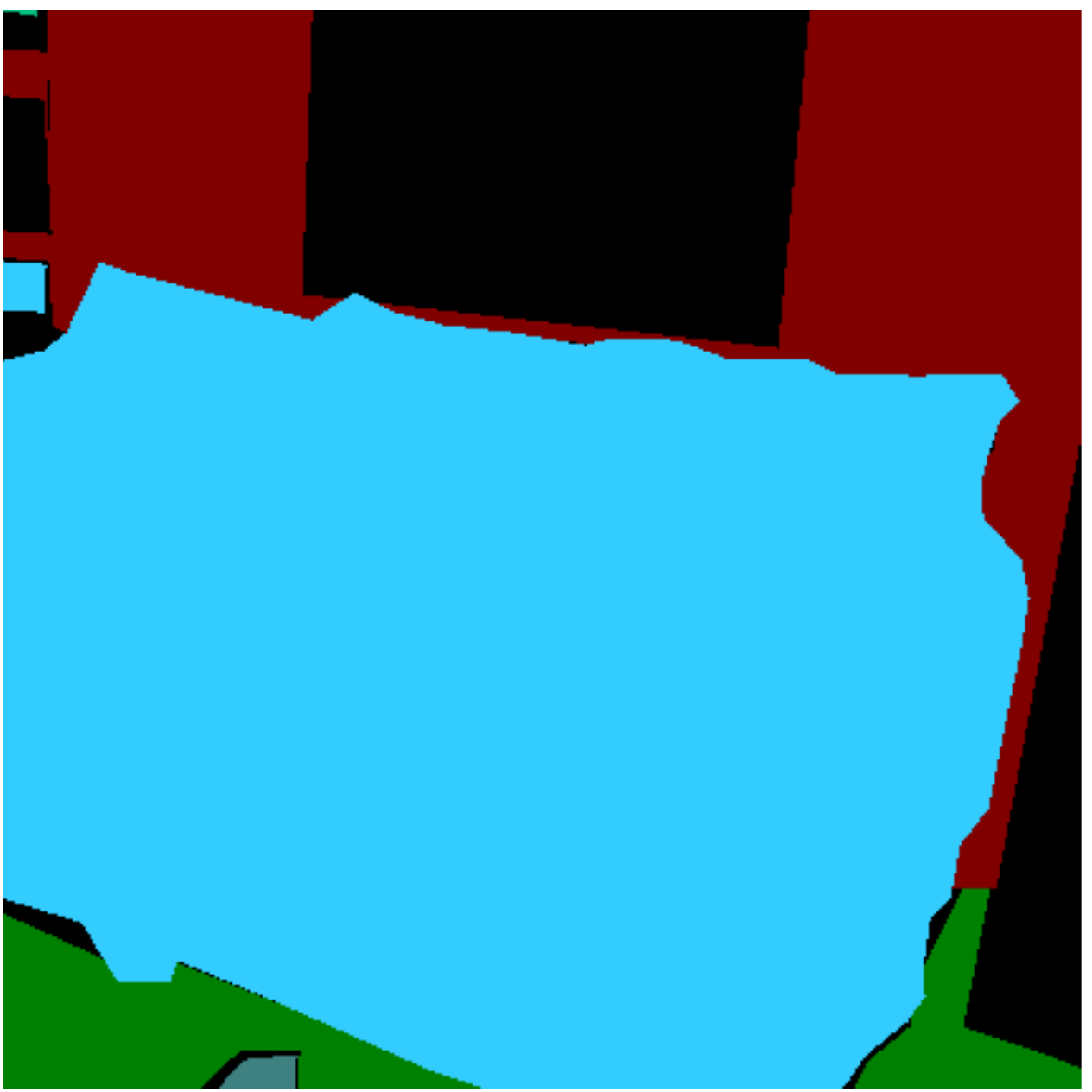} \\
\includegraphics[width=0.9\textwidth]{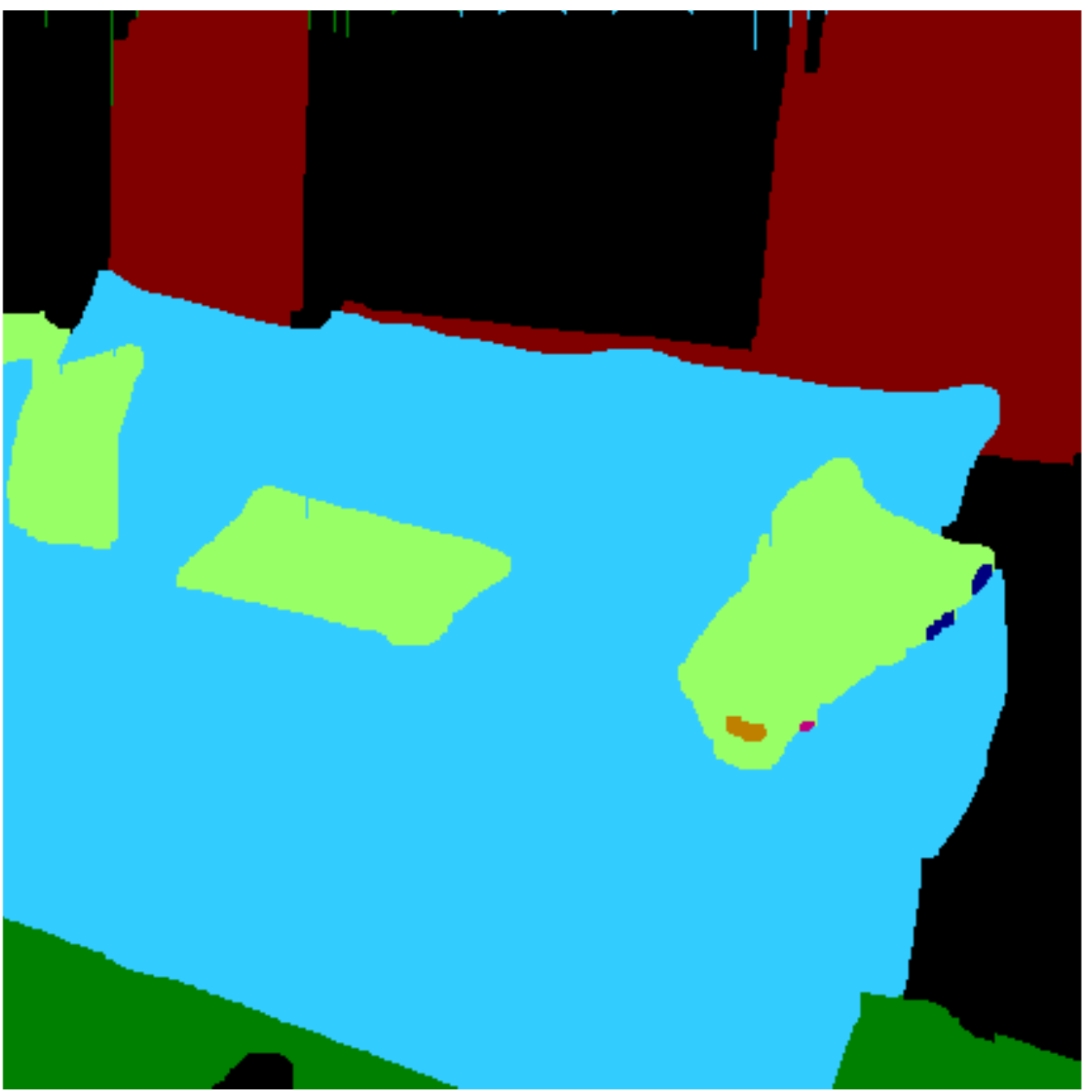}
\end{minipage}
}
\subfigure[]{
\begin{minipage}[b]{0.1\textwidth}
\includegraphics[width=0.9\textwidth]{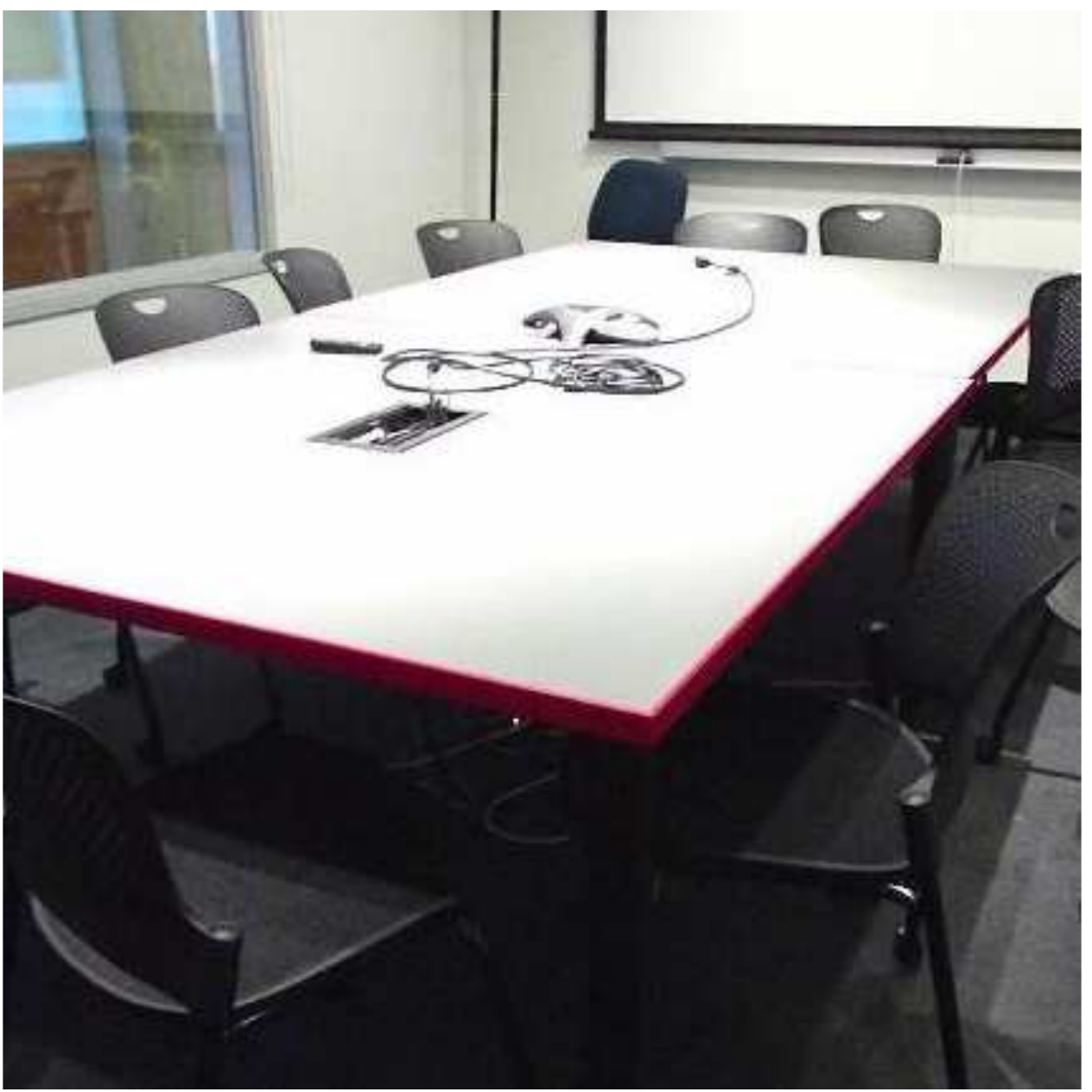} \\
\includegraphics[width=0.9\textwidth]{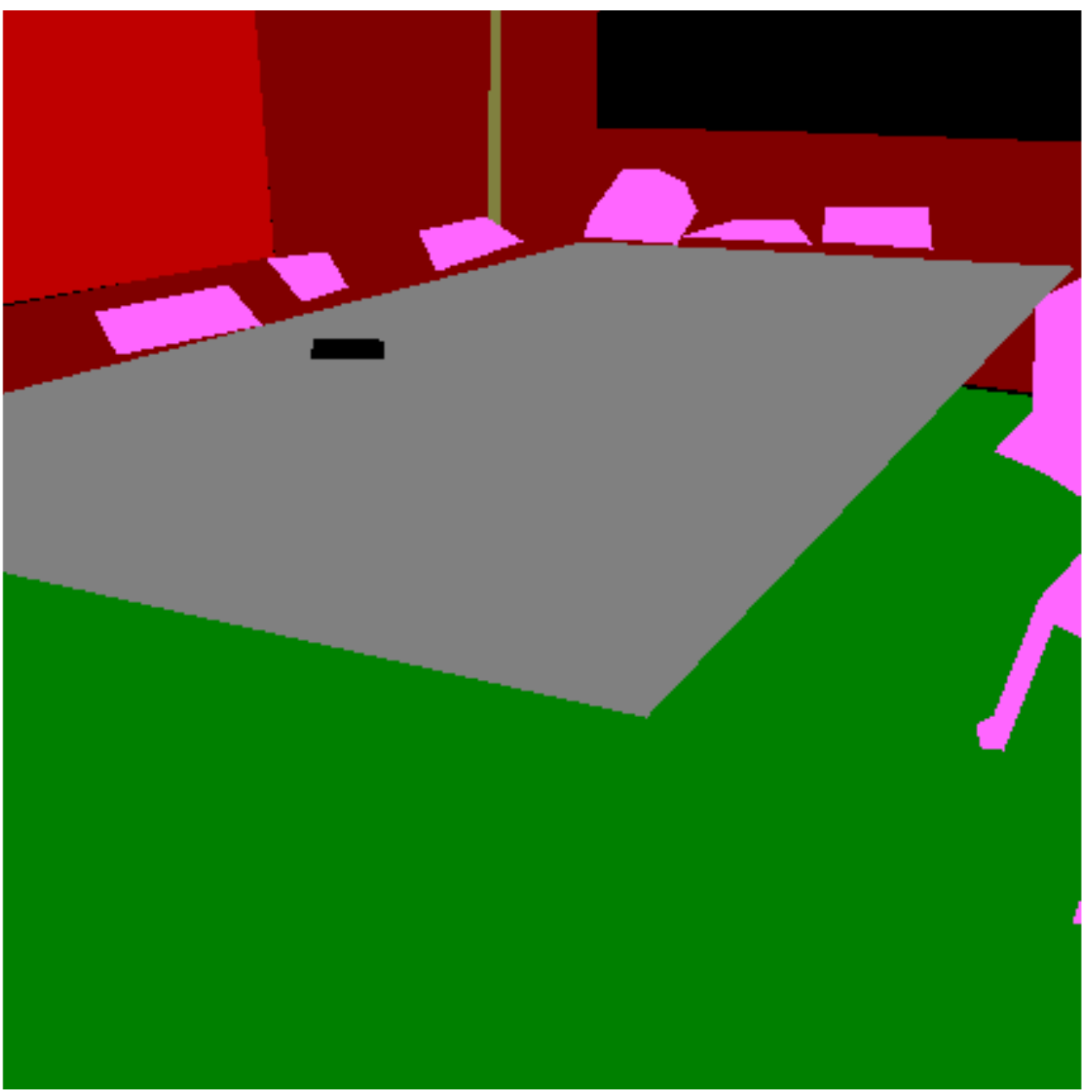} \\
\includegraphics[width=0.9\textwidth]{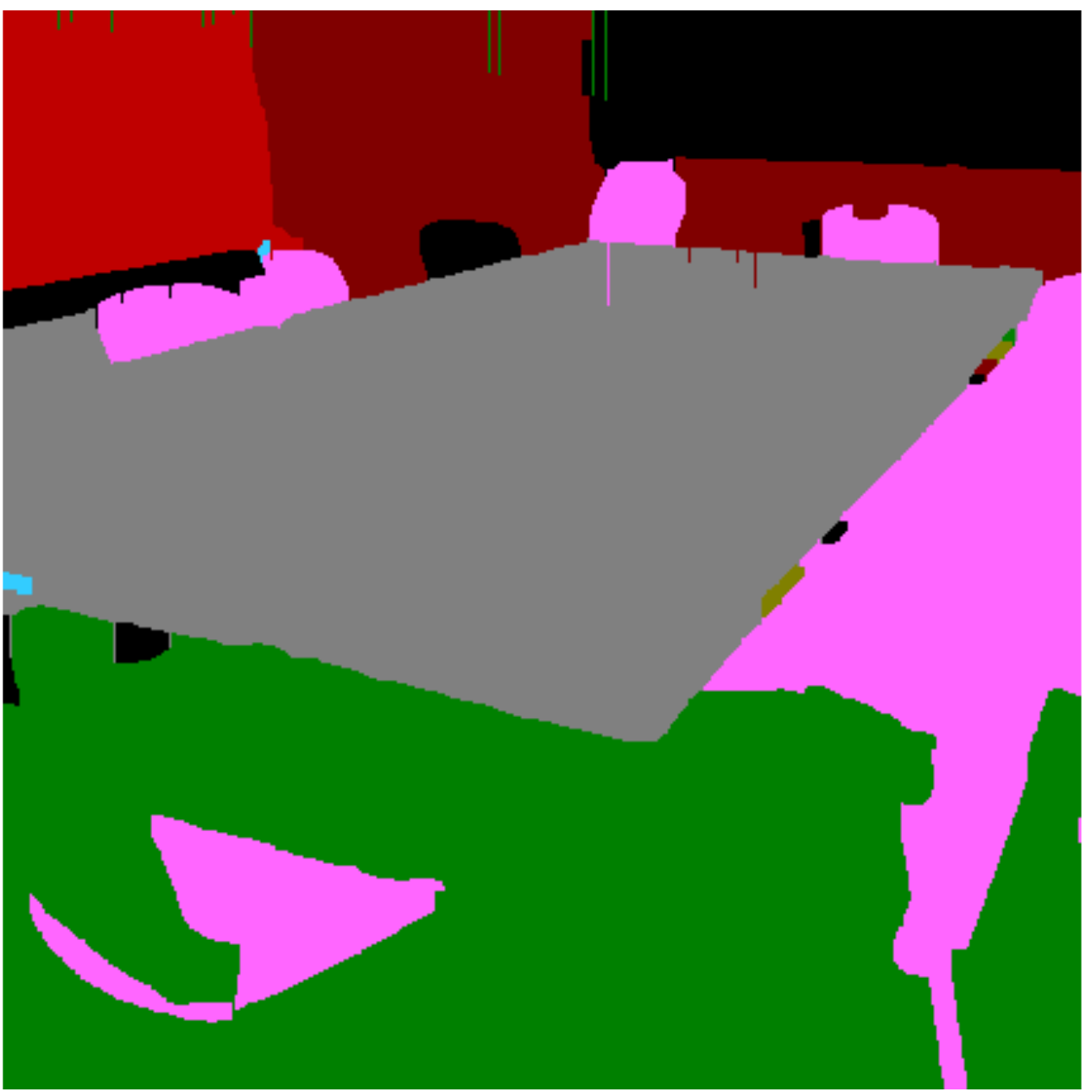}
\end{minipage}
}
\subfigure[]{
\begin{minipage}[b]{0.1\textwidth}
\includegraphics[width=0.9\textwidth]{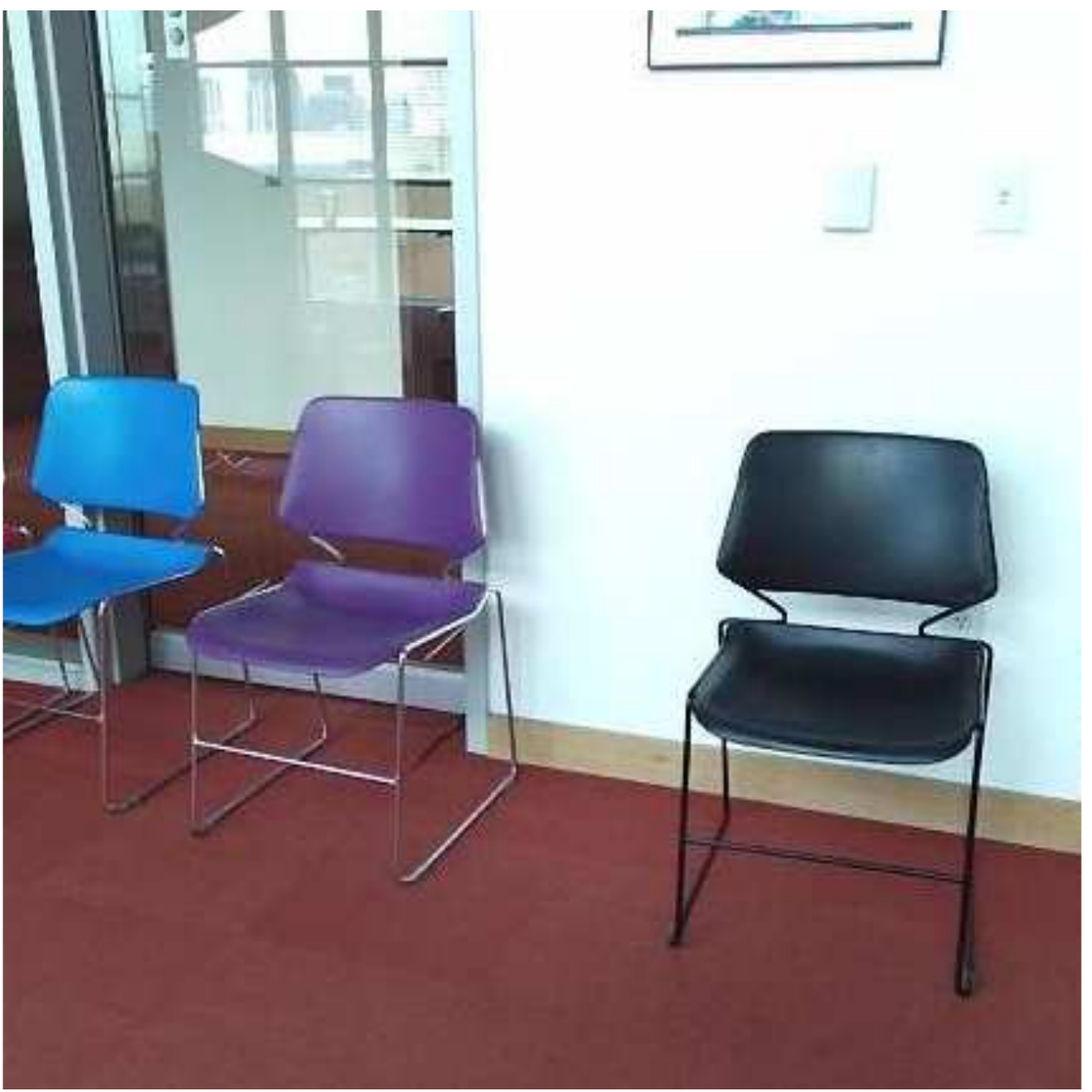} \\
\includegraphics[width=0.9\textwidth]{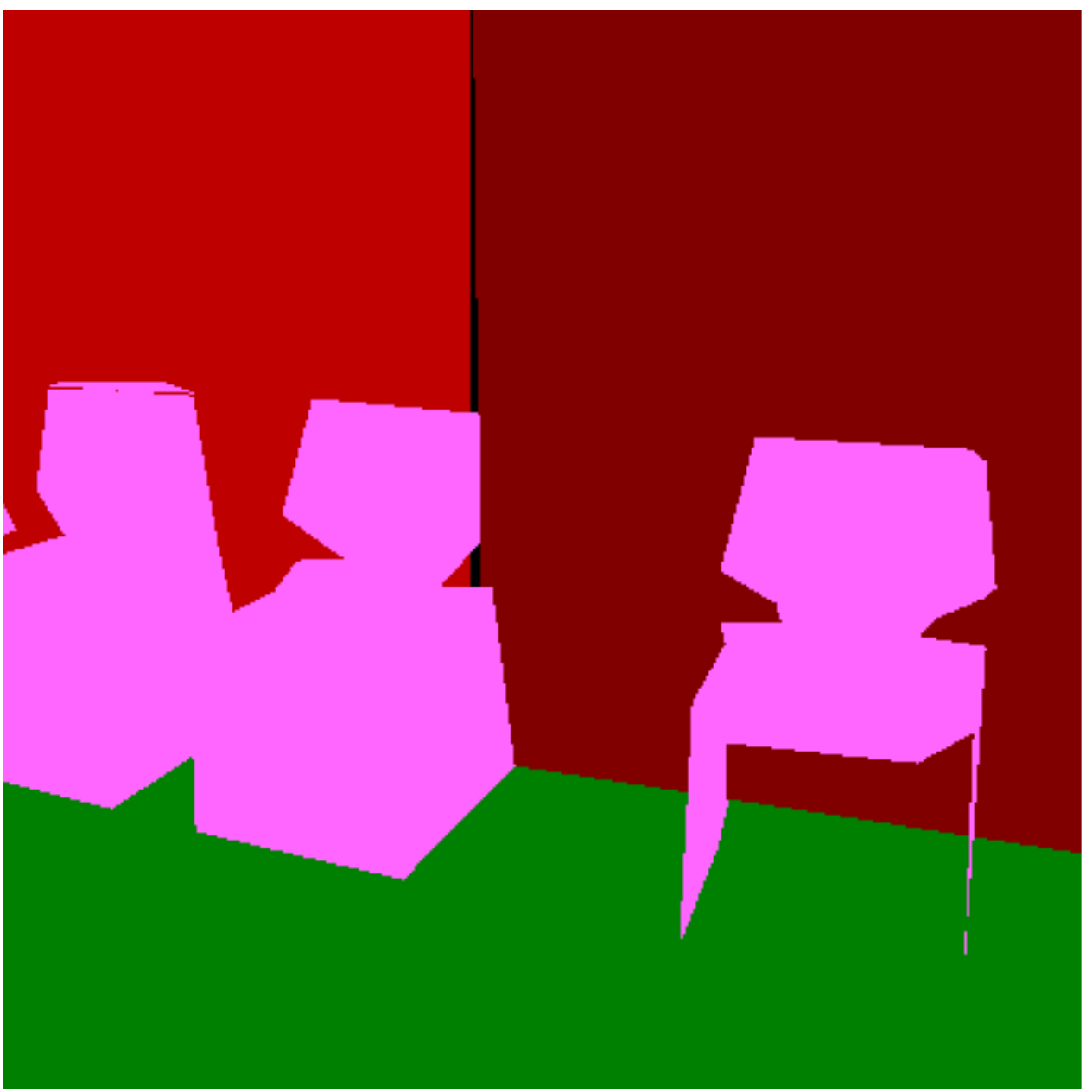} \\
\includegraphics[width=0.9\textwidth]{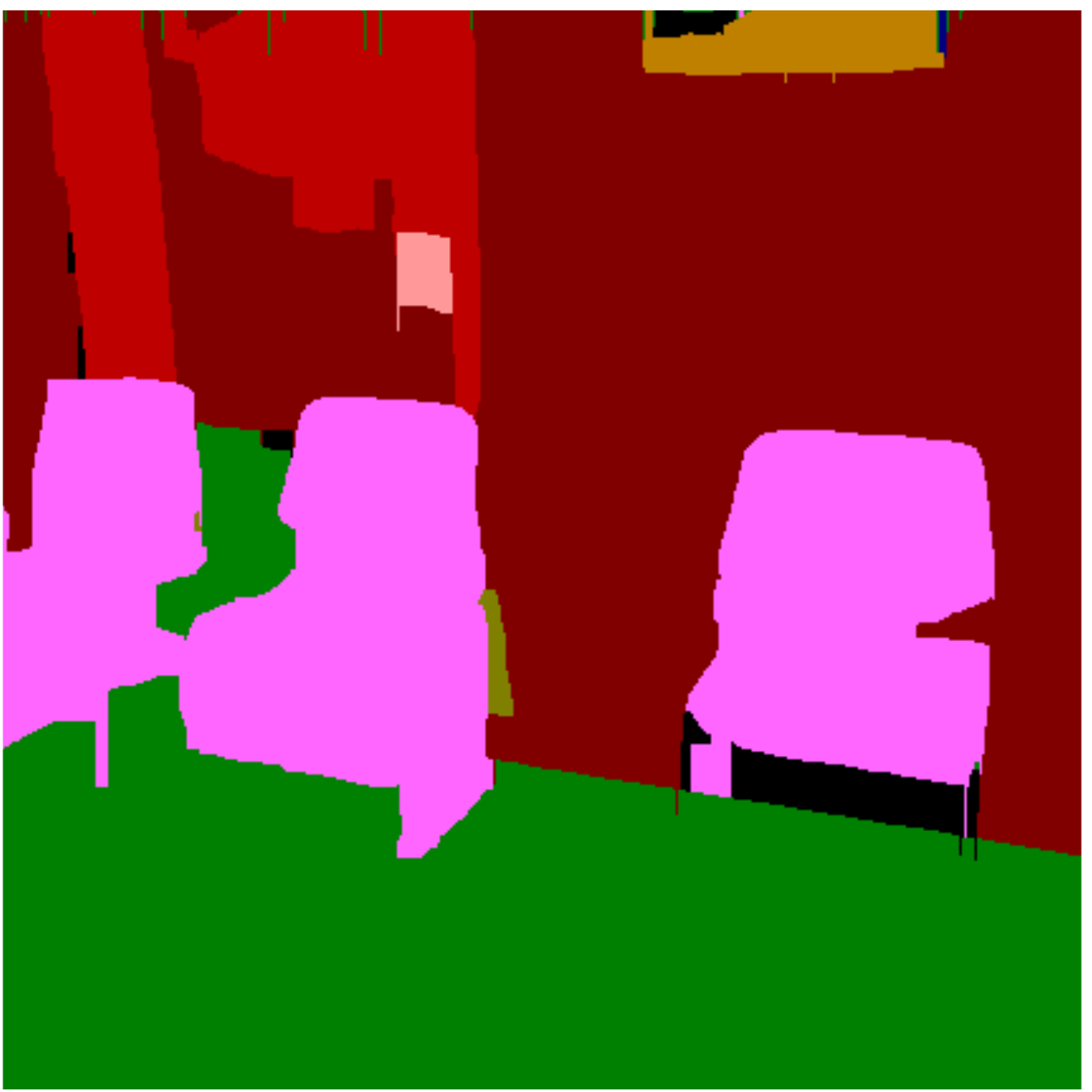}
\end{minipage}
}
\subfigure[]{
\begin{minipage}[b]{0.1\textwidth}
\includegraphics[width=0.9\textwidth]{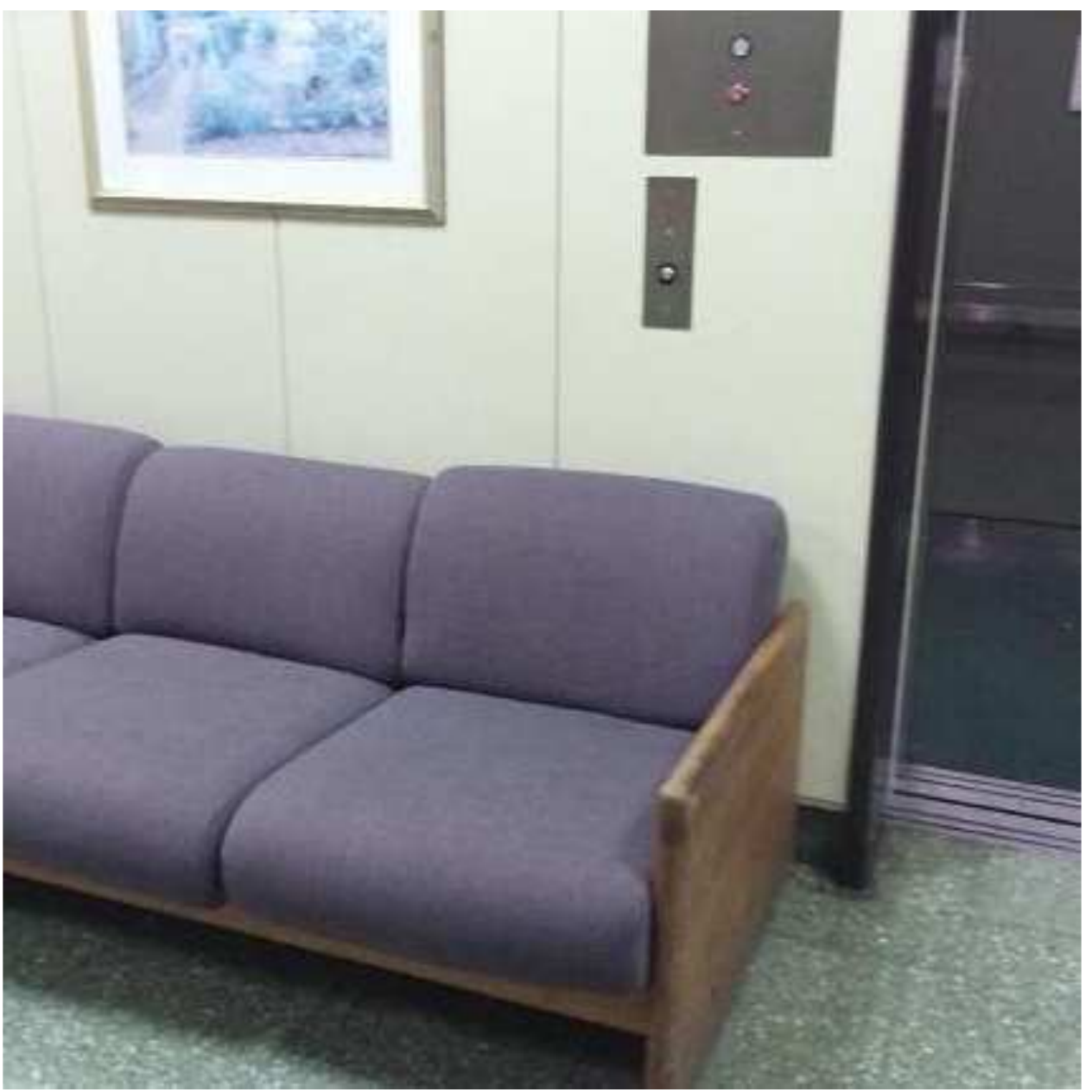} \\
\includegraphics[width=0.9\textwidth]{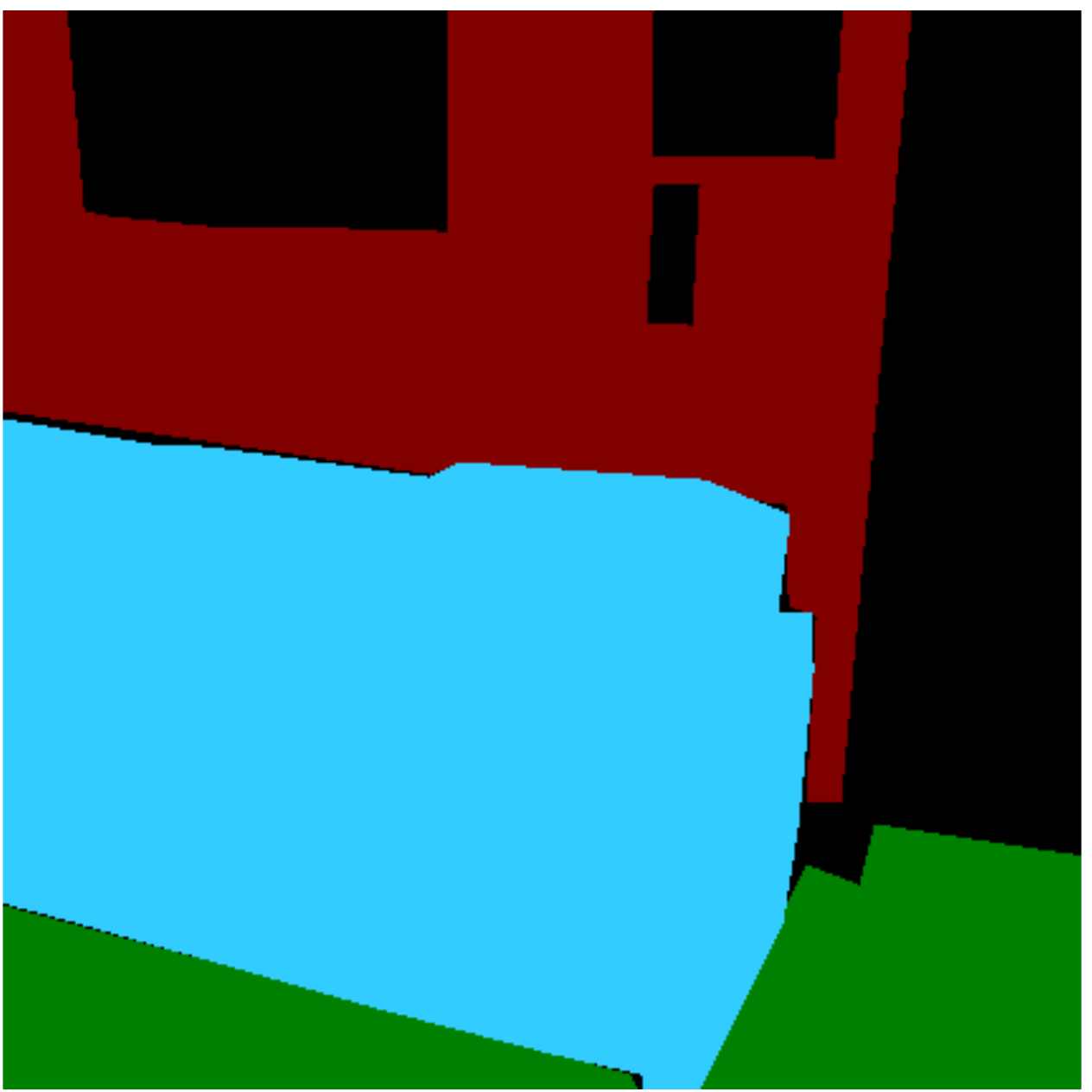} \\
\includegraphics[width=0.9\textwidth]{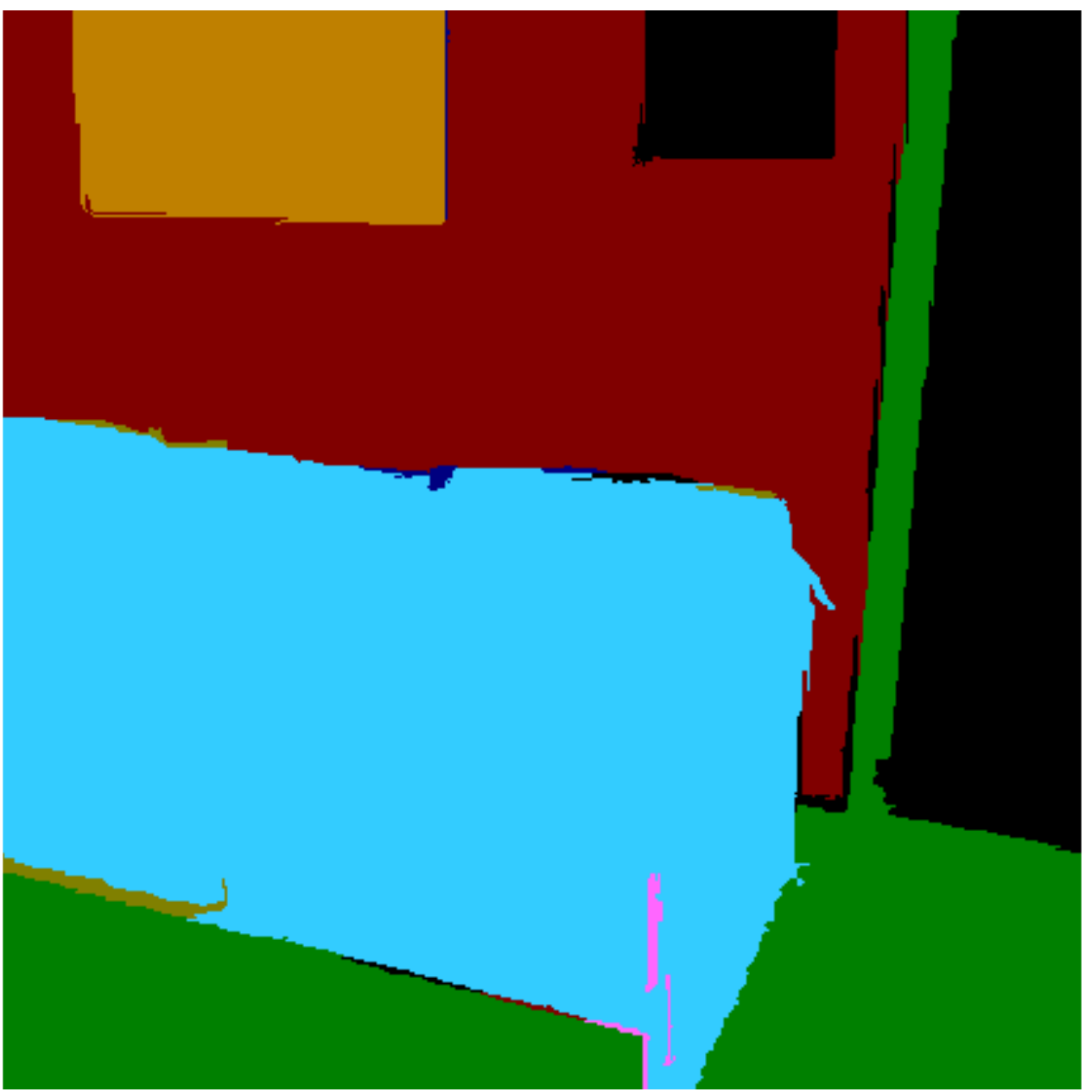}
\end{minipage}
}
\subfigure[]{
\begin{minipage}[b]{0.1\textwidth}
\includegraphics[width=0.9\textwidth]{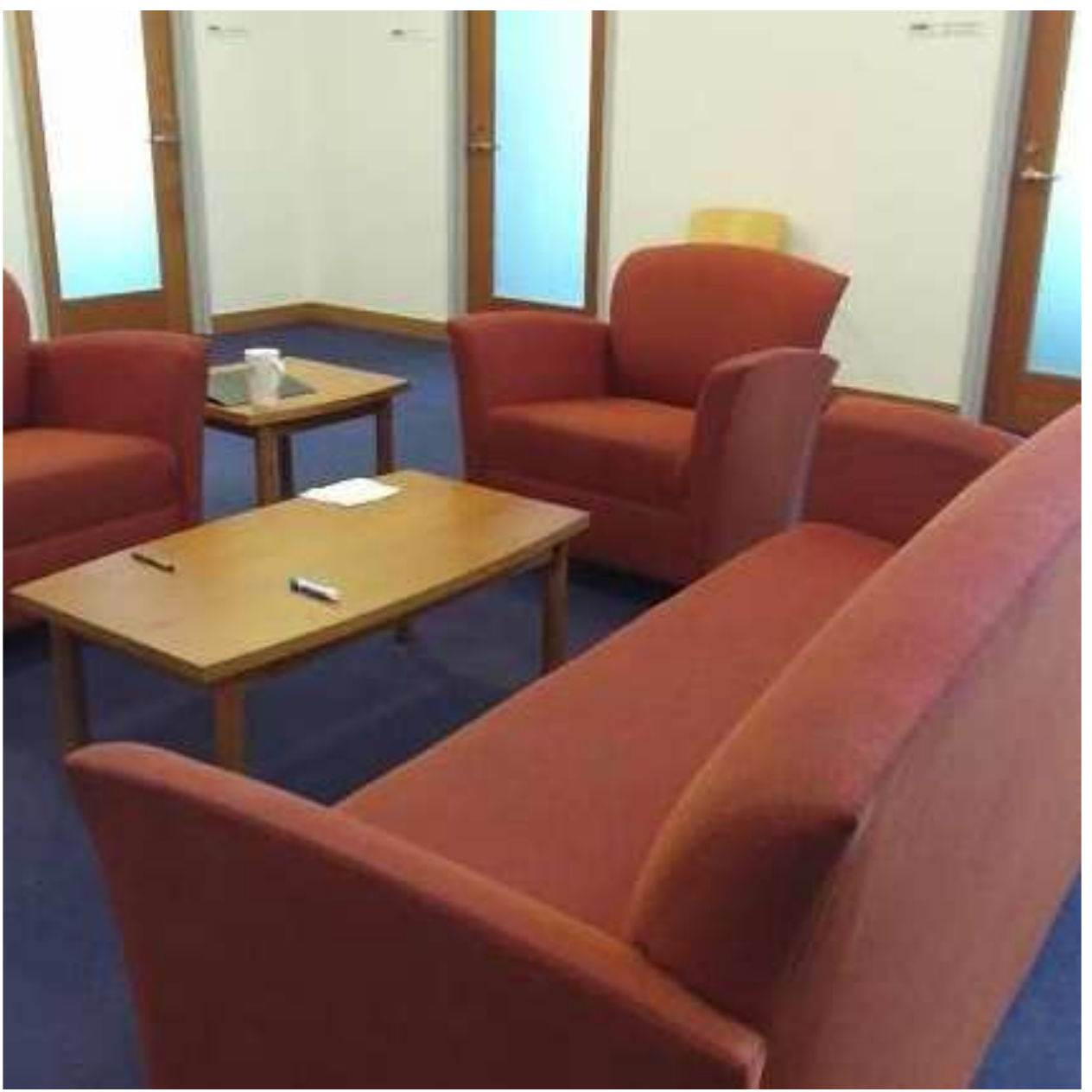} \\
\includegraphics[width=0.9\textwidth]{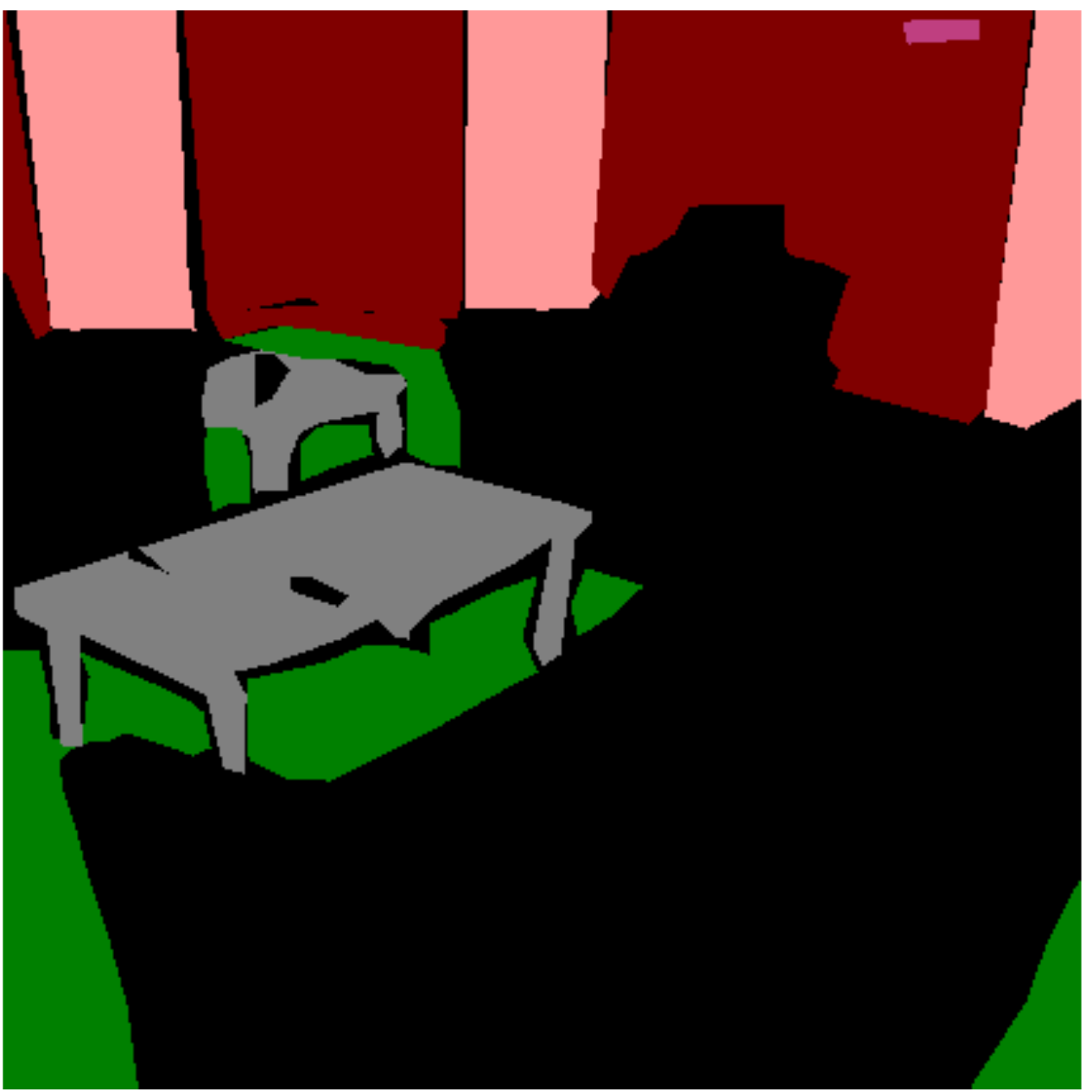} \\
\includegraphics[width=0.9\textwidth]{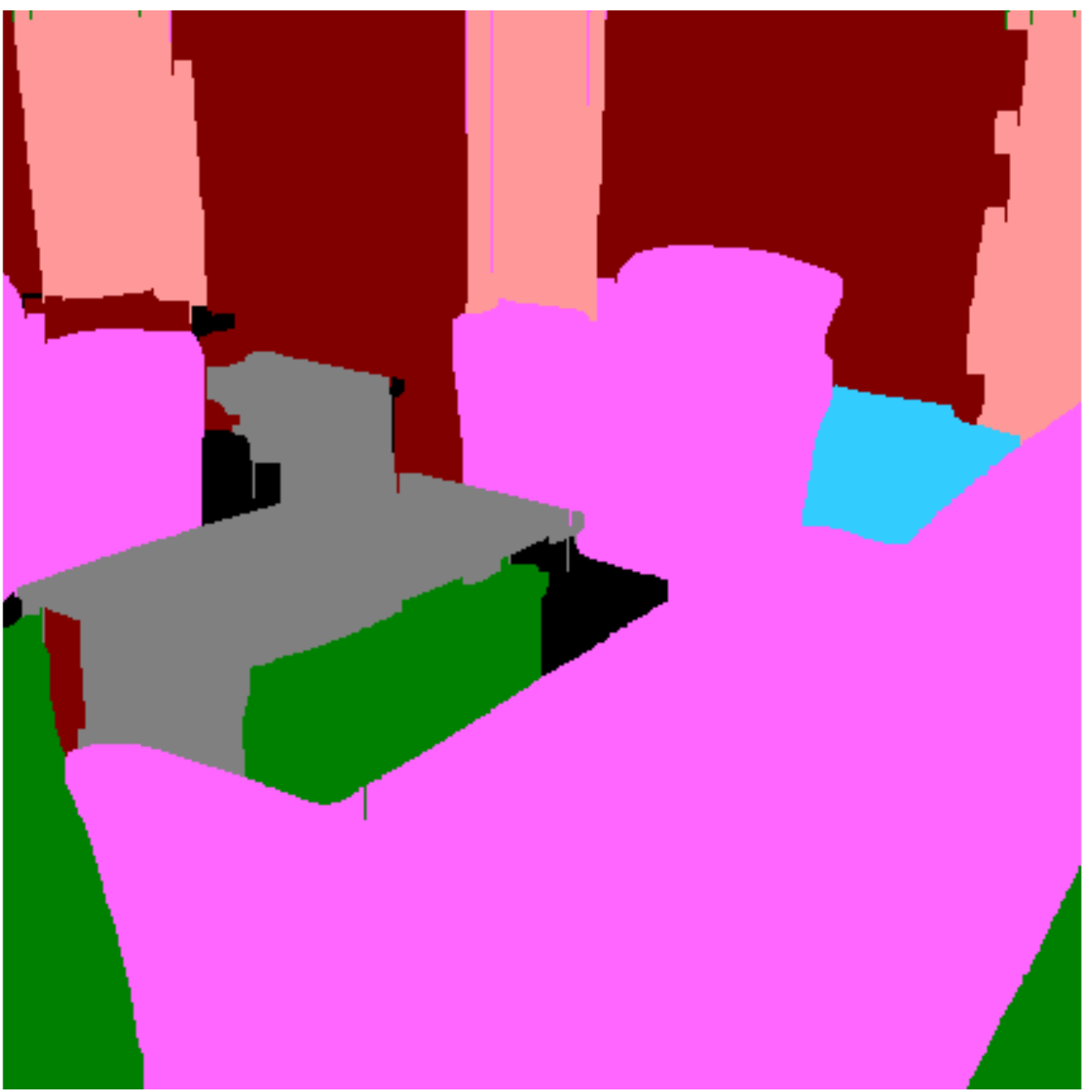}
\end{minipage}
}
\subfigure[]{
\begin{minipage}[b]{0.1\textwidth}
\includegraphics[width=0.9\textwidth]{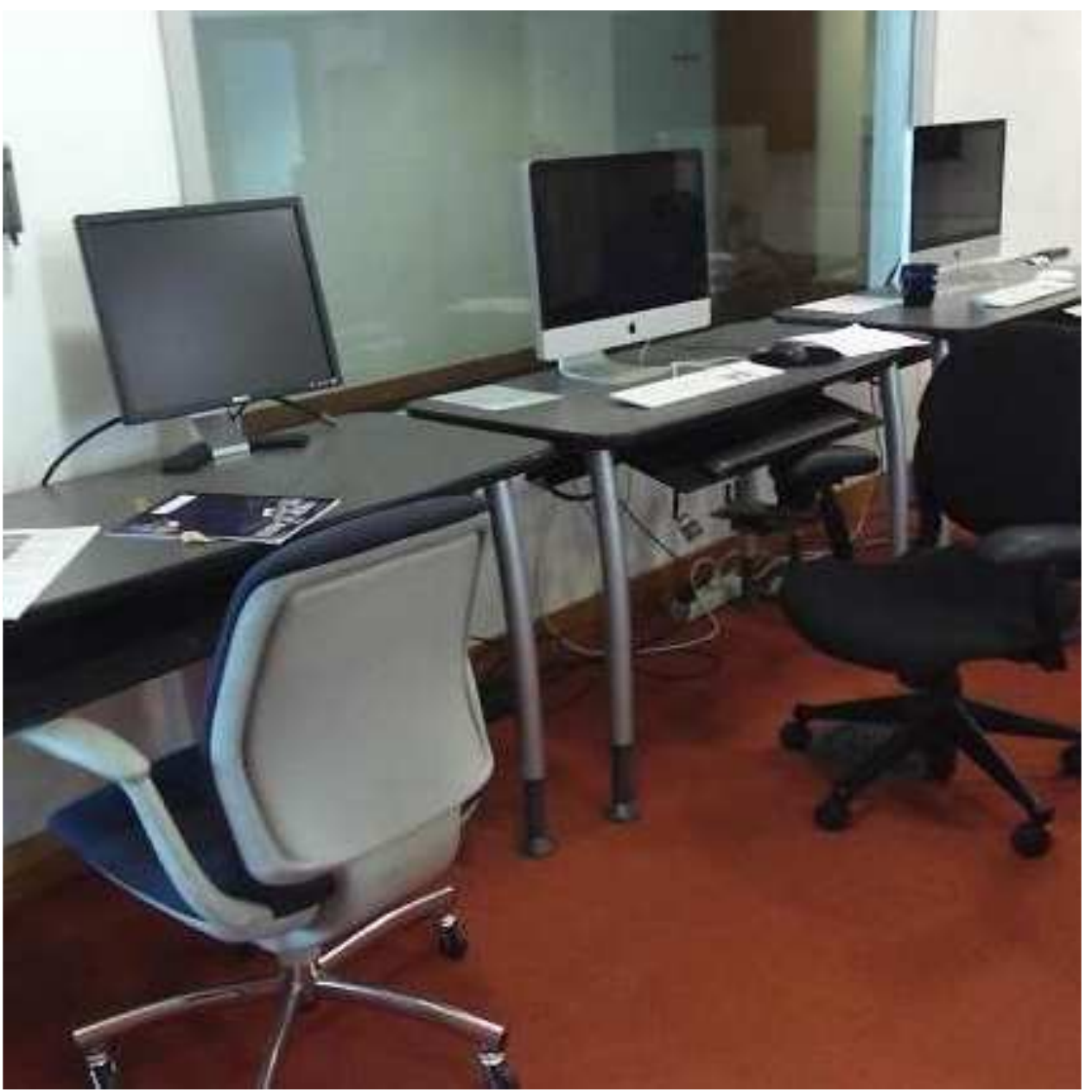} \\
\includegraphics[width=0.9\textwidth]{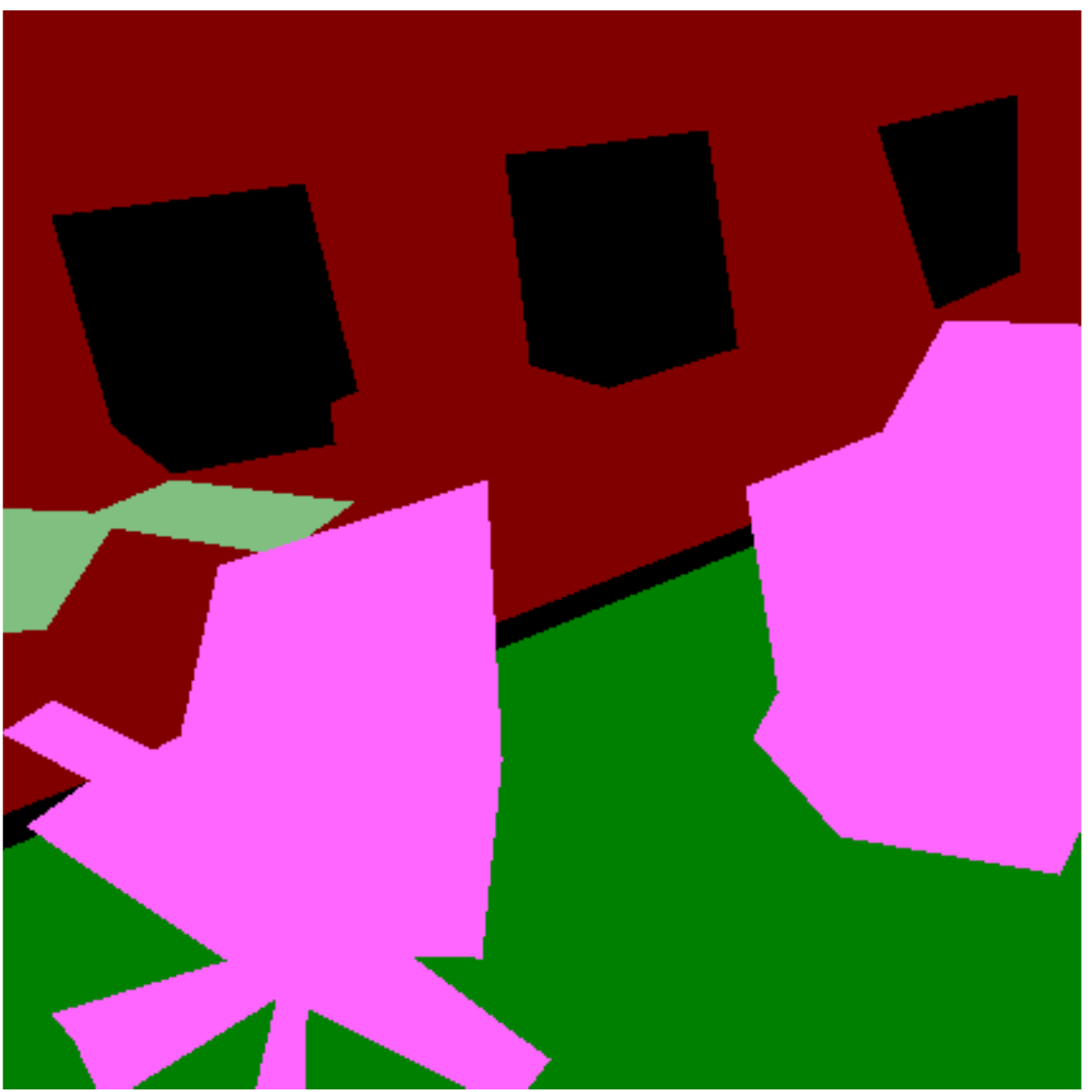} \\
\includegraphics[width=0.9\textwidth]{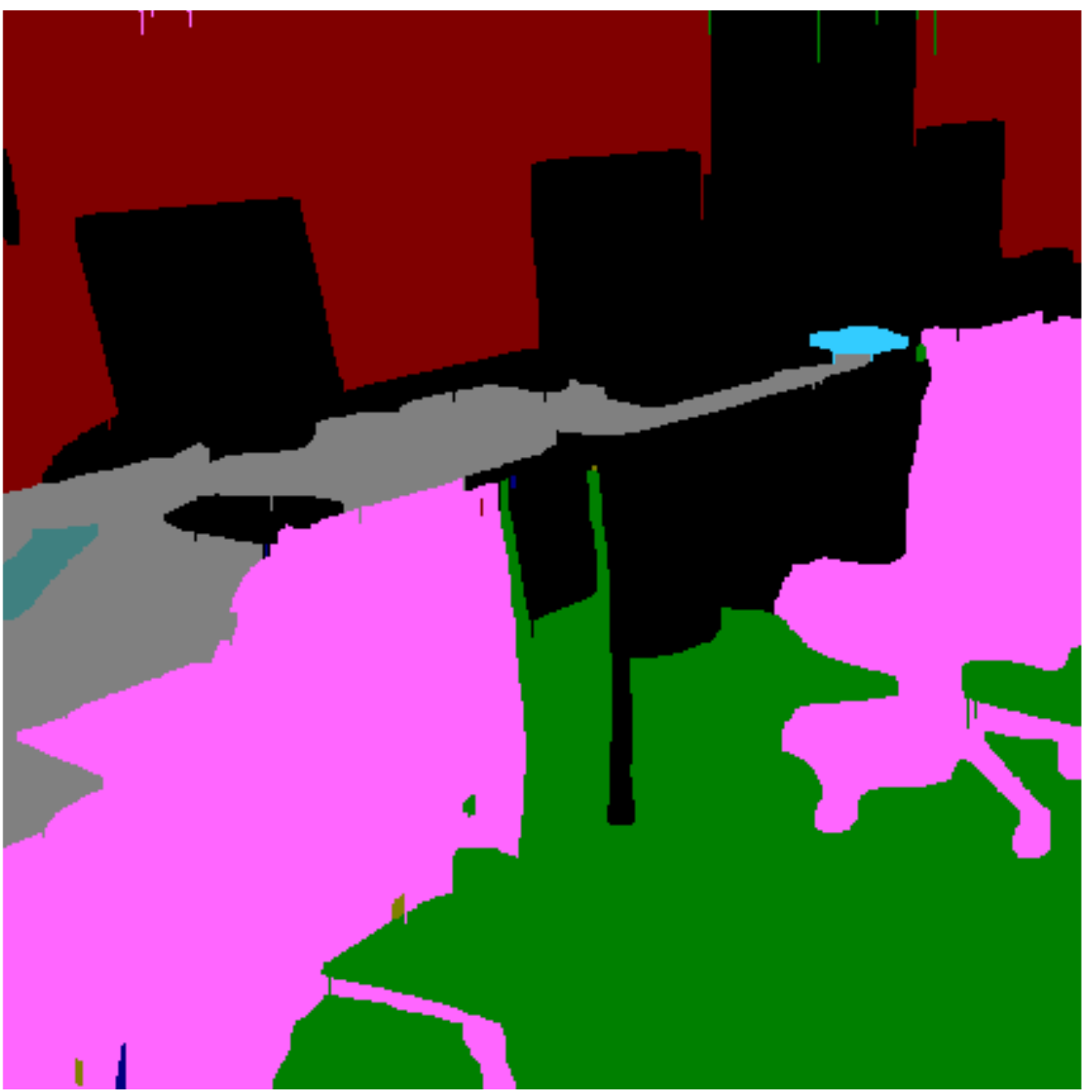}
\end{minipage}
}
\subfigure[]{
\begin{minipage}[b]{0.1\textwidth}
\includegraphics[width=0.9\textwidth]{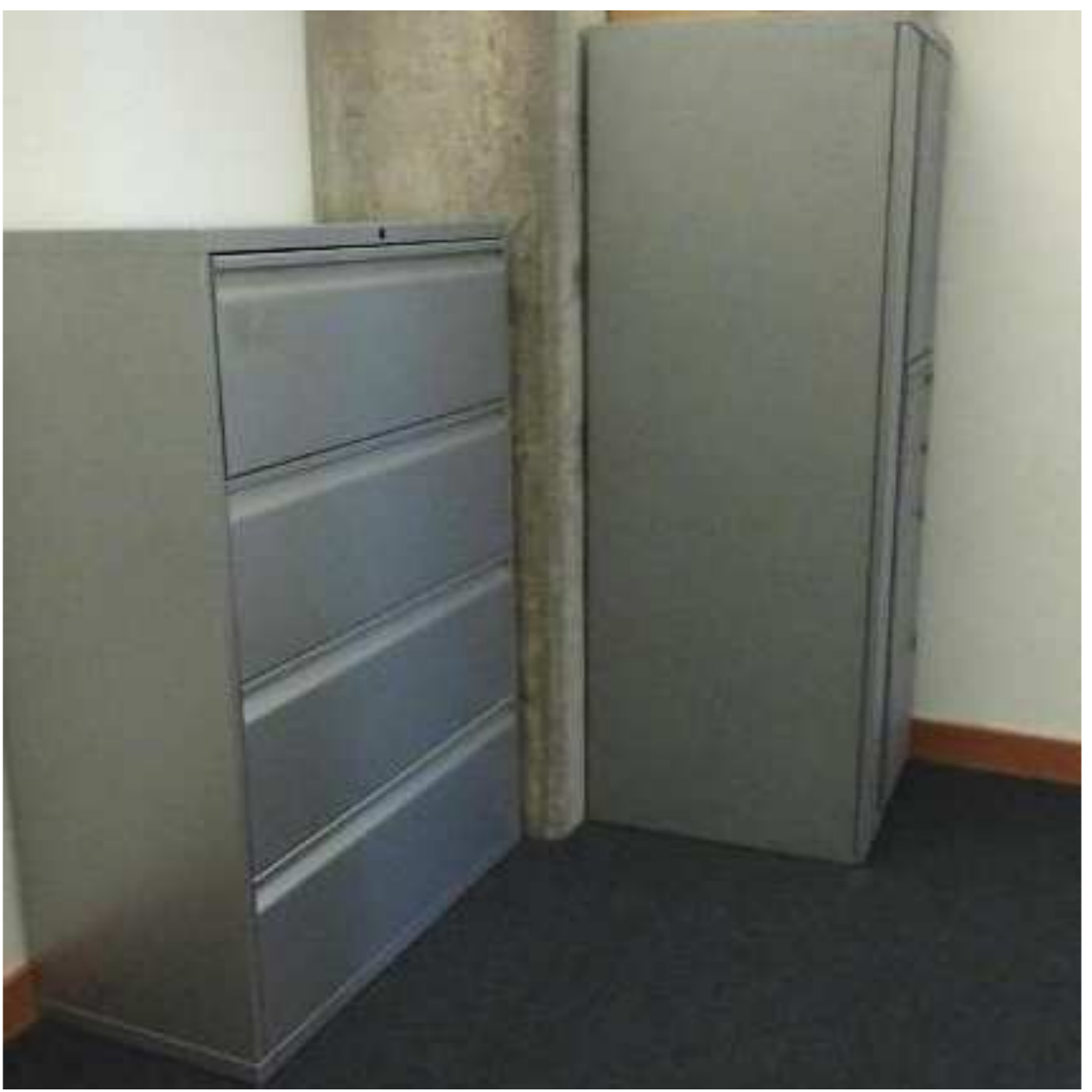} \\
\includegraphics[width=0.9\textwidth]{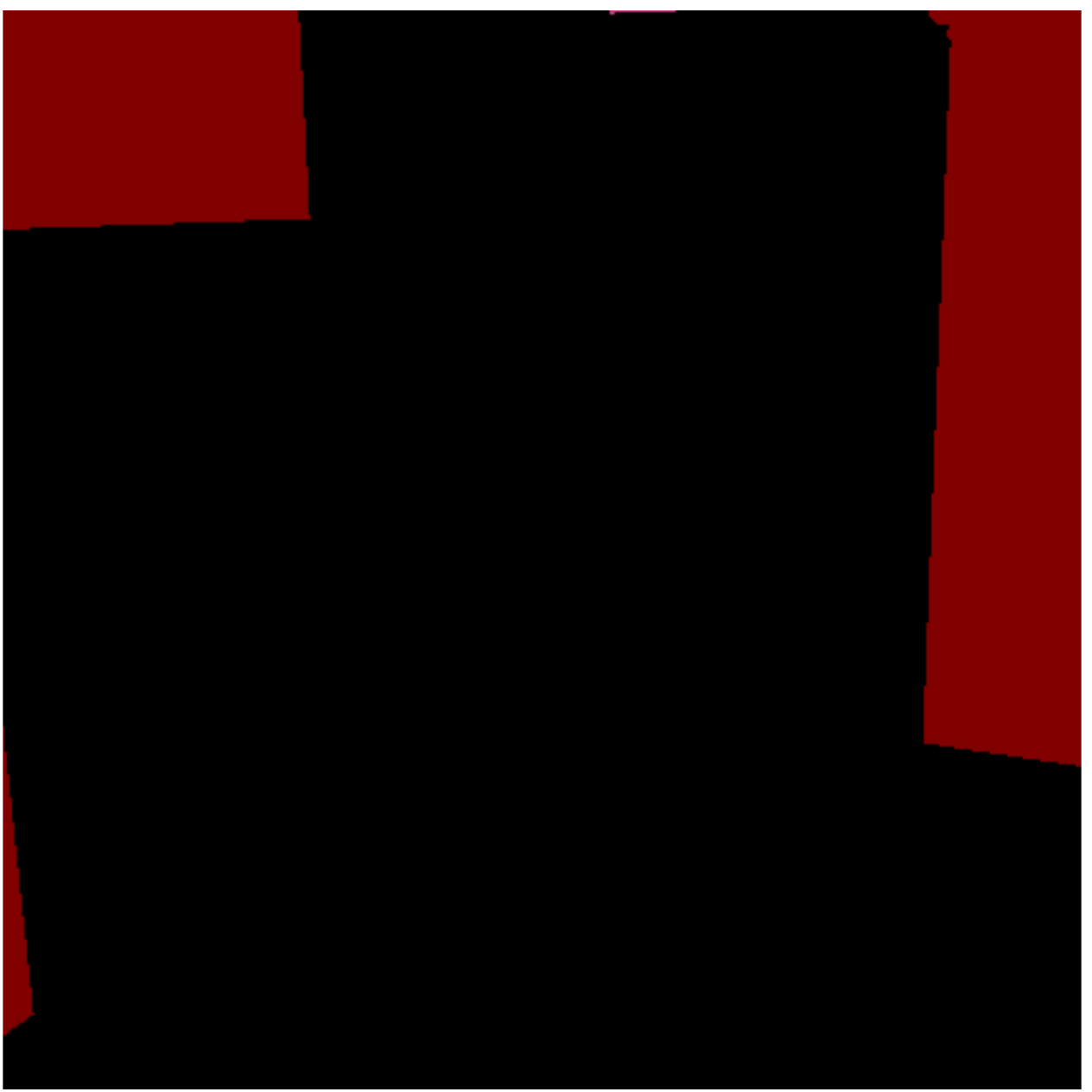} \\
\includegraphics[width=0.9\textwidth]{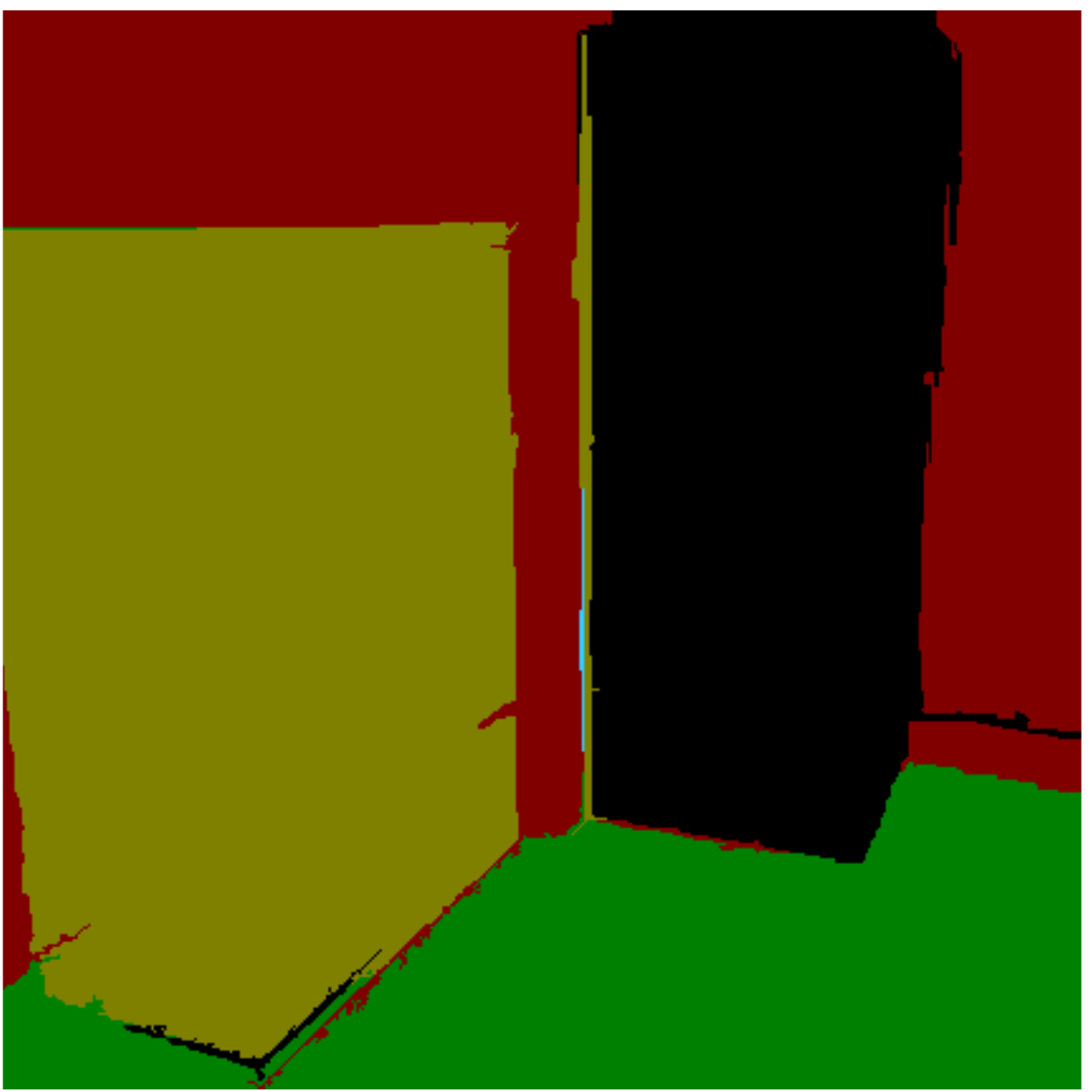}
\end{minipage}
}
\subfigure[]{
\begin{minipage}[b]{0.1\textwidth}
\includegraphics[width=0.9\textwidth]{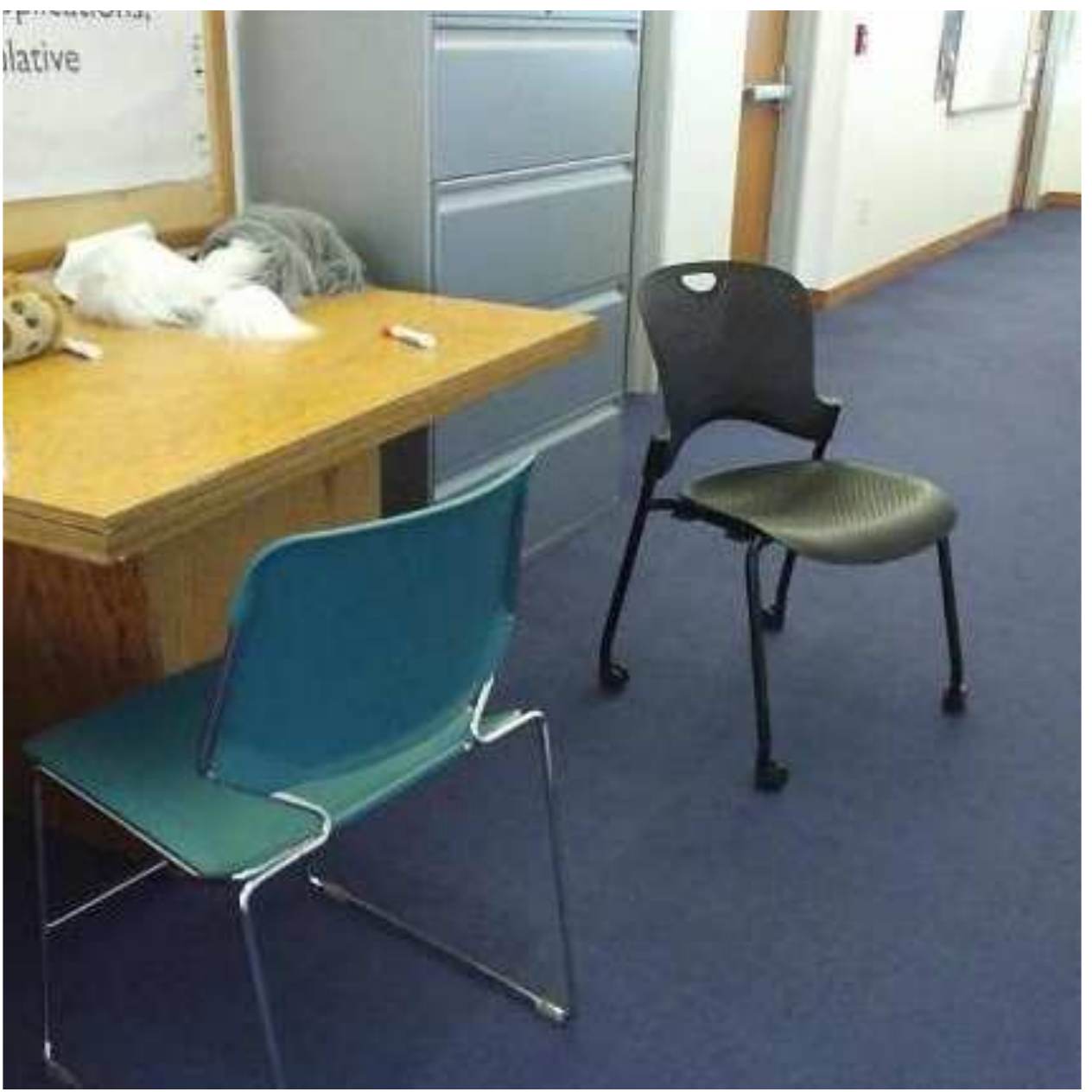} \\
\includegraphics[width=0.9\textwidth]{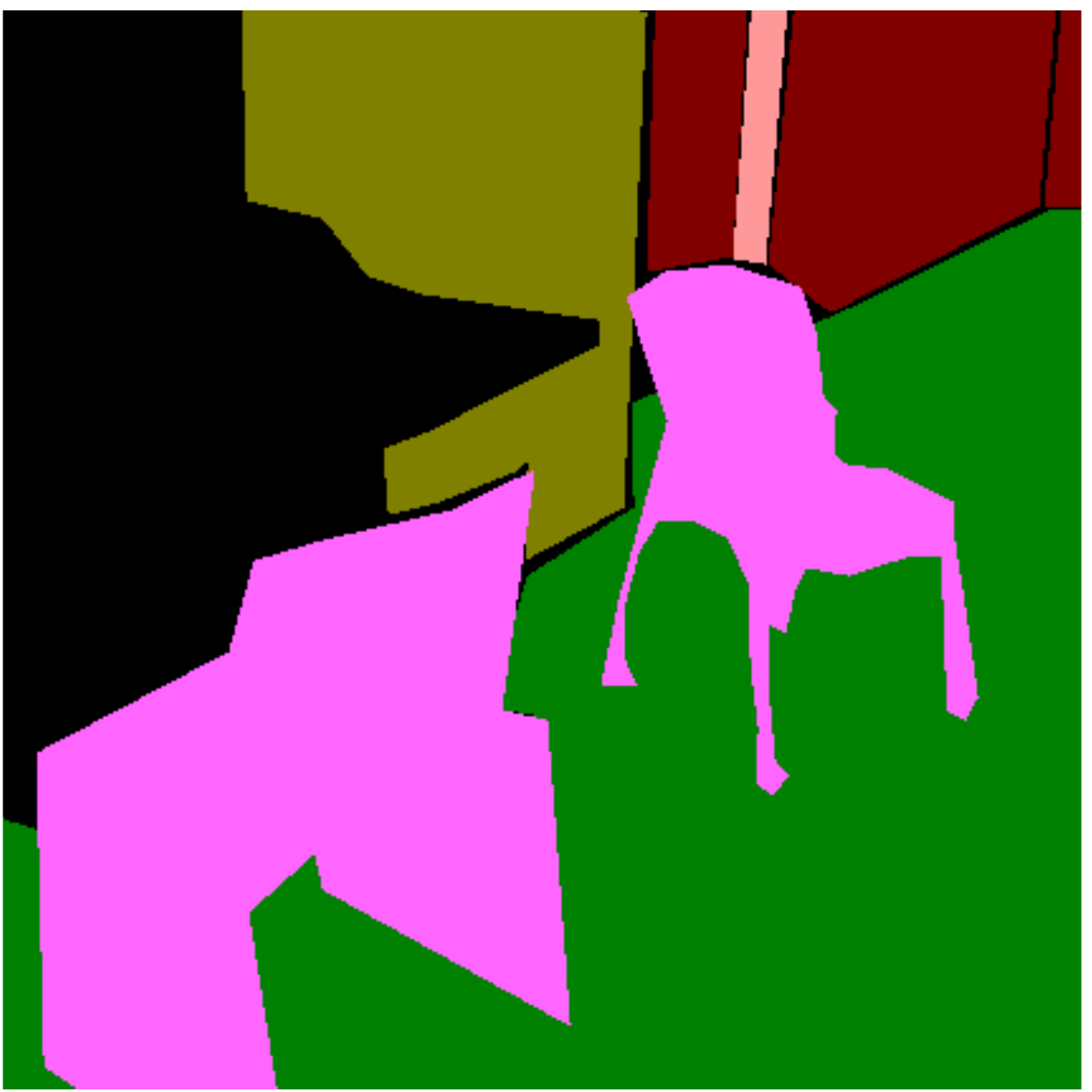} \\
\includegraphics[width=0.9\textwidth]{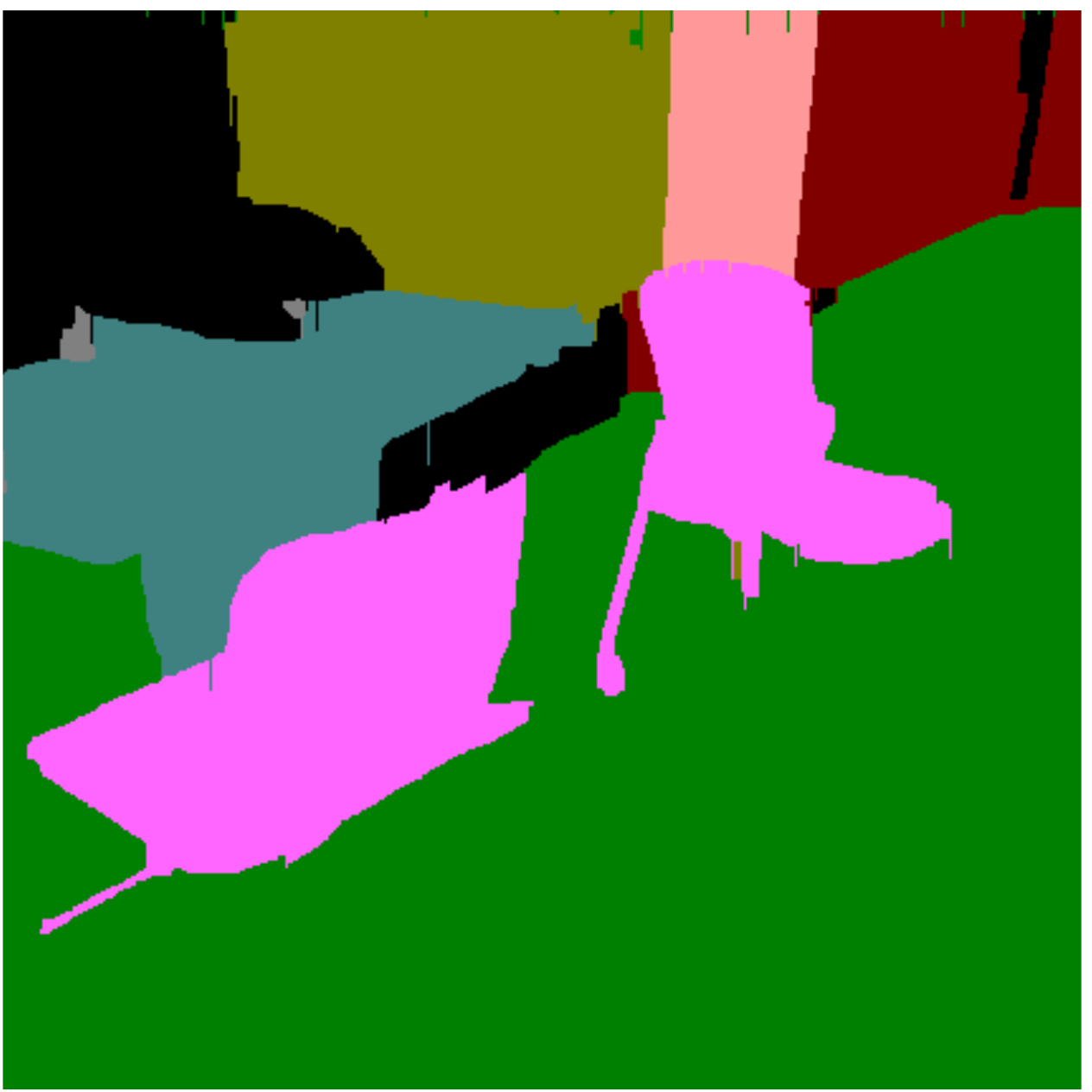}
\end{minipage}
}
\caption{Annotation refinement on the SUNRGBD dataset. The top row shows the input RGB images, the middle row shows the original annotations, and the bottom row shows scene labeling results from our model.}
\label{fig:refine}
\end{figure}

\subsection{Ablation Study}
To discover the vital elements in our proposed model, we conduct an ablation study to remove or replace individual components in our deep network when training and testing on the SUNRGBD dataset. Specifically, we have tested the performance of our model without the RGB path, the depth path, multi-scale RGB feature concatenation, the memorized context layers or the memorized fusion layer. In addition, we also conduct an experiment with a model that does not combine the final convolutional features of photometric channels (i.e., Conv$7$ in Fig. \ref{fig:pipeline}) with the global contexts of the complete RGB-D image to figure out the importance of different components. The results are presented in Table \ref{table.ablation}. From the given results, we find that the final convolutional features of the photometric channels is the most vital information, i.e., the cross-layer combination is the most effective component as the performance drops to $15.2\%$ without it, which is consistent with previously mentioned properties of depth and photometric data. In addition, multi-scale RGB feature concatenation before the memorized context layer also plays a vital role as it directly affects the vertical contexts in the photometric channels and the performance drops to $42.1\%$ without it. It is obvious that performance would be inevitably harmed without the depth path. Among the memorized layers, the memorized fusion layer is more important than the memorized context layers in our pipeline as it accomplishes the fusion of contexts in photometric and depth channels.

\subsection{Visual Comparisons}
{\bf SUNRGBD Dataset:}
We present visual results of RGB-D scene labeling in Fig. \ref{fig:visulization}. Here, we leverage super-pixel based averaging to smooth visual labeling results as being done in \cite{khan2015integrating}. The algorithm in \cite{felzenszwalb2004efficient} is used for performing super-pixel segmentation.  As can be observed in Fig. \ref{fig:visulization}, our proposed deep network produces accurate and semantically meaningful labeling results, especially for large regions and high frequency labels. For instance, our model takes advantage of global contexts when labeling `bed' in Fig. \ref{fig:visulization}(a), `wall' in Fig. \ref{fig:visulization}(e) and `mirror' in Fig. \ref{fig:visulization}(i). Our proposed model can precisely label almost all `chairs' (a high frequency label) by exploiting integrated photometric and depth information, regardless of occlusions.

{\bf NYUDv$2$ Dataset:} We also perform visual comparisons on the NYUDv$2$ benchmark, which has complicated indoor scenes and well-labeled ground truth. We compare our scene labeling results with those publicly released labeling results from \cite{gupta2015indoor}. It is obvious that our results are clearly better than those from \cite{gupta2015indoor} both visually and numerically (under the metric of average Jaccard Index) even though scene labeling in \cite{gupta2015indoor} is based on sophisticated segmentation.

{\bf{Label Refinement:}} Surprisingly, our model can intelligently refine certain region annotations, which might have inaccuracies due to under-segmentation, especially in the newly captured $3943$ RGB-D images, as shown in Fig. \ref{fig:refine}. Specifically, the cabinets in Fig. \ref{fig:refine}(a) were annotated as `background', the pillows in Fig. \ref{fig:refine}(g) as `bed', and the tables in Fig. \ref{fig:refine}(n) as `wall' by mistake. Our model can effectively deal with these difficult regions. For example, the annotation of the picture in Fig. \ref{fig:refine}(e) and that of the pillows in Fig. \ref{fig:refine}(g) have been corrected. Thus, our model can be exploited to refine certain annotations in the SUNRGBD dataset, which is another contribution of our model.

\section{Conclusions}
In this paper, we have developed a novel Long Short-Term Memorized Context Fusion (LSTM-CF) model that captures image contexts from a global perspective and deeply fuses contextual representations from multiple sources (i.e., depth and photometric data) for semantic scene labeling.  In future, we will explore how to extend the memorized layers with an attention mechanism, and refine the performance of our model in boundary labeling.

\clearpage

\bibliographystyle{splncs}
\bibliography{eccv16lz}
\end{document}